\documentclass[10pt,twocolumn,letterpaper]{article}

\usepackage{cvpr}              

\usepackage{makecell}
\usepackage{multirow}
\usepackage[table]{xcolor}
\usepackage[hang,flushmargin]{footmisc}
\usepackage{pifont}
\usepackage[numbers]{natbib}
\usepackage[figuresright]{rotating}
\usepackage{amsmath}

\definecolor{cvprblue}{rgb}{0.21,0.49,0.74}
\usepackage[pagebackref,breaklinks,colorlinks,allcolors=cvprblue]{hyperref}
\def\paperID{3126} 
\def\confName{CVPR}
\def\confYear{2026}
\definecolor{firebrick}{RGB}{178,34,34}
\newcommand{\supp}{Appendix\xspace}
\newcommand{\hytt}[1]{\texttt{\hyphenchar\font=\defaulthyphenchar #1}}

\title{How far have we gone in Generative Image Restoration?\\A study on its capability, limitations and evaluation practices}

\author{
Xiang Yin$^1$, Jinfan Hu$^{23}$, Zhiyuan You$^{24}$, Kainan Yan$^{23}$, Yu Tang$^2$, Chao Dong$^{25}$, Jinjin Gu$^6$\\
\small $^1$Fudan University\quad $^2$Shenzhen Institutes of Advanced Technology, Chinese Academy of Sciences\\
\small$^3$University of the Chinese Academy of Sciences\quad$^4$Multimedia Laboratory, The Chinese University of Hong Kong\\
\small$^5$Shenzhen University of Advanced Technology\quad$^6$INSAIT, Sofia University ``St. Kliment Ohridski''\\
\tt \small \url{https://github.com/yxyuanxiao/how-far-have-we-gone}\\
}

\begin{document}

\twocolumn[{
\renewcommand\twocolumn[1][]{#1}
\maketitle
\thispagestyle{empty}
\vspace{-25pt}
\begin{center}
    \centering
    \includegraphics[width=0.95\linewidth]{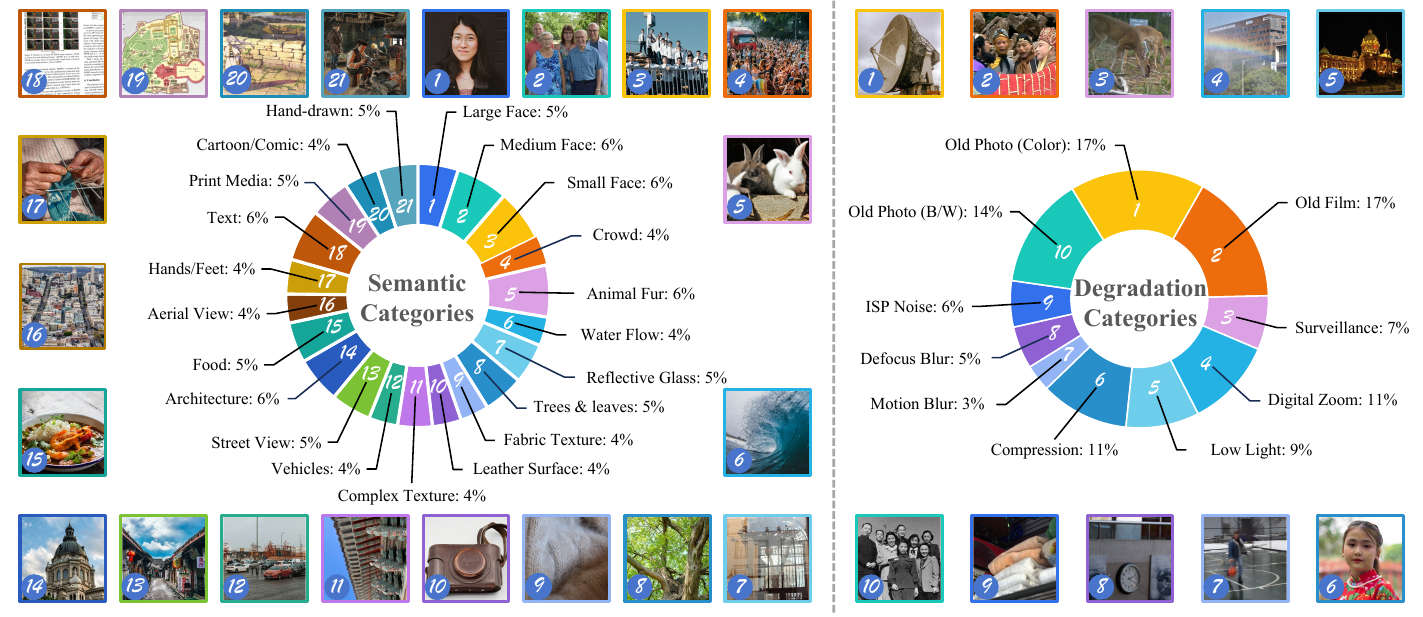}
    \vspace{-10pt}
    \captionof{figure}{
    Overview of the proposed dataset composition across \textbf{Semantic} and \textbf{Degradation} dimensions.  
    The semantic categories (left) cover diverse visual contents to reveal the perceptual and structural challenges faced by generative restoration models.  
    The degradation categories (right) reflect real-world conditions where restoration models are typically applied.  
    }
\label{fig:teaser}
\end{center}
}]

\begin{abstract}
Generative Image Restoration (GIR) has achieved impressive perceptual realism, but how far have its practical capabilities truly advanced compared with previous methods? 
To answer this, we present a large-scale study grounded in a new multi-dimensional evaluation pipeline that assesses models on detail, sharpness, semantic correctness, and overall quality.
Our analysis covers diverse architectures, including diffusion-based, GAN-based, PSNR-oriented, and general-purpose generation models, revealing critical performance disparities.
Furthermore, our analysis uncovers a key evolution in failure modes that signifies a paradigm shift for the perception-oriented low-level vision field.
The central challenge is evolving from the previous problem of detail scarcity (under-generation) to the new frontier of detail quality and semantic control (preventing over-generation).
We also leverage our benchmark to train a new IQA model that better aligns with human perceptual judgments. 
Ultimately, this work provides a systematic study of modern generative image restoration models, offering crucial insights that redefine our understanding of their true state and chart a course for future development.

\end{abstract}
\vspace{-10pt}

\vspace{-4pt}
\section{Introduction}
\label{sec:intro}
\vspace{-2pt}

Generative image restoration (GIR) has brought remarkable progress to the field of image restoration.
By leveraging generative priors, recent models~\cite{srgan,esrgan,sr3,stablesr,supir,hypir} achieve visually realistic and perceptually pleasing restoration results, producing fine textures and details unattainable by conventional methods~\cite{diffusion_survey}.
At the same time, these methods seem to exhibit strong generalization, performing surprisingly well even on unseen or mixed degradations.
However, this powerful generative capacity, while beneficial, also raises fundamental questions about the fidelity and controllability of restored content, especially when applied to real-world degradations and diverse semantic categories.
These issues, though increasingly prevalent, have not been systematically studied or adequately reflected. 
Focusing on GIR methods, particularly diffusion-based approaches, we aim to provide a comprehensive understanding of both the remarkable advances and the emerging risks of GIR.


Generative priors, the foundation of recent progress, can also be a double-edged sword.
In earlier paradigms, restoration models lacked the ability to generate new content. 
By contrast, modern GIR models with strong generative capabilities can enhance realism but also lead to hallucinated details, structural distortions, and semantic inconsistencies.
Controlling this generative capability has thus become a critical and challenging problem: the stronger the generative prior, the harder it is to balance realism and fidelity.
Although modern GIR models perform the aforementioned tasks, yet a comprehensive understanding of when and why GIR models fail under different conditions is still missing.

Limitations lie in how current GIR models are evaluated.
Existing benchmarks~\cite{isrgenqa,afine} and IQA methods largely overlook semantic dependency, treating all scenes as equally difficult.
Moreover, most existing datasets~\cite{pieapp,pipal} focus only on conventional synthetic degradations, such as downsampling, noise, compression, or blur.
These degradations do not reflect the real-world conditions in which GIR models are commonly applied, such as Internet images, old films, and old photos.
Furthermore, most assessments collapse image quality into a single numerical score, making it difficult to identify the underlying causes of failure.

In this paper, we present a systematic and fine-grained study of state-of-the-art image restoration models.
Specifically, (1) we perform detailed analyses across both semantic scenes and degradation types, constructing a carefully balanced dataset with fine-grained semantic and degradation annotations;  
(2) we quantify image quality at a fine-grained level and go beyond simple scoring by categorizing the diverse failure patterns observed in GIR models
; and (3) we use this dataset to train an image quality assessment (IQA) model capable not only of predicting perceptual quality but also of diagnosing the specific failure modes present in a restored image.
Our study thus enables a more comprehensive and diagnostic understanding of GIR model performance and provides a foundation for future evaluation and the development of agent-based image restoration systems.

\vspace{-3pt}
\section{Related Work}
\label{sec:related}
\vspace{-3pt}

\textbf{Image Restoration} (IR) aims to recover high-quality images from their degraded counterparts. 
Early methods focused on specific degradations such as denoising~\cite{DnCNNs,mask_denoise,deep_graph_denoise}, deblurring~\cite{deepcnn_deblur,HI-Diff,deblurgan}, deraining~\cite{clearing,slacking_off}, and super-resolution~\cite{SRCNN,san,esrgan,edsr,hat,swinir}. 
These models typically relied on hand-crafted priors or task-specific convolutional architectures~\cite{diffusion_survey}, thus exhibiting limited generalization to unseen degradations.
With the rise of generative modeling, GAN-based IR methods~\cite{bsrgan,calgan,realesrgan} introduced adversarial objectives to enhance perceptual realism beyond pixel-level fidelity. 
Although GANs improved visual quality, they often generated unnatural textures and suffered from unstable training~\cite{diffusion_survey}. 
To overcome these limitations, diffusion-based generative models have recently emerged as a more stable and expressive alternative, demonstrating superior performance in both fidelity and diversity~\cite{diffusion_survey}. 
Recently, diffusion models~\cite{refusion,resshift} have demonstrated remarkable generative capability and surpassed GANs in both fidelity and diversity.
Recent studies~\cite{supir,diffbir,pisasr,seesr,OSEDiff,PASD,hypir} have further leveraged large-scale diffusion backbones for real-world image restoration, achieving photorealistic and semantically consistent results across diverse degradations.

\vspace{2pt}\noindent\textbf{Image Quality Assessment} (IQA) plays a vital role in evaluating the perceptual quality of restored or generated images. 
IQA methods can be categorized into \textit{full-reference} and \textit{non-reference} types. 
Full-reference IQA approaches compute a similarity-based quality score between a distorted image and its corresponding high-quality reference. 
Classical methods like PSNR and SSIM~\cite{ssim} rely on hand-crafted perceptual metrics~\cite{fsim,vif}. 
Pioneered by LPIPS~\cite{lpips} and PieAPP~\cite{pieapp}, and supported by large-scale IQA datasets~\cite{lpips, pipal, kadid, live}, deep-learning methods~\cite{WaDIQaM, JSPL, dists, A-DISTS, ghildyal2022stlpips, CVRKD, SRIF} have achieved high accuracy in regressing perceptual quality scores. 
Non-reference IQA methods directly predict a quality score without requiring any reference. 
Early works relied on hand-crafted natural image statistics~\cite{ma2017learning, brisque, niqe, moorthy2010two, DIIVINE, saad2012blind, tang2011learning}, 
while later deep-learning-based methods~\cite{CNNIQA, RankIQA, BPSQM, HyperIQA, graphiqa, CKDN, MetaIQA, zhang2022continual, paq2piq, unique, musiq, maniqa, liqe} replaced these handcrafted features with learned perceptual priors from human-annotated datasets~\cite{koniq, spaq, kadid, agiqa}. 
Recent methods incorporate vision-language models~\cite{clipiqa,qalign,deqa_score,depictqa,depictqav2,qinsight,visualqualityr1}, enabling multimodal reasoning and richer perceptual understanding through textual semantics.

\vspace{2pt}\noindent\textbf{IQA for model-restored images}. 
Most existing IQA models still struggle to capture subtle perceptual or semantic discrepancies introduced by modern generative restoration methods, as indicated by benchmarks like FGResQ~\cite{FGResQ}, SRIQA-Bench~\cite{afine}, and ISRGen-QA~\cite{isrgenqa}. 
Nonetheless, the coverage of these benchmarks remains limited, lacking coverage of the semantic and content-level variations introduced by modern generative restoration or generation models. 
In this work, we focus on the emerging challenge of evaluating \textit{generative image restoration}, where both traditional and multimodal IQA methods fail to effectively capture hallucination tendencies and semantic shifts. 

\vspace{-3pt}
\section{Evaluation Design}
\label{sec:method}
\vspace{-3pt}

\begin{figure*}[tp]
\scriptsize
    \centering
    \includegraphics[width=0.95\linewidth]{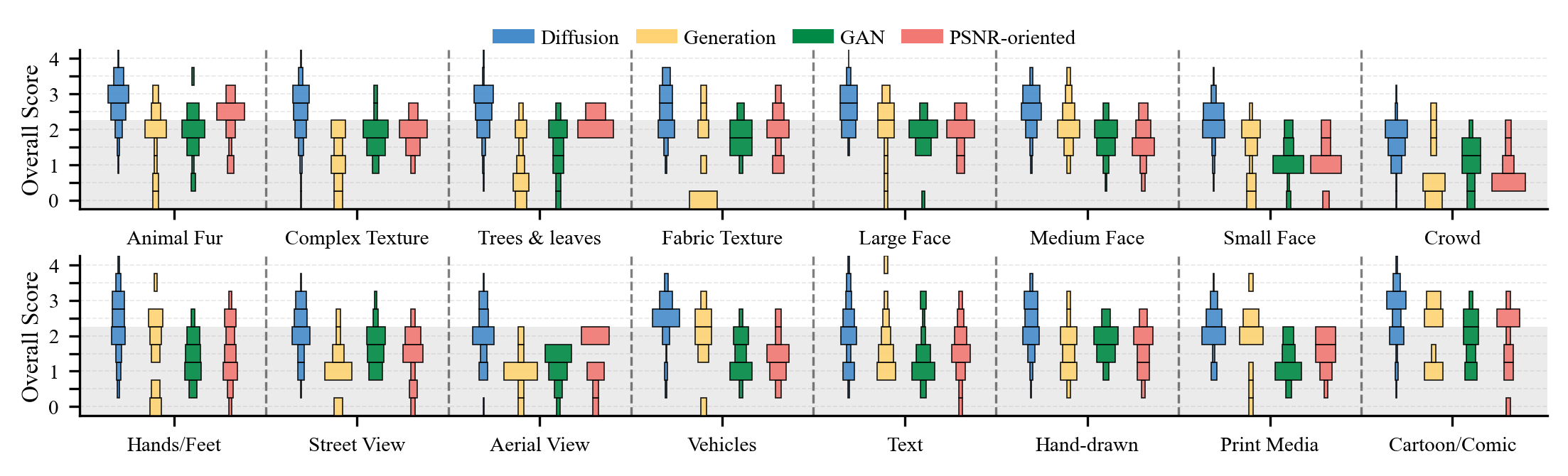}
    \vspace{-14pt}
    \caption{
    Distribution of annotated total scores across semantic scene groups.  
    The horizontal width of each box indicates the percentage of samples within each score interval.  
    The light gray region indicates low overall scores, representing generally unacceptable results.
    }
    \label{fig:semantics_total_score}
    \vspace{-18pt}
\end{figure*}

\subsection{Dataset Collection}
\vspace{-3pt}
The datasets employed in previous benchmarks are typically constructed by mixing a wide variety of images from many categories, while treating the semantic content of each image as incidental rather than a controlled factor.
As a result, these benchmarks largely support only holistic analyses based on average performance metrics~\cite{afine,isrgenqa}, and offer little insight into how GIR models behave on different kinds of images.
Such aggregate analysis is no longer sufficient for contemporary GIR models.
Their integration of semantic understanding and generative capability leads to strongly semantic sensitive behavior, where performance can vary dramatically across categories such as text, hands, and faces~\cite{chen2023textdiffuser,zhang2025hand1000,liang2025authface}.
Yet there is still no dedicated dataset or benchmark that systematically examines how semantic content and degradation type jointly affect model behavior.
To fill this gap, we construct a new testset that deliberately balances diverse semantic categories and degradation types, with a particular focus on those that are known to be challenging for current GIR models.

\vspace{2pt}\noindent\textbf{Semantic categories}.
The semantic dimension of our dataset is curated with two principles:
(1) the selected scenes should be challenging for current GIR models; and  
(2) they should cover content that is both perceptually salient to humans and diagnostically valuable for evaluating generative restoration quality.
As illustrated in~\cref{fig:teaser}, our semantic categories span a wide range of human-centric, structural, texture-related, and symbolic scenes that comprehensively reflect real-world visual diversity.
Human-centric categories such as \textit{faces} and \textit{hands/feet} are included.
Their distinctive geometric structure and familiarity to humans, where even slight distortions or inconsistencies are easily noticeable and thus serve as clear indicators of restoration fidelity.
Smaller faces suffer from severe degradation and rely more heavily on generative priors to reconstruct plausible features.
While in the \textit{crowd} category, numerous small, variably oriented faces further challenge a model’s ability to infer orientation and maintain facial geometry.
Structure-related categories such as \textit{vehicles, architecture, street view, aerial view} and \textit{food} are included to evaluate a model’s capacity for geometric layout and spatial consistency.
These categories cover both rigid and non-rigid structures, ranging from buildings and cars to food, and reflect how well models preserve perspective relations, proportions, and scene layouts under degradation.
Texture-rich materials such as \textit{animal fur, water flow, fabric, leather} and  \textit{trees \& leaves} are included to assess whether models can generate fine-grained and coherent textures that appear both natural and physically plausible.
We also include stylized and symbolic content such as \textit{text, printed media, cartoons/comics,} and \textit{hand-drawn}.
These scenes emphasize the challenge of preserving semantic and stylistic fidelity, where text demands pixel-level precision for legibility, and artistic depictions require consistent style maintenance.

\begin{figure}[tp]
\scriptsize
\centering
    \includegraphics[width=\linewidth]{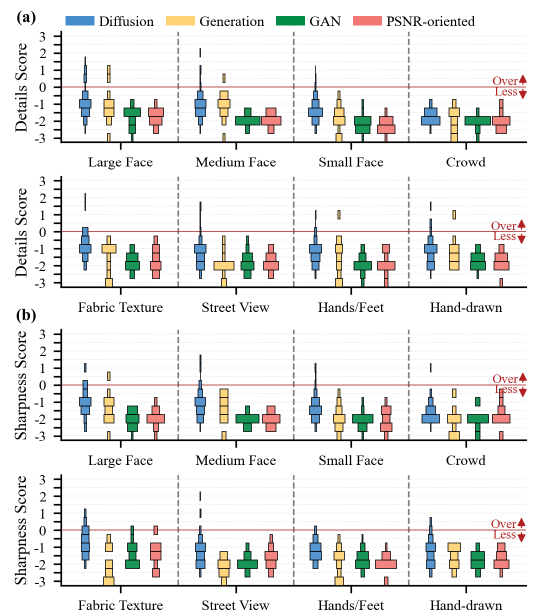}
    \vspace{-15pt}
    \caption{
        Detail and sharpness score distributions across semantic scenes. The width of each box corresponds to the percentage of samples. The red line indicates the balance point; scores above it (\textcolor{firebrick}{Over $\uparrow$}) denote over-generation, and scores below it (\textcolor{firebrick}{Less $\downarrow$}) denote under-generation.
    }
\label{fig:detail_sharpness_8scenes}
\vspace{-18pt}
\end{figure}

\vspace{2pt}\noindent\textbf{Degradation types}.
The treatment of degradation in previous testsets is also rather limited.
Most benchmarks adopt only a small set of simple degradations, with little deliberate design or diversity.
This narrow coverage can be traced back to the weak generalization of earlier learning-based restoration models.
Since these models rarely improved images corrupted by unfamiliar degradations, evaluating them on richer or more realistic degradation types was considered unnecessary or impractical \cite{10947624,2021Discovering}. 
With the emergence of diffusion-based GIR models, this assumption no longer holds.
Their strong content priors substantially enhance generalization and allow the model to recover natural content instead of merely removing specific corruptions \cite{hu2025revisiting}. 
Moreover, modern GIR models also exhibit highly nonuniform behavior across degradation types, performing well on some degradations while failing on others, making degradation-specific evaluation essential.
To match this new capability, the evaluation protocol must also expand to include a broader and more realistic spectrum of degradations.
We therefore construct a degradation suite that intentionally covers both synthetic and authentic real-world scenarios, as illustrated in~\cref{fig:teaser}.
Our final dataset consists of two complementary components: a synthetic subset of 147 high-quality images across 21 semantic categories, degraded using the RealESRGAN \cite{realesrgan} pipeline, and a real-world subset of 207 degraded images.
Together, these components provide both semantic and degradation diversity, enabling a more comprehensive evaluation of GIR models.

\begin{table}[tp]
\centering
\footnotesize
\setlength\tabcolsep{2.9pt}
\begin{tabular}{cc|cc}
\toprule
Model & Architecture & Model & Architecture \\
\midrule
HYPIR~\cite{hypir} & Diffusion-based & S3Diff~\cite{s3diff} & Diffusion-based \\
SUPIR~\cite{supir} & Diffusion-based & TSD-SR~\cite{tsdsr} & Diffusion-based \\
PiSA-SR~\cite{pisasr} & Diffusion-based & ResShift~\cite{resshift} & Diffusion-based \\
SeeSR~\cite{seesr} & Diffusion-based & FLUX~\cite{flux} & Generation \\
OSEDiff~\cite{OSEDiff} & Diffusion-based & Nano Banana~\cite{nanobanana} & Generation\\
CCSR~\cite{ccsr} & Diffusion-based & BSRGAN~\cite{bsrgan} & GAN-based \\
DiffBIR~\cite{diffbir} & Diffusion-based & CAL-GAN~\cite{calgan} & GAN-based \\
StableSR~\cite{stablesr} & Diffusion-based & RealESRGAN~\cite{realesrgan} & GAN-based \\
PASD~\cite{PASD} & Diffusion-based & HAT~\cite{hat} & PSNR-oriented \\
Invsr~\cite{invsr} & Diffusion-based & SwinIR~\cite{swinir} & PSNR-oriented \\
\bottomrule
\end{tabular}
\vspace{-10pt}
\caption{
Restoration or generation models used in the benchmark.
}
\vspace{-18pt}
\label{tab:model_list}
\end{table}

\vspace{-6pt}
\subsection{Restoration Models Selection}
\vspace{-6pt}
In this work, we emphasize a comparative analysis of generative paradigms that prioritize perceptual quality in image restoration. To this end, we select 20 representative models spanning four distinct families, as listed in \cref{tab:model_list}, which together generate 7,080 restored images on our test set.
Rather than exhaustively surveying all methods, our selection captures the diverse design philosophies and generative behaviours exhibited across model families.  
Specifically, we include the following four categories.
\noindent\textbf{(1) Diffusion-based models.} 
This group represents the current mainstream of GIR, leveraging diffusion processes to iteratively refine degraded inputs into high-quality outputs.  
These models include both those built upon large-scale pre-trained diffusion backbones (\textit{e.g.,} Stable Diffusion or Flux) and those trained from scratch specifically for restoration, differing mainly in whether their generative priors originate from external pretraining or task-specific supervision.
\noindent\textbf{(2) General image generation models}. These models (e.g., Nano Banana and Flux Kontext) perform image-to-image translation beyond conventional restoration pipelines, effectively blurring the boundary between image restoration and open-ended generation.
\noindent\textbf{(3) GAN-based models}. 
Generative adversarial networks have long dominated perceptual image restoration before the diffusion era.
Although their generative priors are generally weaker than those of diffusion-based models, they remain crucial baselines for comparison.  
\noindent\textbf{(4) PSNR-oriented models}. 
PSNR-oriented models achieve sharper edges and higher PSNR performance~\cite{pieapp}, representing restoration techniques trained under deterministic supervision.
They rely less on generative priors and more on spatial–contextual modelling, providing a non-generative reference for evaluating realism–fidelity boundaries.
These models demonstrate the evolving boundary between image restoration and general image generation, offering insight into how generative architectures behave in image restoration.



\begin{table}[tp]
\centering
\footnotesize
\setlength\tabcolsep{2pt}
\begin{tabular}{l|cccc}
\toprule
Dataset & \# Images & \# IR Model & \# GIR Models & Multi. Aspect \\
\midrule
SRIQA-Bench~\cite{afine} & 1.1K & 10 & 8 & \ding{55} \\
ISRGen-QA~\cite{isrgenqa} & 0.72K & 15 & 10 & \ding{55} \\
\midrule
\textbf{Ours} & 7K & 20 & 18 & \ding{51} \\
\bottomrule
\end{tabular}
\vspace{-9pt}
\caption{
Comparison of existing datasets for GIR assessment.
}
\vspace{-20pt}
\label{tab:benck_compare}
\end{table}

\vspace{-6pt}
\subsection{Design of the Human Evaluation}
\vspace{-5pt}
The goal of our human evaluation is not only to obtain an overall quality score for each restored image but also to understand why an image is perceived as good or bad.
Unlike previous studies~\cite{afine,isrgenqa} that rely solely on a single holistic score, our design introduces a multi-dimensional scoring framework that captures fine-grained perceptual and semantic aspects of generative restoration.
This allows us to analyze specific failure sources and perceptual biases of different models.
To achieve this, we identify the key perceptual dimensions most relevant to image restoration quality: Detail, Sharpness, Semantics, and Overall score.

\begin{figure}[tp]
\scriptsize
\centering
    \includegraphics[width=\linewidth]{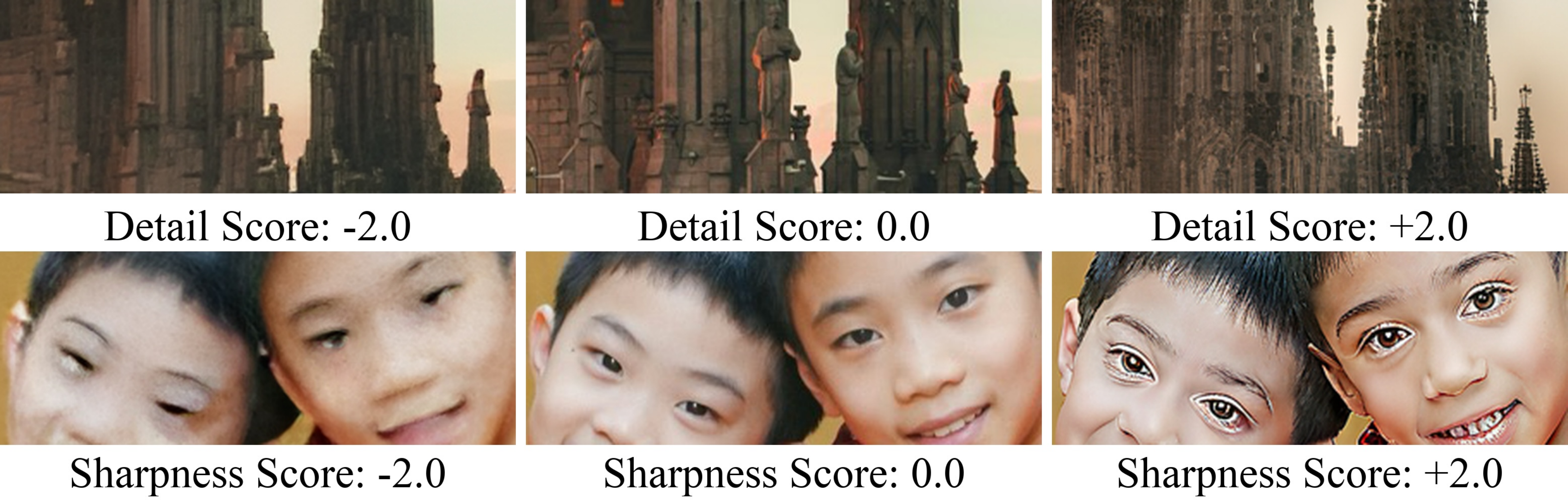}
    \includegraphics[width=\linewidth]{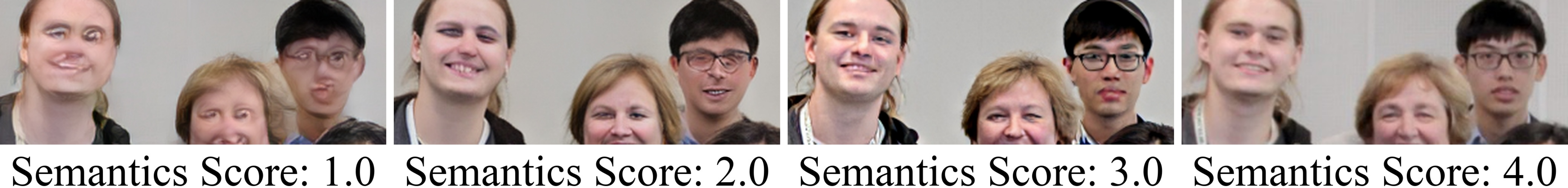}
    \vspace{-16pt}
    \caption{
    Illustration of our annotation criteria. The rows (top to bottom) indicate: Detail (under-generated [-2] to over-generated [+2]), Sharpness (blurred [-2] to over-sharpened [+2]), and Semantic correctness (severe failure [0] to fully consistent [4]).
    }
    \vspace{-10pt}
\label{fig:score_explain}
\end{figure}

\textbf{Detail} is scored on a bipolar scale from \(-3\) to \(+3\), capturing both under-generation and over-generation of details. 
Scores below \(0\) indicate insufficient or overly smoothed detail, while scores above \(0\) reflect excessive or hallucinated textures that deviate from natural appearance. 
\(0\) represents a perceptually balanced level of detail consistent with the input content. 
\textbf{Sharpness} is also evaluated on a \(-3\) to \(+3\) scale, measuring perceptual clarity.
Negative scores correspond to blurry or poorly defined edges, whereas positive scores correspond to over-sharpened or haloed edges.
\(0\) indicates appropriate, naturally sharp object boundaries.
\textbf{Semantics} is assessed using a \(0\) to \(4\) scale, focusing on structural correctness and semantic alignment. 
\(0\) indicates complete semantic failure, which means the object is missing, replaced by irrelevant content, or entirely unrecognizable. 
\(1\) denotes an unsuccessful attempt at generating the intended semantics. 
At \(2\), the object is partly recognizable but exhibits some texture or geometric distortion. 
\(3\) corresponds to minor distortions with an overall recognizable and coherent structure.
\(4\) represents a fully natural and semantically consistent restoration. 
\textbf{Overall} quality score reflects the evaluator’s willingness to accept the restoration as a final output, ranging from completely unacceptable \(0\) to fully satisfactory \(4\).
\cref{fig:score_explain} further illustrates representative examples corresponding to different score levels across the evaluated dimensions.
A total of \(56\) human evaluators then scored these results following our defined evaluation dimensions and detailed rating criteria, enabling consistent and fine-grained perceptual assessment across models.
Compared to previous datasets for generative restoration, our approach is significantly more comprehensive, as summarized in Tab.~\ref{tab:benck_compare}.

\vspace{-12pt}
\begin{figure}[tp]
\scriptsize
    \centering
    \includegraphics[width=\linewidth]{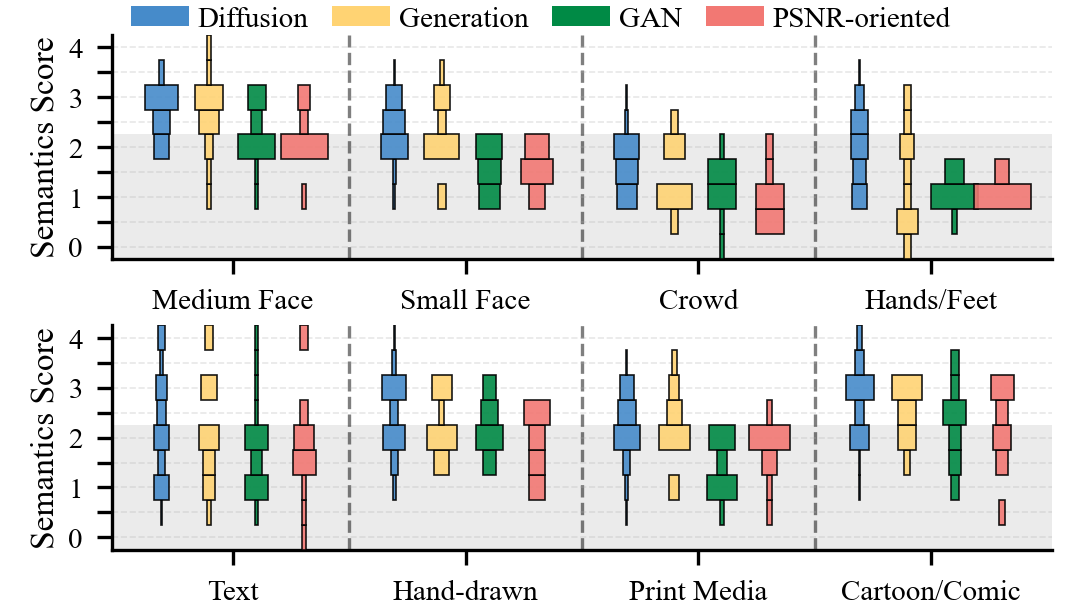}
    \vspace{-18pt}
    \caption{
    Distribution of semantic consistency scores across scenes. Box widths represent the sample percentage at each score. The light gray region marks unacceptable results.  
    }
    \label{fig:semantics_8scenes}
    \vspace{-20pt}
\end{figure}

\section{Results}
\label{sec:results}
\vspace{-4pt}

\subsection{Semantic-Dependent Behavior}
\vspace{-2pt}

\textbf{Semantic specificity across all models}.
A key observation from~\cref{fig:semantics_total_score} is that all restoration models exhibit highly uneven behavior across semantic categories.
Different semantics pose inherently different levels of difficulty: some are consistently well restored, while others remain persistently challenging across all model families.
This asymmetry highlights that current restoration models are strongly influenced by the semantic content of the scene rather than degradation alone.
It also suggests that model robustness is bounded not only by the type of degradation but also by the intrinsic structure and meaning of the visual content itself.

\begin{figure}[tp]
\scriptsize
\centering
    \includegraphics[width=\linewidth]{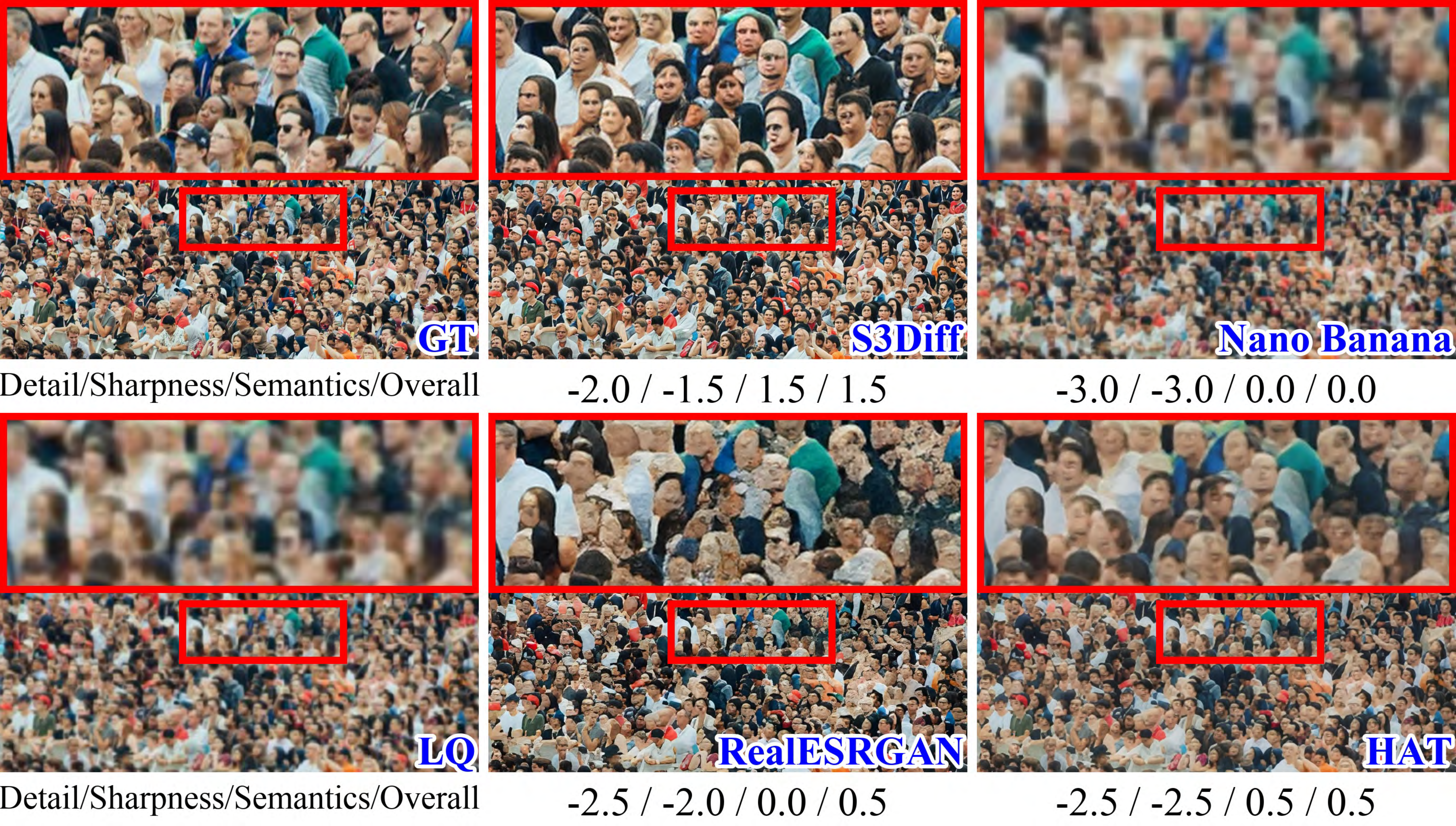}
    \includegraphics[width=\linewidth]{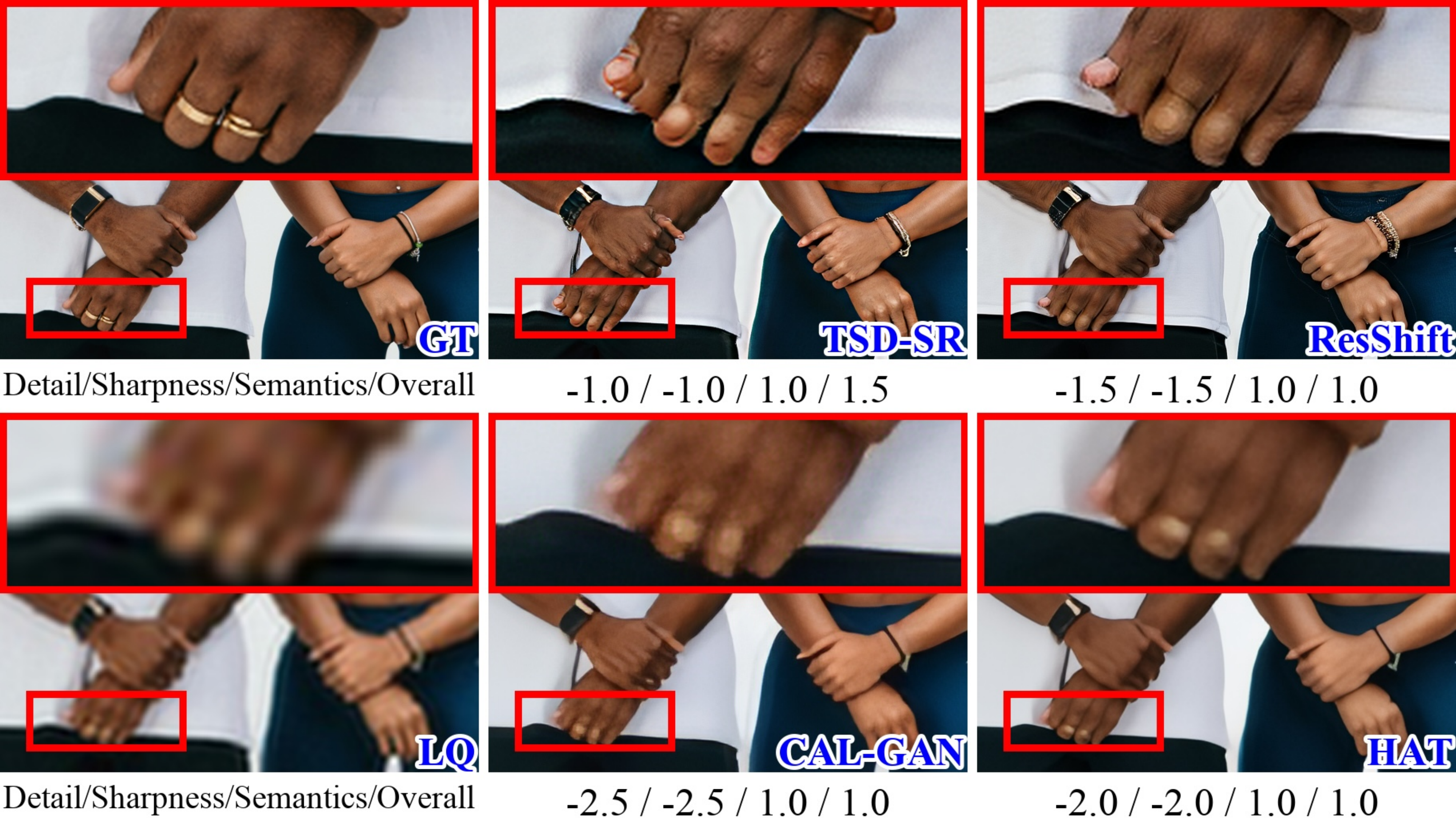}
    \includegraphics[width=\linewidth]{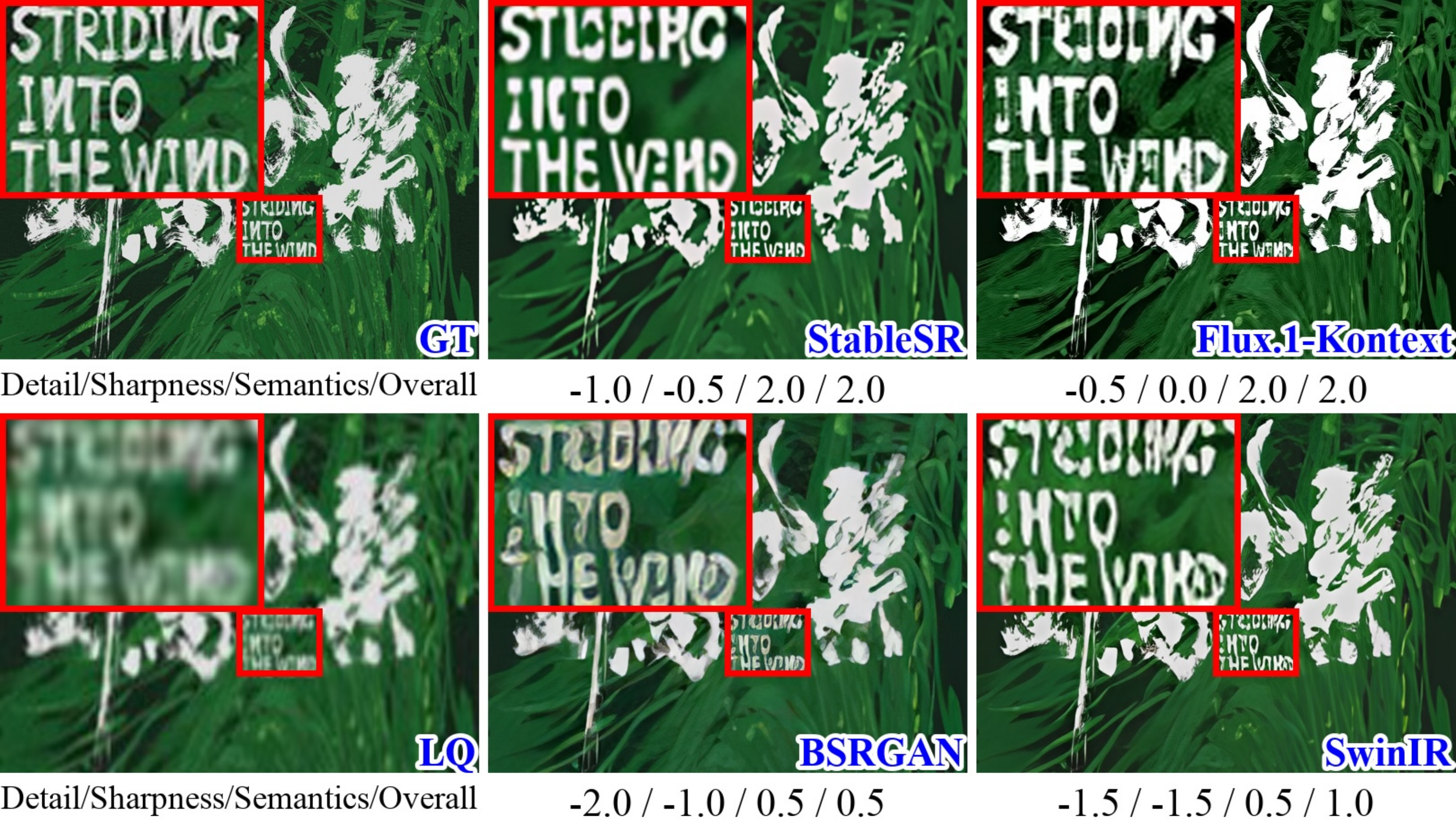}
    \vspace{-17pt}
    \caption{
    Common failure modes in hard cases: distorted faces in \textit{crowd} scenes (top), structural errors in \textit{hands/feet} (middle), and illegible characters in \textit{text} (bottom).
    }
    \vspace{-23pt}
\label{fig:hard_cases}
\end{figure}

As shown in ~\cref{fig:semantics_total_score}, most models perform well on \textit{animal fur} and \textit{cartoon/comic} semantics, with nearly all scores meeting or exceeding our acceptable threshold of 2. This indicates that models find it easier to consistently generate the textures typical of these categories.
Meanwhile, semantics like \textit{small faces}, \textit{crowd}, \textit{hands/feet}, \textit{text} and \textit{print meida} remain difficult to restore, as shown in~\cref{fig:hard_cases}. 
Human perception is acutely sensitive to errors in both the structural coherence and semantic correctness of these particular categories.
However, current models often fail to achieve this satisfactory level of realism, a limitation reflected in~\cref{fig:semantics_total_score} where the majority of scores lie within the unacceptable region (shaded light gray).
Their failures are evident in common artifacts such as distorted facial features, malformed hands, and illegible text, which do not align with human cognitive expectations and are consequently perceived as low-quality restorations.

Most restoration models still tend to under-generate fine details, resulting in overly smooth and textureless appearances, as shown in~\cref{fig:detail_sharpness_8scenes}(a).
While the overall distribution of detail scores is concentrated in the $[-2, -1]$ range.
A closer look reveals a key difference between model families: we observe that GAN and PSNR-oriented models never over-generate details, whereas diffusion-based and generation methods possess a stronger capability for fine-grained synthesis, sometimes even to the point of over-generation. 
This architectural divergence marks a crucial turning point: while \textbf{current GIR models are clearly breaking the previous bottleneck of insufficient detail, they now face the new problem of over-generation}. 
Evaluating this emerging trade-off is precisely what underscores the necessity of our work in such fine-grained evaluation.
This detail generation performance is also highly dependent on the semantic category. 
As illustrated in~\cref{fig:detail_sharpness_8scenes}(a), models can generate abundant fine structures in scenes like \textit{large faces} and \textit{fabric texture}, but struggle with detail of \textit{crowd} images.

We hypothesize two primary reasons for this phenomenon.
First, despite their strong generative priors, existing models struggle to generalize fine-grained textures universally.
Their priors are often biased towards common patterns, leading to inconsistent texture generation in geometrically complex or less-represented scenes. Second, aggressively generating detail risks introducing semantic or structural hallucinations. 
To mitigate this, modern diffusion-based approaches often adopt a conservative strategy, favoring smoother outputs that minimize artifacts.
\begin{figure}[tp]
\scriptsize
\centering
    \includegraphics[width=\linewidth]{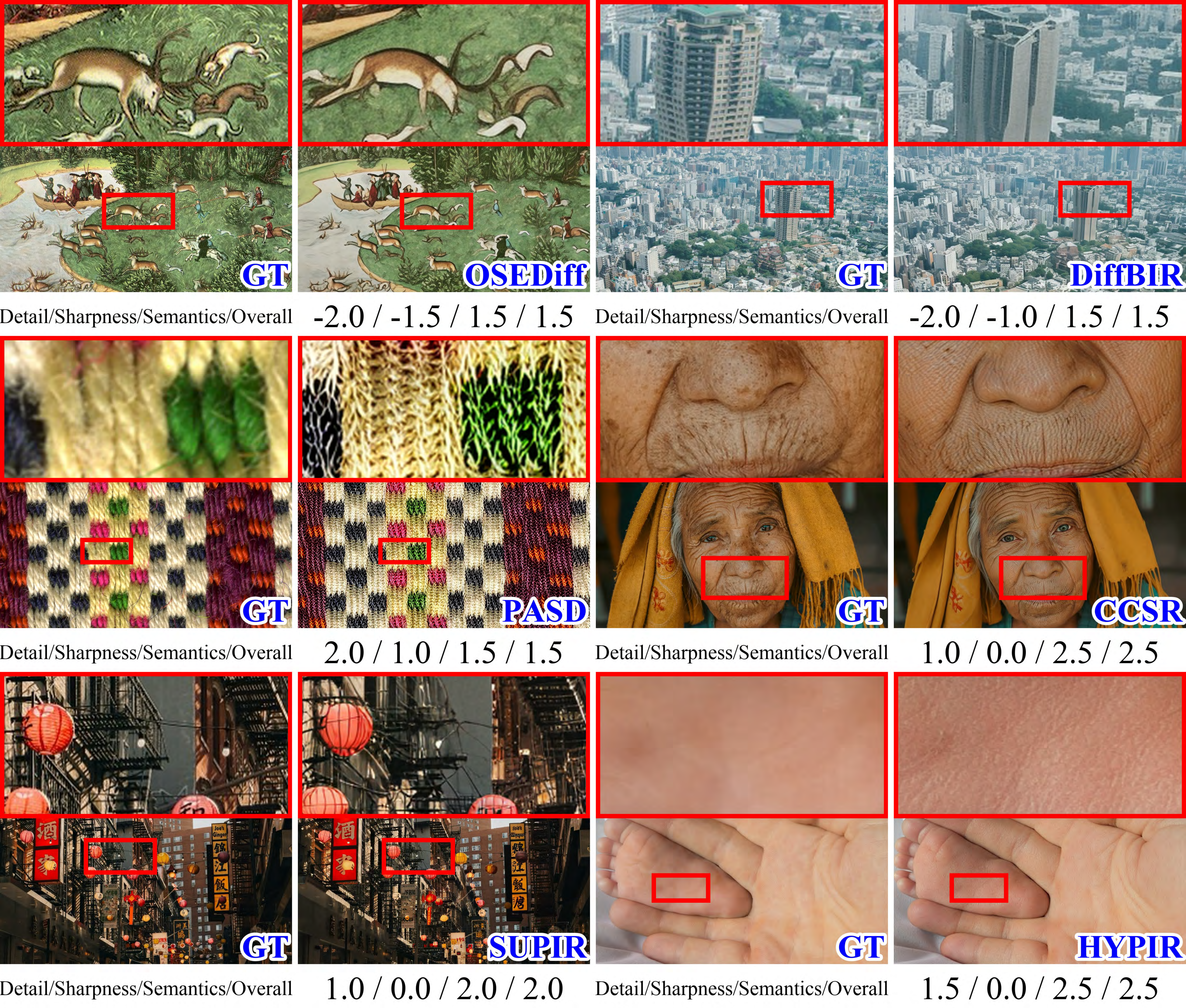}
    \vspace{-16pt}
    \caption{
Diffusion models exhibit a dual tendency in detail generation: sometimes under-generating textures, leading to overly smooth results (top row), and at other times over-generating, creating implausible or redundant details (middle and bottom rows)
    }
\label{fig:detail_cases}
\vspace{-10pt}
\end{figure}

These observations highlight two kinds of difficulty in image restoration:
The first comes from information deficiency in small, low-resolution regions;
The second arises from structural complexity in geometrically or semantically rich scenes. Moreover, humans are particularly sensitive to distortions in these high-level semantic regions, meaning that even subtle geometric or textual inconsistencies can drastically degrade perceived realism. This finding underscores the importance of semantically salient regions for evaluating restoration fidelity and perceptual plausibility.

\begin{figure}[tp]
\scriptsize
\centering
    \includegraphics[width=\linewidth]{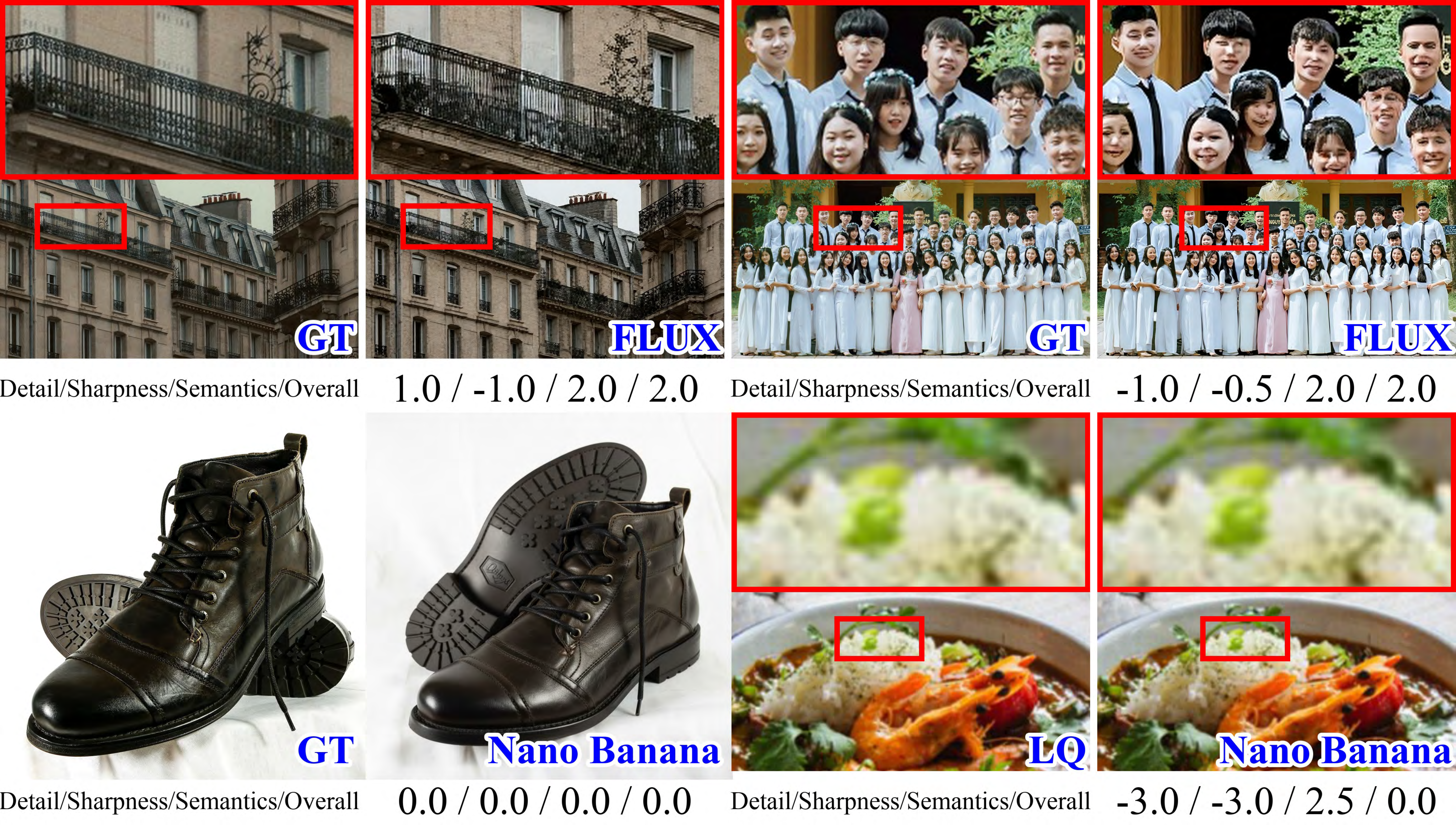}
    \vspace{-16pt}
    \caption{
    Examples illustrating over-generation, semantic infidelity, and degradation unremovement in generation models:
    more whiskers and railings are generated (first),
    the geometry of the face is not restored, and the position of the shoes is completely changed (second and third),
    and artifacts and noise are not eliminated (last).
    }
    \vspace{-11pt}
\label{fig:generation_cases}
\end{figure}

\begin{figure}[tp]
\scriptsize
\centering
    \includegraphics[width=\linewidth]{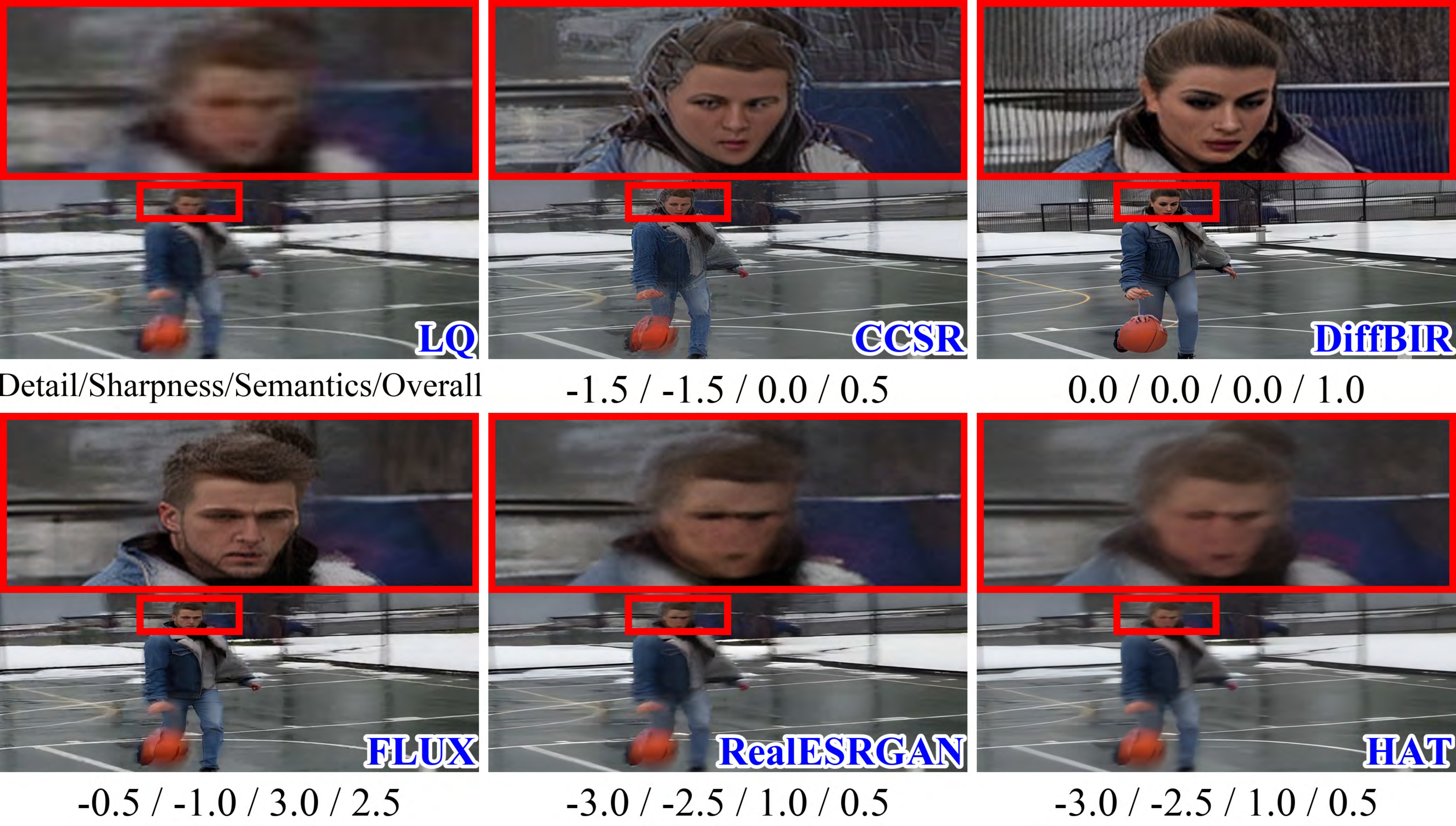}
    \includegraphics[width=\linewidth]{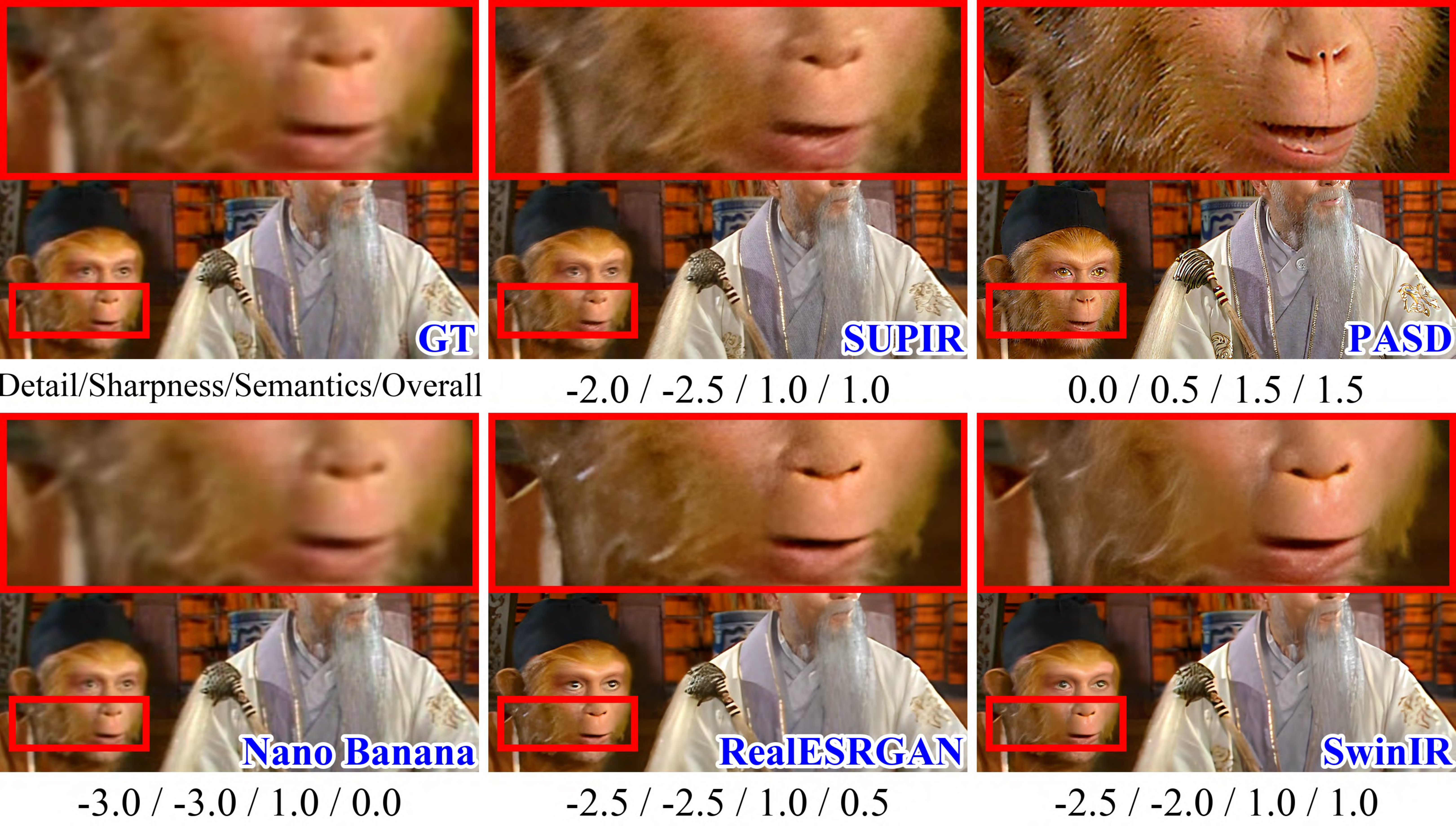}
    \vspace{-17pt}
    \caption{
    Examples of hard degradations (\textit{motion blur}, \textit{old film}).
    }
    \vspace{-20pt}
\label{fig:hard_degradation}
\end{figure}

\vspace{2pt}\noindent\textbf{Model family comparison}. 
Overall, diffusion-based models achieve the highest upper-bound scores across nearly all semantic categories~\cref{fig:semantics_total_score}, demonstrating their promising potential in image restoration. 
This superior performance arises from their dual capability: they effectively remove low-level degradations while leveraging powerful generative priors to generate perceptually rich textures and details.
As shown in~\cref{fig:detail_sharpness_8scenes}, diffusion-based models are more likely to produce clearer and more detailed images, as their detail scores are generally distributed closer to the balanced region (0 score) compared with other model families.
However, diffusion-based models still have challenges in controlling their generative ability.
On one hand, diffusion models may under-generate, producing smooth but visually incomplete restorations; on the other hand, they can easily over-generate, introducing excessive or unrealistic textures that deviate from the original input.
As illustrated in~\cref{fig:detail_sharpness_8scenes}(a), diffusion-based models may produce either insufficient or excessive details.
In the \textit{hand-drawn} example (top left), the model fails to generate enough fine-grained structures, resulting in flat textures.
In texture-rich materials such as \textit{fabric} (middle left), models tend to over-enhance local patterns, generating unnatural details.
For \textit{large face and hand/feet} scenes (middle right and bottom right), they frequently amplify surface textures and wrinkles beyond realism, making the skin appear overly coarse.
In \textit{street view} (bottom left), the models over-generate fine structural elements such as lantern edges and railings, resulting in visually cluttered and chaotic line patterns.
Furthermore, their semantic lower bounds remain comparable to those of other model families.
As shown in~\cref{fig:semantics_total_score}, the performance of diffusion-based models also drops sharply in challenging scenes such as \textit{crowd}, \textit{hands/feet}, and \textit{text}.
\cref{fig:hard_cases} shows that in these scenes, diffusion-based models struggle to maintain geometric consistency and semantic coherence in such complex scenarios.
Moreover, as shown in~\cref{fig:detail_sharpness_8scenes}(a), excessive or exaggerated texture generation also undermines fidelity.
This suggests that their performance limitations arise partly from the inherent difficulty of certain scenes, and partly from the challenge of controlling the generative process to produce details consistent with the input.

Generation models demonstrate limited robustness in image restoration.
They sometimes match the overall score of diffusion-based models, yet the score fluctuates drastically across scenes.
As shown in~\cref{fig:semantics_total_score,fig:detail_sharpness_8scenes,fig:semantics_8scenes}, their upper bounds sometimes rival diffusion-based models, but their lower bounds drop sharply, revealing unstable restoration abilities and poor adaptability to diverse restoration conditions.
This instability manifests in several ways.
As illustrated in~\cref{fig:detail_sharpness_8scenes}(a), generation models predominantly suffer from under-generation, often failing to produce sufficient fine details or restore complex textures.
Occasionally, they exhibit an over-generating tendency, similar to diffusion-based models, as shown in the first example of~\cref{fig:generation_cases}.
Generation models also suffer from semantic infidelity.
Beyond struggling to maintain geometric consistency and semantic coherence in difficult scenes, they sometimes alter the identity of the whole restored content, as shown in the second and third examples of~\cref{fig:generation_cases}.
They further show limited ability to remove degradations: their sharpness scores remain low across many scenes, and the last examples of~\cref{fig:generation_cases} reveal frequent residual blur and artifacts.
These findings indicate that \textbf{simply applying general-purpose generative models to image restoration is insufficient}. 
Their lack of robustness and strong dependence on scene content highlight the need for the development of task-specific architectures.

\begin{figure}[tp]
    \scriptsize
    \centering
    \includegraphics[width=0.8\linewidth]{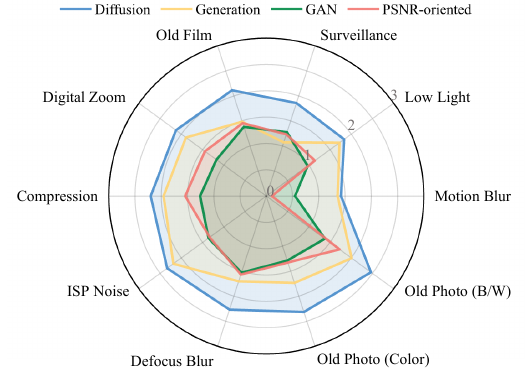}
    \vspace{-14pt}
    \caption{
    Averaged overall scores of different model families across various degradation types. 
    }
    \label{fig:total_degradation}
    \vspace{-10pt}
\end{figure}

\newcommand{\rankone}{\cellcolor{gray!40}}   
\newcommand{\ranktwo}{\cellcolor{gray!20}}   
\newcommand{\rankthree}{\cellcolor{gray!10}}  

\begin{table}[t]
\scriptsize
\centering
\begin{minipage}{0.32\linewidth}
\centering
\setlength\tabcolsep{2.1pt}
\begin{tabular}{ccc}
\toprule
\makecell{Config\\ID}& \makecell{Hands/\\Feet} & \makecell{Aerial\\View} \\
\midrule
1 & 2.59 & \textbf{2.40} \\
2 & \textbf{2.82} & 1.80 \\
\bottomrule
\end{tabular}
\subcaption{DiffBIR}
\end{minipage}
\hfill
\begin{minipage}{0.32\linewidth}
\centering
\setlength\tabcolsep{2.1pt}
\begin{tabular}{ccc}
\toprule
\makecell{Config\\ID} &  Crowd & \makecell{Street\\View}\\
\midrule
1 & \textbf{1.60} & 2.30 \\
2 & 1.30 &  \textbf{2.40}  \\
\bottomrule
\end{tabular}
\subcaption{PiSA-SR}
\end{minipage}
\hfill
\begin{minipage}{0.32\linewidth}
\centering
\setlength\tabcolsep{2.1pt}
\begin{tabular}{ccc}
\toprule
\makecell{Config\\ID} &  \makecell{Digital\\Zoom} & \makecell{OP\\(Color)} \\
\midrule
1 & \textbf{2.09} & 1.65 \\
2 & 1.76 & \textbf{1.81} \\
\bottomrule
\end{tabular}
\subcaption{SUPIR}
\end{minipage}
\vspace{-10pt}
\caption{
Diffusion-based models' parameter sensitivity. Mean scores show that the optimal configuration is scene- and degradation-dependent.
}
\label{tab:params_result}
\vspace{-20pt}
\end{table}


PSNR-oriented and GAN-based approaches generally underperform diffusion-based models across most semantic categories and evaluation dimensions.
As illustrated in~\cref{fig:detail_sharpness_8scenes}(a), their upper-bound scores are lower than those of diffusion-based methods, reflecting limited generative capability and weaker texture generation.
\cref{fig:detail_sharpness_8scenes}(b) shows that they exhibit an advantage in robustness over generation models.
Their lower-bound scores are typically higher, suggesting a relatively reliable baseline.

\vspace{-3pt}
\subsection{Degradation-Dependent Behavior}
\vspace{-3pt}

Current restoration models still exhibit clear limitations when handling information-deficient degradations.
In~\cref{fig:total_degradation}, degradation types such as \textit{motion blur}, \textit{old film}, \textit{low light}, and \textit{surveillance}  exhibit noticeably lower average overall scores.
\cref{fig:hard_degradation} shows the restoration results of these degradations.
These degradations remain challenging for existing GIR models.
Their common characteristic lies in information loss and irreversible corruption: \textit{motion blur} introduces overlapping spatial information; \textit{surveillance} footage tends to be extremely low-resolution and blurry; \textit{low-light} scenes provide limited visible cues; \textit{old film} often suffers from extensive detail degradation.
These degradations may require task-specific modeling beyond current general restoration frameworks.  

Among all model families, diffusion-based models demonstrate relatively stronger robustness and generalization across different degradation types.
As shown in~\cref{fig:total_degradation}, they achieve noticeably higher average overall scores than other architectures.
However, their performance still drops in extreme cases, indicating that generative capability alone is insufficient when the input lacks recoverable information.

\begin{figure}[tp]
\scriptsize
\centering
    \includegraphics[width=\linewidth]{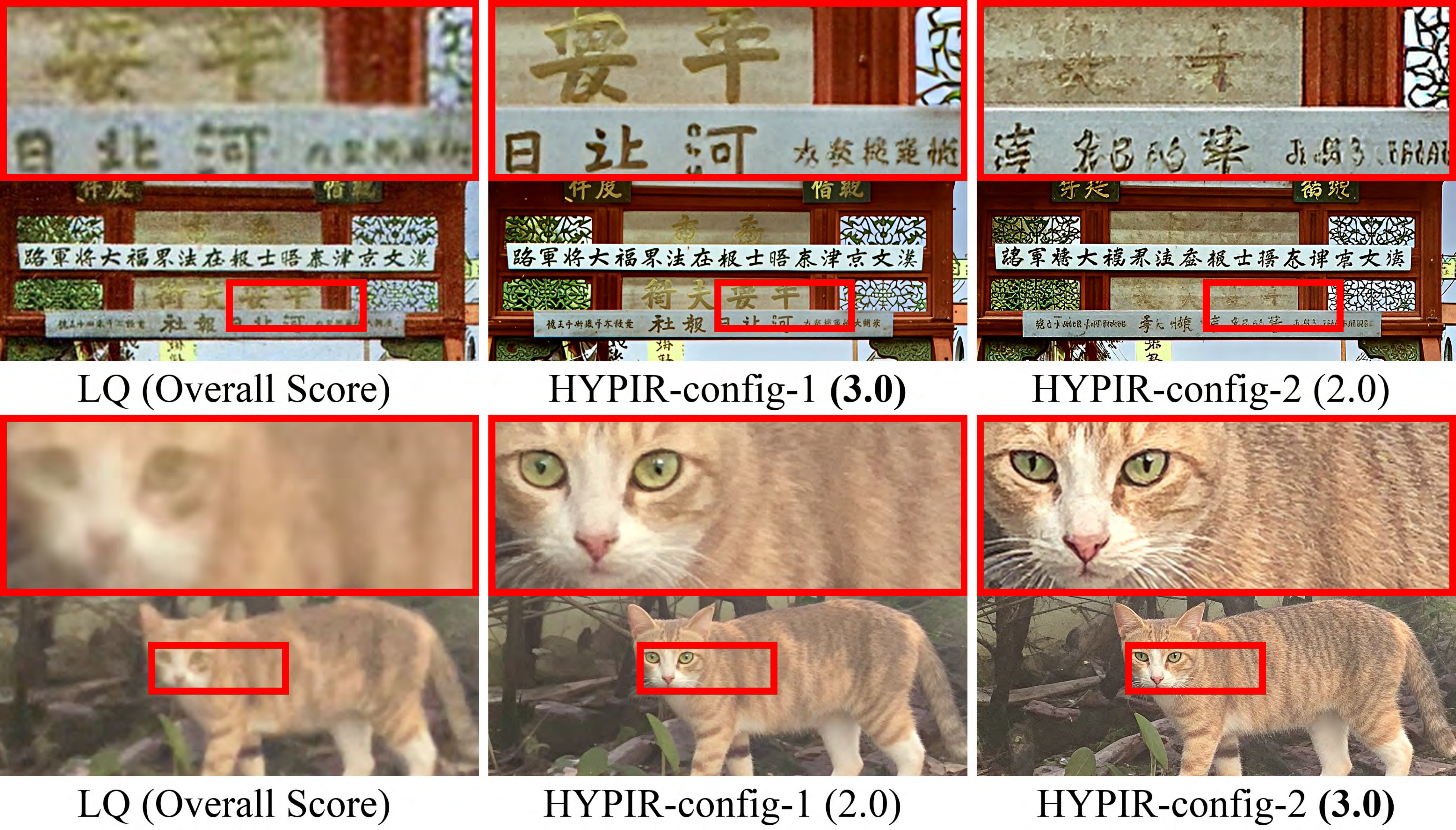}
    \vspace{-16pt}
    \caption{
    Effect of parameter configurations on restoration behavior in diffusion-based models (HYPIR as an example).
    }
    \vspace{-15pt}
\label{fig:param_cases}
\end{figure}

\vspace{-3pt}
\subsection{Sensitivity to Parameter Configuration}
\vspace{-3pt}

For diffusion-based models, parameter configuration proves to be a critical factor influencing restoration quality.
As shown in~\cref{tab:params_result}, different parameter settings yield the highest overall scores in different semantic scenes and degradations.
\cref{fig:param_cases} further demonstrates diffusion-based models require scene- and degradation-specific parameter choices to achieve optimal restoration performance.
While this sensitivity introduces additional complexity in model use, it also provides valuable flexibility: by properly tuning parameters, users can selectively emphasize realism or fidelity according to task requirements.
Therefore, careful and context-aware parameter adjustment is essential to fully exploit the potential of diffusion-based models and achieve optimal restoration performance. 
To further enhance model usability and performance, we expect that future diffusion-based GIR frameworks should not only retain parameter-controllable flexibility but also incorporate adaptive mechanisms that automatically adjust generative strength based on scene semantics and degradation conditions.

\begin{table}[tp]
\centering
\scriptsize
\setlength\tabcolsep{3.2pt}
\begin{tabular}{l|cccccc|c}
\toprule
 & NIQE & PI & MANIQA & MUSIQ & CLIP-IQA & DeQA-Score & \textbf{Ours} \\
\midrule
SRCC & 0.164 & 0.178 & 0.361 & 0.376 & 0.389 & 0.409 & \textbf{0.662} \\
PLCC & 0.206 & 0.230 & 0.391 & 0.427 & 0.420 & 0.464 & \textbf{0.677} \\ 
\bottomrule
\end{tabular}
\vspace{-8pt}
\caption{
Performance comparison of existing IQA methods and our trained IQA model on our dataset.
}
\label{tab:iqa_comparison}
\vspace{2pt}
\centering
\scriptsize
\setlength\tabcolsep{4.8pt}
\begin{tabular}{c|ccccc}
\toprule
Metric & Sharpness & Detail & Semantic & Total Score  \\
\midrule
SRCC / PLCC & 0.633 / 0.645 & 0.575 / 0.603 & 0.585 / 0.622 & 0.662 / 0.677 \\
\bottomrule
\end{tabular}
\vspace{-8pt}
\caption{
Performance on different aspects of our dataset.
}
\vspace{-20pt}
\label{tab:iqa_aspect}
\end{table}

\vspace{-3pt}
\section{Advances in Image Quality Assessment}
\label{sec:main_iqa}
\vspace{-3pt}

Despite the rapid development of IQA techniques, existing methods~\cite{afine,isrgenqa,pieapp} remain inadequate for evaluating the fine-grained behaviors of GIR models. 
Conventional IQA metrics are limited in two key aspects. 
First, they often fail to produce accurate predictions in semantically inconsistent cases, especially when the restored image deviates from the original semantic details.
Second, most IQA models provide only a single score, which cannot capture the multi-dimensional nature of GIR quality, where sharpness, detail, semantics, and overall score may vary independently.

Our work represents the first effort to perform multi-dimensional, fine-grained analysis of GIR models.
Leveraging our human-annotated dataset, which provides scores along multiple dimensions, we train an IQA model capable of both holistic and dimension-specific quality prediction. 
This enables diagnostic evaluation of restored images that existing single-score IQA frameworks cannot provide.
We thus split our dataset by 9:1 for training and evaluation, then train a baseline IQA model using the DeQA-Score framework~\cite{deqa_score}. We use Spearman Rank Correlation Coefficient (SRCC) and Pearson Linear Correlation Coefficient (PLCC) to measure alignment with human annotations.

We compare our IQA model with leading no-reference methods (NIQE~\cite{niqe}, PI~\cite{pi}, MANIQA~\cite{maniqa}, MUSIQ~\cite{musiq}, CLIP-IQA~\cite{clipiqa}, DeQA-Score~\cite{deqa_score}) on our dataset, which is specifically designed to challenge models with diverse semantics and degradations. 
As shown in~\cref{tab:iqa_comparison}, the inability of existing methods to generalize to these conditions is evident in their poor performance. 
In contrast, our model achieves SRCC of \textbf{0.662} and PLCC of \textbf{0.677}.
We also evaluate the IQA models trained on each dimension in~\cref{tab:iqa_aspect}.
The experimental results demonstrate that conventional IQA methods are insufficient for analyzing the complex behaviors of modern GIR models.
Our trained IQA framework, grounded in fine-grained perceptual and semantic annotations, allows not only accurate overall quality prediction but also interpretable diagnosis of specific aspects.
This multi-dimensional approach provides a new foundation for perceptually aligned evaluation and could guide the development of next-generation restoration metrics and intelligent evaluation agents.

\vspace{-4pt}
\section{Conclusion}
\label{sec:conclusion}
\vspace{-2pt}

In this work, we present the first fine-grained framework for understanding GIR across both semantic scenes and degradation types. 
By constructing a carefully balanced test set, introducing multi-dimensional human annotations, and analyzing 20 representative models, we reveal the strengths and limitations of modern GIR systems. 
We hope this work provides a foundation for more reliable GIR evaluation and inspires future research toward controllable, robust, and semantically faithful restoration models.

\paragraph{Acknowledgment.} This work was supported by the National Natural Science Foundation of China (Grant No.62276251) and the Joint Lab of CAS-HK. This research was partially funded by the Ministry of Education and Science of Bulgaria (support for INSAIT, part of the Bulgarian National Roadmap for Research Infrastructure).

{
    \small
    \bibliographystyle{unsrt}
    \bibliography{main}
}


\maketitlesupplementary
\appendix
\section*{Appendix}

This \supp provides additional details, quantitative and qualitative results that complement the findings presented in the main paper. 
It is organized as follows:
Appendix~\ref{sec:evaluation_design_detail} provides comprehensive details of our dataset construction and implementation details of GIR models.
Appendix~\ref{sec:add_results} shows additional score distributions across semantic scenes and degradation types.
The section also reports extended quantitative and qualitative results for parameter sensitivity analysis.
Appendix~\ref{sec:iqa} presents qualitative demonstrations of the limitations of existing IQA methods.
Appendix~\ref{sec:failure_cases} provides additional failure cases of current GIR models.
Appendix~\ref{sec:model_details} reports detailed per-model performance across all evaluation dimensions.
Appendix~\ref{sec:training} procides training details and more qualitative results of our IQA model.

\section{Details of Evaluation Design}
\label{sec:evaluation_design_detail}
\subsection{Details of the dataset}
The images in dataset are collected from both open-access online repositories~\cite{WikimediaCommons,Unsplash,Pixabay,InternationalPhotoMag}, our own captured photographs and existing degradation datasets~\cite{UFDD_face_dataset,renoir,realdof,blur_detection,defocus_deblur,ntire2025,nighttime,div2k,urban100,Movie-Poster,uhdm,wikiart,products10k,UCF}.
Together, these sources provide rich semantic diversity and a broad range of real-world degradations, providing a solid foundation for evaluating the behavior of GIR models under different conditions.

All images are processed to a unified resolution of $1024 \times 1024$.
Each restoration model receives a $1024 \times 1024$ degraded image as input and produces a restored output of the same resolution, ensuring consistent comparison across all GIR models.

For the synthetic subset of 147 high-quality images across 21 semantic categories, we apply degradations using the official RealESRGAN pipeline~\cite{realesrgan}. 
The two-stage degradation framework of Real-ESRGAN can span an extremely severe degradation space, so severe that the entire image becomes unrecognizable to humans.
This goes beyond realistic image degradation and loses practical meaning for restoration.
To ensure that input images remain meaningful for evaluation, we have made several modifications to the default degradation settings.
We fix the blur kernel size to $17 \times 17$ for both stages and constrain the blur standard deviations to $[0.2, 1.5]$ and $[0.2, 0.8]$ respectively. 
The probability of applying the final sinc filter is reduced from $0.8$ to $0.4$. 
In both degradation stages, only Gaussian noise is added, with noise levels set to $[1, 10]$ in stage 1 and $[1, 5]$ in stage 2. 
The resize operation is applied within scale ranges of $[0.5, 1.5]$ in stage 1 and $[0.8, 1.2]$ in stage 2. 
JPEG compression levels are restricted to $[50, 95]$ and $[60, 95]$, producing moderate compression artifacts compared with the original broader ranges.
Overall, these adjustments intentionally lighten the default RealESRGAN degradation strength, allowing us to introduce diverse but not overly destructive degradations. 
This ensures that the synthetic inputs remain realistic and challenging while still preserving enough visual information for meaningful generative restoration.

\begin{figure*}[tp]
\scriptsize
\centering
    \includegraphics[width=0.47\linewidth]{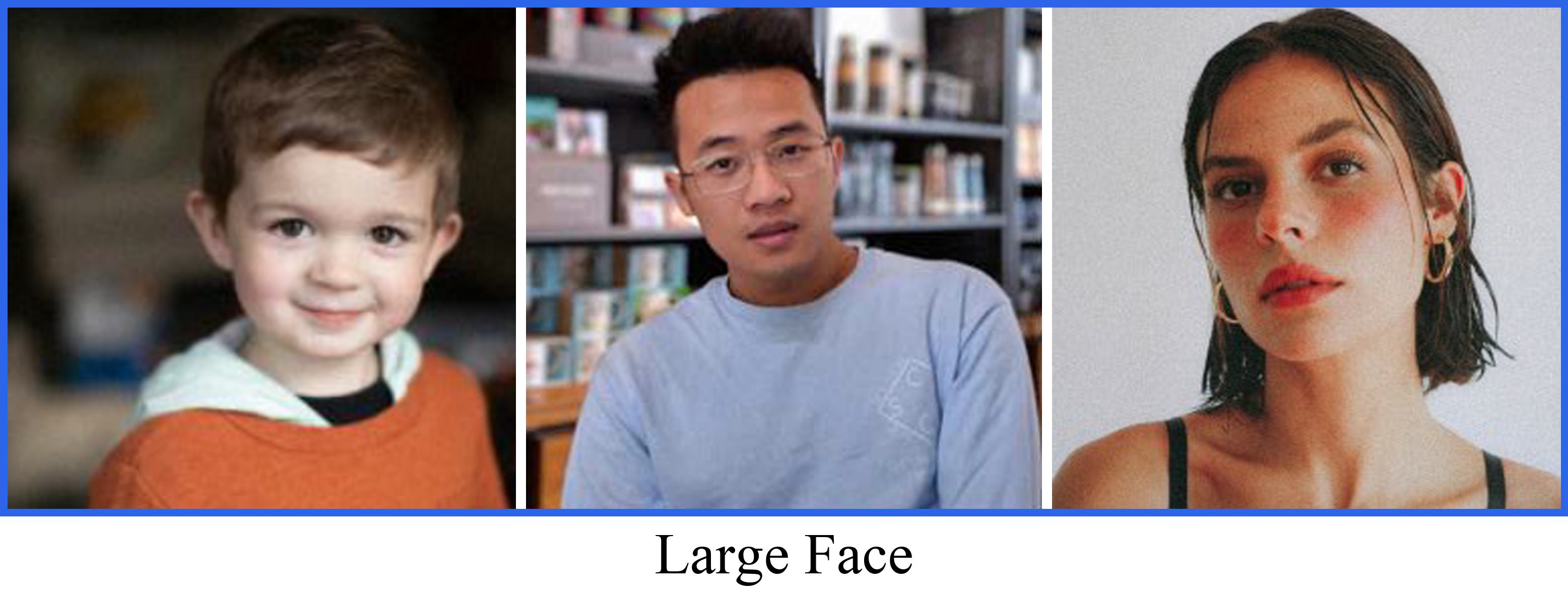}
    \hfill
   \includegraphics[width=0.47\linewidth]{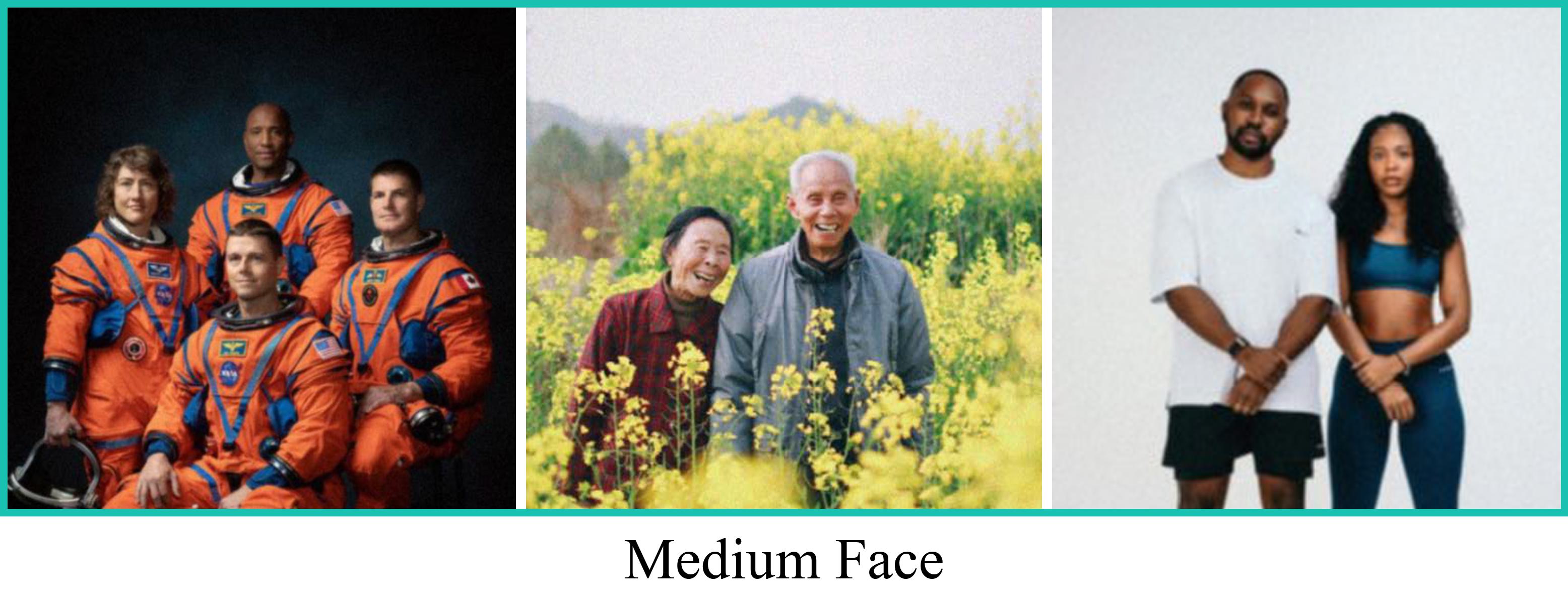}
   \hfill
   \includegraphics[width=0.47\linewidth]{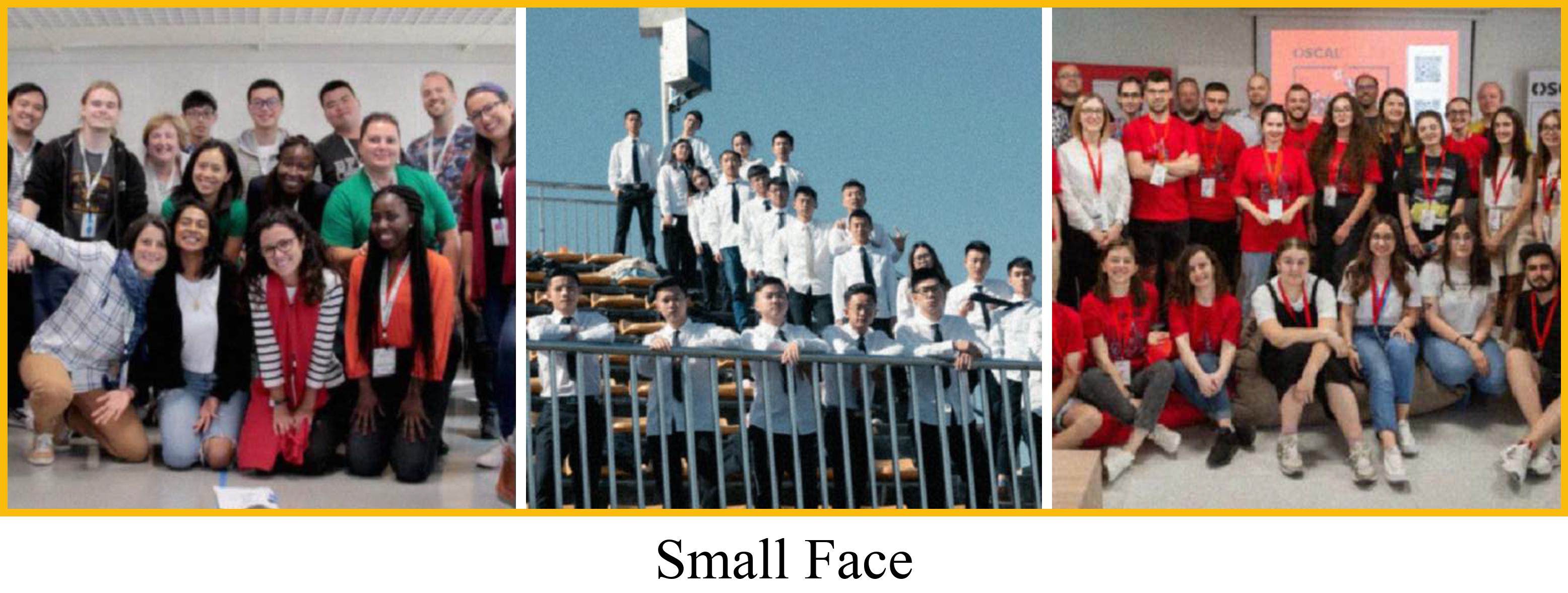}
   \hfill
   \includegraphics[width=0.47\linewidth]{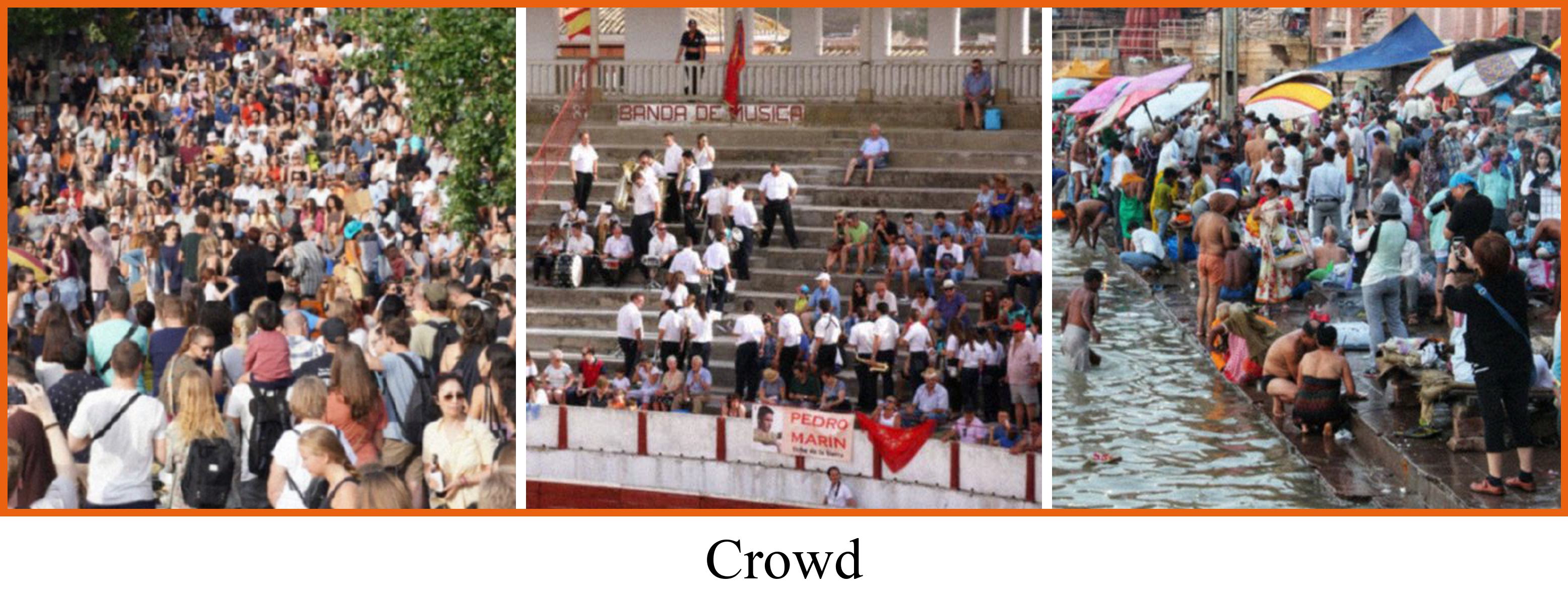}
   \hfill
   \includegraphics[width=0.47\linewidth]{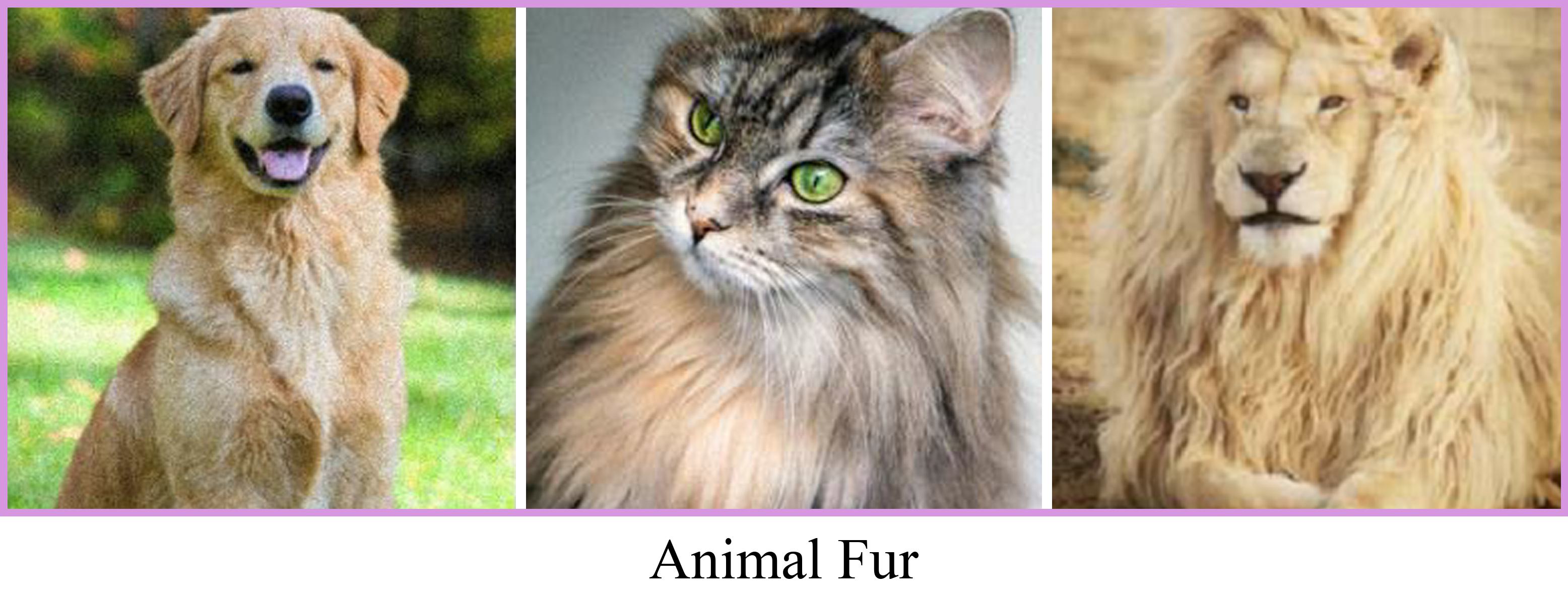}
   \hfill
   \includegraphics[width=0.47\linewidth]{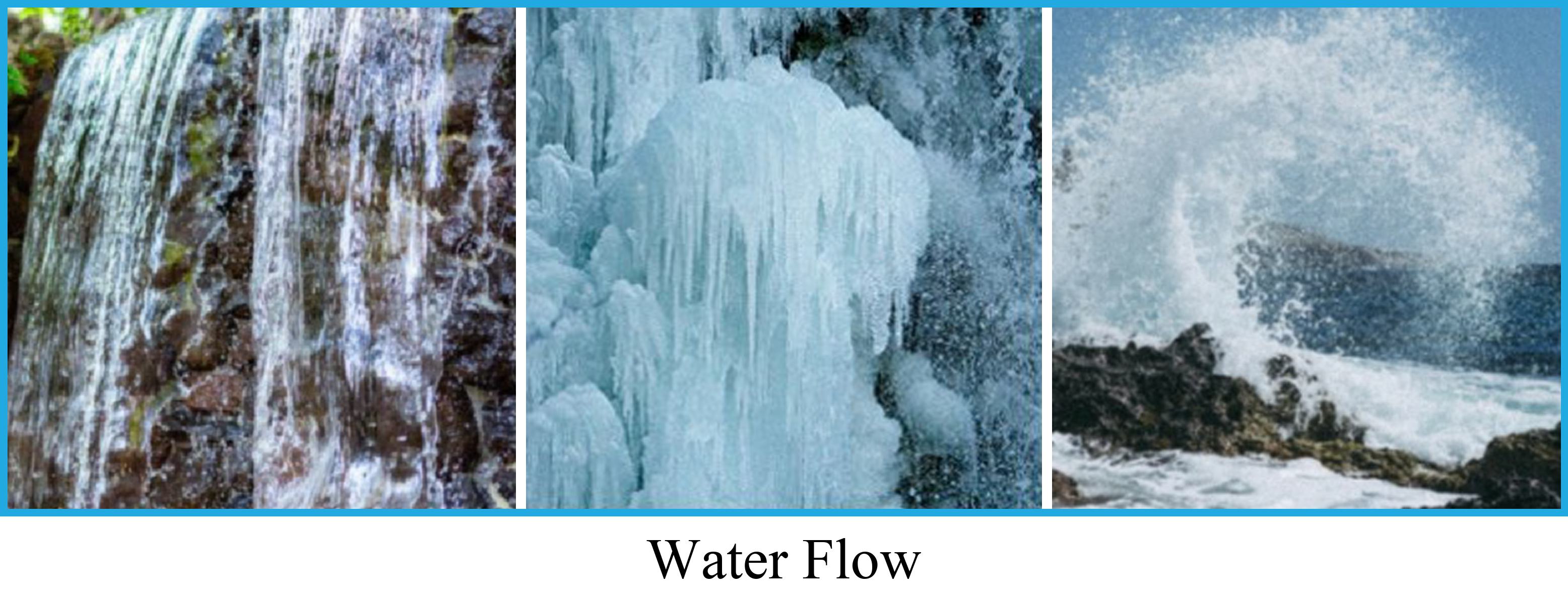}
   \hfill
   \includegraphics[width=0.47\linewidth]{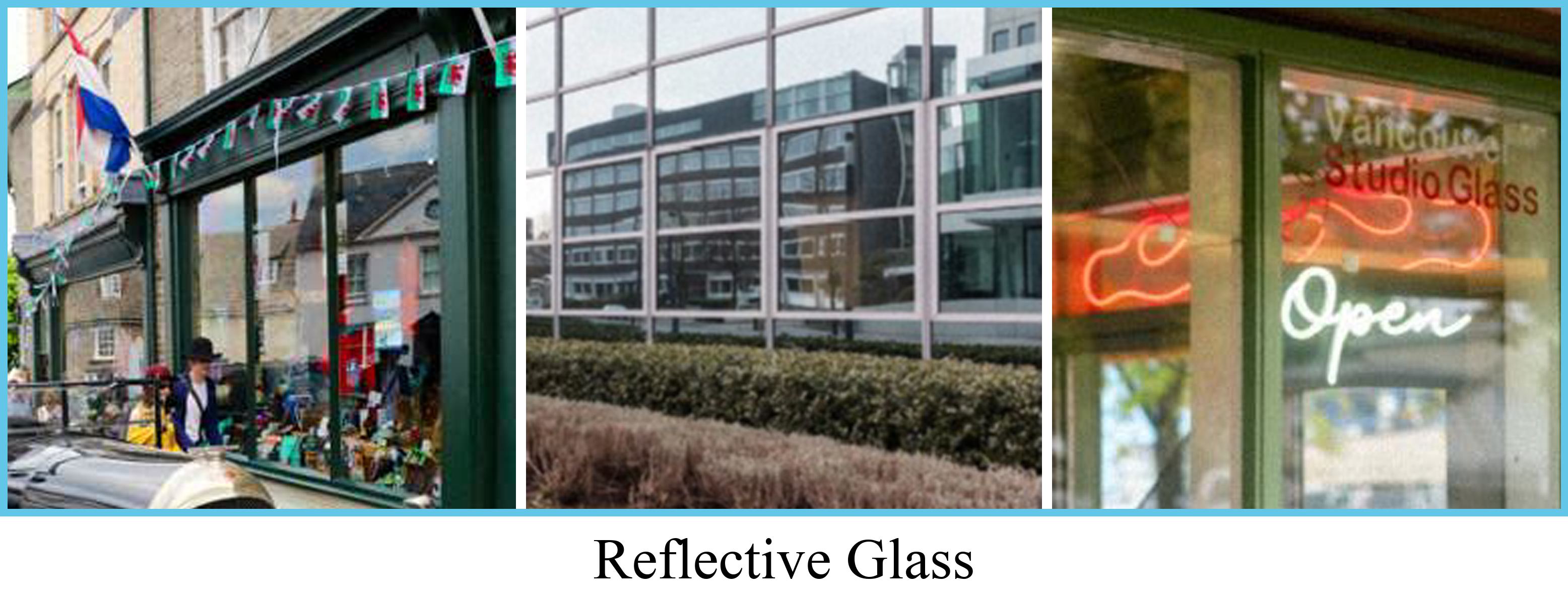}
   \hfill
   \includegraphics[width=0.47\linewidth]{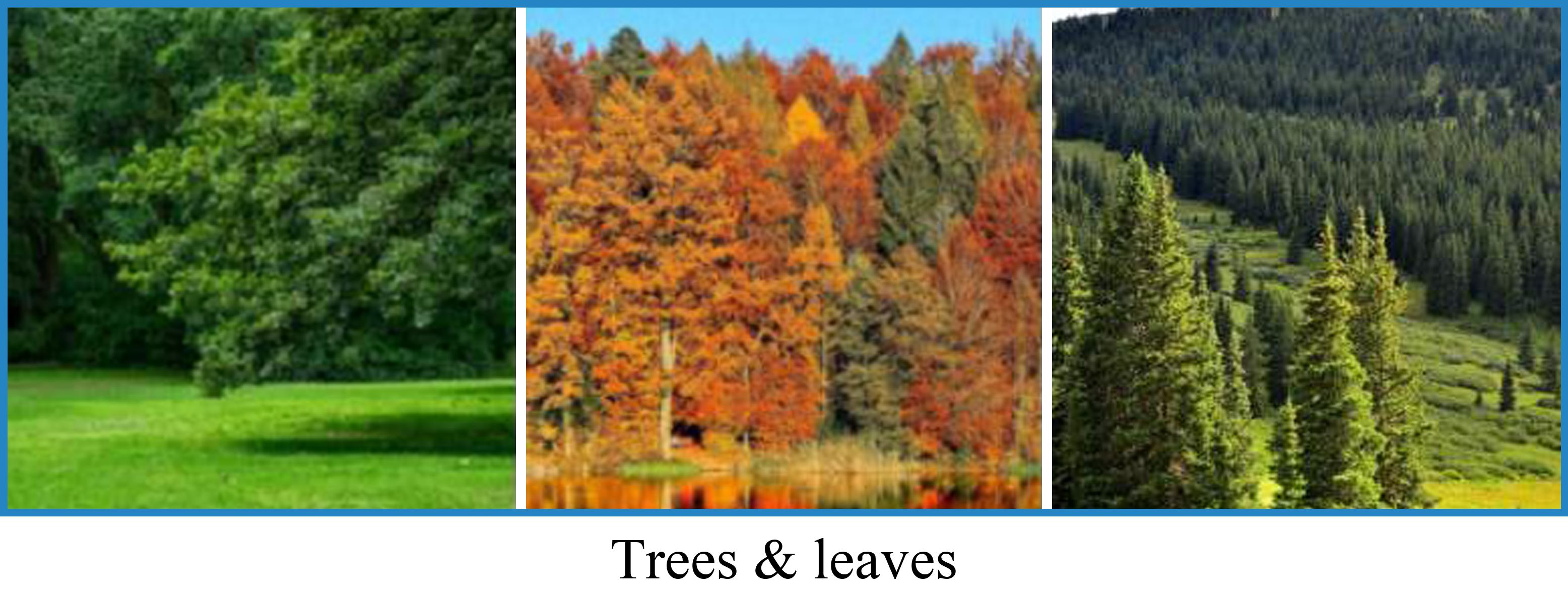}
   \hfill
   \includegraphics[width=0.47\linewidth]{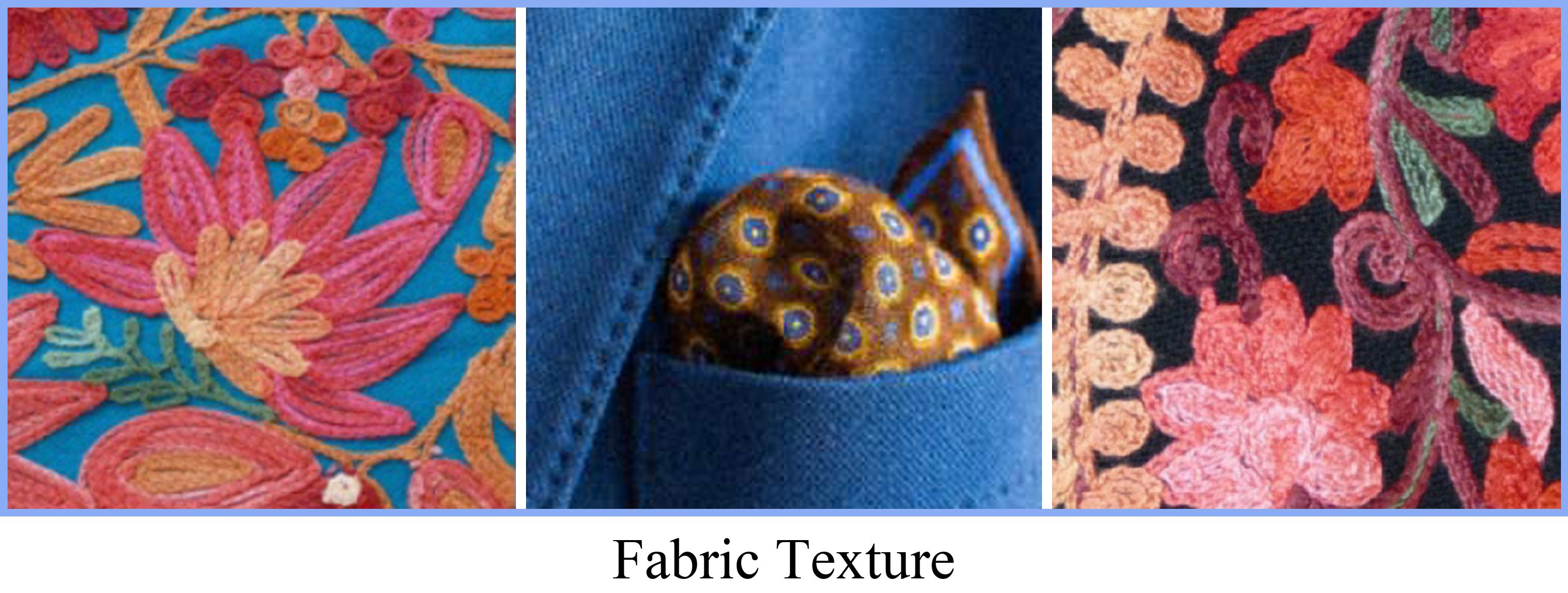}
   \hfill
   \includegraphics[width=0.47\linewidth]{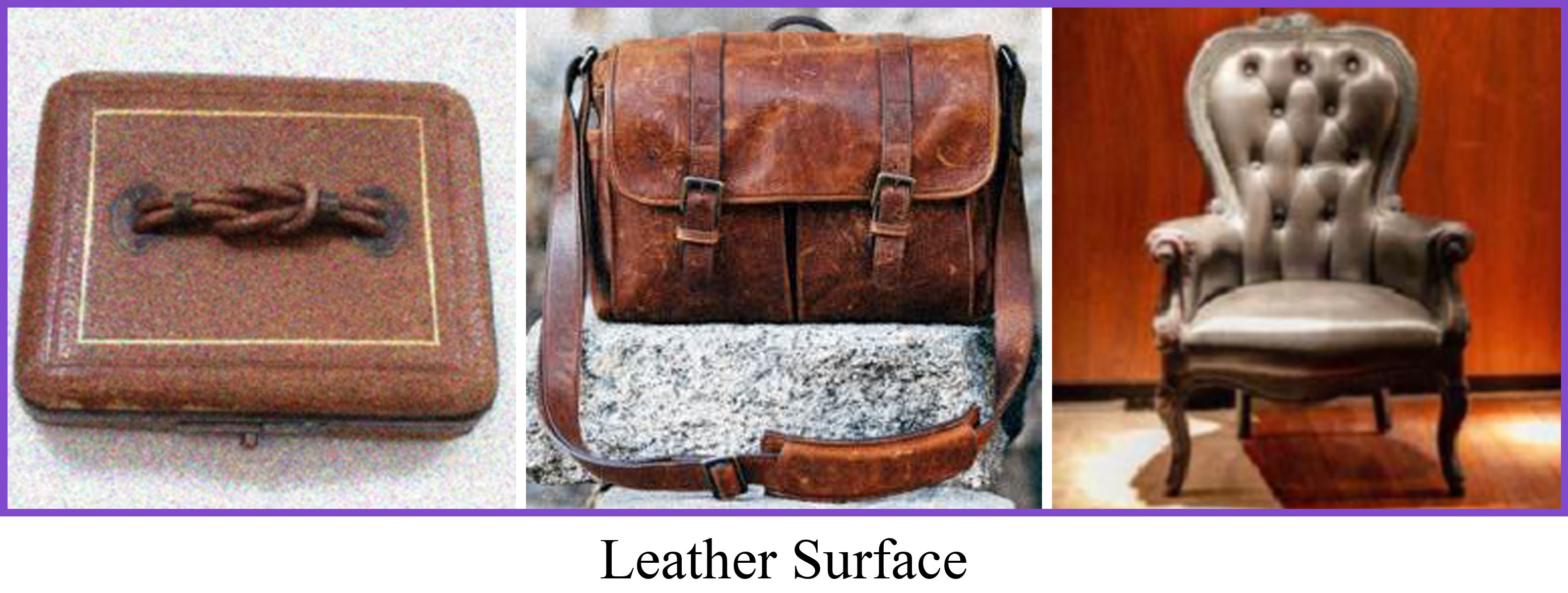}
   \hfill
   \includegraphics[width=0.47\linewidth]{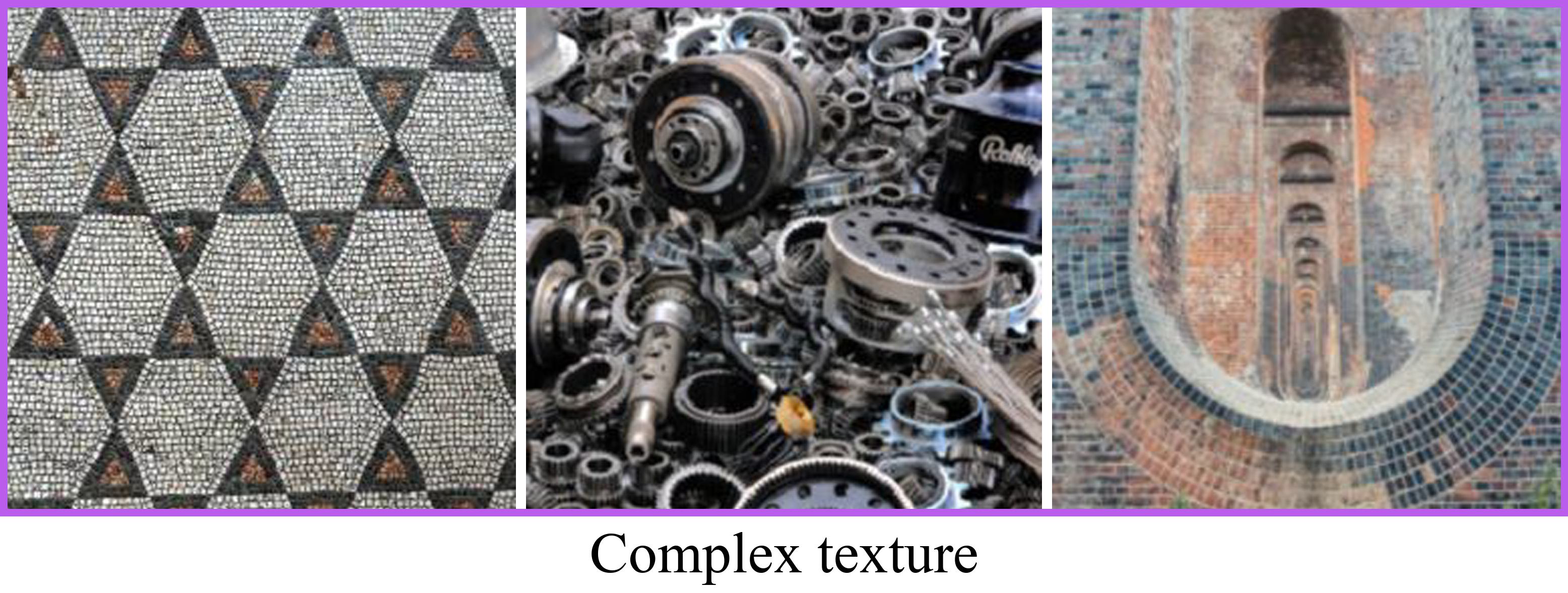}
   \hfill
   \includegraphics[width=0.47\linewidth]{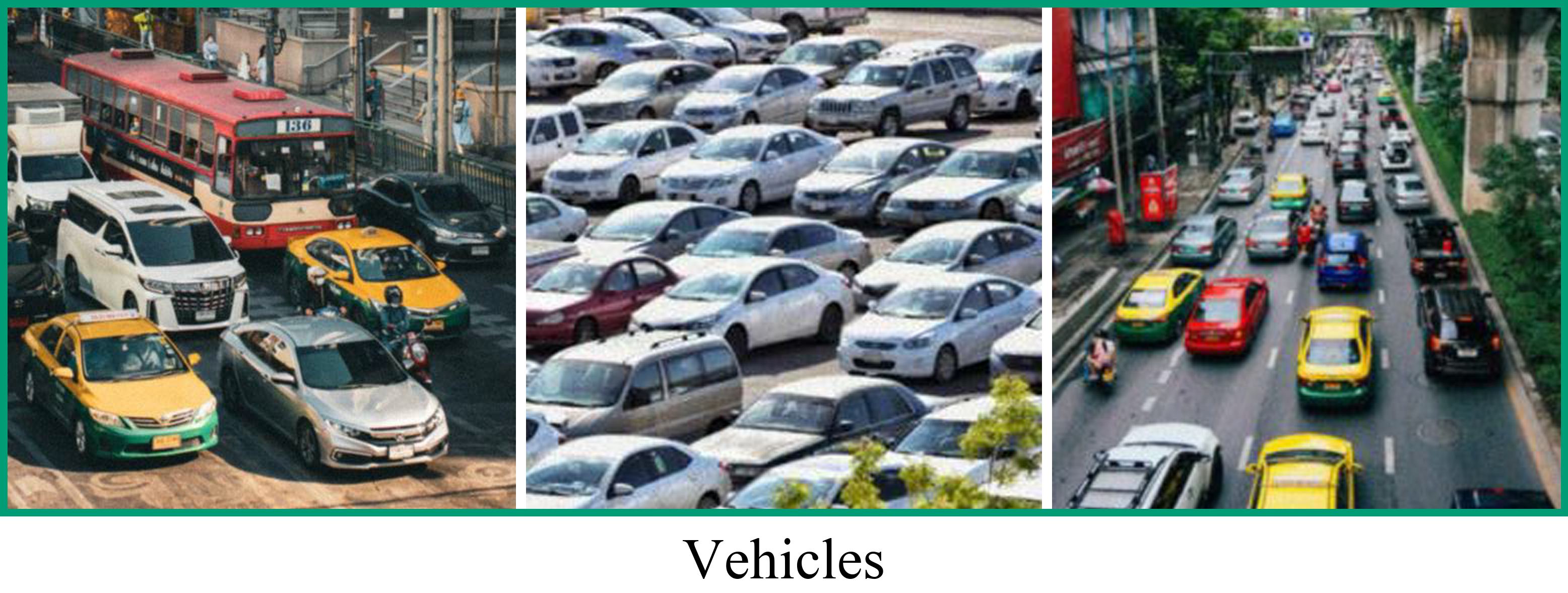}
   \hfill
   \includegraphics[width=0.47\linewidth]{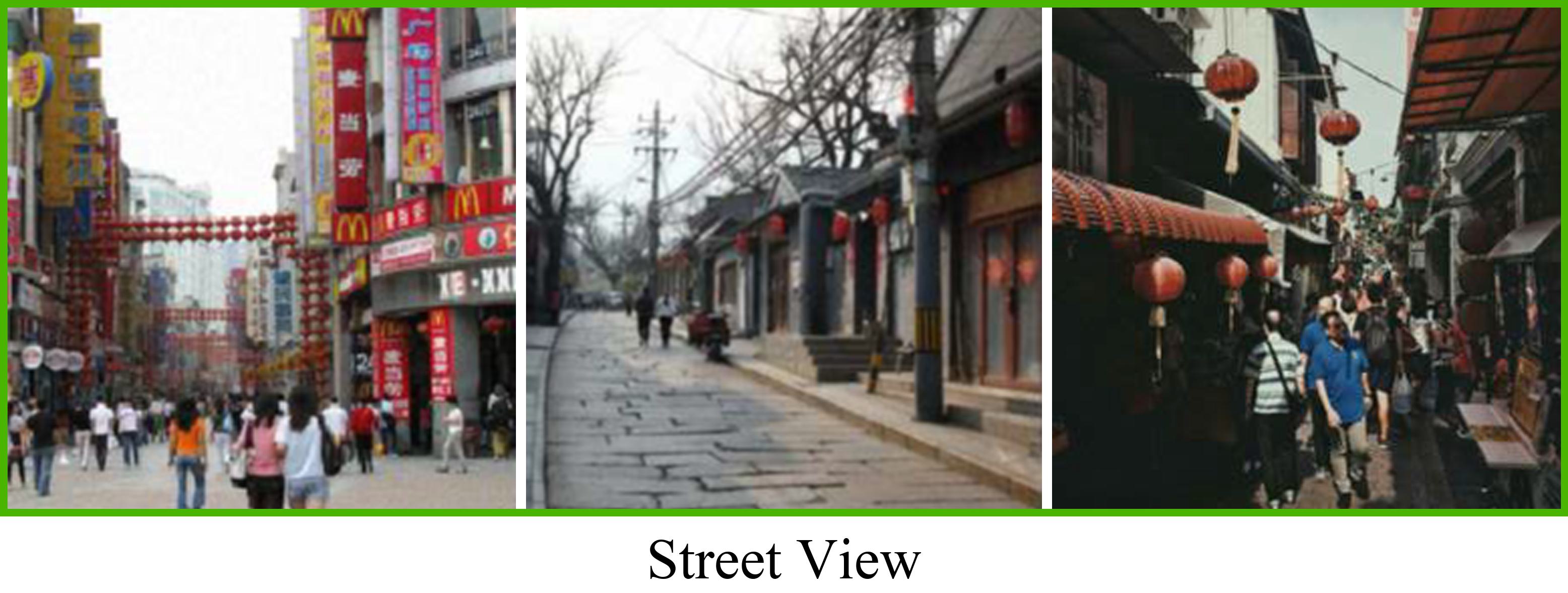}
   \hfill
   \includegraphics[width=0.47\linewidth]{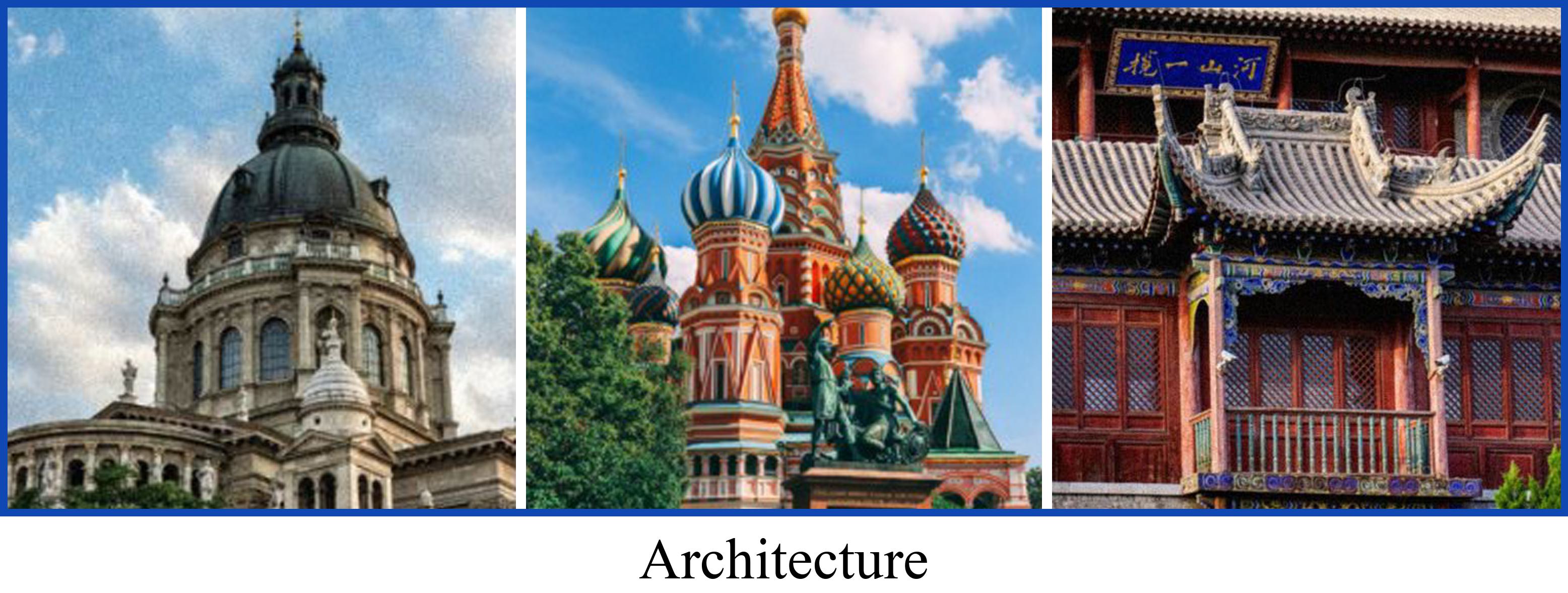}
   \vspace{-5pt}
    \caption{
    More low-quality examples of different semantic categories. Zoom in for a better view.
    }
    \label{fig:dataset_semantics_supp1}
\end{figure*}

\begin{figure*}[tp]
\scriptsize
\centering
    \includegraphics[width=0.47\linewidth]{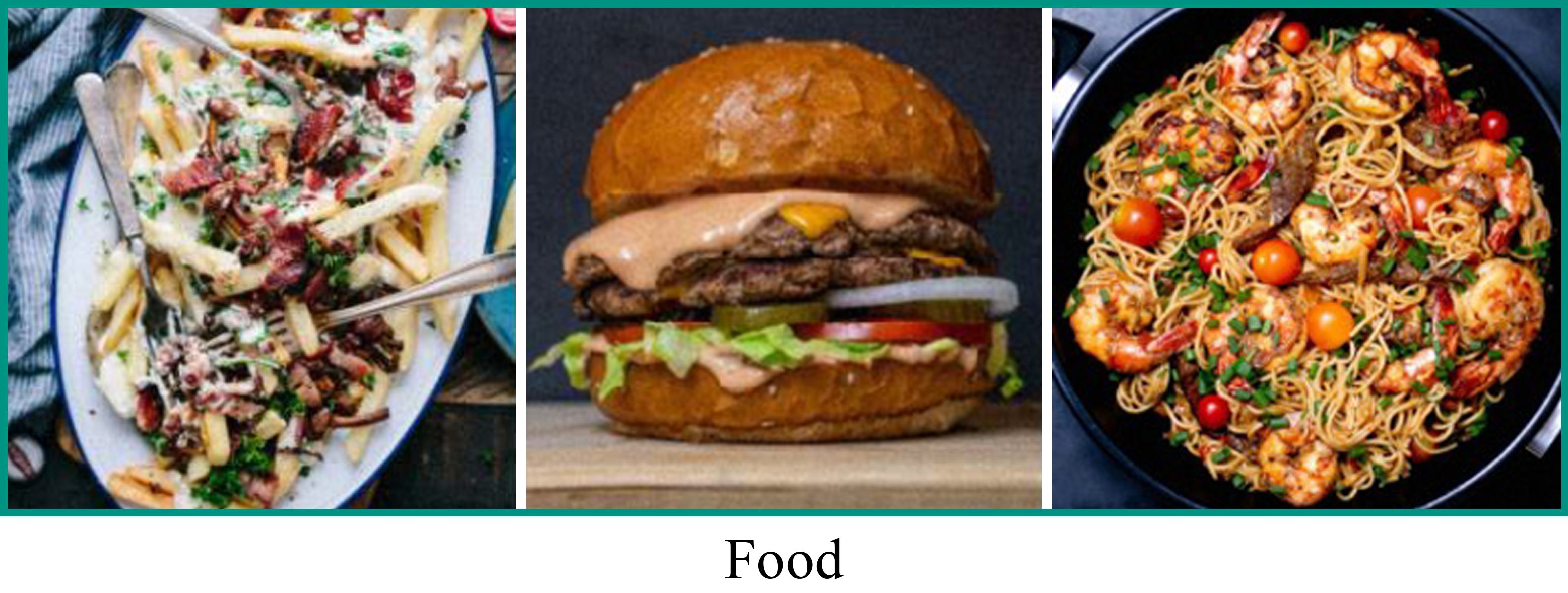}
    \hfill
   \includegraphics[width=0.47\linewidth]{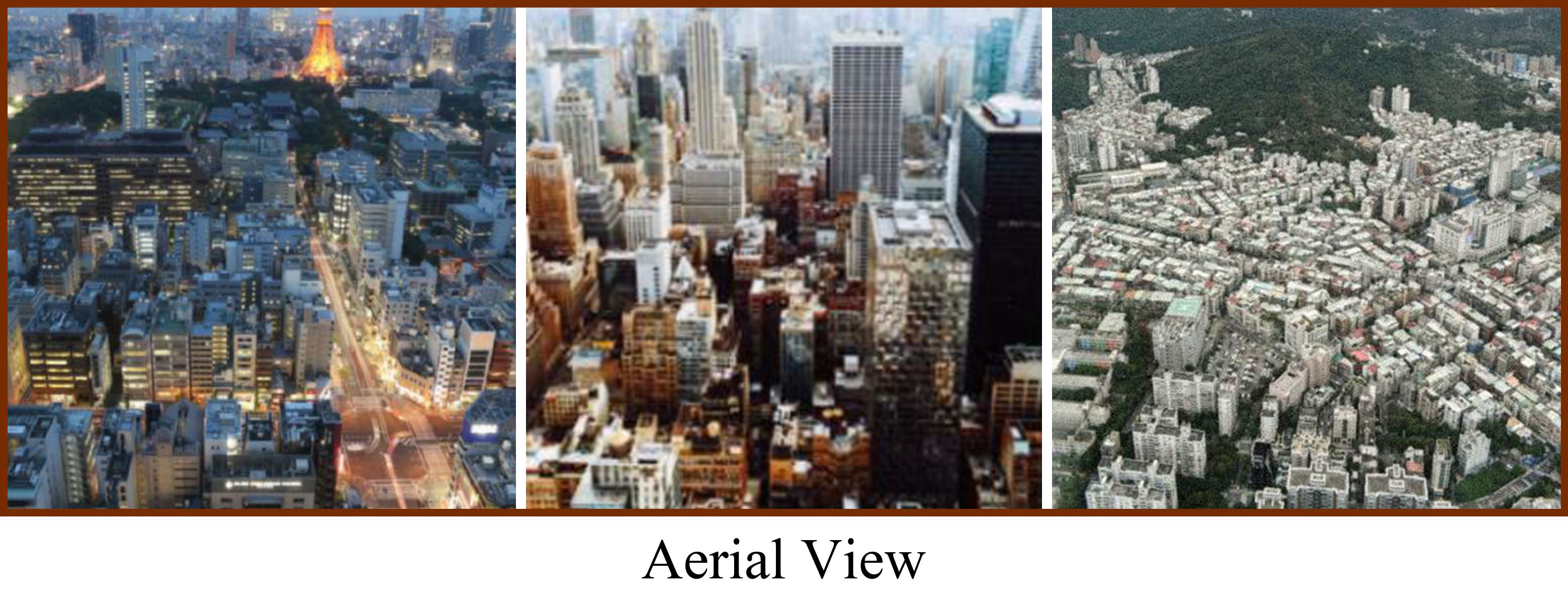}
   \hfill
   \includegraphics[width=0.47\linewidth]{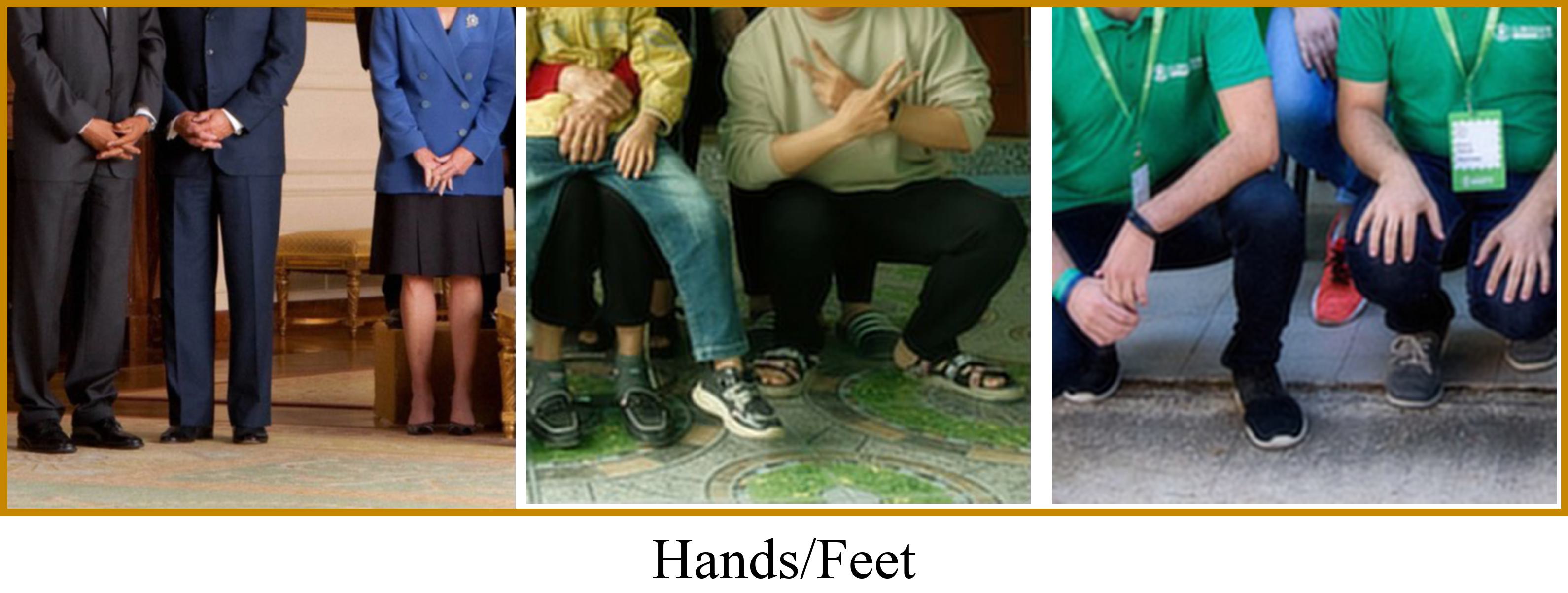}
   \hfill
   \includegraphics[width=0.47\linewidth]{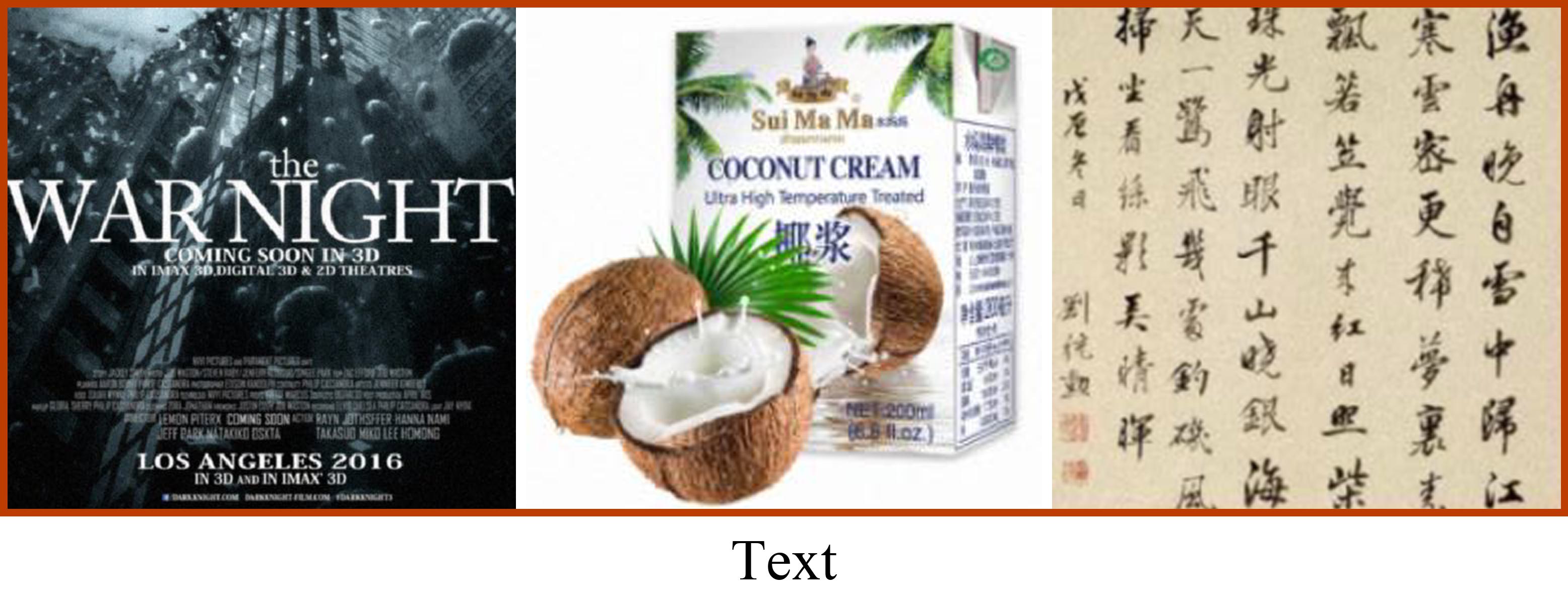}
   \hfill
   \includegraphics[width=0.47\linewidth]{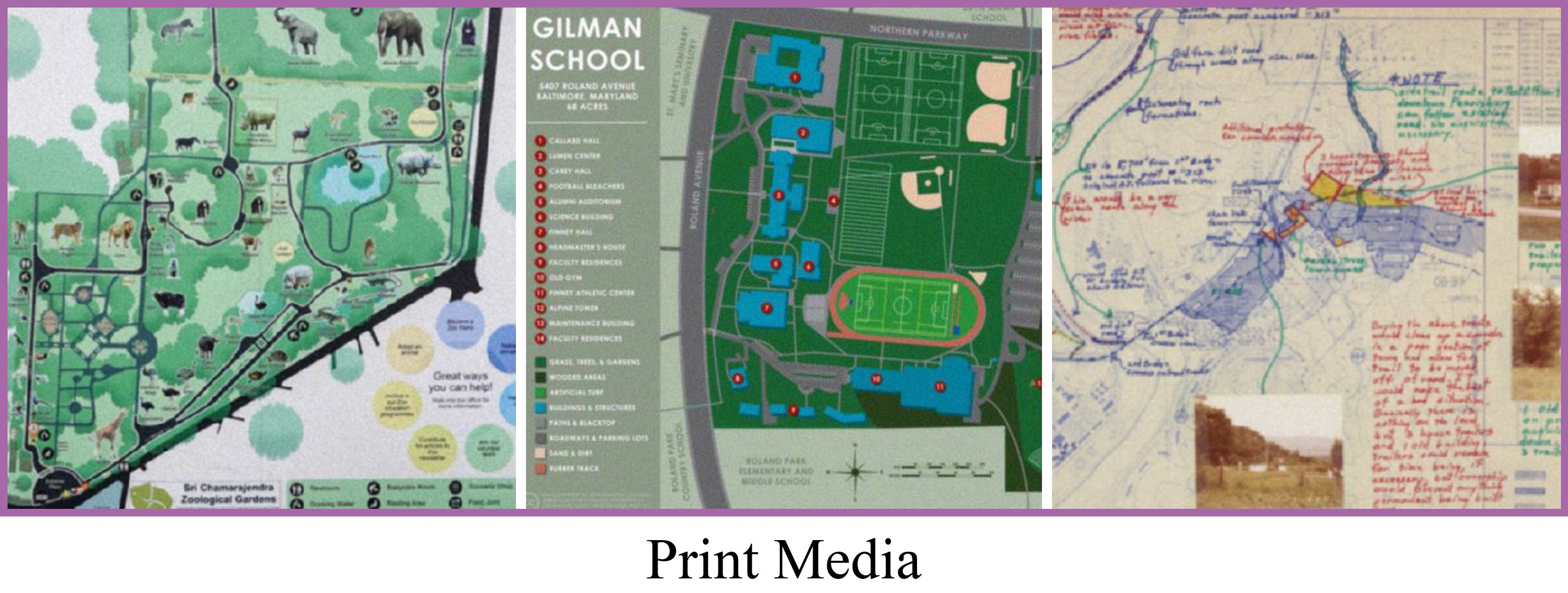}
   \hfill
   \includegraphics[width=0.47\linewidth]{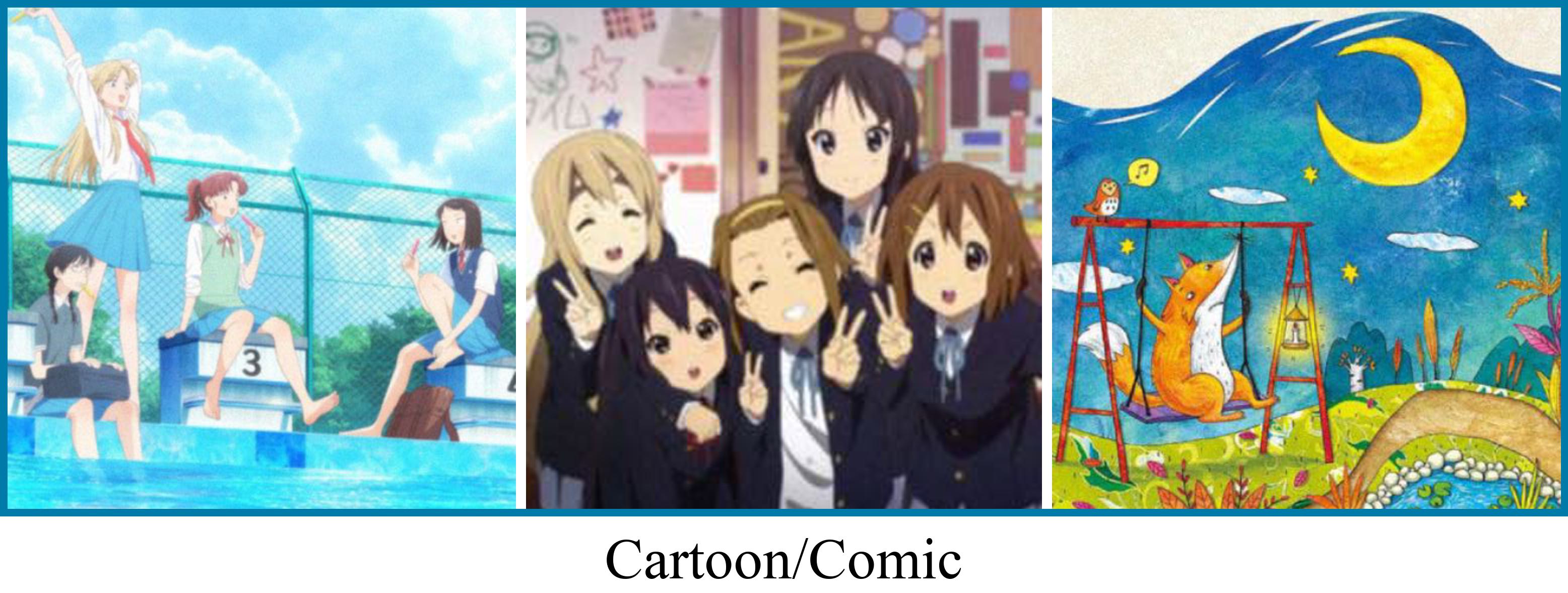}
   \hfill
   \includegraphics[width=0.47\linewidth]{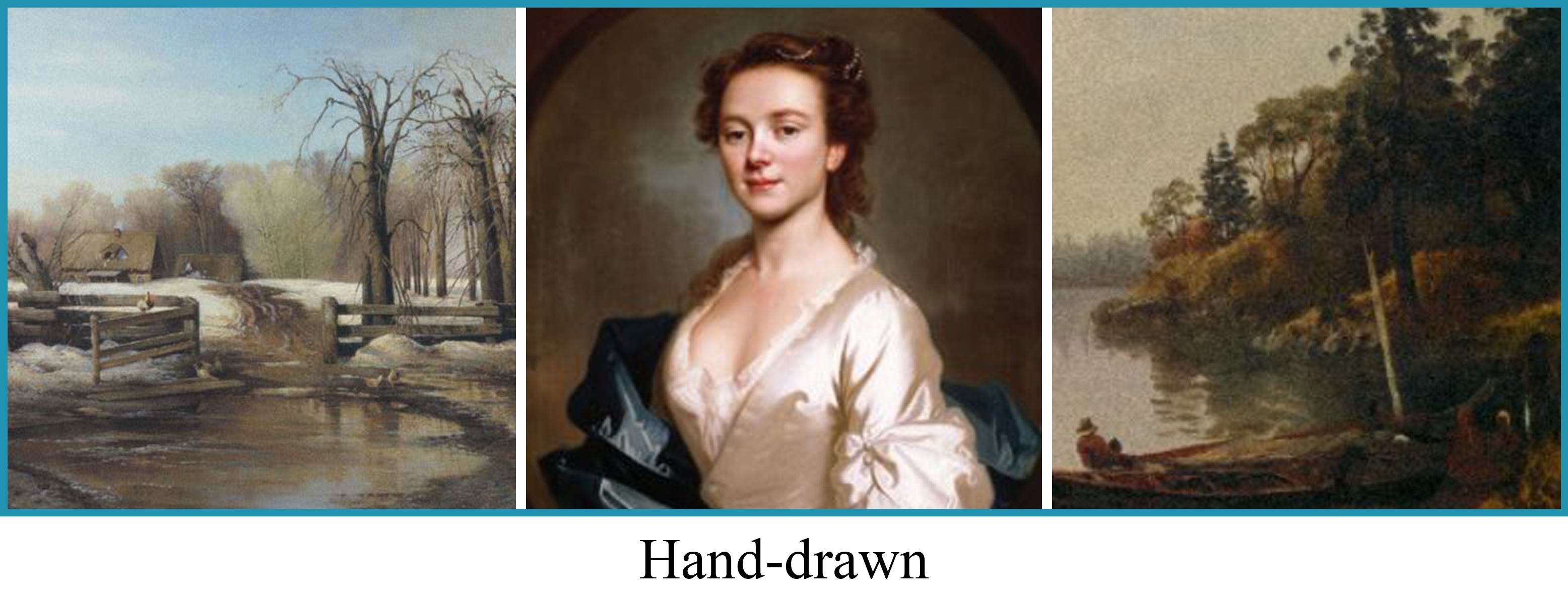}
    \caption{
    More low-quality examples of different semantic categories. Zoom in for a better view.
    }
    \label{fig:dataset_semantics_supp2}
\end{figure*}

\begin{figure*}[tp]
\scriptsize
\centering
    \includegraphics[width=0.48\linewidth]{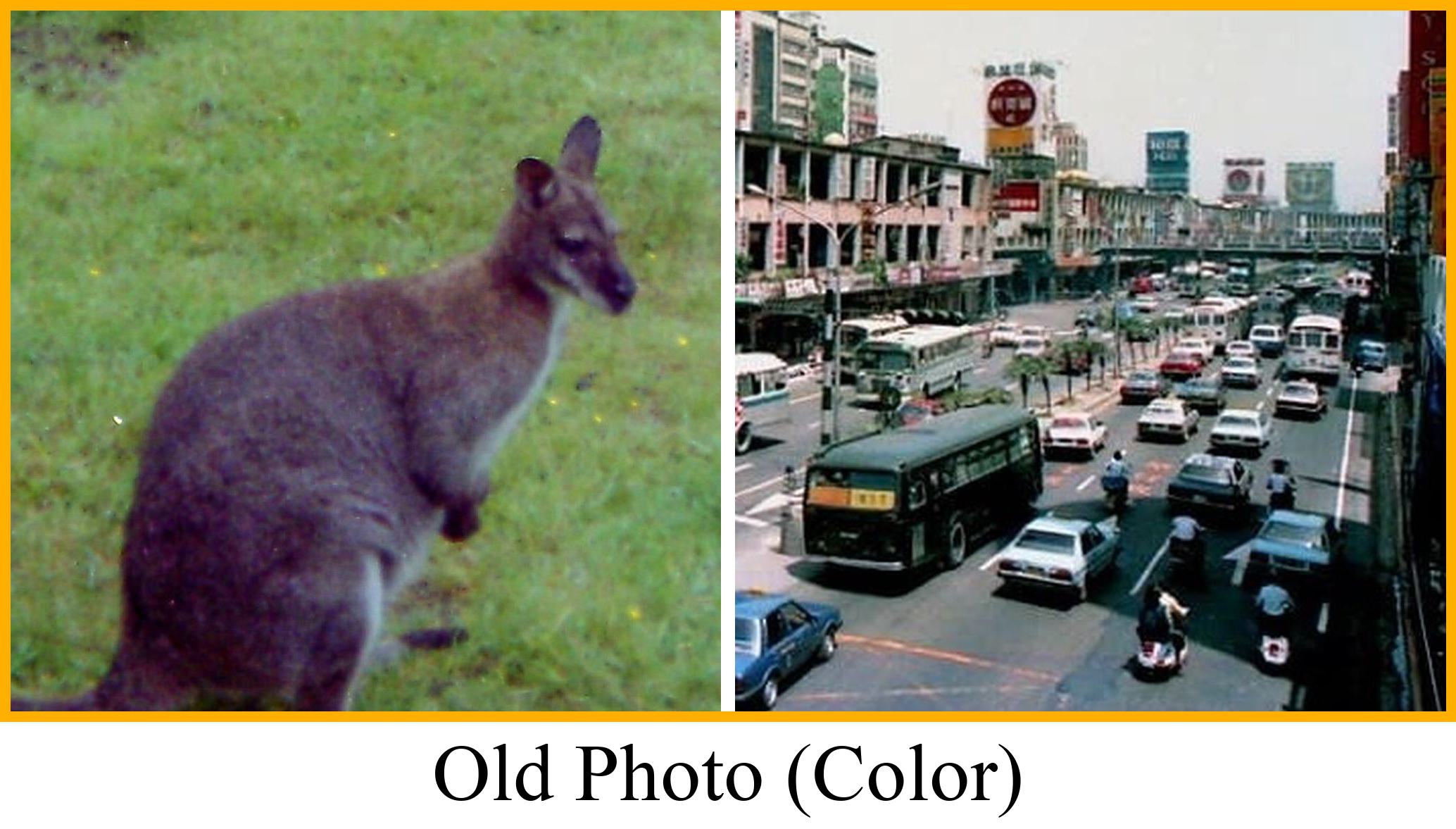}
    \hfill
   \includegraphics[width=0.48\linewidth]{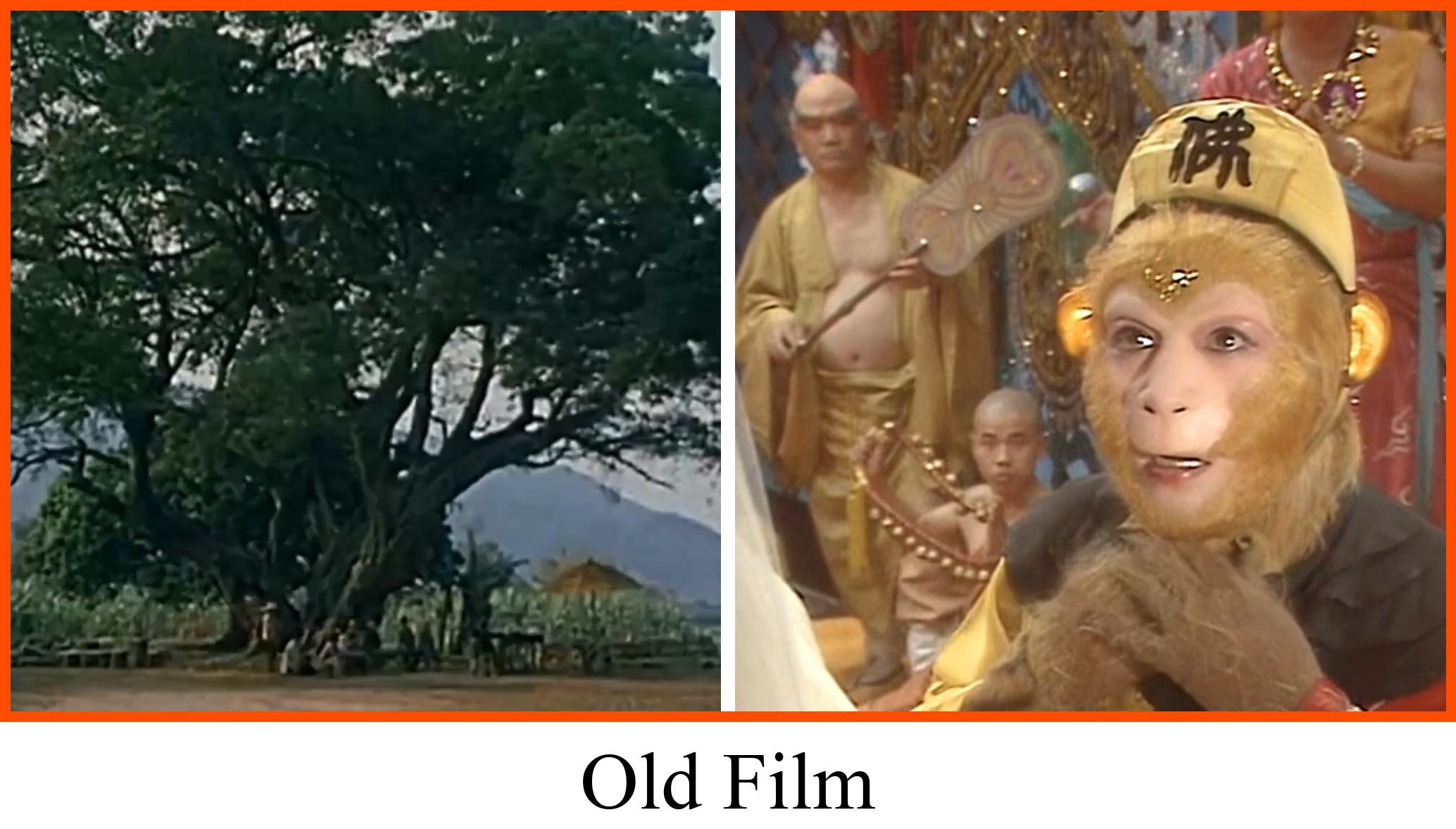}
    \caption{
    More visual examples of different degradation categories. Zoom in for a better view.
    }
    \label{fig:dataset_degradation_supp1}
\end{figure*}

\begin{figure*}[tp]
\scriptsize
\centering
    \includegraphics[width=0.48\linewidth]{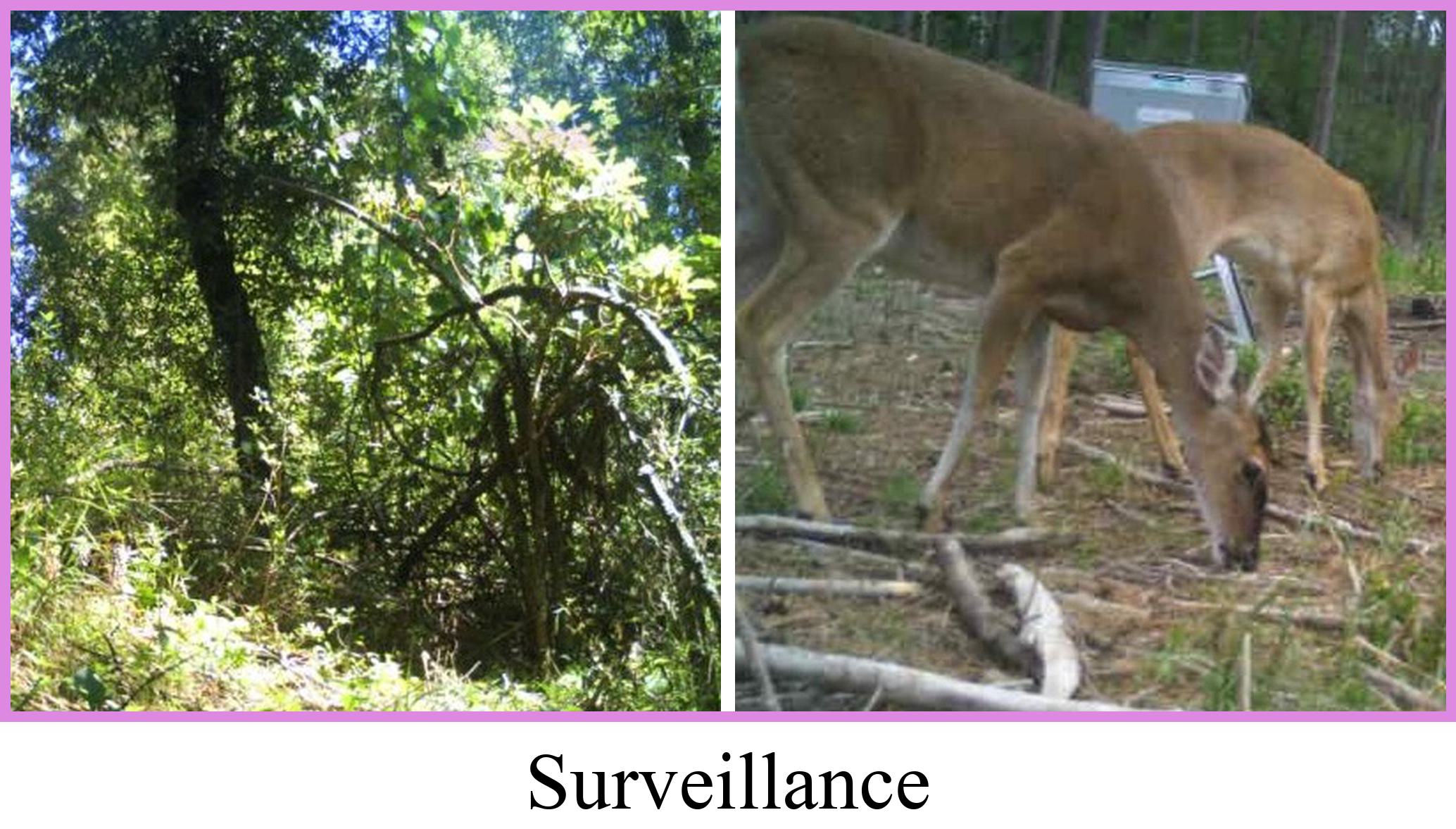}
   \hfill
   \includegraphics[width=0.48\linewidth]{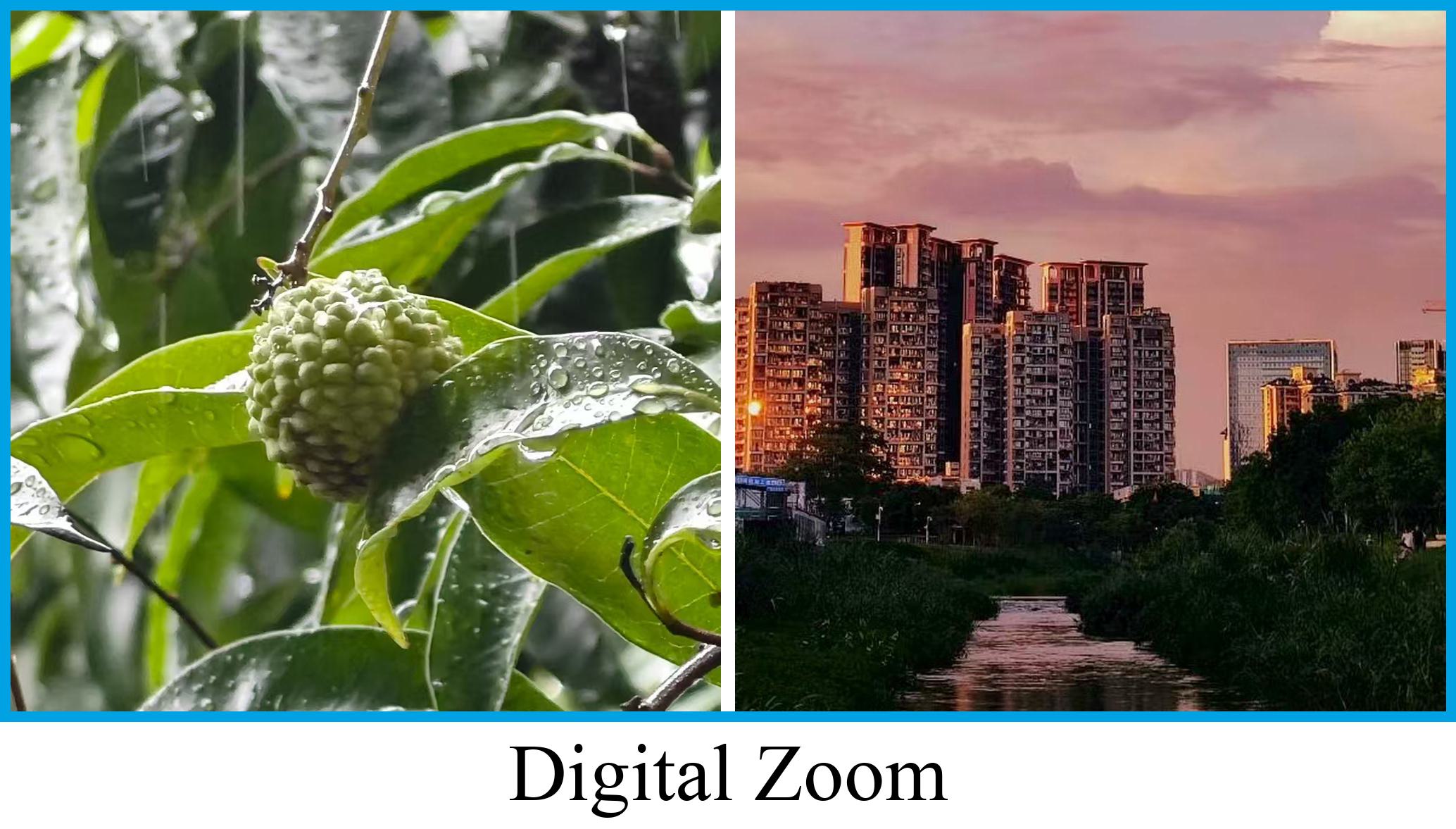}
   \hfill
   \includegraphics[width=0.48\linewidth]{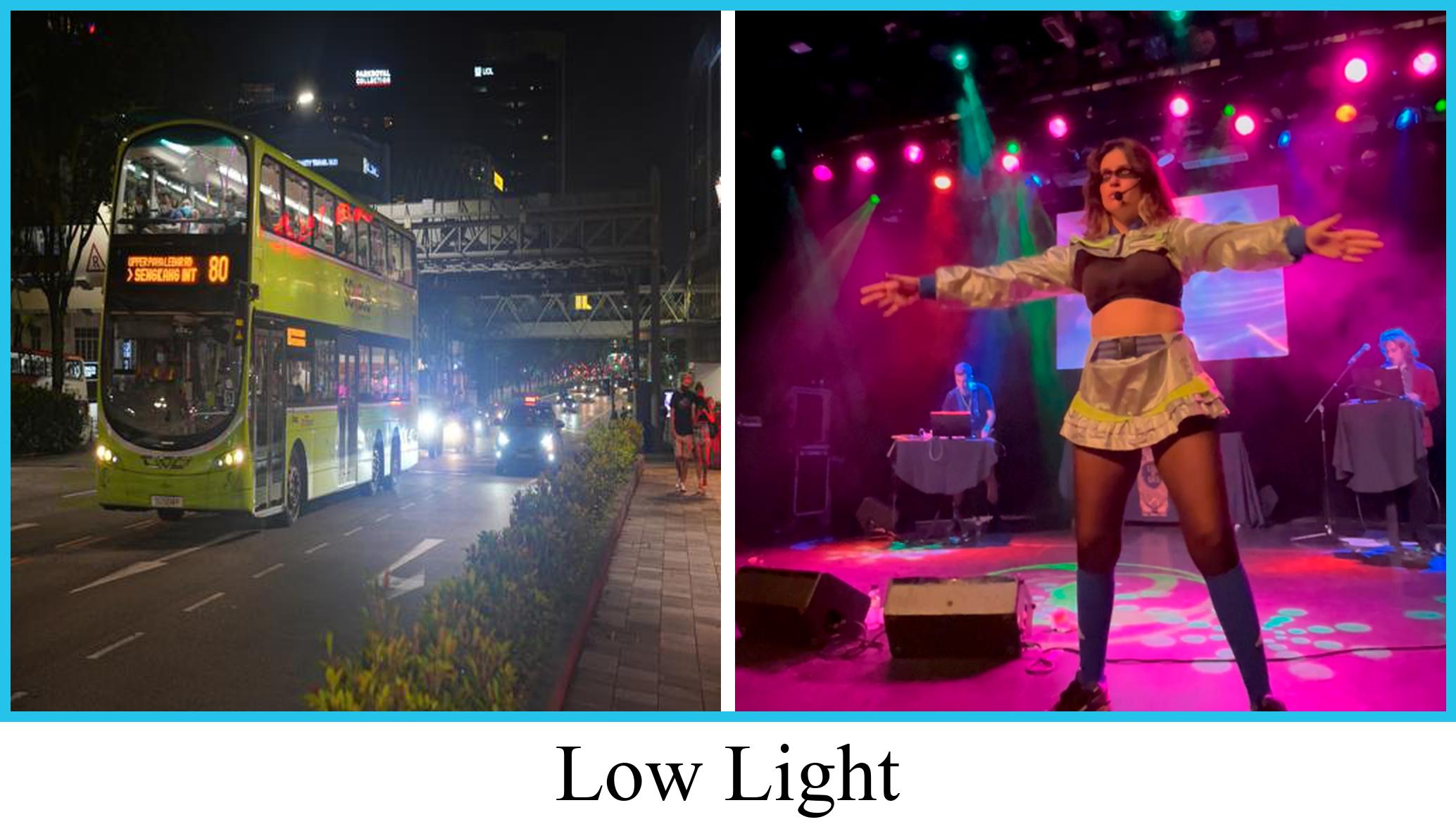}
   \hfill
   \includegraphics[width=0.48\linewidth]{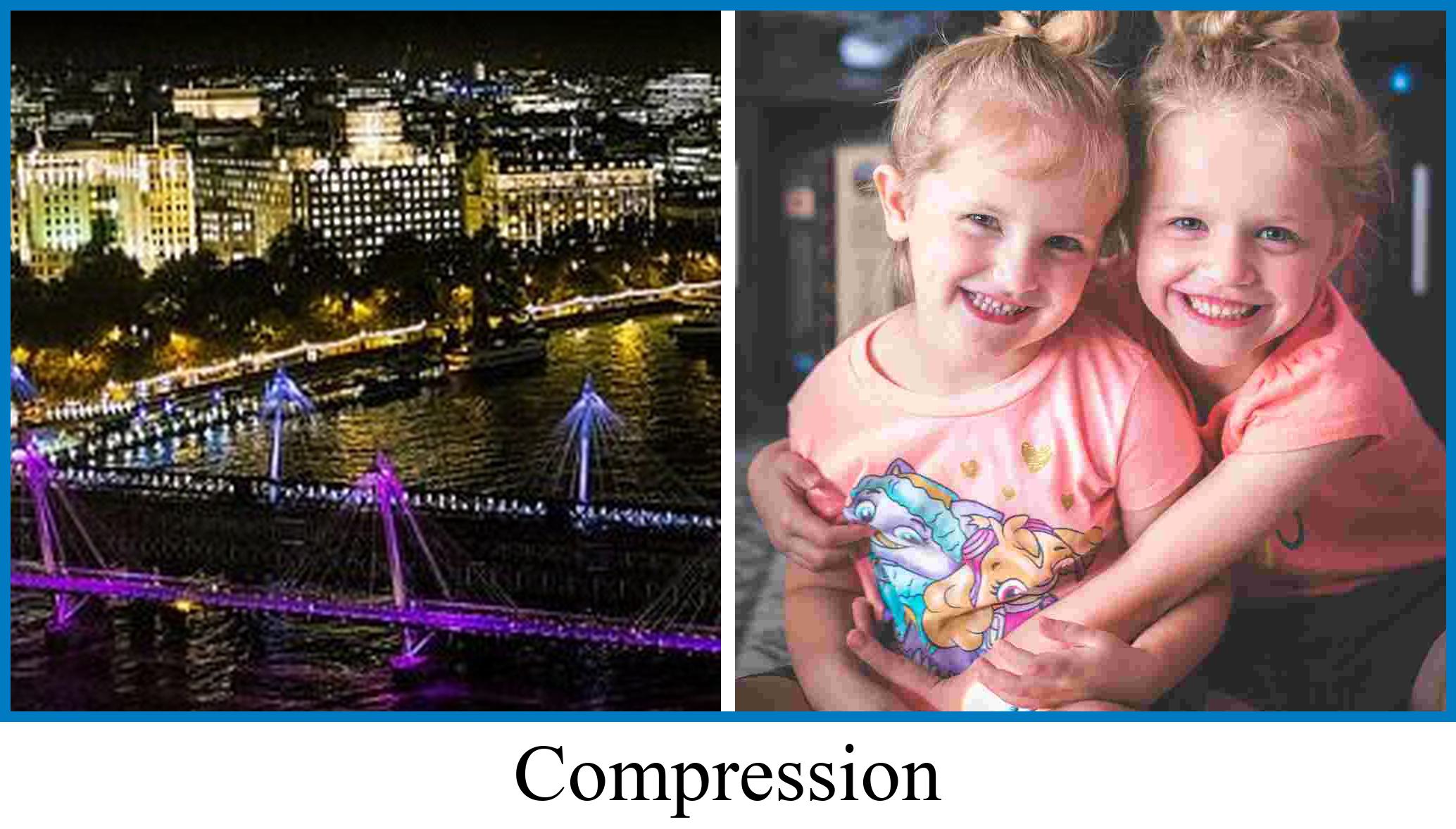}
   \hfill
   \includegraphics[width=0.48\linewidth]{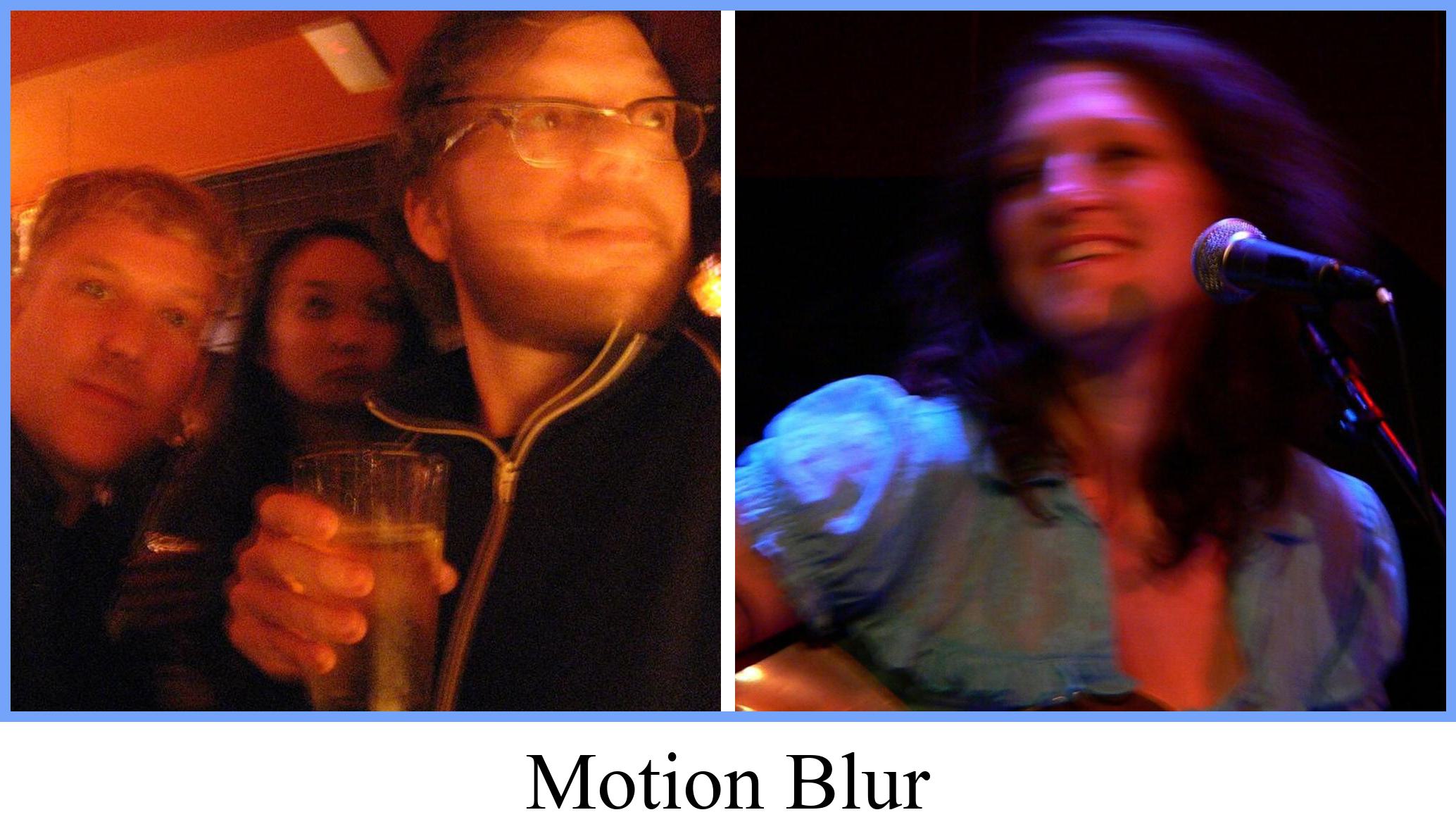}
   \hfill
   \includegraphics[width=0.48\linewidth]{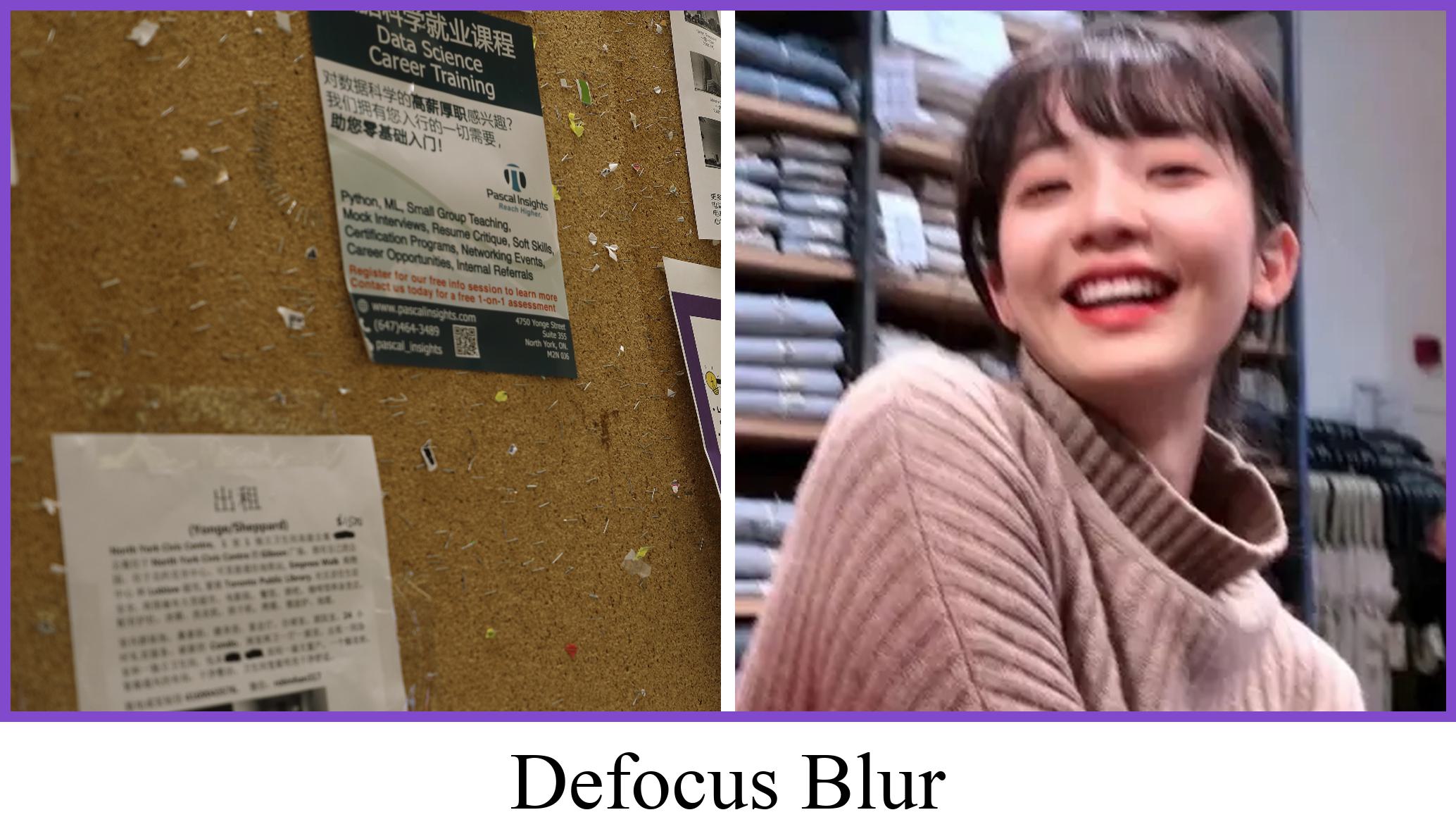}
   \hfill
   \includegraphics[width=0.48\linewidth]{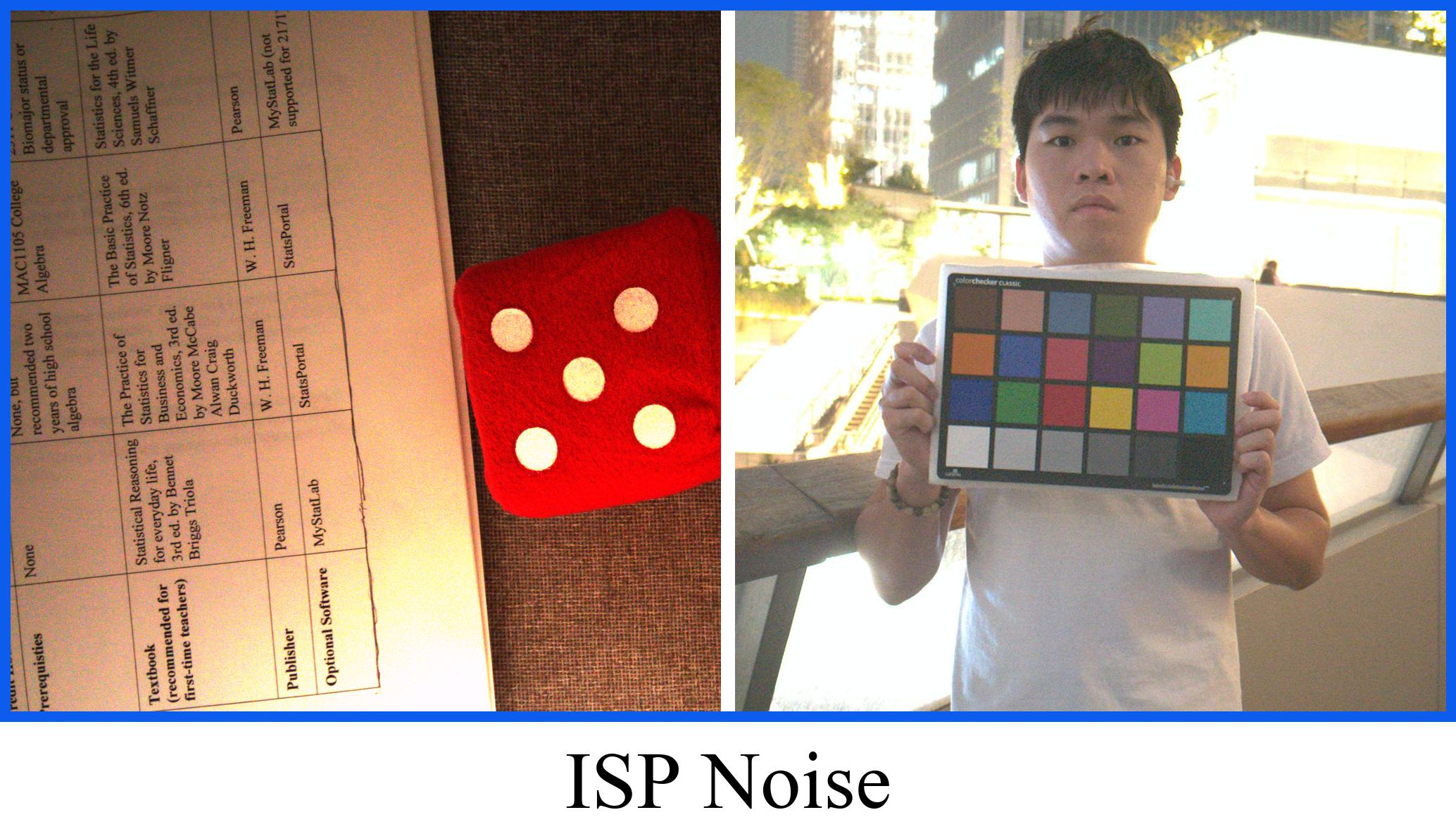}
   \hfill
   \includegraphics[width=0.48\linewidth]{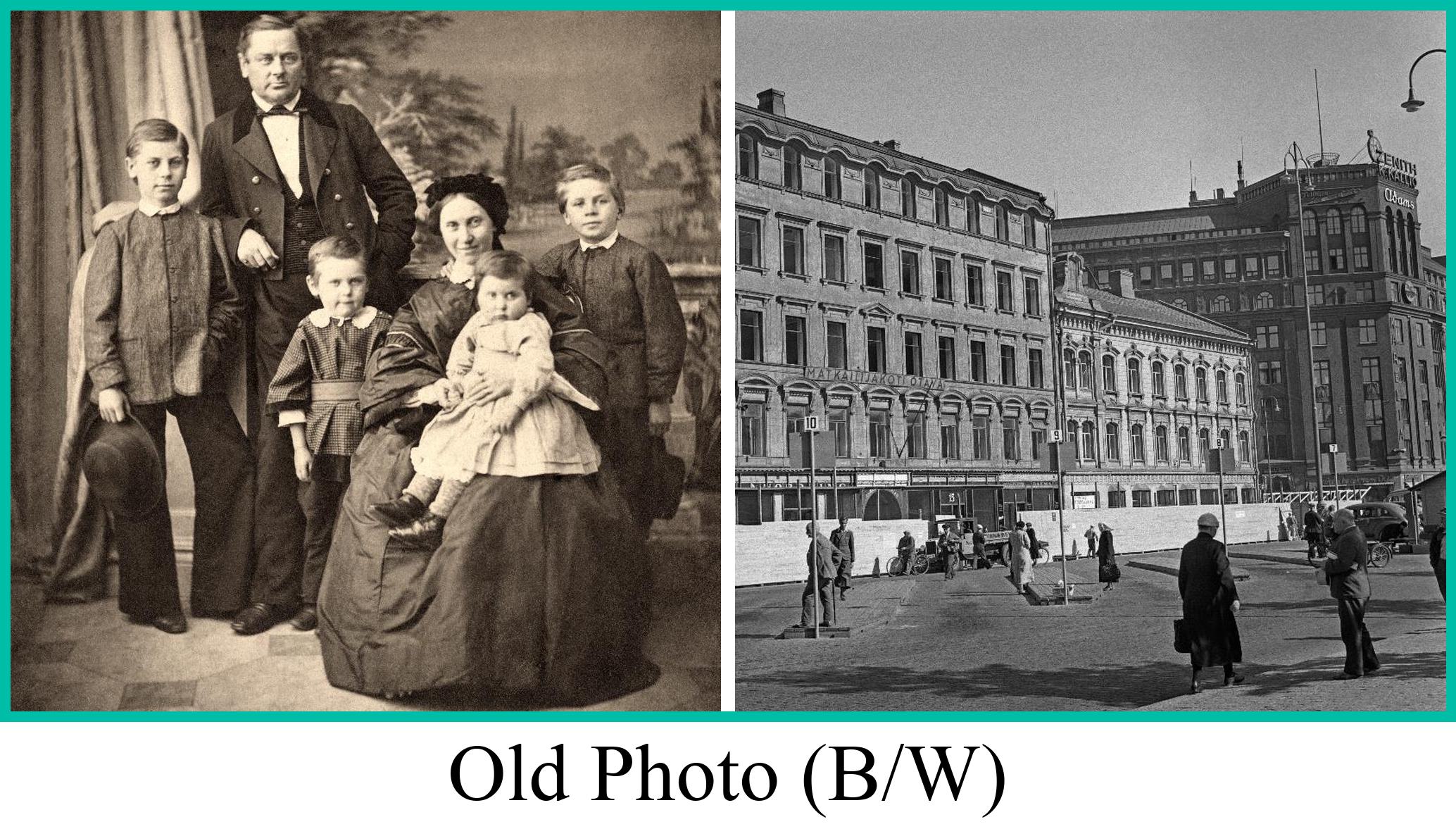}
    \caption{
    More visual examples of different degradation categories. Zoom in for a better view.
    }
    \label{fig:dataset_degradation_supp2}
\end{figure*}

To better demonstrate the semantic and degradation diversity covered in our benchmark, we include extended visual examples in~\cref{fig:dataset_semantics_supp1,fig:dataset_semantics_supp2,fig:dataset_degradation_supp1,fig:dataset_degradation_supp2}.

\subsection{Implementation Details of Restoration Models}
Most models were evaluated using their official default configurations, including pretrained weights and recommended inference hyperparameters, to ensure fair comparison and reproducibility.
For SUPIR, we adopt the configuration \hytt{s\_stage2 = 0.85}, \hytt{s\_cfg = 4.0}, \hytt{spt\_linear\_CFG = 1.0}, and \hytt{s\_noise = 1.007}.
For HYPIR, we use the Flux-based variant and set \hytt{sharpness = 800} and \hytt{noise = 200}, without any prompt provided.
For PASD, we evaluate the model using its official SD-XL version.

Generation models require prompts for image-to-image restoration.
To ensure consistency, we use the same prompt across both generation models:
\begin{quote}
\hytt{Perform high-quality image restoration and super-resolution in this image: enhance clarity, remove noise and scratches, reduce blur, and sharpen fine details, while preserving a natural, realistic appearance without over-processing or artificial artifacts.}
\end{quote}

\begin{figure*}[tp]
\scriptsize
\centering
    \includegraphics[width=\linewidth]{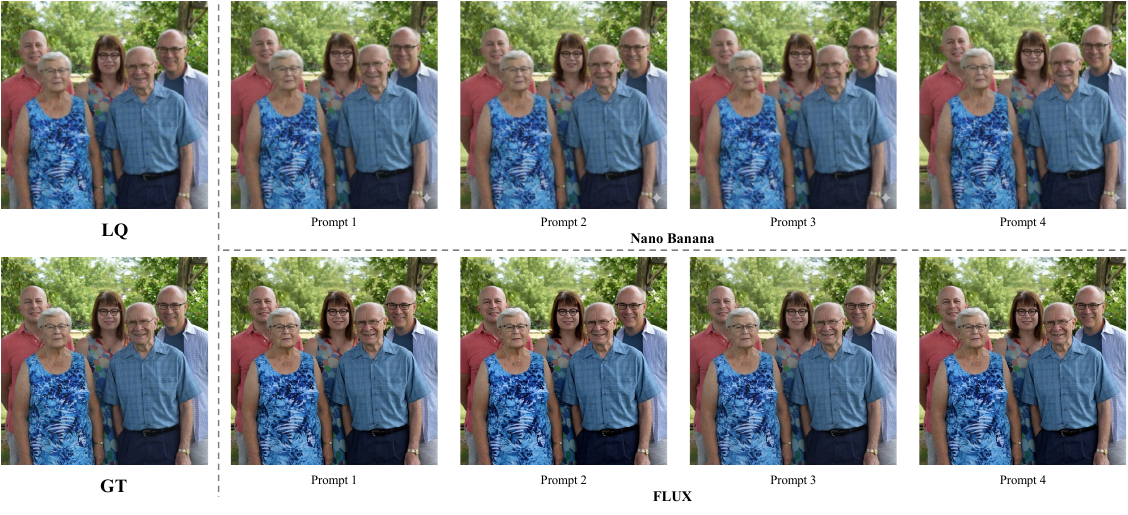}
    \includegraphics[width=\linewidth]{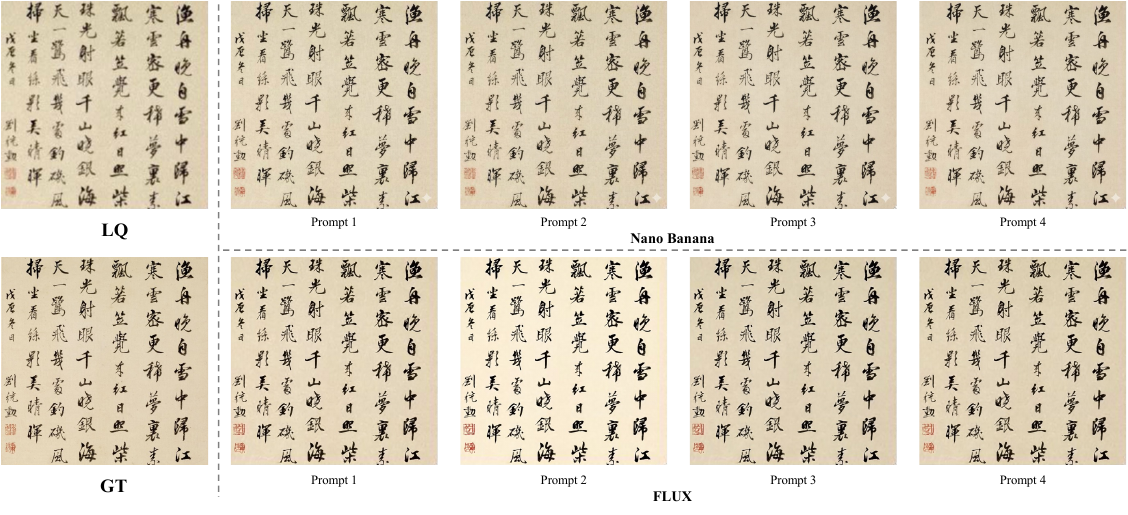}
    \caption{
    Results of different prompts used in generation models. Zoom in for a better view. \\
    Prompt 1: \hytt{Perform high-quality image restoration and super-resolution in this image: enhance clarity, remove noise and scratches, reduce blur, and sharpen fine details—while preserving a natural, realistic appearance without over-processing or artificial artifacts.}\\
    Prompt 2: \hytt{Restore this image with enhanced clarity and sharpness. Reduce noise, fix compression artifacts, and improve detail while strictly preserving the original content, style, and semantics.}\\
    Prompt 3: \hytt{Clean and restore the image by removing artifacts, noise, and blur. Maintain full fidelity to the original composition and avoid creating new visual elements.}\\
    Prompt 4: \hytt{Improve this image’s perceptual quality: refine edges, correct degradation, and enhance fine details. Do not alter identity, expression, or any semantic attributes.}
    }
    \label{fig:prompts_results1}
\end{figure*}

\begin{figure*}[tp]
\scriptsize
\centering
    \includegraphics[width=\linewidth]{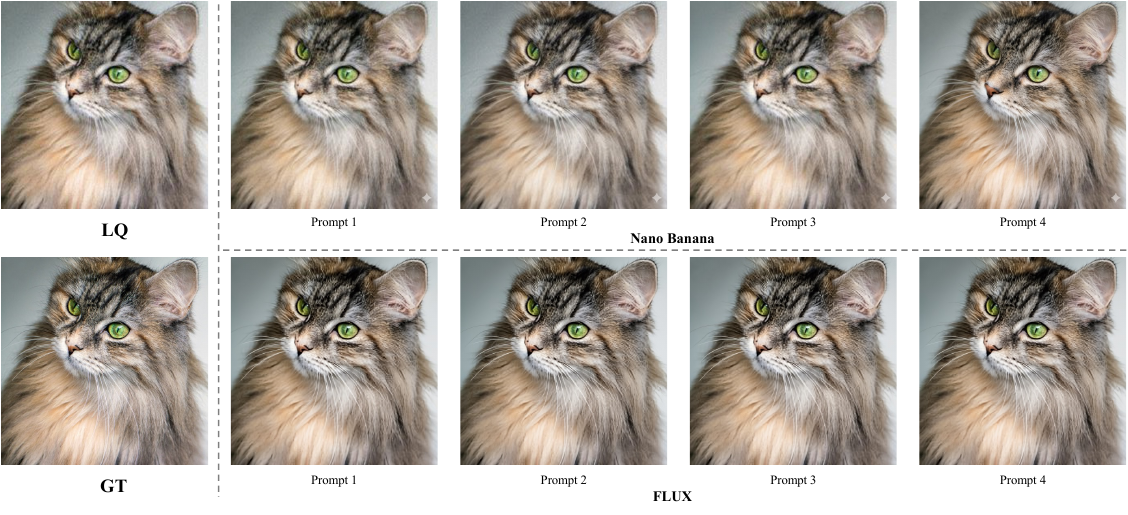}
    \includegraphics[width=\linewidth]{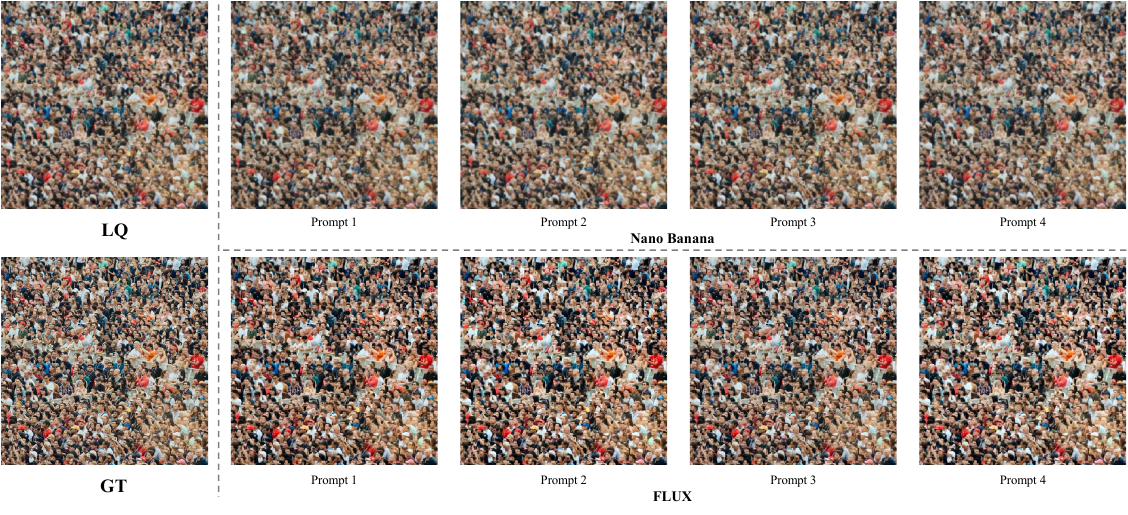}
    \caption{
    Results of different prompts used in generation models. Zoom in for a better view.\\
    Prompt 1: \hytt{Perform high-quality image restoration and super-resolution in this image: enhance clarity, remove noise and scratches, reduce blur, and sharpen fine details—while preserving a natural, realistic appearance without over-processing or artificial artifacts.}\\
    Prompt 2: \hytt{Restore this image with enhanced clarity and sharpness. Reduce noise, fix compression artifacts, and improve detail while strictly preserving the original content, style, and semantics.}\\
    Prompt 3: \hytt{Clean and restore the image by removing artifacts, noise, and blur. Maintain full fidelity to the original composition and avoid creating new visual elements.}\\
    Prompt 4: \hytt{Improve this image’s perceptual quality: refine edges, correct degradation, and enhance fine details. Do not alter identity, expression, or any semantic attributes.}
    }
    \label{fig:prompts_results2}
\end{figure*}

To verify that our choice of prompt does not bias the evaluation, we further tested each generation model using several alternative prompts. 
Across all tested prompts, we observed only minor variations in overall perceptual quality and no systematic improvement over the main prompt we used. 
Representative comparisons of different prompts are shown in the~\cref{fig:prompts_results1,fig:prompts_results2}, demonstrating that our chosen prompt provides stable and representative restoration behavior.

\subsection{Detailed Criteria of Human Evaluation}
Here, we provide the full detailed criteria used by annotators during manual evaluation, expanding upon the definitions of \textbf{Detail}, \textbf{Sharpness}, and \textbf{Semantics}.

\textbf{Detail.} This dimension evaluates the amount of fine-grained textures.  
-3: 
The restored image exhibits almost no improvement in local or global detail. 
Fine patterns, textures, or surface structures are largely absent, and most objects appear flat and featureless.
-2:
The restoration introduces limited additional detail compared to the degraded input. 
Fine textures and local variations remain largely missing or underdeveloped.
-1: 
The image shows moderate enhancement in detail, but fine textures are still partially missing or over-smoothed in certain regions.
0:
The level of detail is appropriate and consistent with the content of the input image. 
Textures are well-balanced, yielding a natural and realistic appearance.
1:
Additional details beyond the original content are generated. These details slightly deviate from the ground truth and reduce overall realism.
2:
A considerable amount of excessive or inconsistent detail is introduced, producing cluttered textures or unnatural structures that degrade visual authenticity.
3:
The restoration introduces an overwhelming number of artificial and incoherent details, leading to an overly complex, chaotic, and unrealistic appearance.

\begin{figure}[tp]
\scriptsize
\centering
    \includegraphics[width=\linewidth]{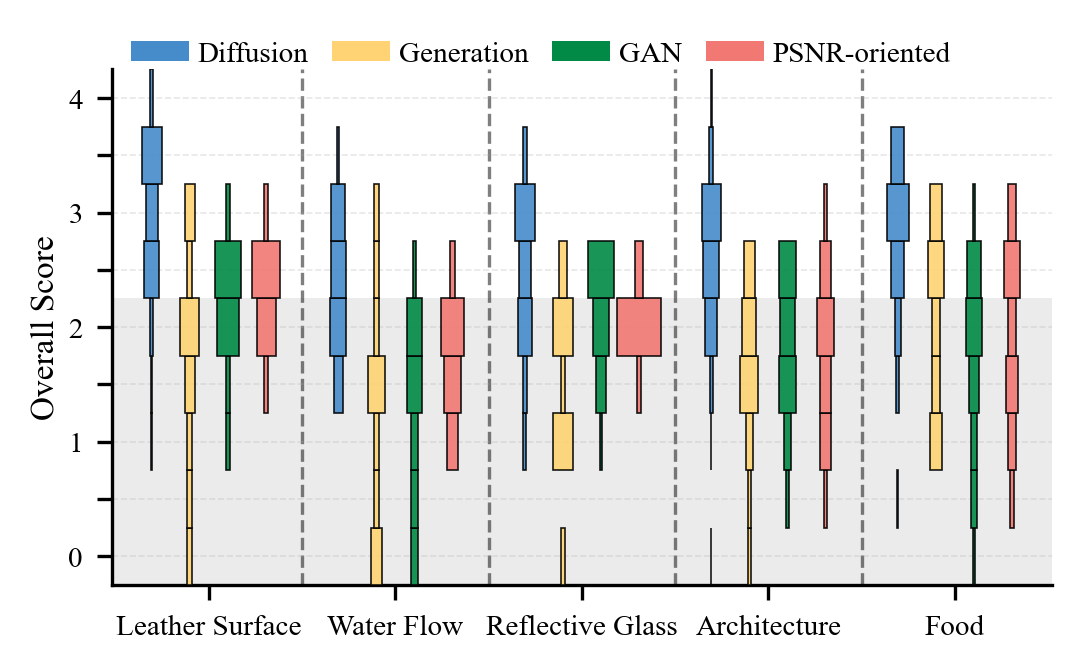}
    \caption{
    Distribution of annotated overall scores across semantic scene groups. The horizontal width of each box indicates the percentage of samples within each score interval. The light gray region indicates low overall scores, representing generally unacceptable results.
    }
    \label{fig:scenes_total_supp}
\end{figure}

\textbf{Sharpness.}
-3:
The restored image remains as blurry as the low-quality input, showing no perceptual improvement in clarity.
-2:
The image shows a slight overall improvement in clarity but remains noticeably blurry across most regions. 
-1:
The image appears generally clear, yet minor blurriness persists in specific local regions 
0:
Sharpness is appropriate; object boundaries appear clear, continuous, and natural.
1:
Object boundaries are slightly sharper than those in natural scenes. The excessive edge enhancement slightly degrades perceptual quality.
2:
Object boundaries appear overly sharp and unnatural, sometimes producing artifacts such as white halos or excessive contrast.
3:
The entire image appears excessively sharp and highly unnatural, with very strong white edges and exaggerated contrast.

\textbf{Semantics.}
0:
Total absence of the required category's semantics. 
Object completely missing, unrecognizable, or the content is pure hallucination relative to the input
1:
Semantic generation was attempted but resulted in a highly chaotic structure, making the intended object unidentifiable or leading to critical structural failure.
2:
Semantic structure is recognizable, but significant texture or geometric distortions are present, severely impacting the perception of the semantic meaning.
3:
Semantic structure is largely correct and fully recognizable. 
Minor texture or structural distortions exist, but they do not lead to misinterpretation, and there are no major structural errors.
4:
Generated texture and structure are natural, reasonable, and fully consistent with the input image.

\begin{table}[tp]
\centering
\footnotesize
\begin{tabular}{>{\centering\arraybackslash}m{1.1cm}|>{\centering\arraybackslash}m{0.8cm}|>{\centering\arraybackslash}m{5cm}}
\toprule
\multirow{2}{*}{Model} & Config  & \multirow{2}{*}{Configurations} \\
& ID & \\
\midrule
\multirow{3}{*}{HYPIR} & 1 &  \hytt{sharpness:800, noise:600}\\
& 2 &  \hytt{sharpness:4000, noise:200}\\
& 3 & \hytt{sharpness:4000, noise:600}\\
\midrule
\multirow{3}{*}{PiSA-SR} & 1 & \hytt{lambda\_pix:1.0, lambda\_sem:0.3} \\
& 2 & \hytt{lambda\_pix:2.0, lambda\_sem:1.0}\\
& 3 & \hytt{lambda\_pix:0.3, lambda\_sem:1.0} \\
\midrule
\multirow{3}{*}{SeeSR} & 1 & \hytt{conditioning\_scale:1.0}\\
& 2 & \hytt{conditioning\_scale:1.5}\\
& 3 & \hytt{conditioning\_scale:2.0}\\
\midrule
\multirow{4}{*}{DiffBIR} & 1 & \hytt{cfg\_scale:1, noise\_aug:0} \\
& 2 & \hytt{cfg\_scale:10, noise\_aug:0} \\
& 3 &  \hytt{cfg\_scale:1, noise\_aug:5} \\
& 4 &  \hytt{cfg\_scale:10, noise\_aug:5} \\
\midrule
\multirow{8}{*}{SUPIR} & \multirow{2}{*}{1} & \hytt{s\_stage2:1.0, s\_noise:1.030} \\
&&\hytt{s\_cfg:6.5, spt\_linear\_CFG:3.5}\\
& \multirow{2}{*}{2} & \hytt{s\_stage2:0.7, s\_noise:1.030}\\
&&\hytt{s\_cfg:6.5, spt\_linear\_CFG:3.5}
\\
& \multirow{2}{*}{3} & \hytt{s\_stage2:0.58, s\_noise:1.007}\\
&&\hytt{s\_cfg:4.0, spt\_linear\_CFG:1.0}
\\
& \multirow{2}{*}{4} & \hytt{s\_stage2:1.0, s\_noise:1.007}\\
&&\hytt{s\_cfg:4.0, spt\_linear\_CFG:1.0}
\\
\bottomrule
\end{tabular}
\caption{
Configurations used in our parameter sensitivity analysis.
All other parameters remain at their official defaults.
}
\label{tab:param_setting}
\end{table}
 
\begin{table*}[tp]
\centering
\footnotesize
\setlength\tabcolsep{0.3cm}
\begin{tabular}{
l|c|cccccccccc
}
\toprule
\multirow{2}{*}{Model} & Config & SmallFace & \multirow{2}{*}{Crowd} & \multirow{2}{*}{Vehicles} & Street & Aerial & Hands/ & Complex & \multirow{2}{*}{Text} & Digital & OP\\
& ID & Face & & & View & View & Feet & Texture & & Zoom & (Color)\\
\midrule
\multirow{3}{*}{HYPIR} & 1 &2.83 &2.00 &3.00 &2.71 &\textbf{2.90} &2.68 &2.58 &2.50 &2.17 &2.23  \\
& 2 &\textbf{2.94} &\textbf{2.50} &3.10 &\textbf{2.86} &\textbf{2.90} &\textbf{3.05} &\textbf{2.75} &\textbf{2.61} &\textbf{2.61} &\textbf{2.43} \\
& 3 &2.61 &2.10 &\textbf{3.20} &2.71 &2.70 &2.73 &2.58 &2.50 &2.26 &2.14 \\
\midrule
\multirow{3}{*}{PiSA-SR} & 1 & \textbf{1.78} &\textbf{1.60} &2.30 &2.00 &\textbf{1.80} &2.64 &\textbf{2.17} &\textbf{2.22} &\textbf{2.15} &1.67  \\
& 2 & 1.67 &1.30 &\textbf{2.40} &\textbf{2.21} &1.70 &\textbf{2.68} &2.00 &2.11 &2.04 &\textbf{1.94} \\
& 3 &1.28 &1.10 &2.00 &1.79 &1.40 &2.41 &2.08 &2.06 &1.96 &1.44 \\
\midrule
\multirow{3}{*}{SeeSR} & 1 &\textbf{2.28} & \textbf{1.70} &\textbf{2.20} &\textbf{2.21} &\textbf{1.70} &2.30 &\textbf{2.17} &1.78 &\textbf{2.20} &\textbf{2.30} \\
& 2 &1.78 & 1.30 &2.10 &1.79 &\textbf{1.70} &\textbf{2.59} &2.08 &\textbf{1.94} &1.65 &1.66\\
& 3 &1.72 & 1.40 &2.00 &1.71 &\textbf{1.70} &2.41 &1.92 &\textbf{1.94} &1.57 &1.54 \\
\midrule
\multirow{4}{*}{DiffBIR} & 1 & \textbf{2.11} &1.30 &2.50 &2.36 &1.80 &\textbf{2.82} &\textbf{2.58} &\textbf{2.39} &1.70 &2.11 \\
& 2 & 2.06 &\textbf{1.40} &2.70 &2.29 &\textbf{2.40} &2.59 &2.17 &2.17 &1.98 &\textbf{2.26}\\
& 3 & \textbf{2.11} &1.30 &2.50 &2.21 &1.90 &2.77 &\textbf{2.58} &\textbf{2.39} &1.65 &2.01\\
& 4 & 2.06 &\textbf{1.40} &\textbf{2.80} &\textbf{2.50} &2.30 &2.50 &2.17 &2.22 &\textbf{2.07} &\textbf{2.26} \\
\midrule
\multirow{4}{*}{SUPIR} & 1 & 2.44 &1.40 &\textbf{2.80} &\textbf{2.71} &2.50 &2.68 &2.08 &2.17 &1.85 &1.77 \\ 
& 2 & \textbf{2.50} &1.30 &\textbf{2.80} &2.36 &2.10 &2.32 &1.75 &1.78 &1.50 &\textbf{1.87}\\
& 3 & \textbf{2.50} &\textbf{1.70} &2.70 &2.64 &\textbf{2.70} &\textbf{2.86} &\textbf{2.33} &\textbf{2.33} &\textbf{1.87} &1.44 \\
& 4 & 2.33 &1.10 &2.30 &2.14 &1.60 &2.32 &1.83 &1.89 &1.33 &1.20\\
\bottomrule
\end{tabular}
\caption{
Average overall scores of different parameter configurations across semantic categories.
Bold numbers indicate the best-performing configuration within each model for a given category
}
\label{tab:quantity_params}
\end{table*}

\subsection{Consistency of Human Evaluation}
To ensure annotation consistency, we recruited professional annotators from the digital visual art industry, standardized the annotation workflow via a custom Gradio system with unified training, and implemented expert inspection to filter anomalies.
To analyze inter-annotator agreement, a random 10\% subset of the dataset was re-annotated.
The Krippendorff's Alpha coefficients~\cite{Krippendorff} were 0.804/0.729/0.780/0.807 for Detail/Sharpness/Semantics/Overall dimensions.
Statistically, these scores are well above the standard acceptance threshold of 0.667, demonstrating that our data reflects rigorous expert consensus rather than subjective noise.

\section{Additional Results}
\label{sec:add_results}

\subsection{More Score Distributions}
\begin{figure*}[tp]
\scriptsize
\centering
    \begin{subfigure}{\linewidth}
        \centering
        \includegraphics[width=\linewidth]{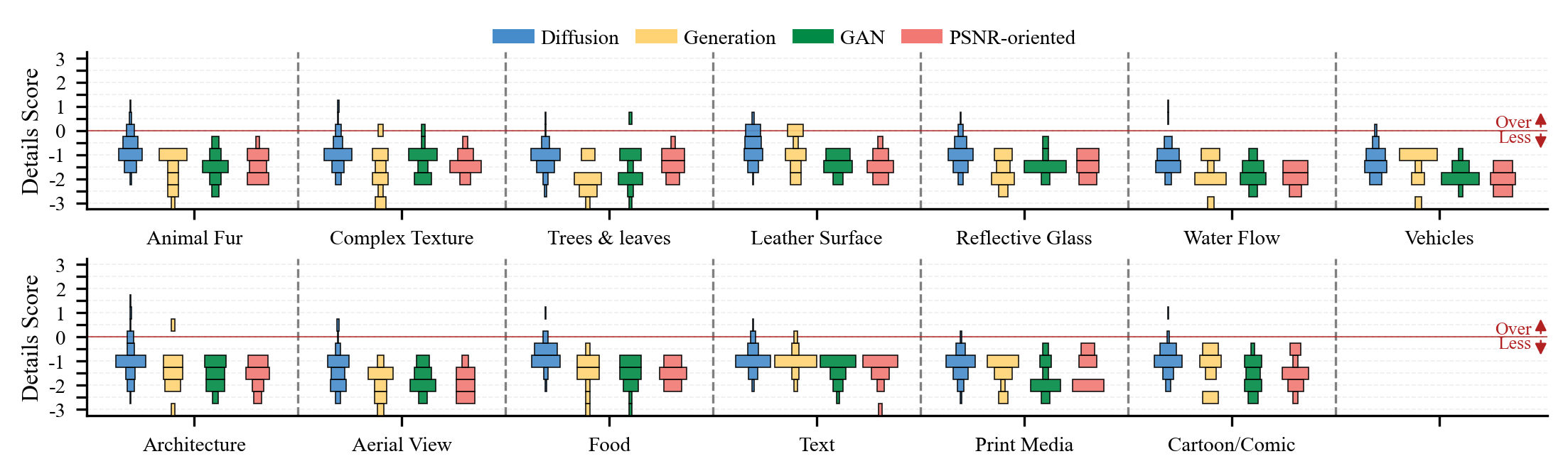}
        \caption{Distribution of detail scores across semantic scene groups.}
        \label{fig:detail_scene_supp}
    \end{subfigure}
    \begin{subfigure}{\linewidth}
        \centering
        \includegraphics[width=\linewidth]{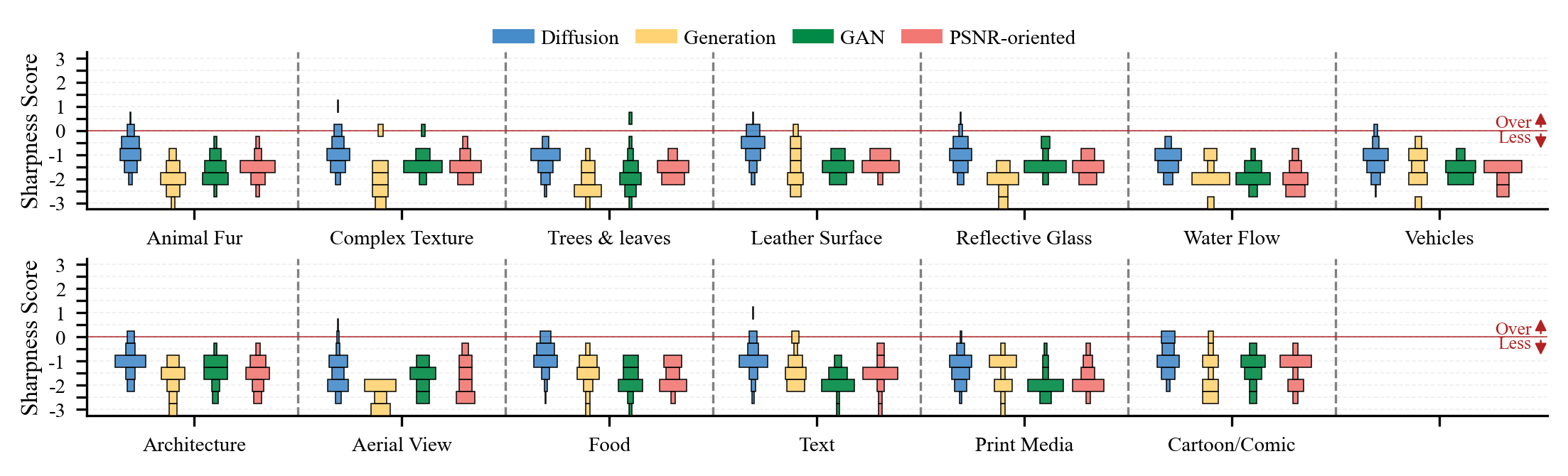}
        \caption{Distribution of sharpness scores across semantic scene groups.}
        \label{fig:sharpness_scene_supp}
    \end{subfigure}
    \begin{subfigure}{\linewidth}
        \centering
        \includegraphics[width=\linewidth]{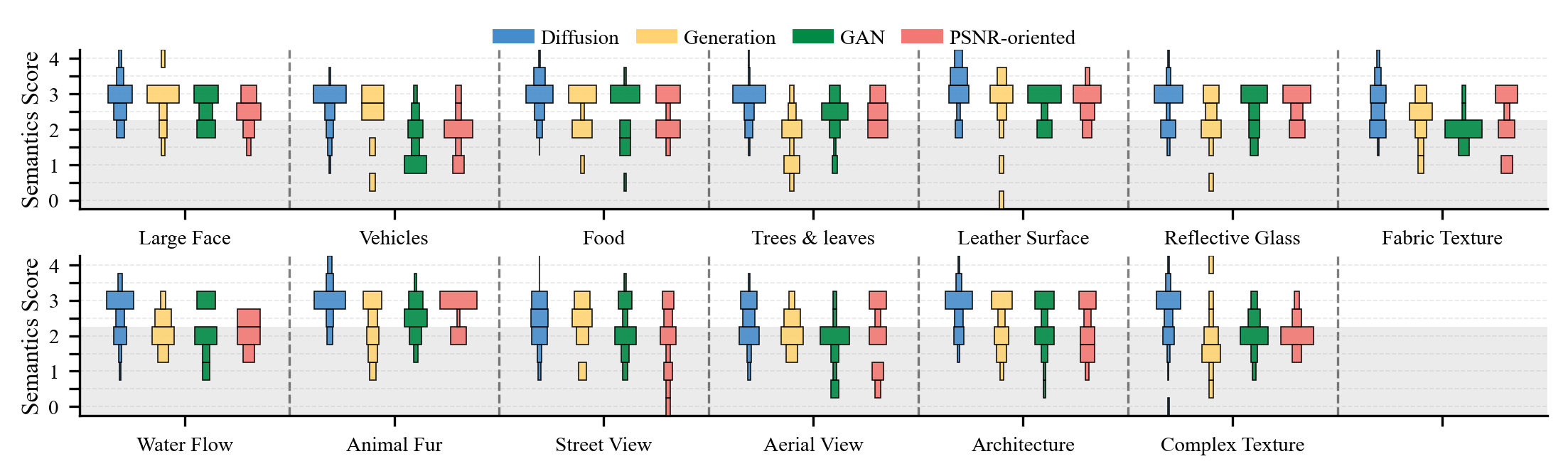}
        \caption{Distribution of semantic scores across semantic scene groups.}
        \label{fig:semantics_scene_supp}
    \end{subfigure}
    \caption{
    Distribution of different scores across various semantic scene groups. The horizontal width of each box indicates the percentage of samples within each score interval. 
    The red line indicates the balance point; scores above it (\textcolor{firebrick}{Over $\uparrow$}) denote over-generation, and scores below it (\textcolor{firebrick}{Less $\downarrow$}) denote under-generation.
    The light gray region indicates low overall scores, representing generally unacceptable results.
    }
    \label{fig:scenes_detail_sharpness_semantics_supp}
\end{figure*}

\begin{figure*}[tp]
\scriptsize
\centering
    \begin{subfigure}{0.48\linewidth}
        \centering
        \includegraphics[width=\linewidth]{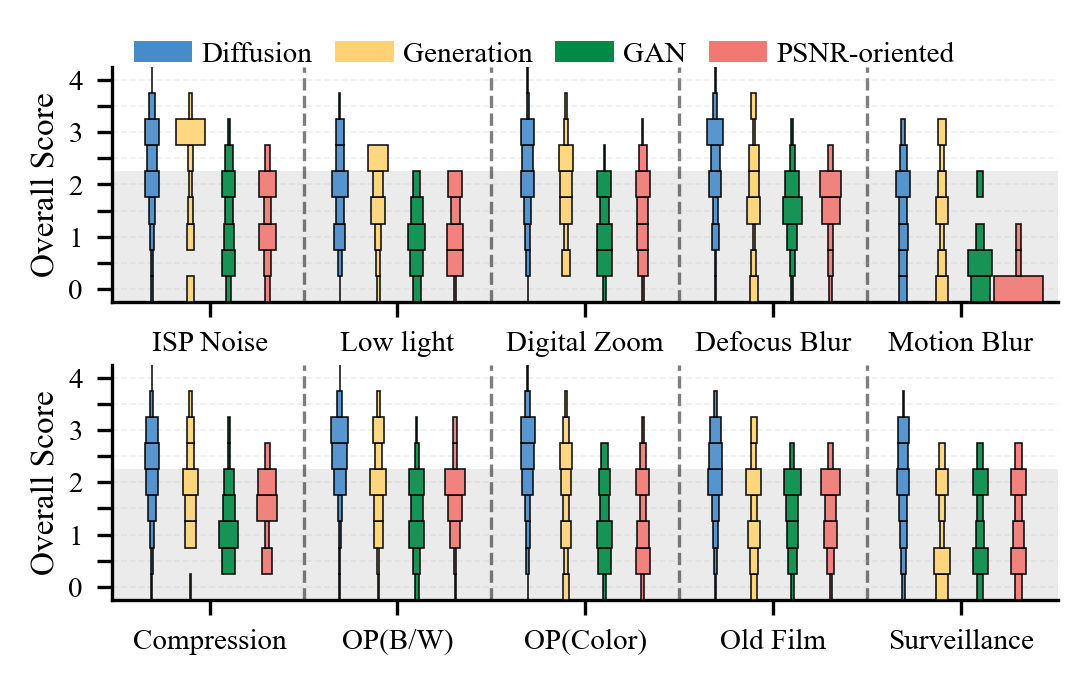}
        \caption{Distribution of overall scores across degradations.}
        \label{fig:total_deg_supp}
    \end{subfigure}
    \hfill
    \begin{subfigure}{0.48\linewidth}
        \centering
        \includegraphics[width=\linewidth]{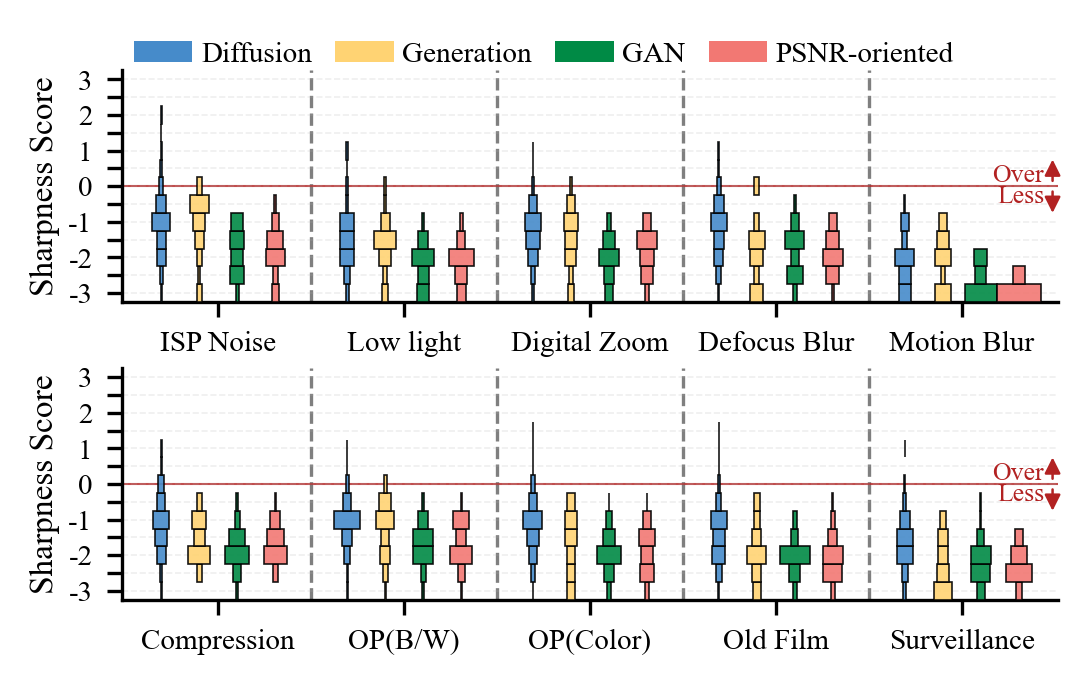}
        \caption{Distribution of semantic scores across degradations.}
        \label{fig:sharpness_deg_supp}
    \end{subfigure}
    \begin{subfigure}{0.48\linewidth}
        \centering
        \includegraphics[width=\linewidth]{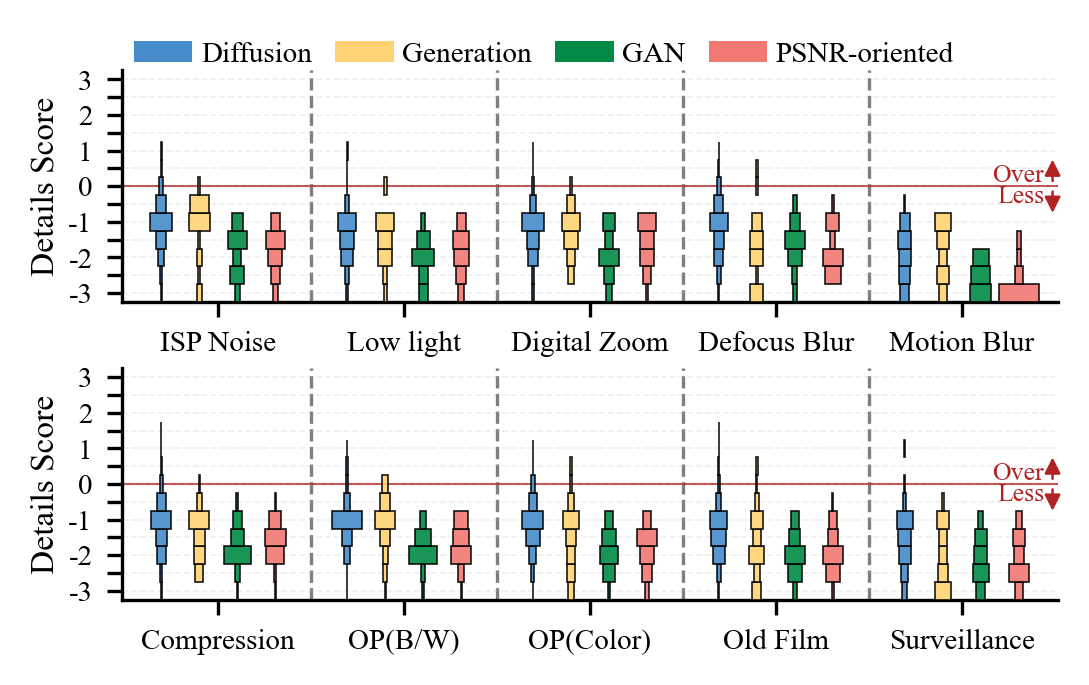}
        \caption{Distribution of detail scores across degradations.}
        \label{fig:detail_deg_supp}
    \end{subfigure}
    \begin{subfigure}{0.48\linewidth}
        \centering
        \includegraphics[width=\linewidth]{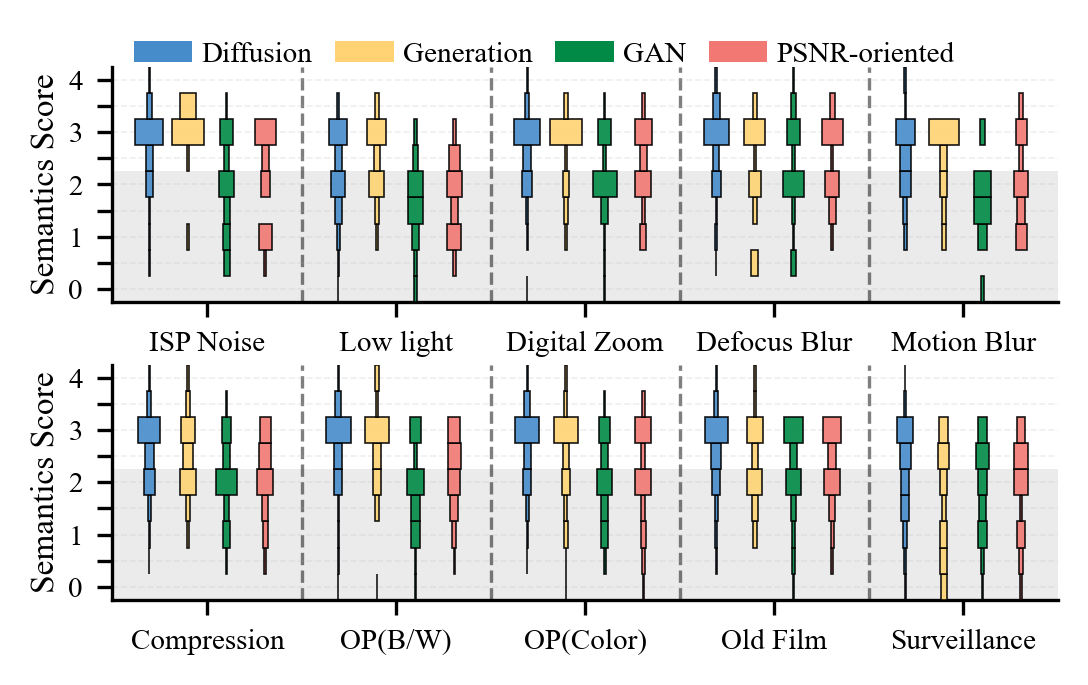}
        \caption{Distribution of sharpness scores across degradations.}
        \label{fig:semantics_deg_supp}
    \end{subfigure}
    \caption{
    Distribution of different scores across various degradation types. The horizontal width of each box indicates the percentage of samples within each score interval. The light gray region indicates low overall scores, representing generally unacceptable results. The red line indicates the balance point; scores above it (\textcolor{firebrick}{Over $\uparrow$}) denote over-generation, and scores below it (\textcolor{firebrick}{Less $\downarrow$}) denote under-generation.
    }
    \label{fig:distribution_degradation_supp}
\end{figure*}

We further provide score distributions for all remaining semantic categories that were not visualized in the main paper.
Results for Overall Detail, Sharpness, and Semantics are shown in~\cref{fig:scenes_detail_sharpness_semantics_supp,fig:scenes_total_supp}.
We also report complete score distributions across all degradation types in~\cref{fig:distribution_degradation_supp}.

\subsection{More Results of Parameter Configurations}
\begin{figure*}[tp]
\scriptsize
\centering
    \includegraphics[width=0.48\linewidth]{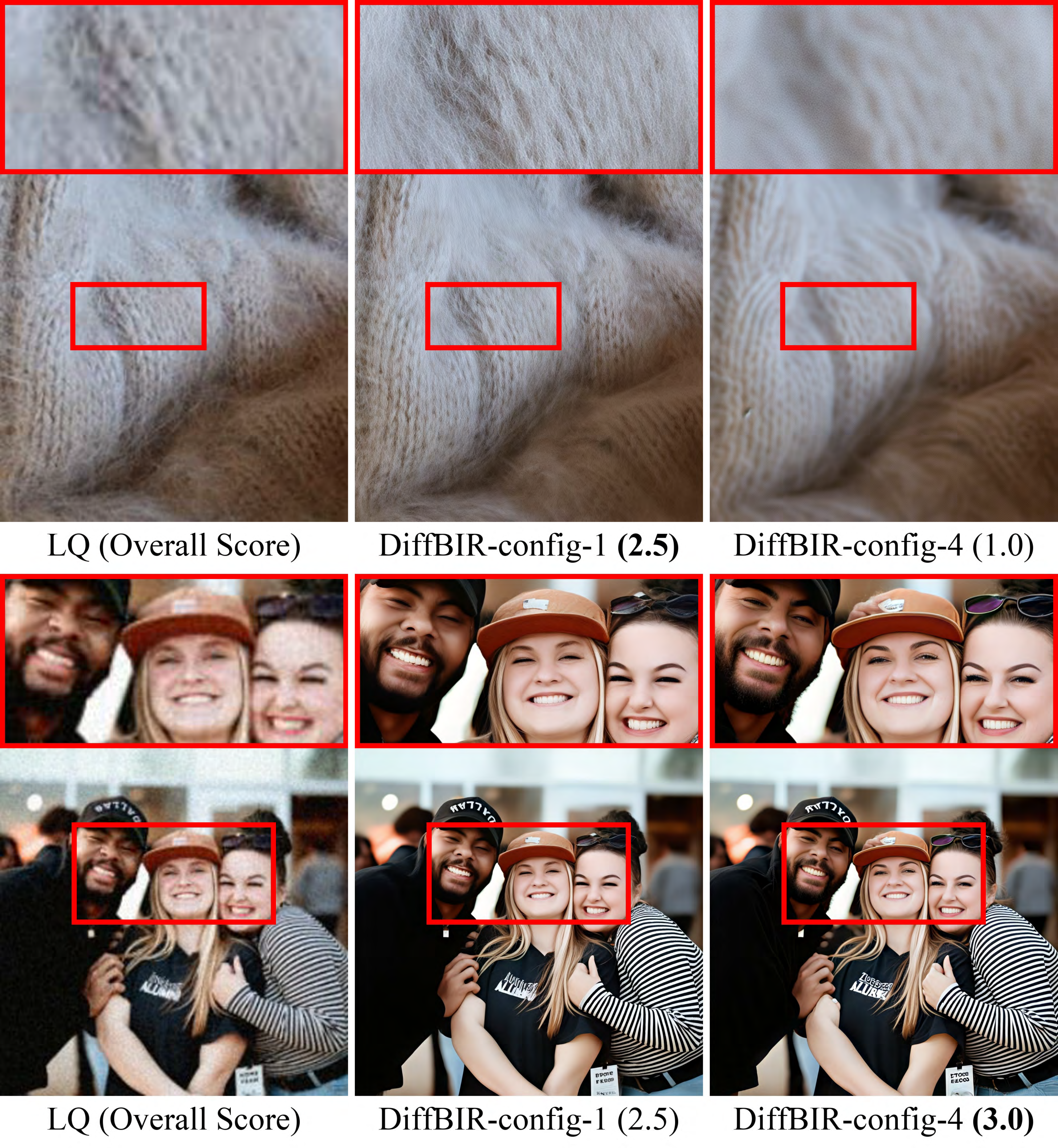}
    \hfill
    \includegraphics[width=0.48\linewidth]{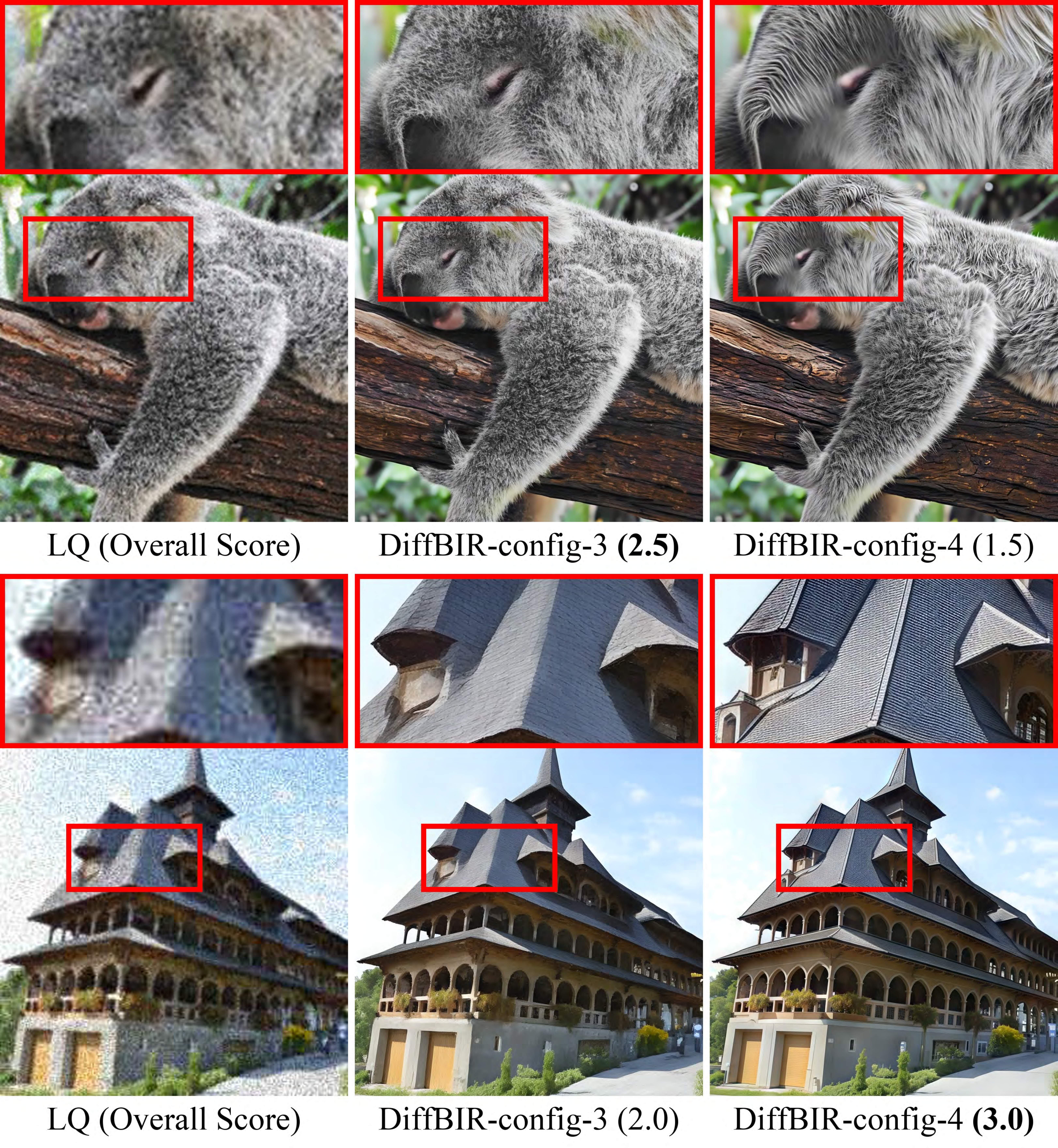}
    \hfill
    \includegraphics[width=0.48\linewidth]{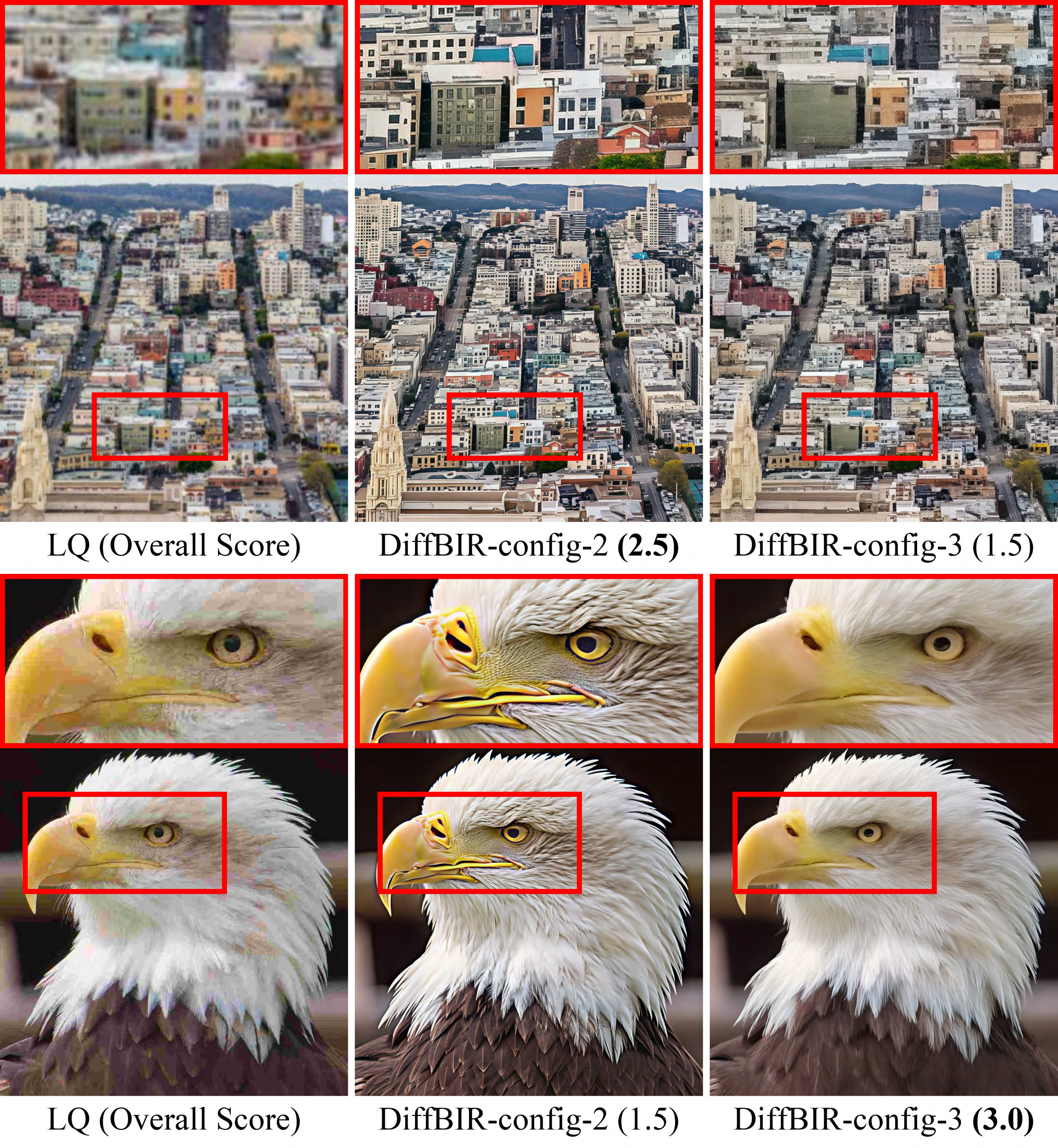}
    \hfill
    \includegraphics[width=0.48\linewidth]{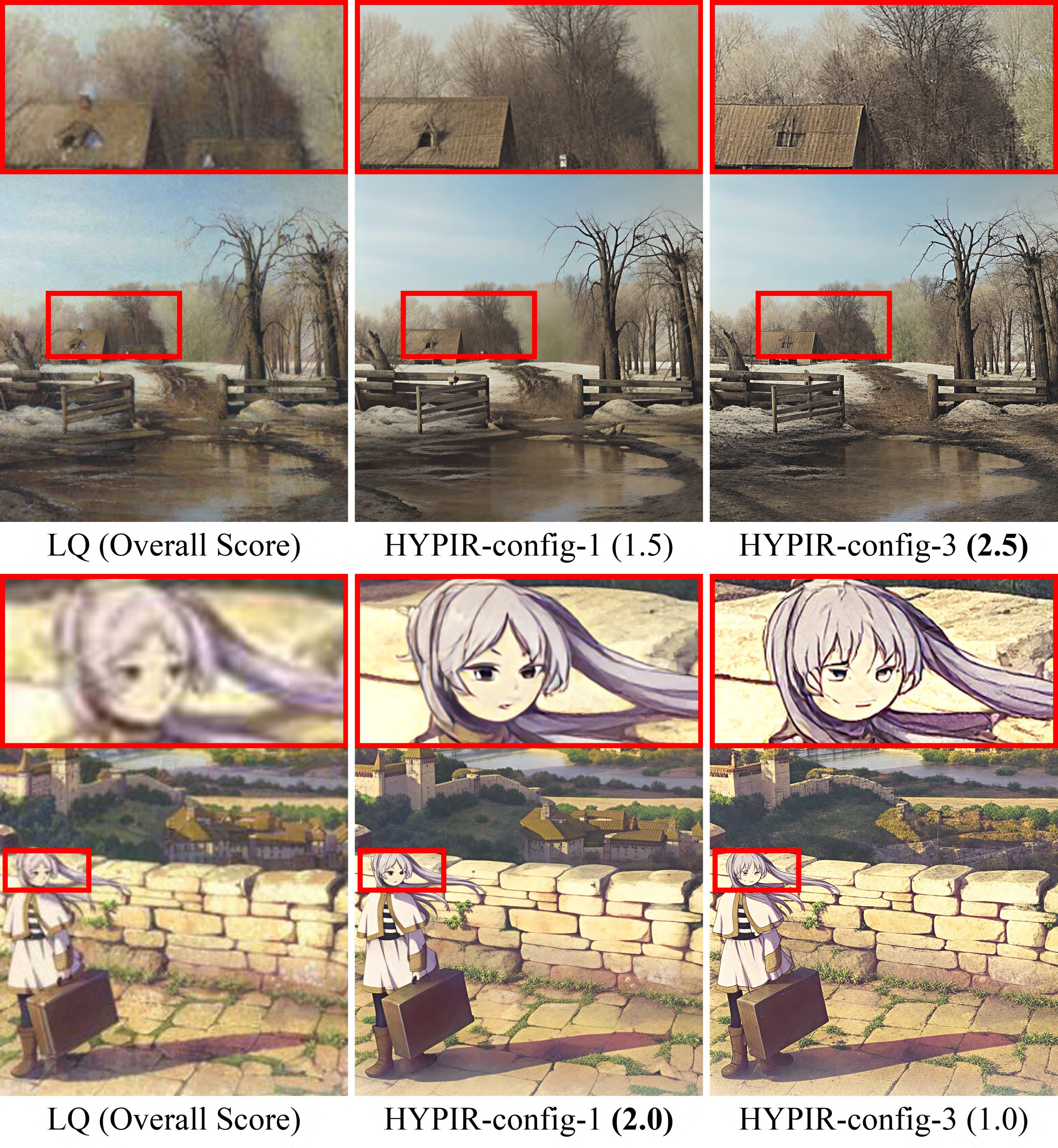}
    \caption{
    Effect of parameter configurations on restoration behavior in diffusion-based models. Zoom in for a better view.
    }
\label{fig:param_example_supp1}
\end{figure*}

\begin{figure*}[tp]
\scriptsize
\centering
    \includegraphics[width=0.48\linewidth]{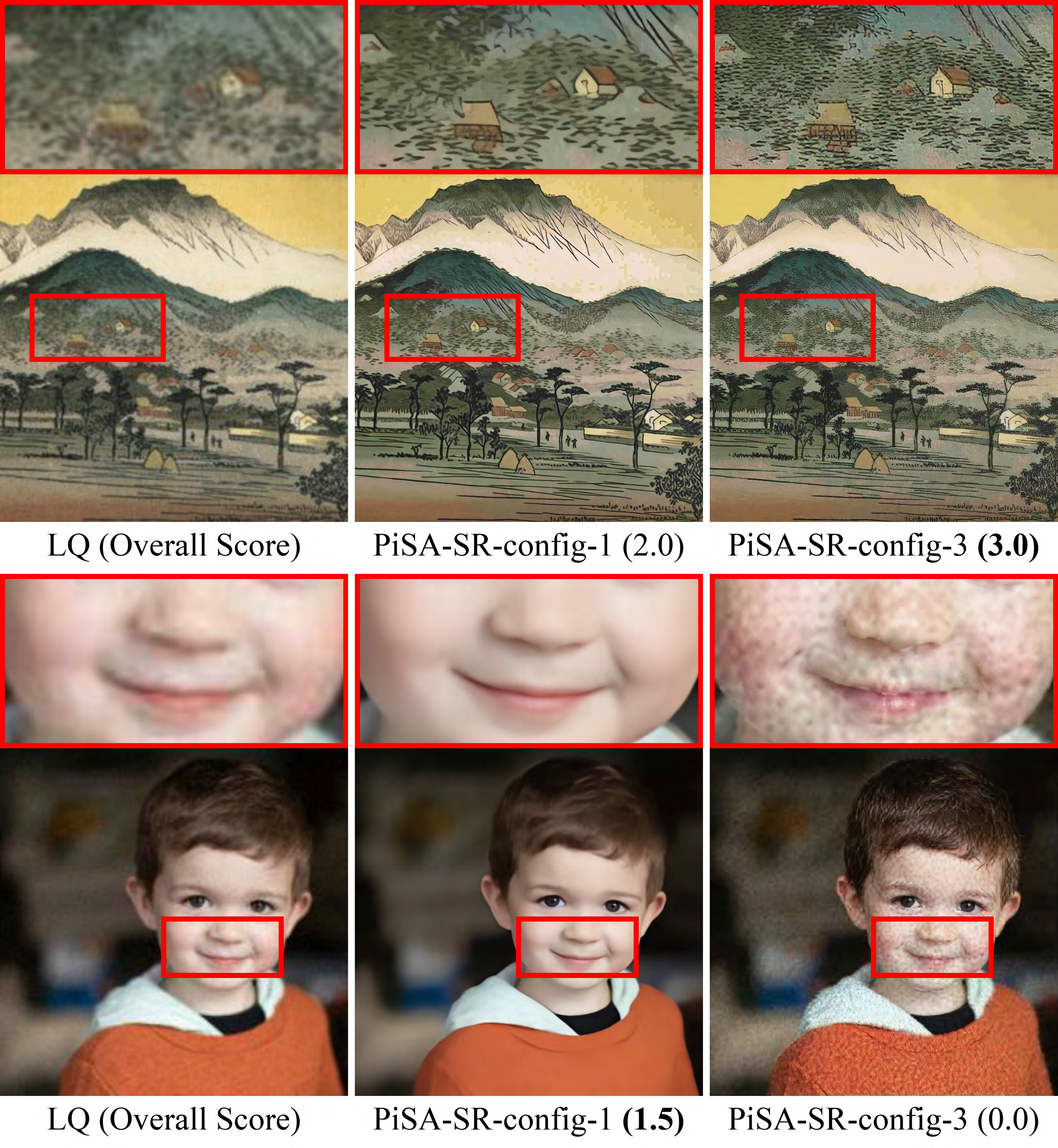}
    \hfill
    \includegraphics[width=0.48\linewidth]{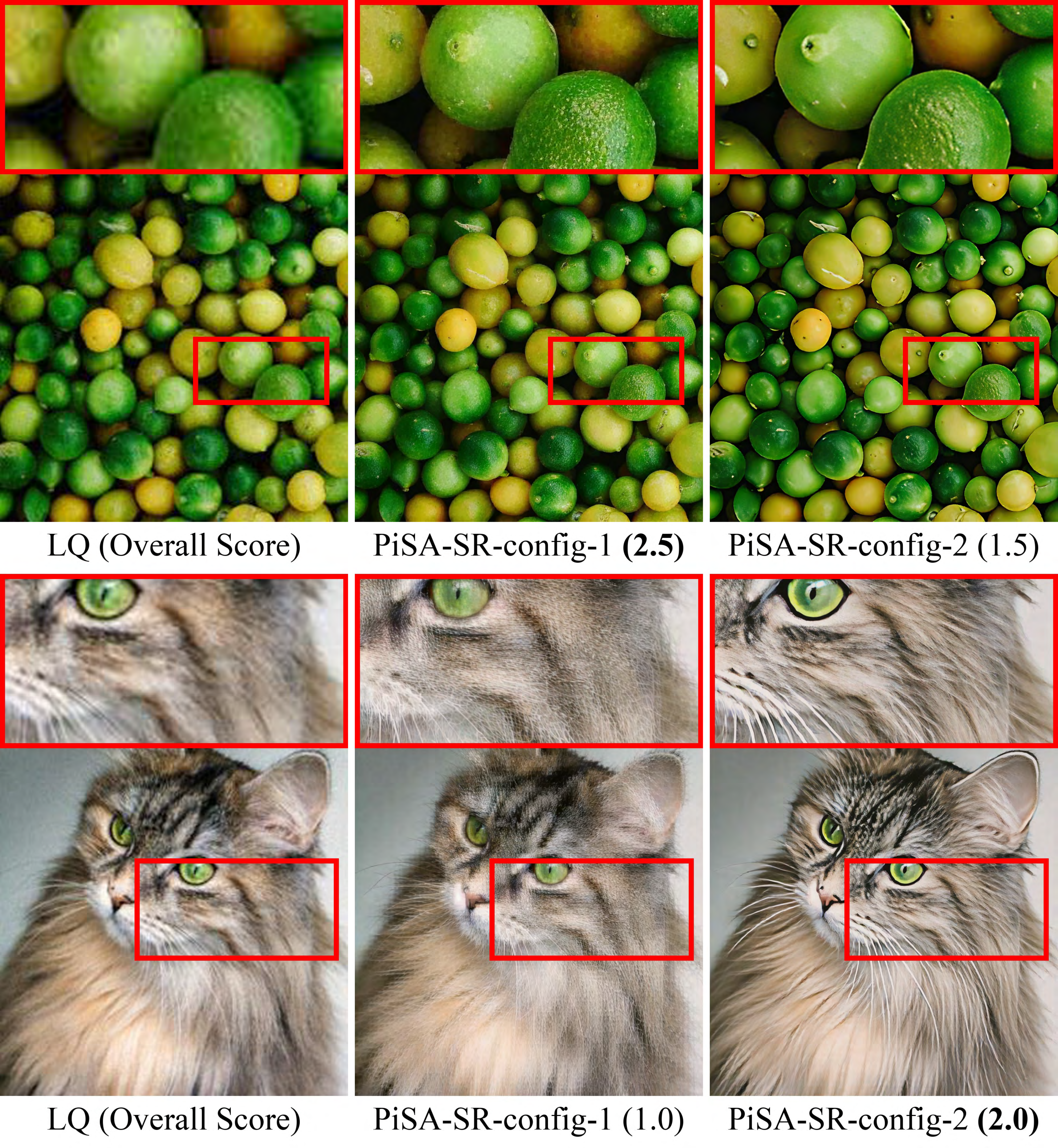}
    \hfill
    \includegraphics[width=0.48\linewidth]{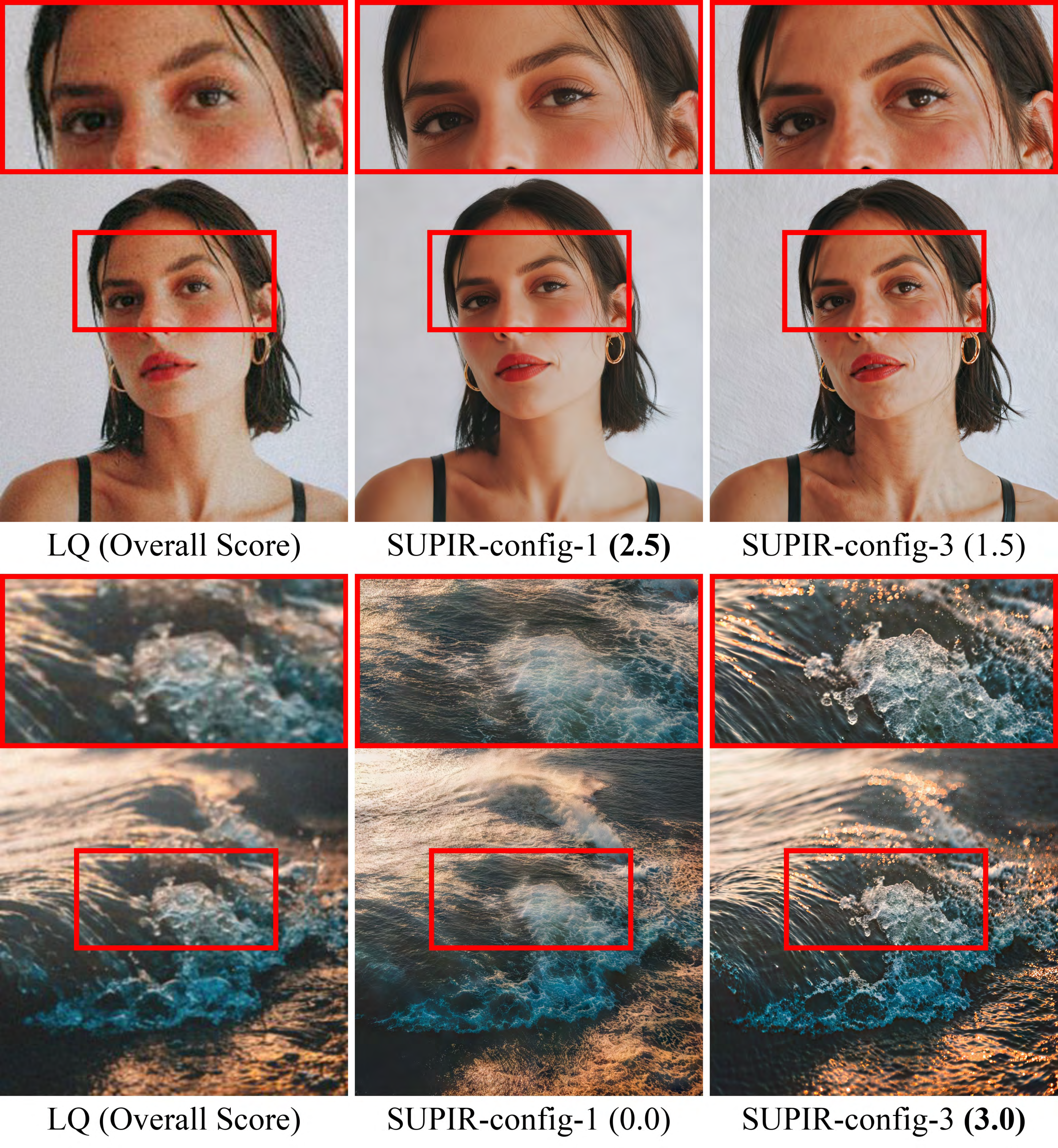}
    \hfill
    \includegraphics[width=0.48\linewidth]{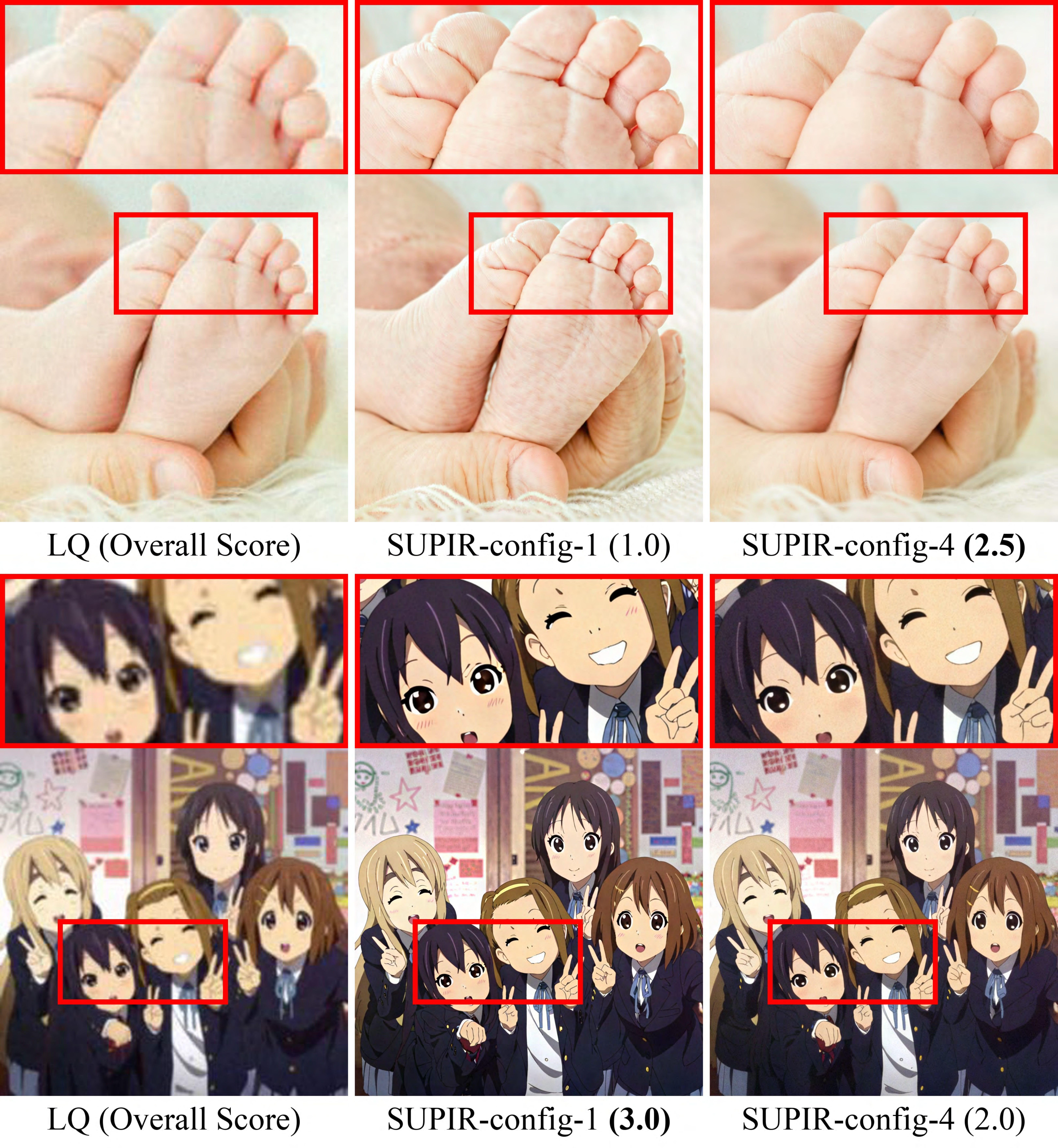}
    \caption{
    Effect of parameter configurations on restoration behavior in diffusion-based models. Zoom in for a better view.
    }
\label{fig:param_example_supp2}
\end{figure*}

To support the parameter sensitivity study in the main paper, we evaluate multiple inference configurations for several diffusion-based GIR models.
These configurations, as shown in~\cref{tab:param_setting} are chosen to intentionally vary the level of fidelity and generation, allowing us to obtain a richer and more diverse set of restoration outcomes.
We further report the performance of each configuration across different semantic scenes and degradation types, as summarized in~\cref{tab:quantity_params}.
Qualitative comparisons illustrating how different parameters affect restoration behavior can be found in~\cref{fig:param_example_supp1,fig:param_example_supp2}.
All configuration identifiers exactly match those reported in the main paper.

\section{Qualitative Analysis of IQA}
\label{sec:iqa}
To complement the quantitative results in the main paper, we provide qualitative examples illustrating the limitations of existing IQA methods when evaluating generative image restoration.
Although recent learning-based IQA models achieve strong performance on traditional distortions, they remain unreliable for GIR scenarios, particularly when semantic inconsistencies or generative artifacts are present.

\begin{figure*}[tp]
\scriptsize
\centering
    \includegraphics[width=0.48\linewidth]{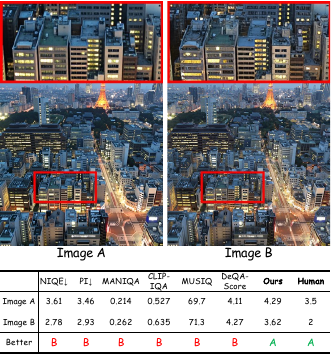}
    \hfill
   \includegraphics[width=0.48\linewidth]{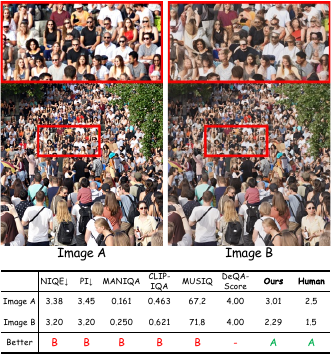}
   \hfill
   \includegraphics[width=0.48\linewidth]{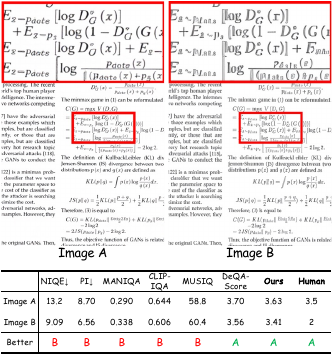}
   \hfill
   \includegraphics[width=0.48\linewidth]{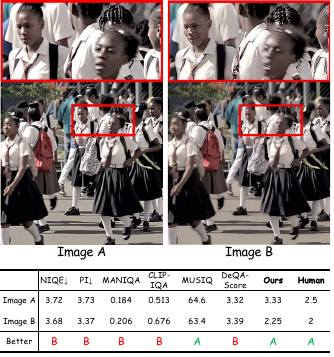}
   
    \caption{
    Qualitative analysis of image quality assessment methods. Zoom in for a better view.
    }
    \label{fig:iqa_examples1}
\end{figure*}

\begin{figure*}[tp]
\scriptsize
\centering
    \includegraphics[width=0.48\linewidth]{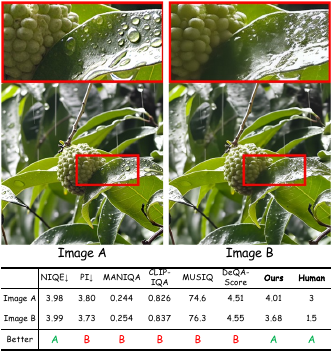}
    \hfill
   \includegraphics[width=0.48\linewidth]{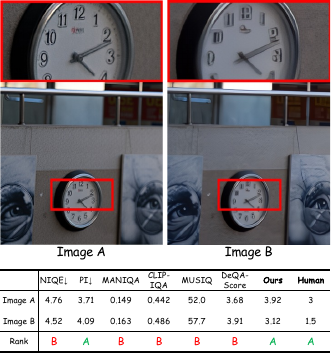}
   \hfill
   \includegraphics[width=0.48\linewidth]{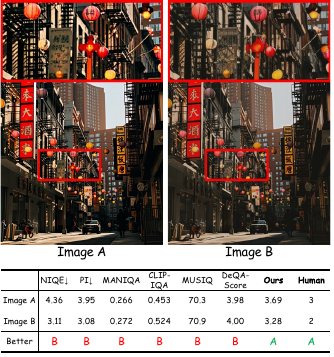}
   \hfill
   \includegraphics[width=0.48\linewidth]{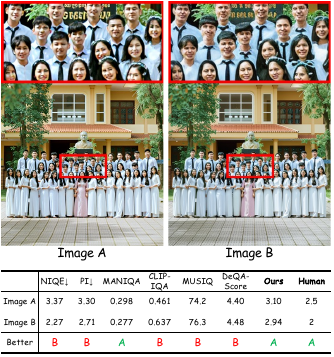}
   
    \caption{
    Qualitative analysis of image quality assessment methods. Zoom in for a better view.
    }
    \label{fig:iqa_examples2}
\end{figure*}

\begin{figure*}[tp]
\scriptsize
\centering
    \includegraphics[width=0.48\linewidth]{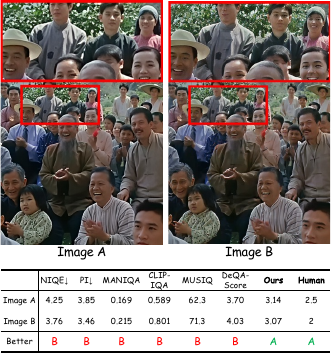}
    \hfill
   \includegraphics[width=0.48\linewidth]{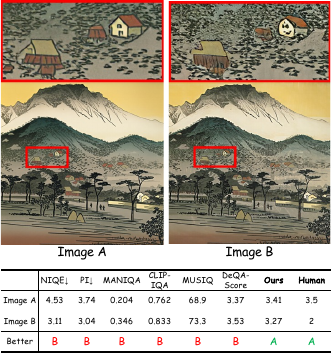}
   \hfill

    \caption{
    Qualitative analysis of image quality assessment methods. Zoom in for a better view.
    }
    \label{fig:iqa_examples3}
\end{figure*}

In the~\cref{fig:iqa_examples1,fig:iqa_examples2,fig:iqa_examples3}, each case includes two restored outputs produced by different GIR models.
For each restored image, we report the human-annotated overall scores, along with the predictions from several representative IQA models and our IQA model.
These examples provide side-by-side comparisons of how IQA models and human annotators evaluate the same pair of restored images, allowing to directly observe the differences in their judgments.

\section{More Failure Cases}
\label{sec:failure_cases}
\begin{figure*}[tp]
\scriptsize
\centering
   \includegraphics[width=0.48\linewidth]{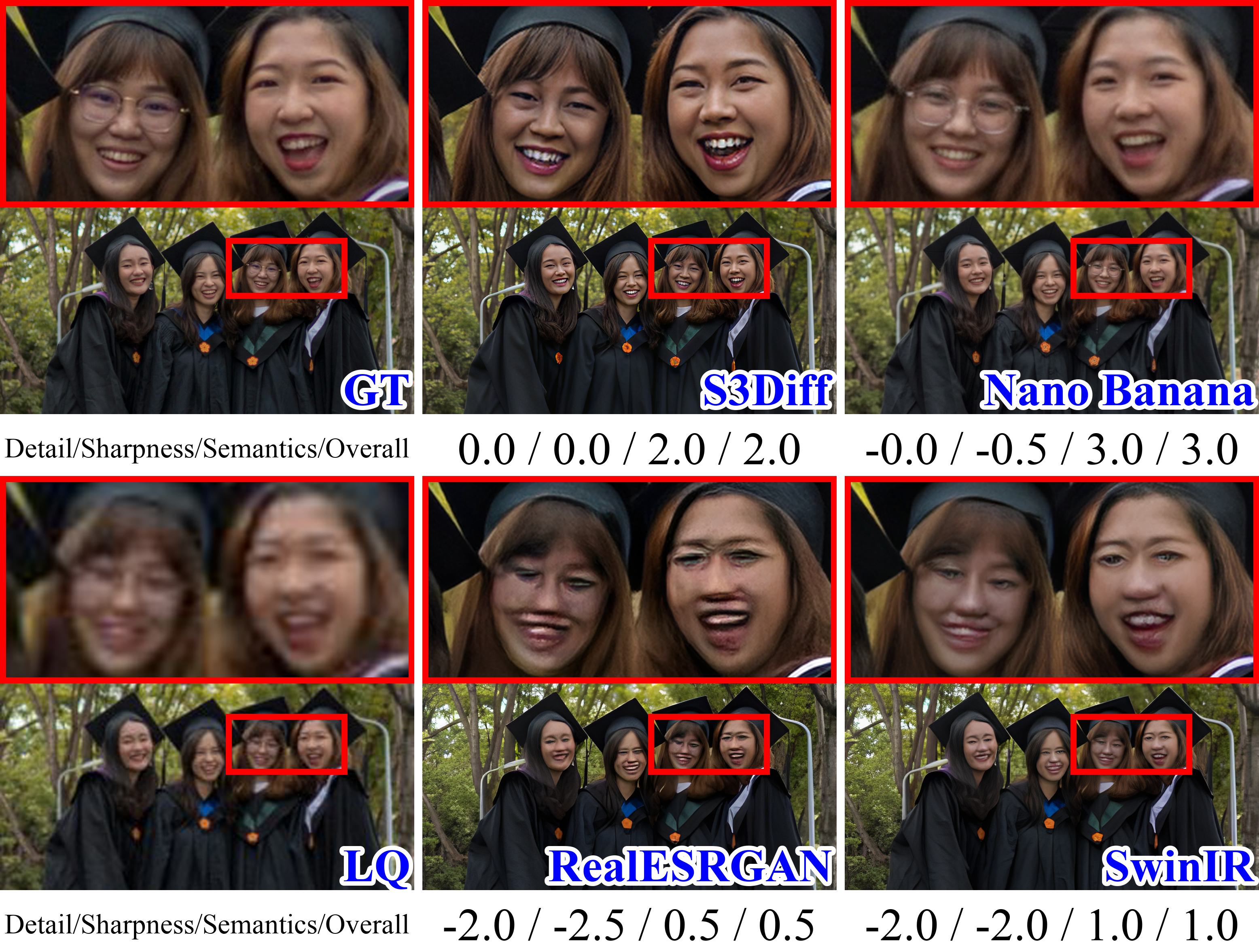}
   \hfill
   \includegraphics[width=0.48\linewidth]{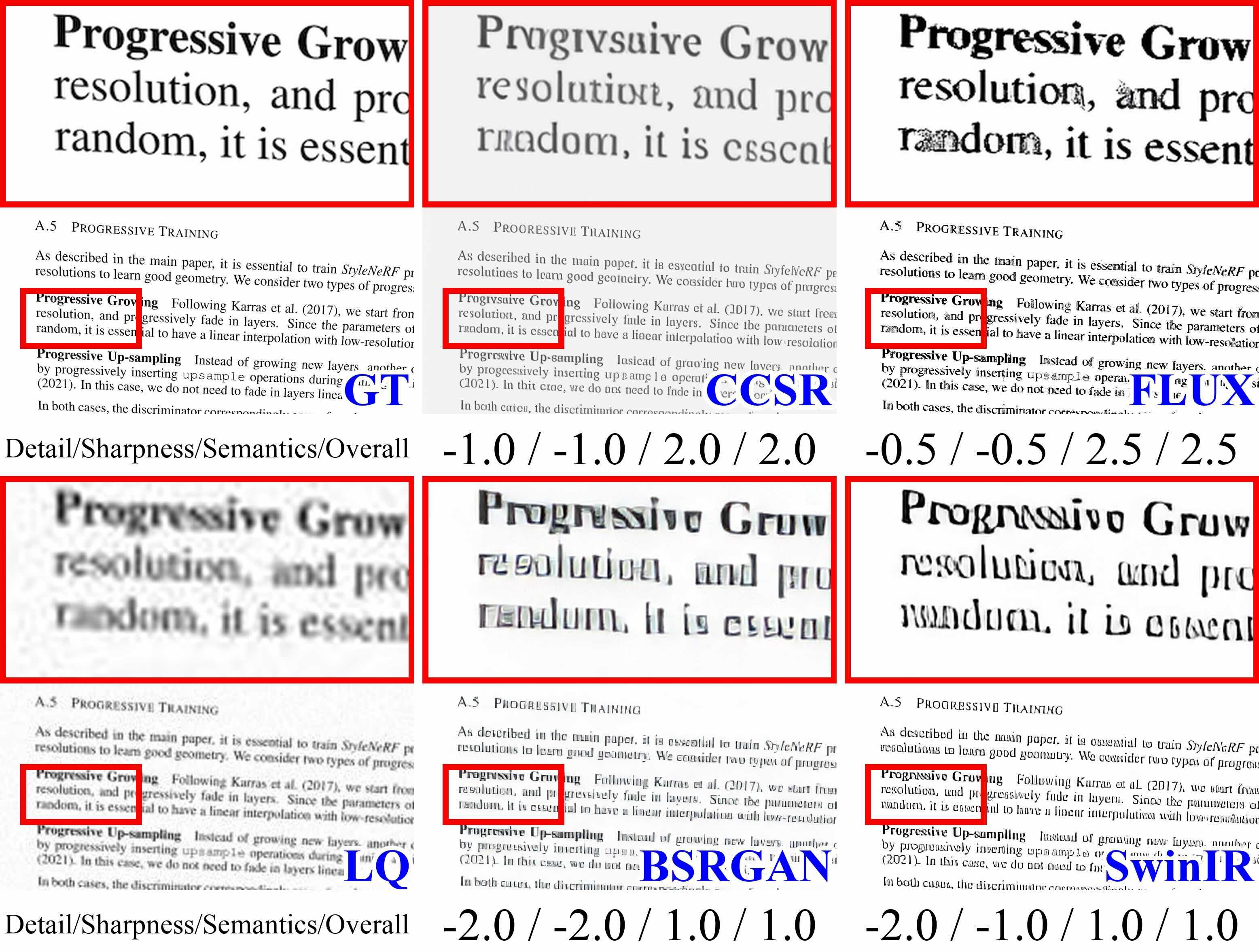}
    \caption{
    Failure cases of semantic errors. Zoom in for a better view.
    }
    \label{fig:failure_cases_supp3}
\end{figure*}

\begin{figure*}[tp]
\scriptsize
\centering
   \includegraphics[width=0.48\linewidth]{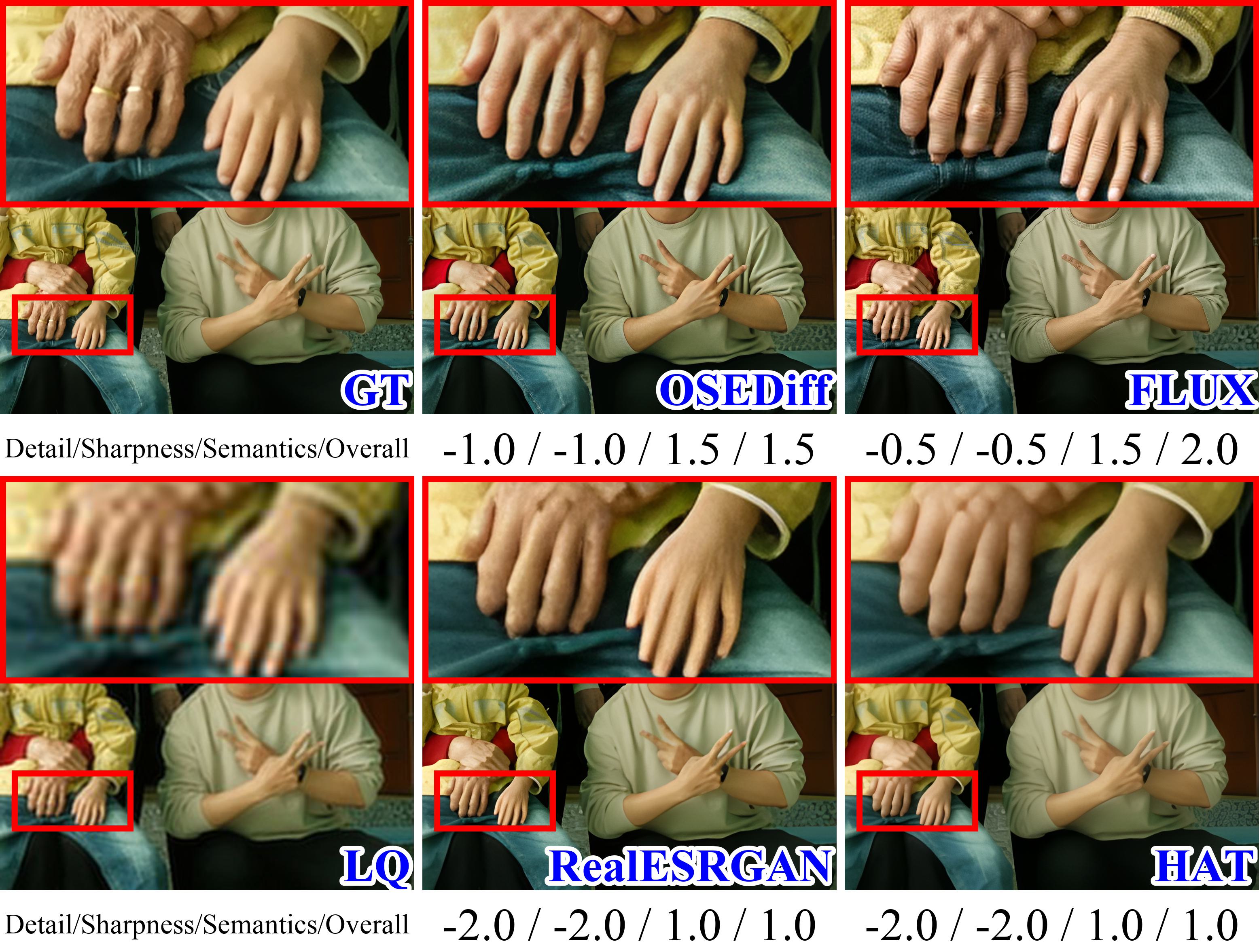}
   \hfill
   \includegraphics[width=0.48\linewidth]{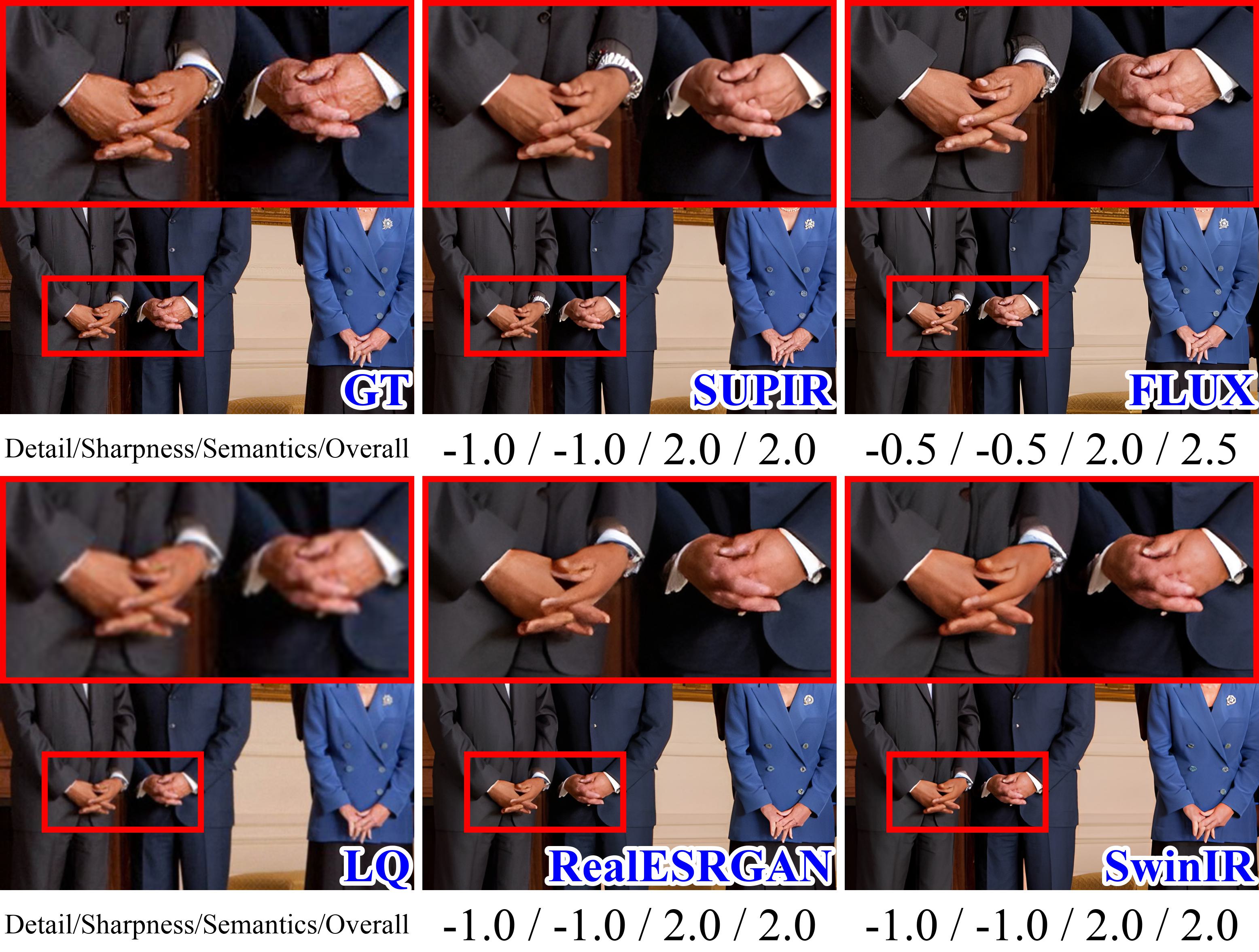}
   \hfill
   \includegraphics[width=0.48\linewidth]{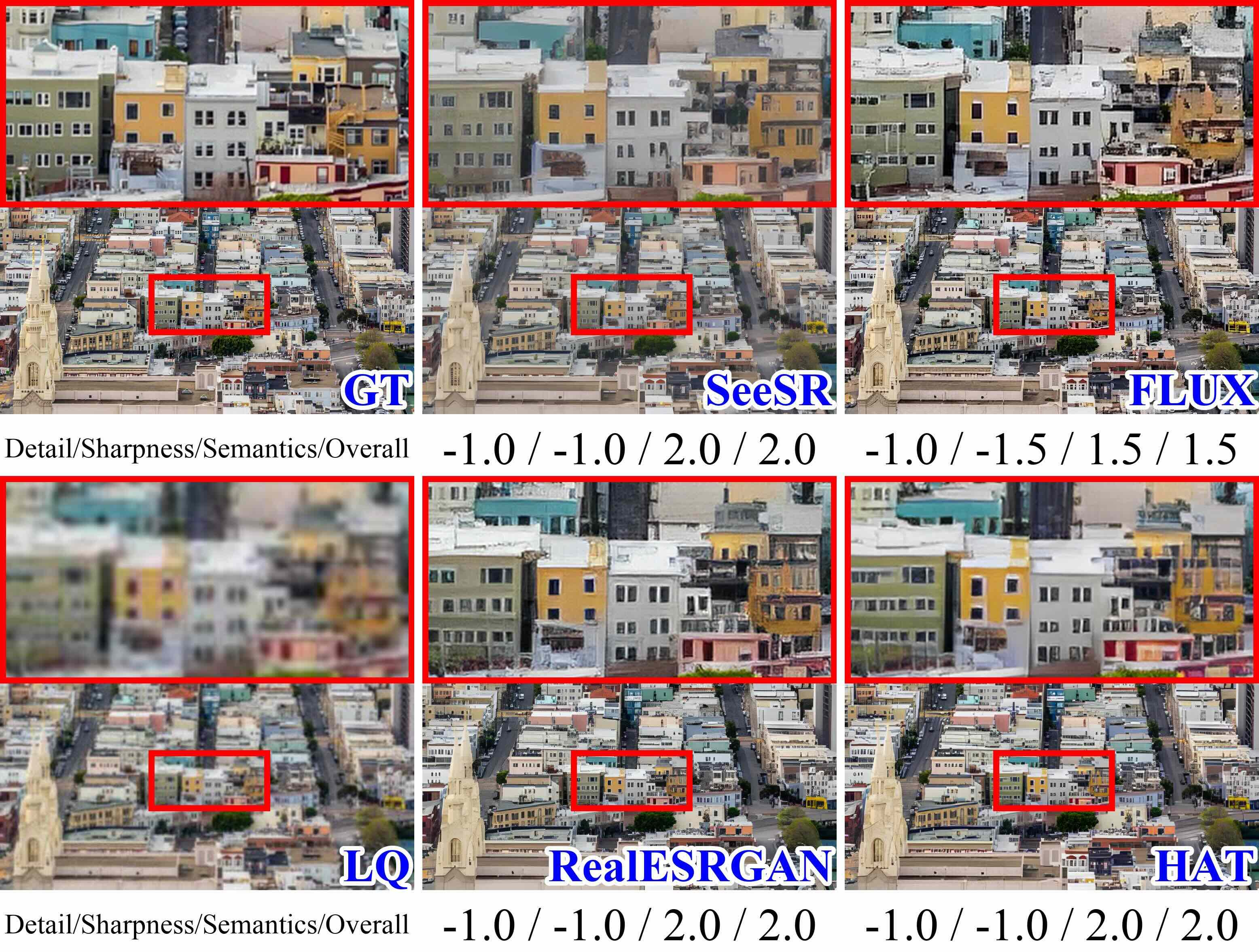}
   \hfill
   \includegraphics[width=0.48\linewidth]{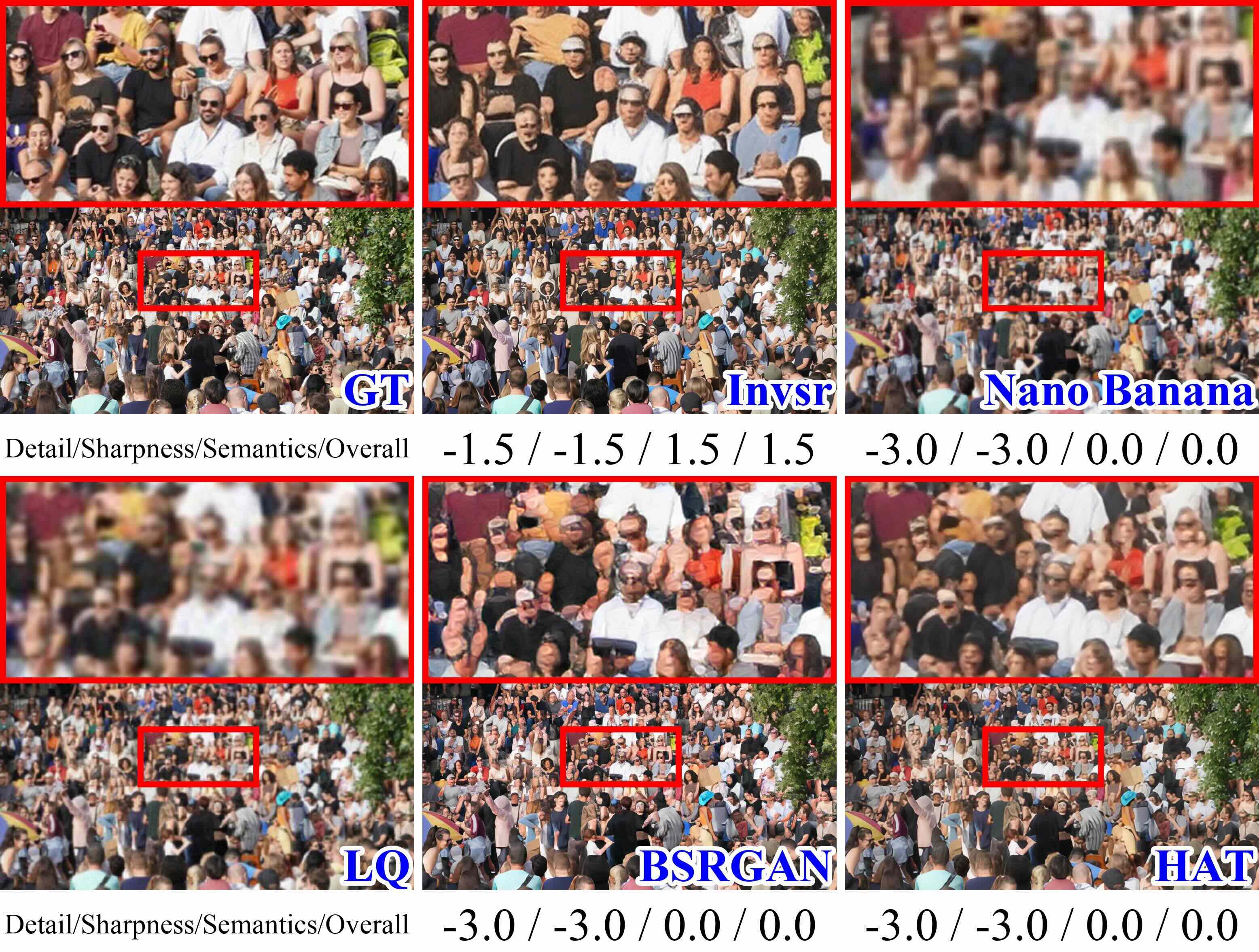}
   \hfill
   \includegraphics[width=0.48\linewidth]{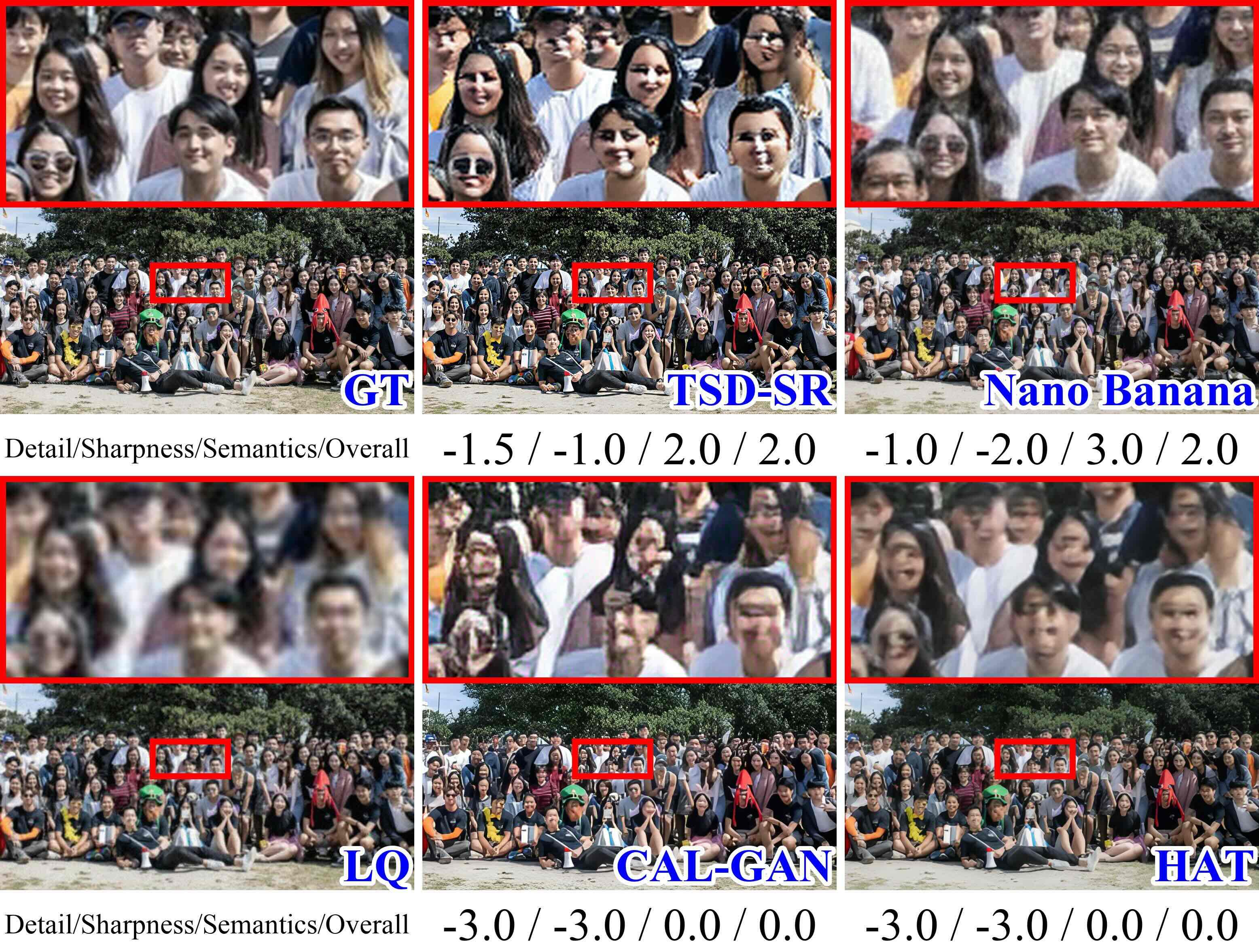}
   \hfill
   \includegraphics[width=0.48\linewidth]{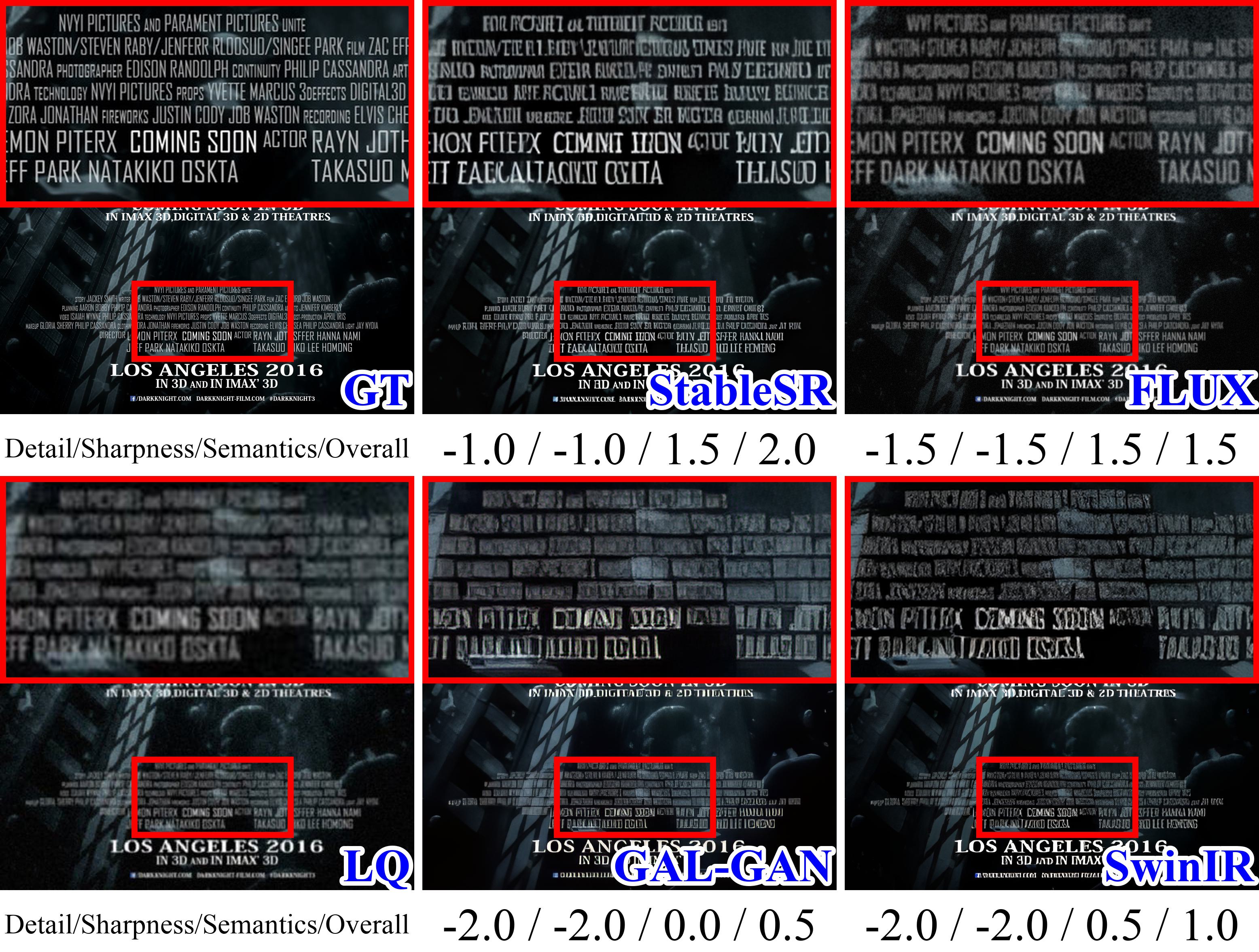}
    \caption{
    Failure cases of semantic errors. Zoom in for a better view.
    }
    \label{fig:failure_cases_supp2}
\end{figure*}

\begin{figure*}[tp]
\scriptsize
\centering
    \includegraphics[width=0.48\linewidth]{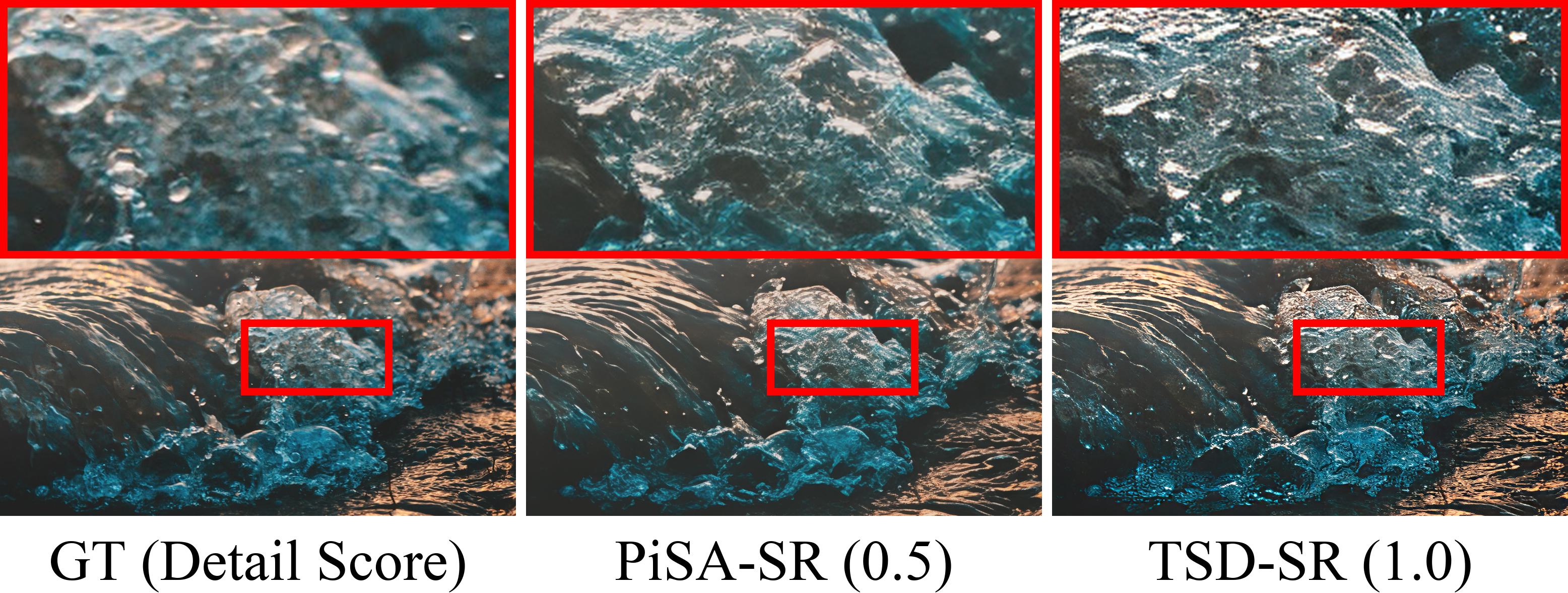}
   \hfill
   \includegraphics[width=0.48\linewidth]{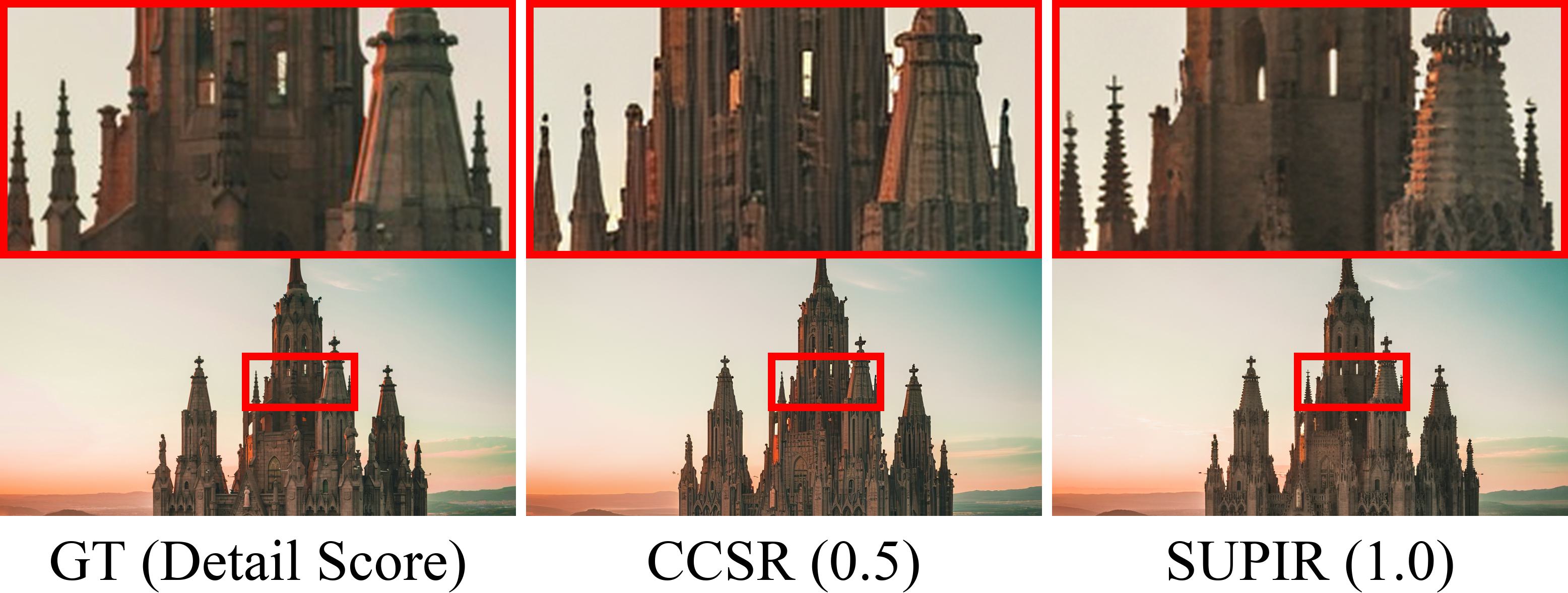}
   \hfill
   \includegraphics[width=0.48\linewidth]{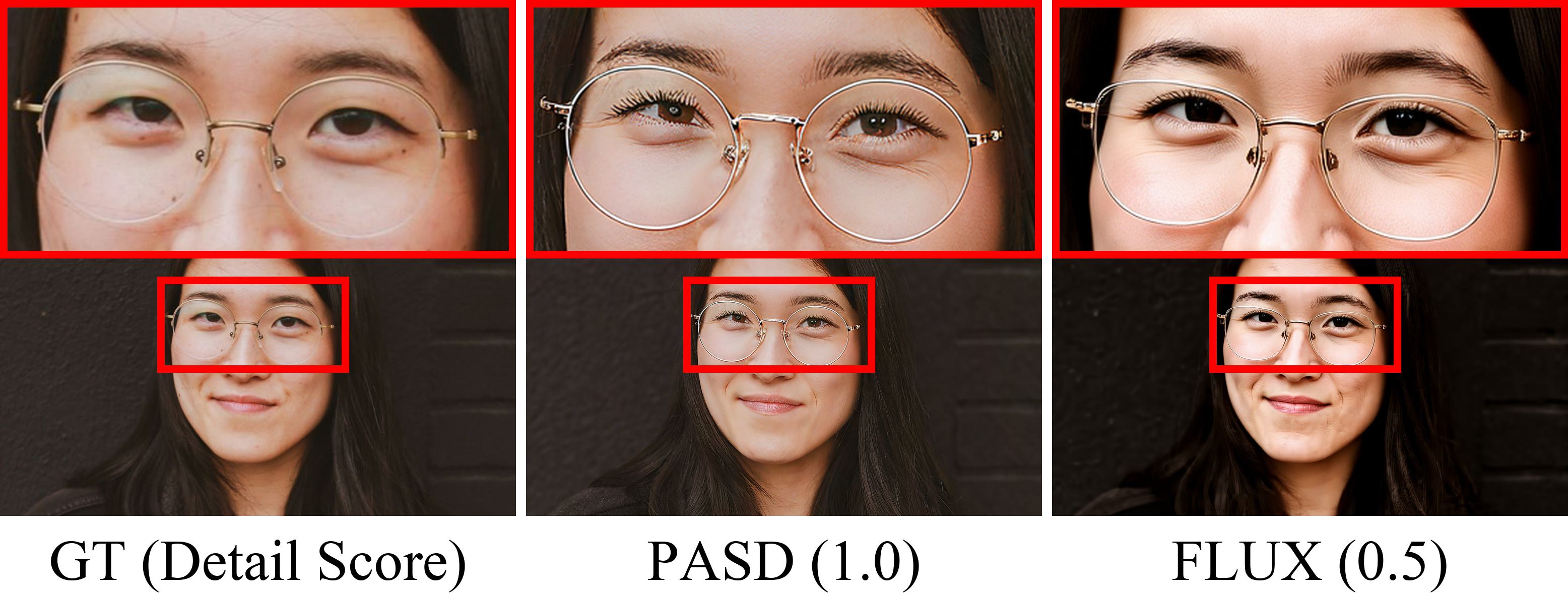}
  \hfill
   \includegraphics[width=0.48\linewidth]{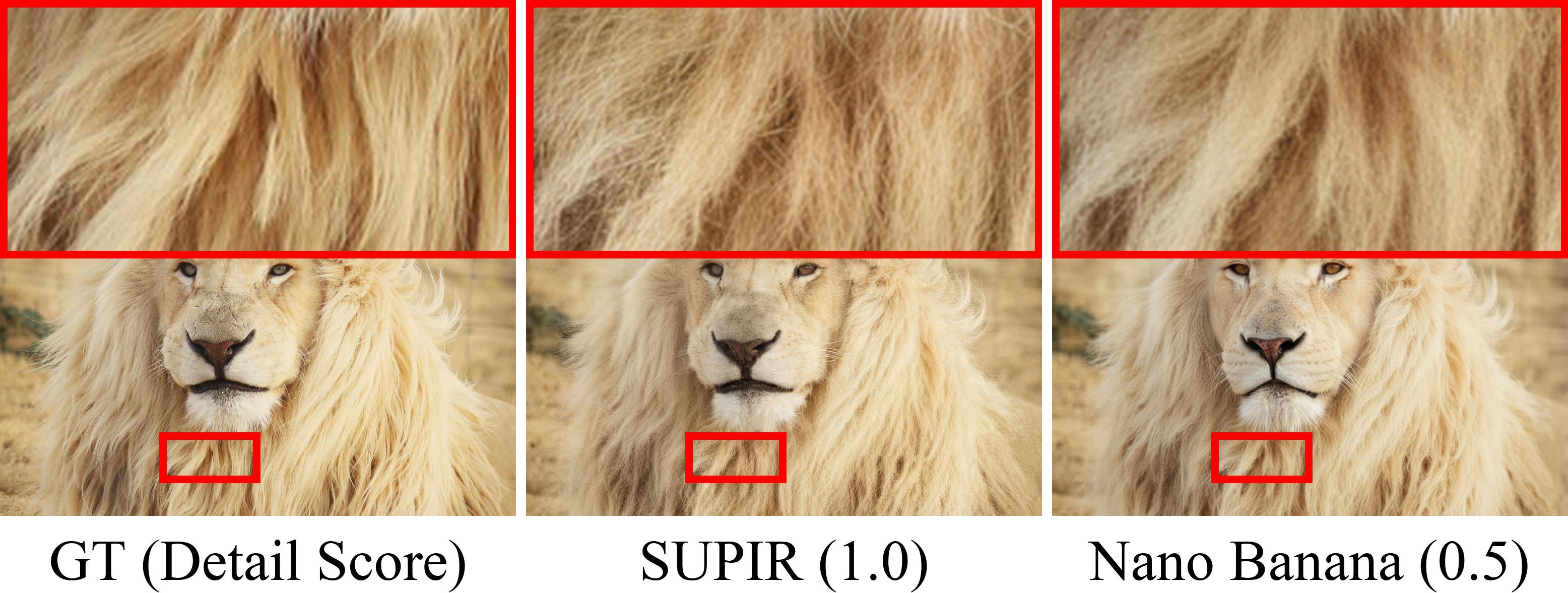}
   \hfill
   \includegraphics[width=0.32\linewidth]{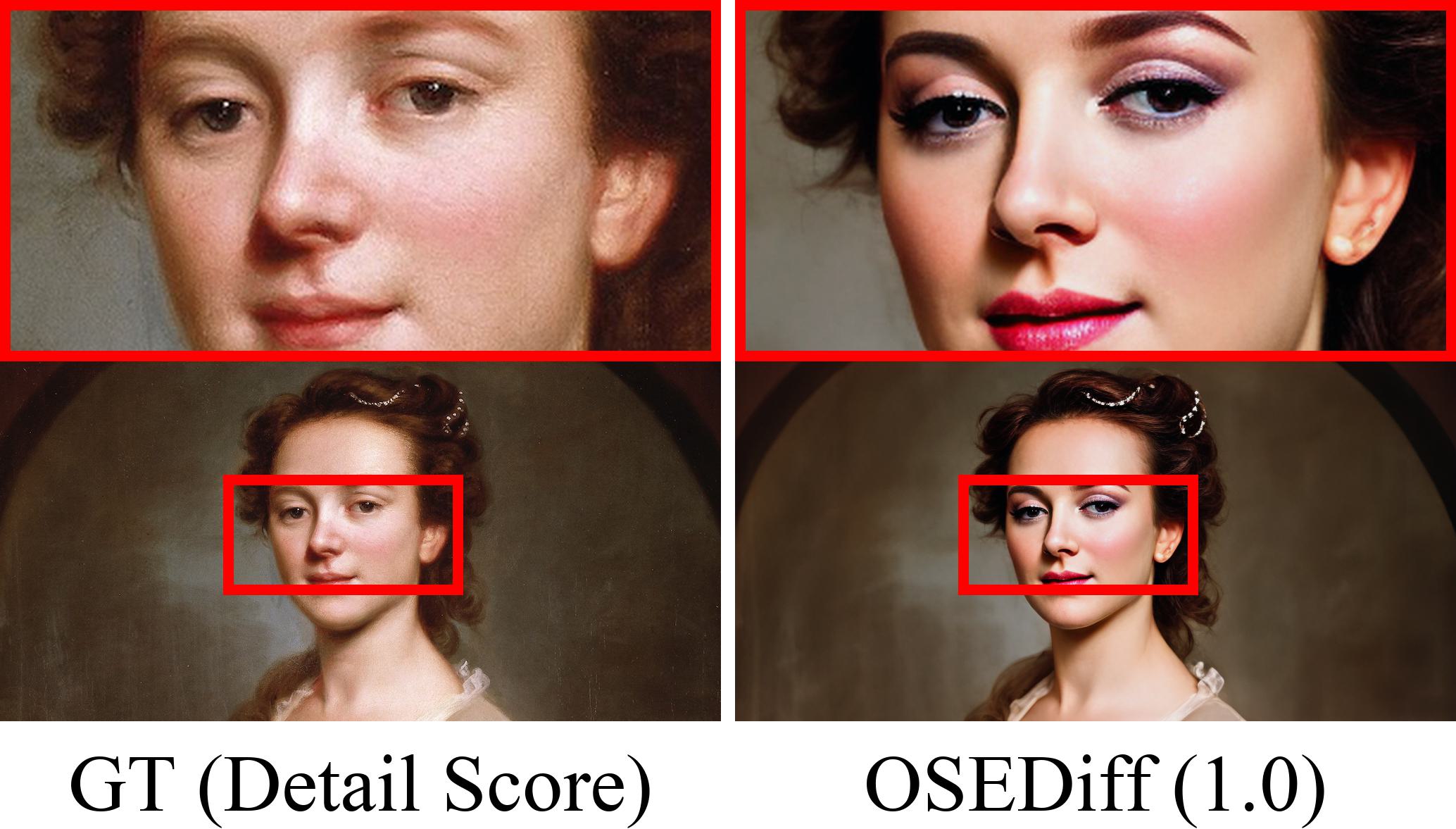}
   \hfill
   \includegraphics[width=0.32\linewidth]{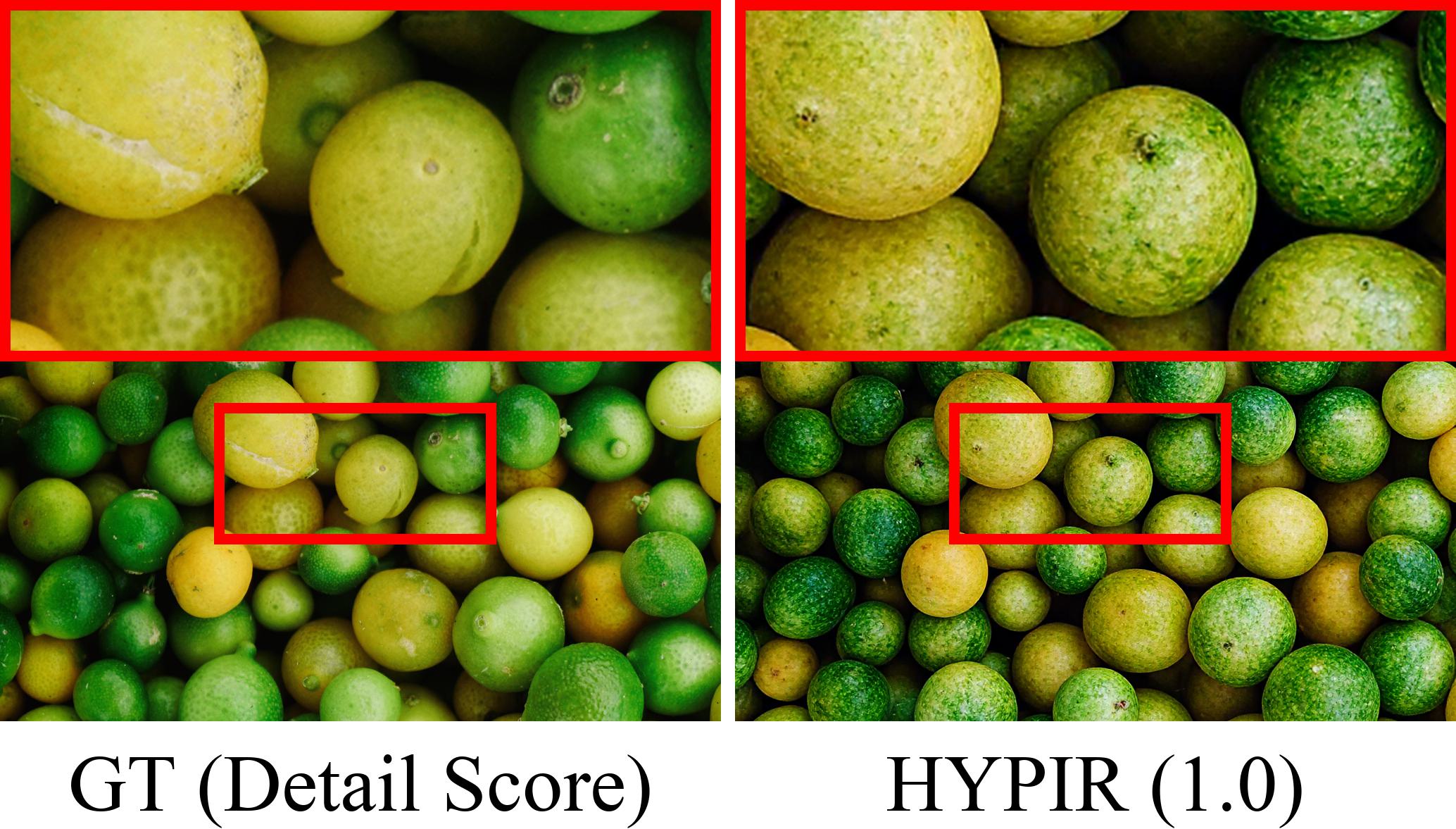}
   \hfill
   \includegraphics[width=0.32\linewidth]{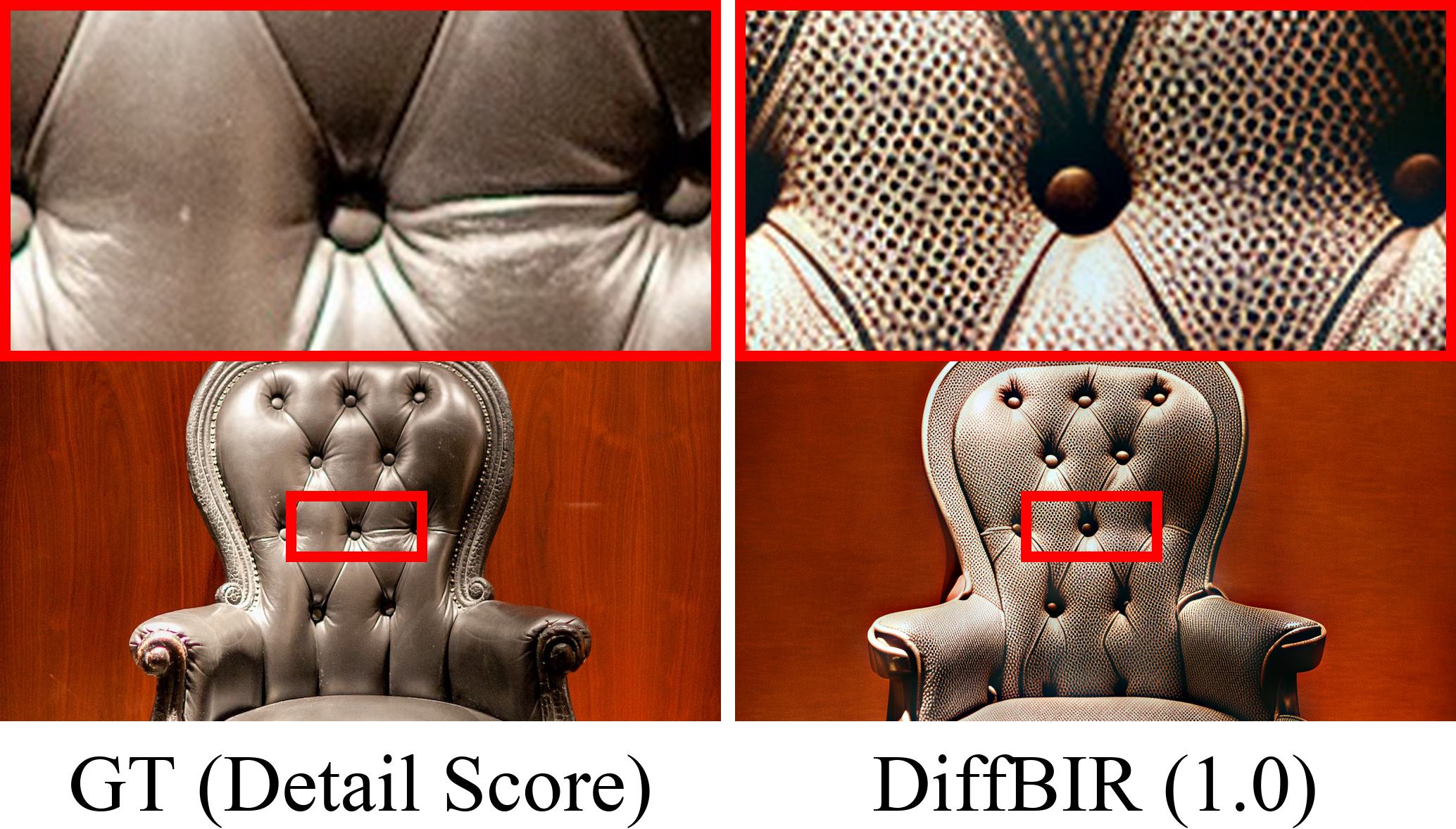}
   \hfill
   \includegraphics[width=0.32\linewidth]{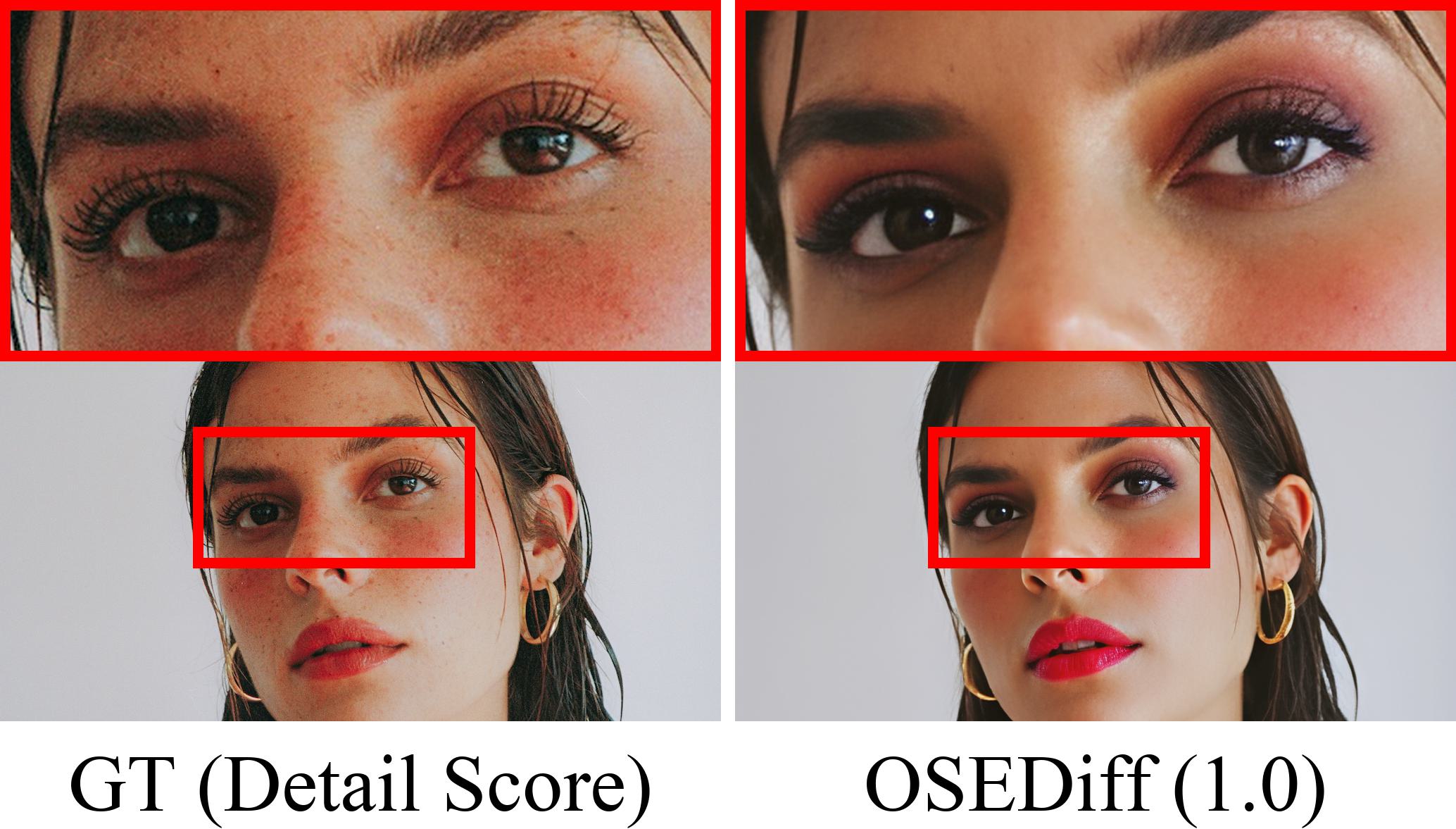}
   \hfill
   \includegraphics[width=0.32\linewidth]{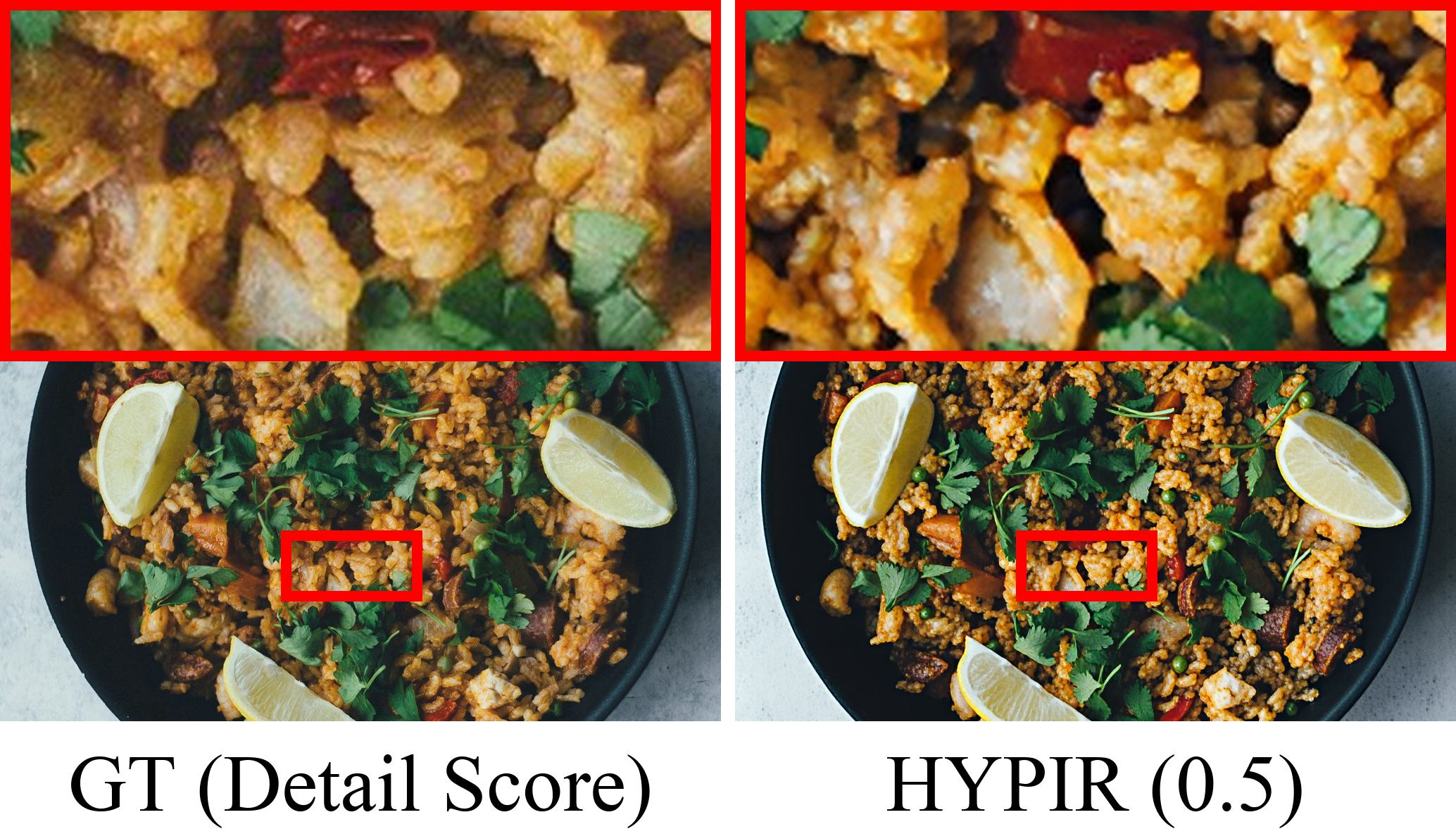}
   \hfill
   \includegraphics[width=0.32\linewidth]{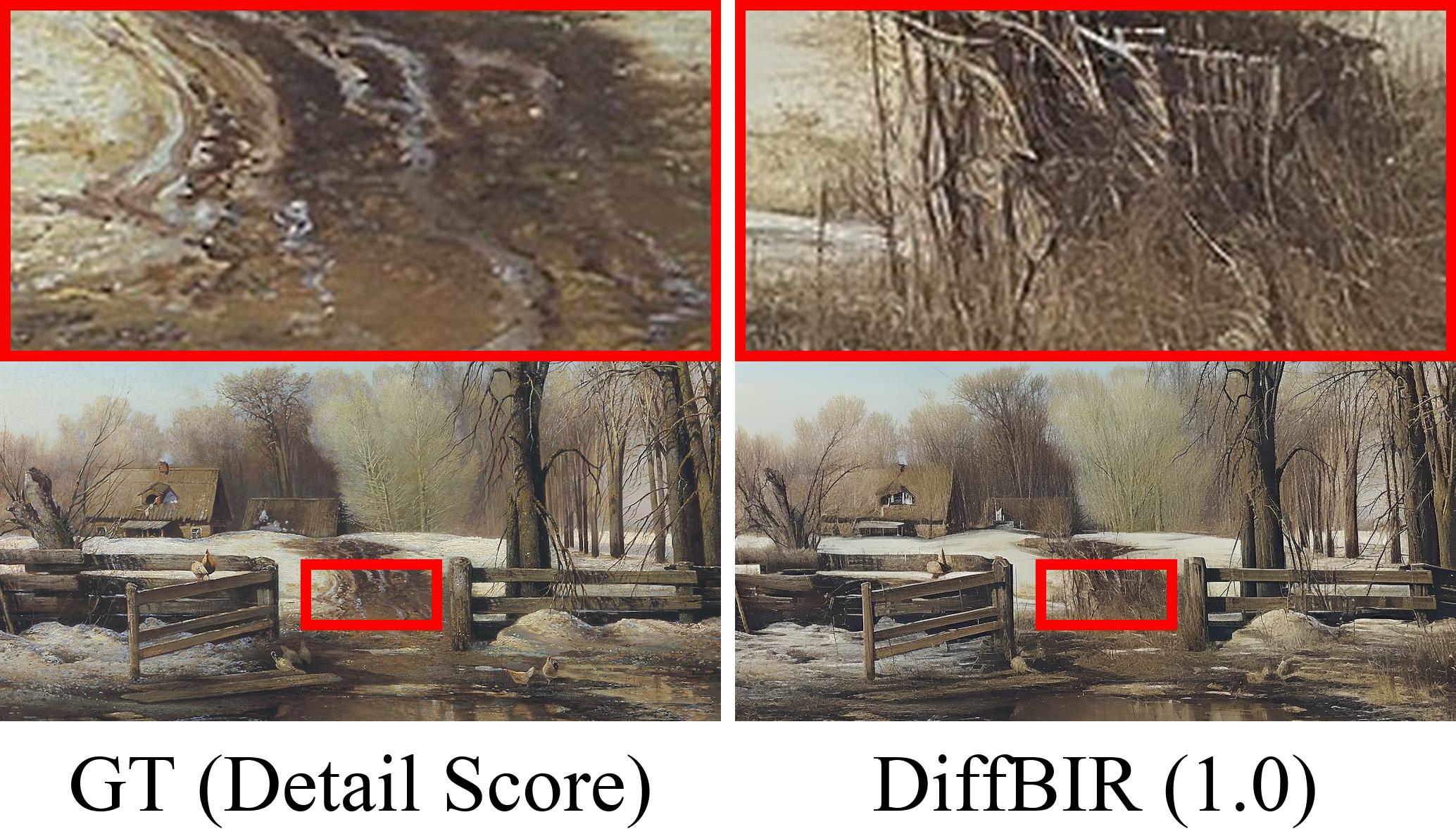}
   \hfill
   \includegraphics[width=0.32\linewidth]{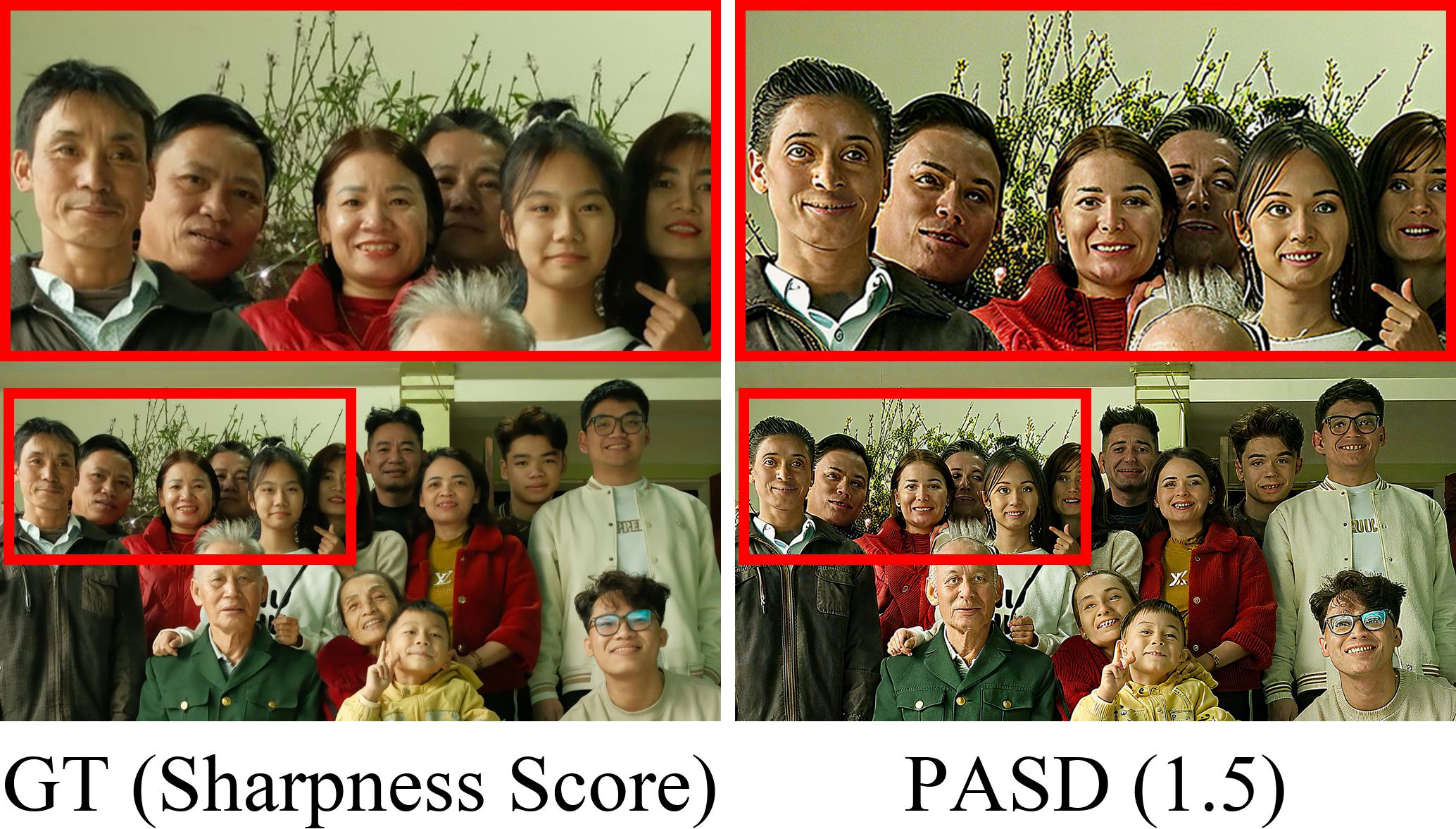}
   \hfill
   \includegraphics[width=0.32\linewidth]{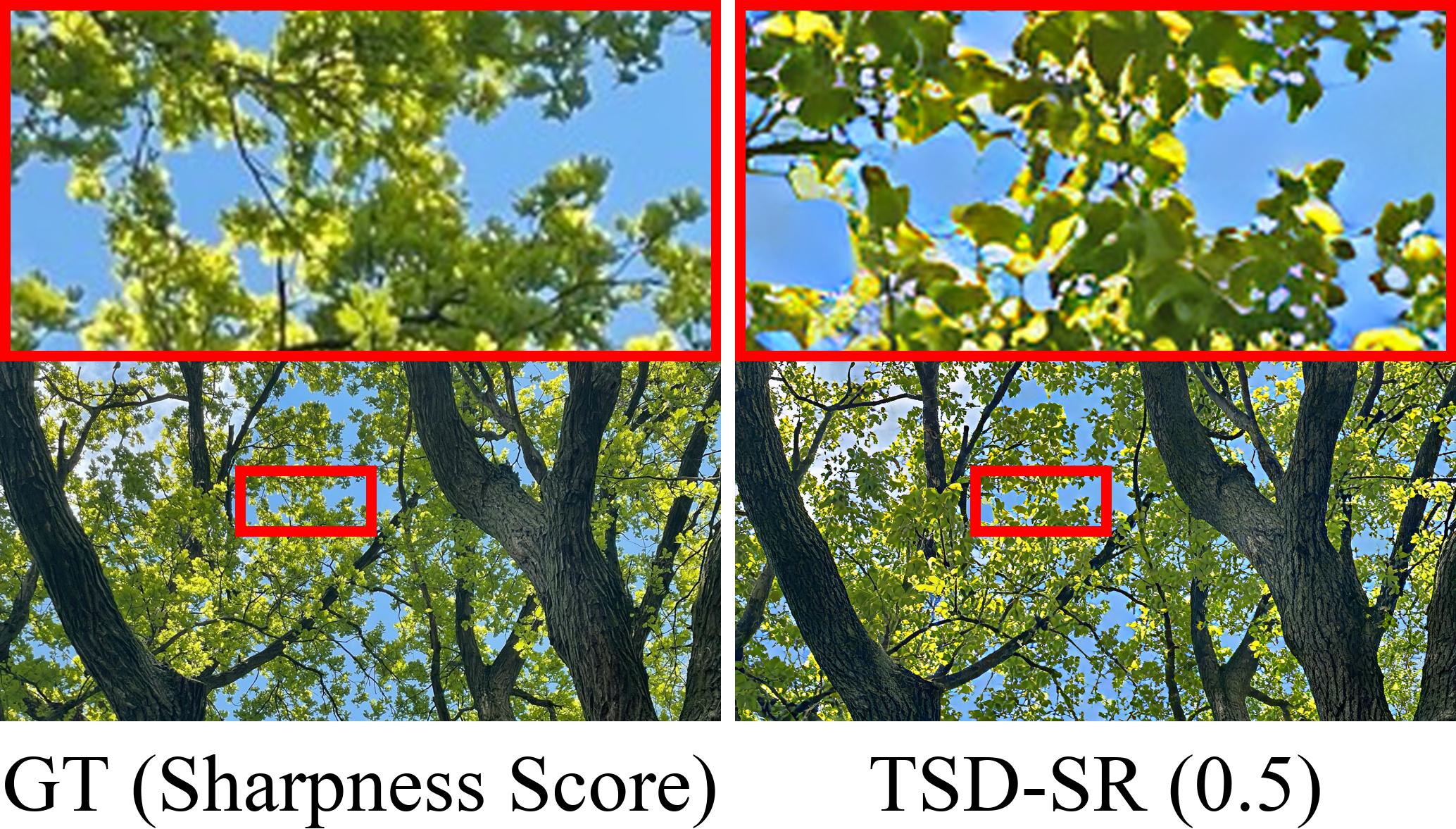}
   \hfill
   \includegraphics[width=0.32\linewidth]{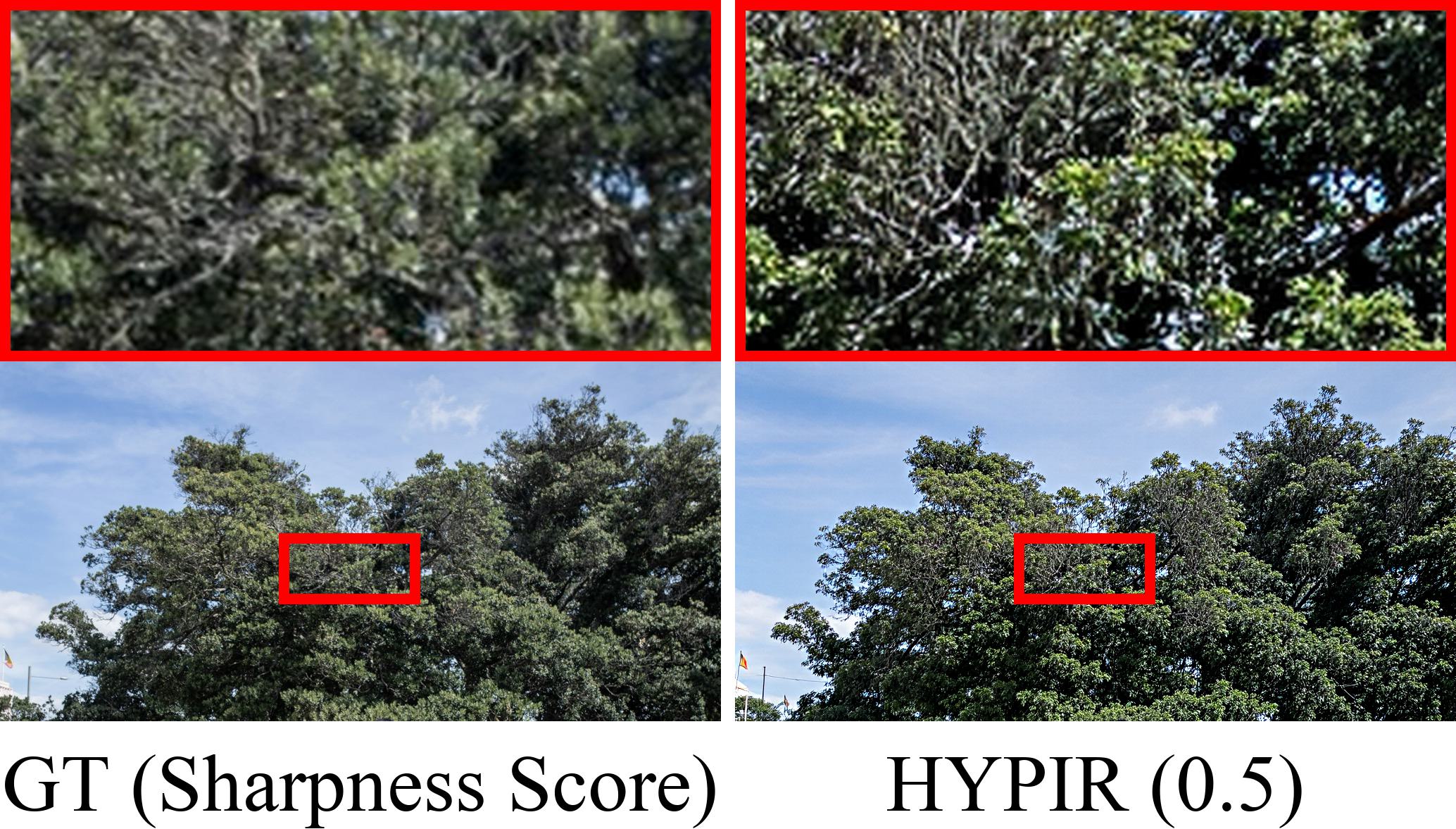}
   \hfill
   \includegraphics[width=0.32\linewidth]{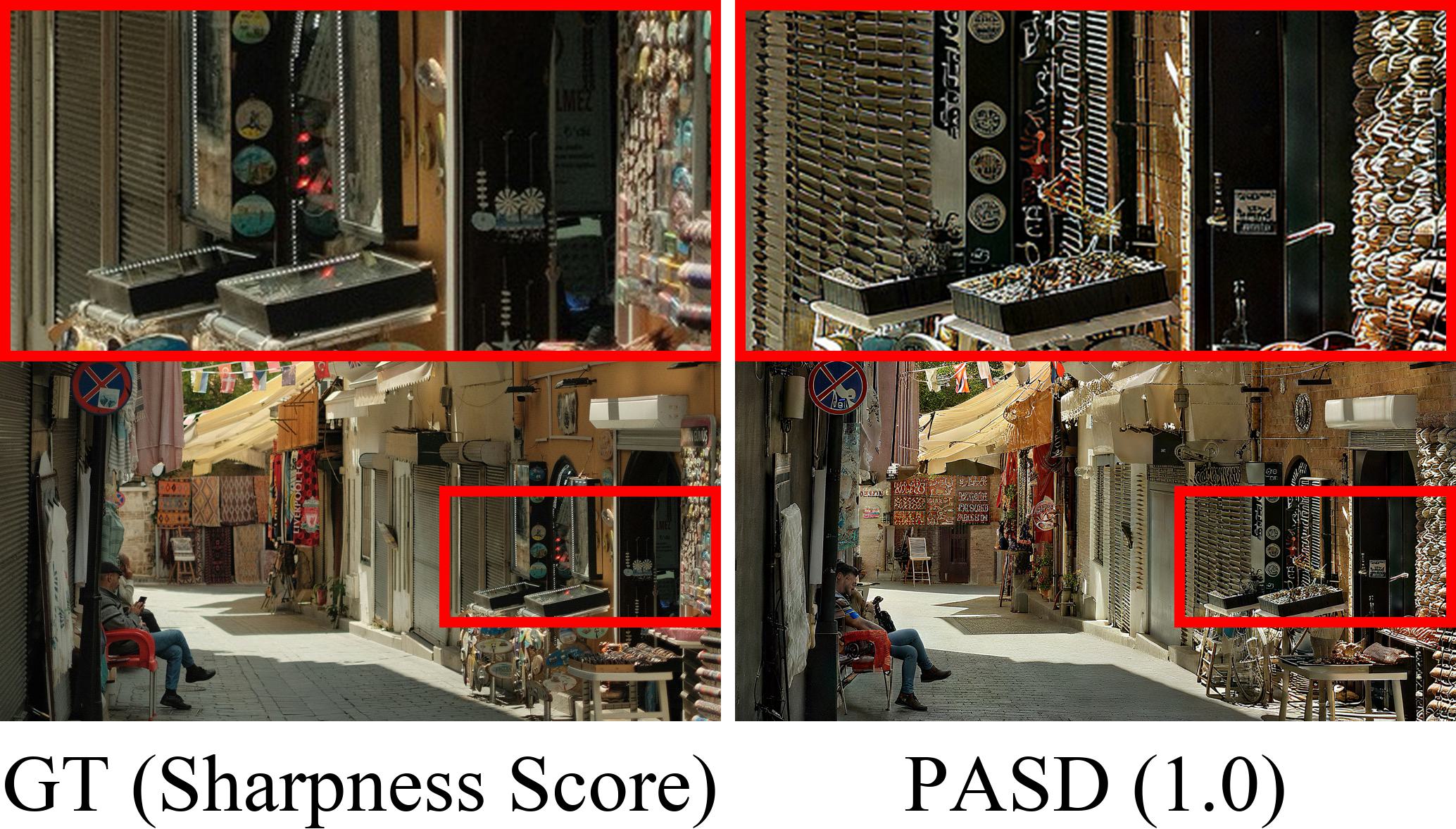}
   \hfill
   \includegraphics[width=0.32\linewidth]{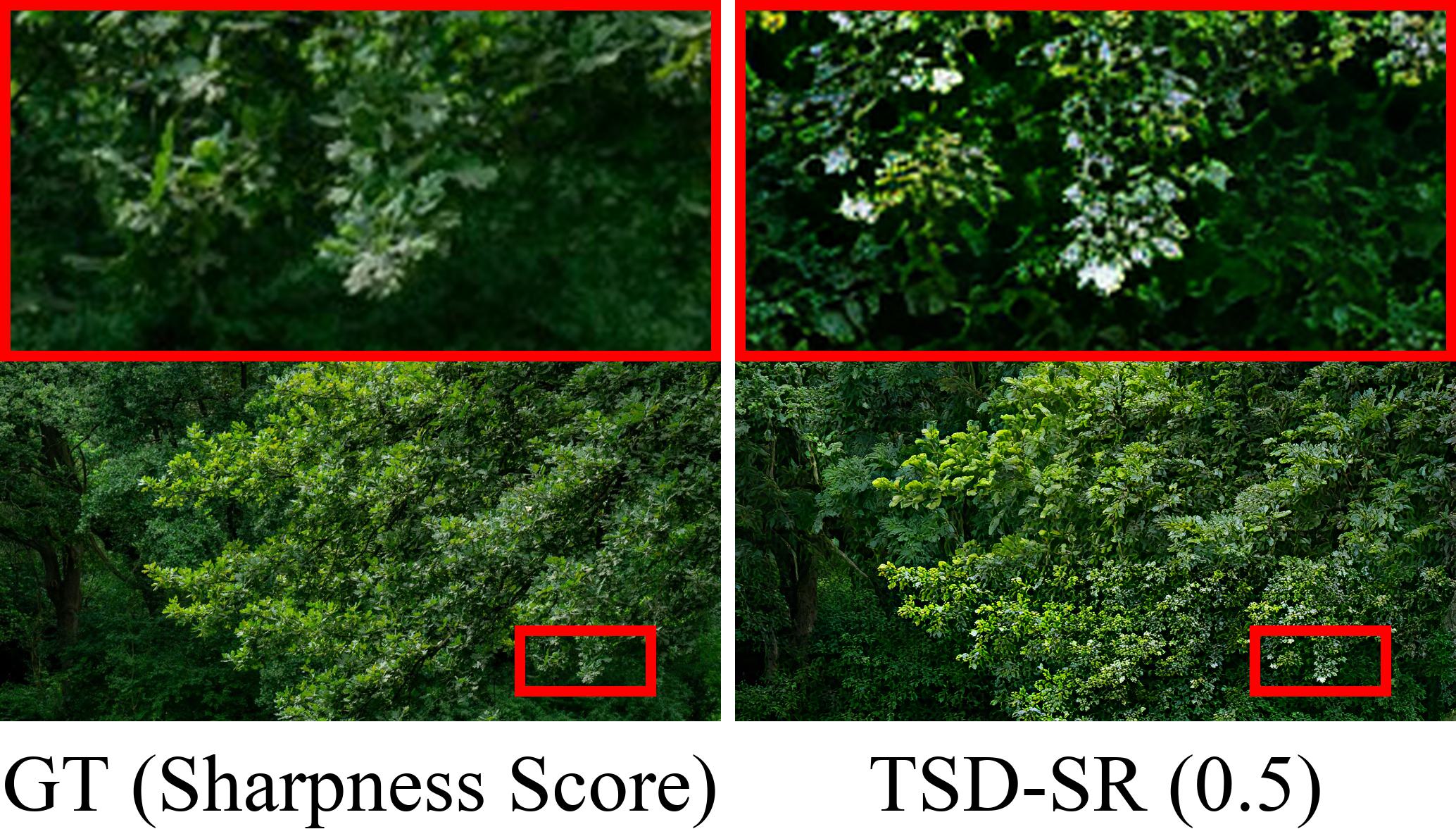}
   \hfill
   \includegraphics[width=0.32\linewidth]{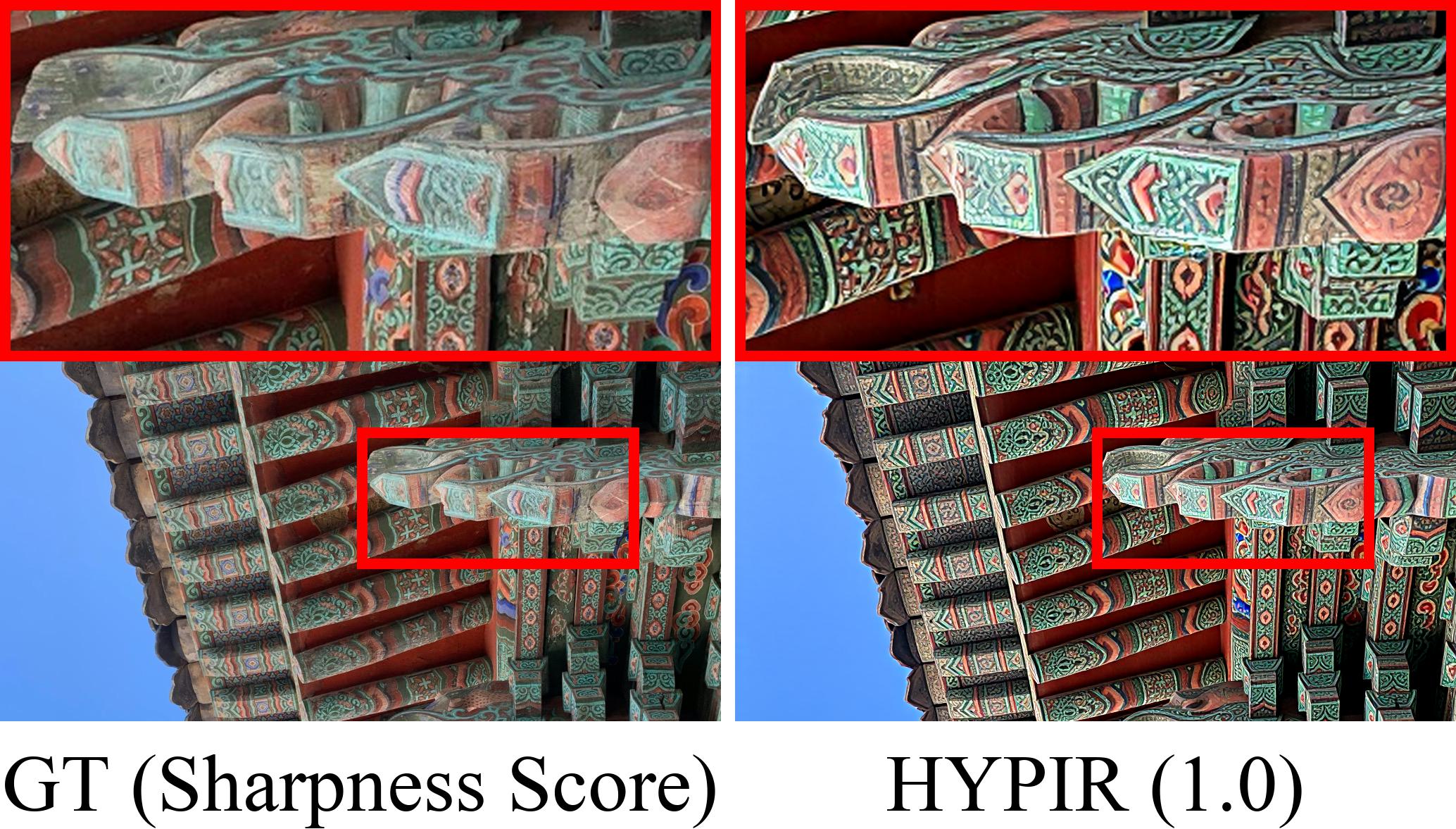}
    \caption{
    Failure cases of over-generation. Zoom in for a better view.
    }
    \label{fig:failure_cases_supp1}
\end{figure*}

\begin{figure*}[tp]
\scriptsize
\centering
   \includegraphics[width=0.48\linewidth]{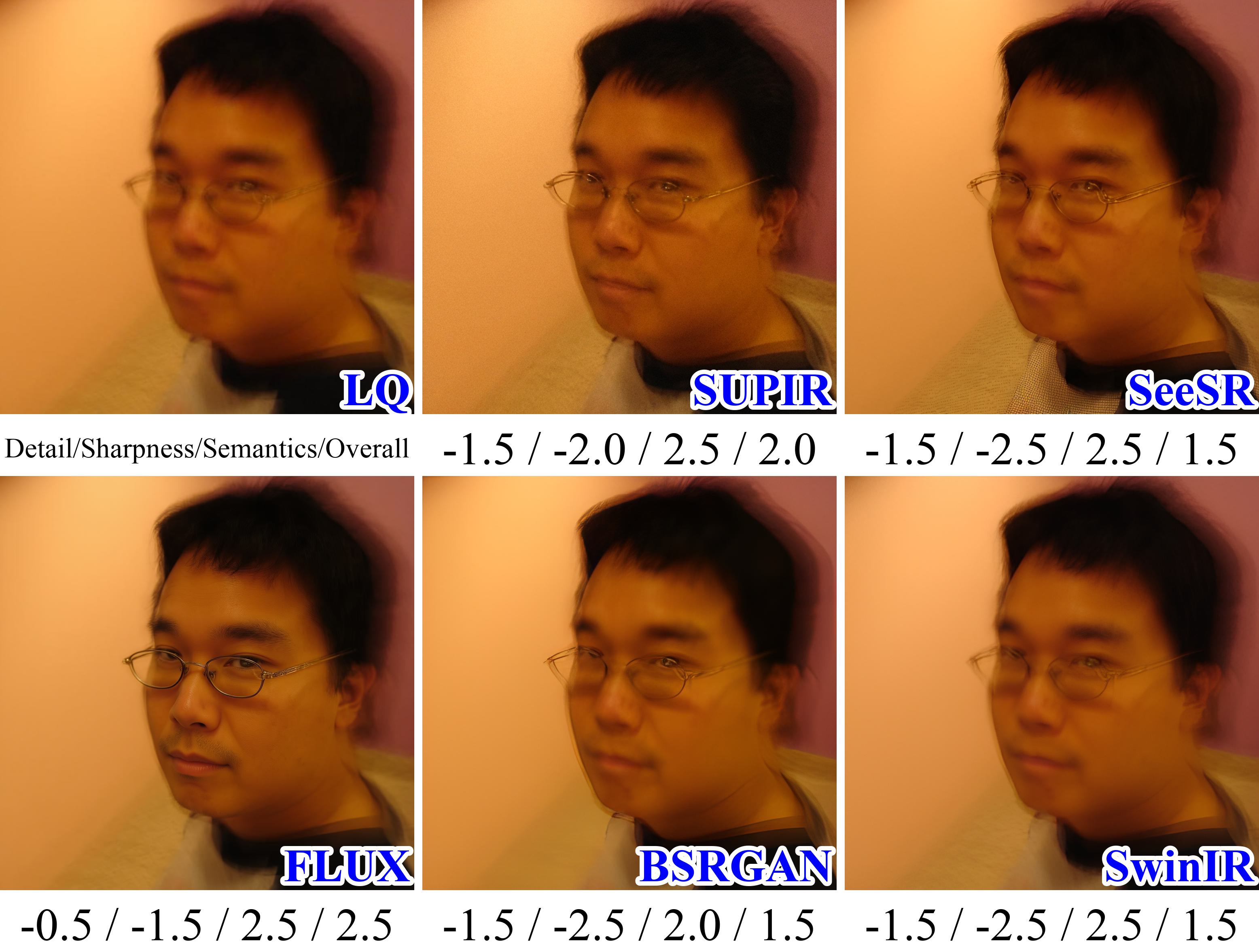}
   \hfill
   \includegraphics[width=0.48\linewidth]{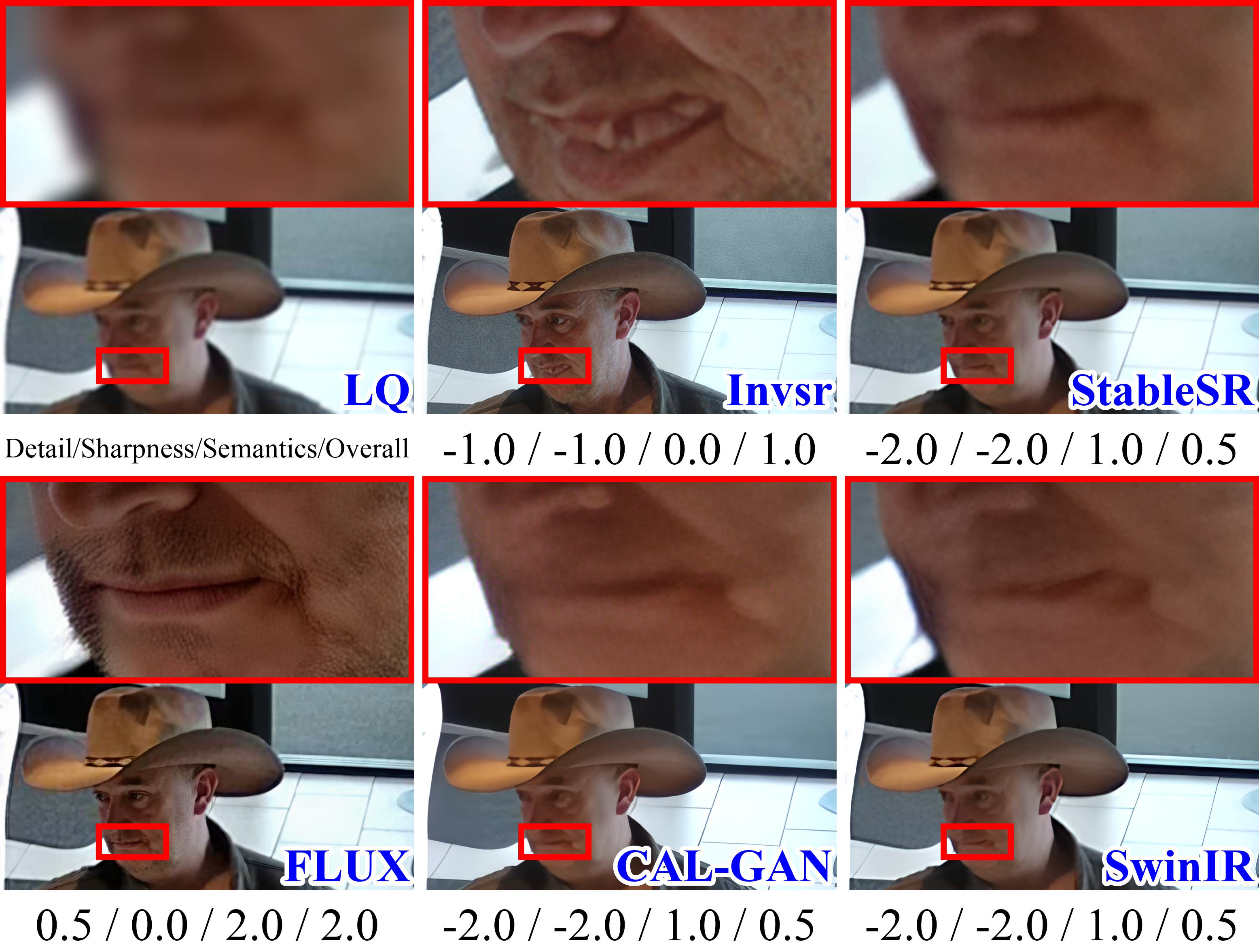}
   \hfill
   \includegraphics[width=0.48\linewidth]{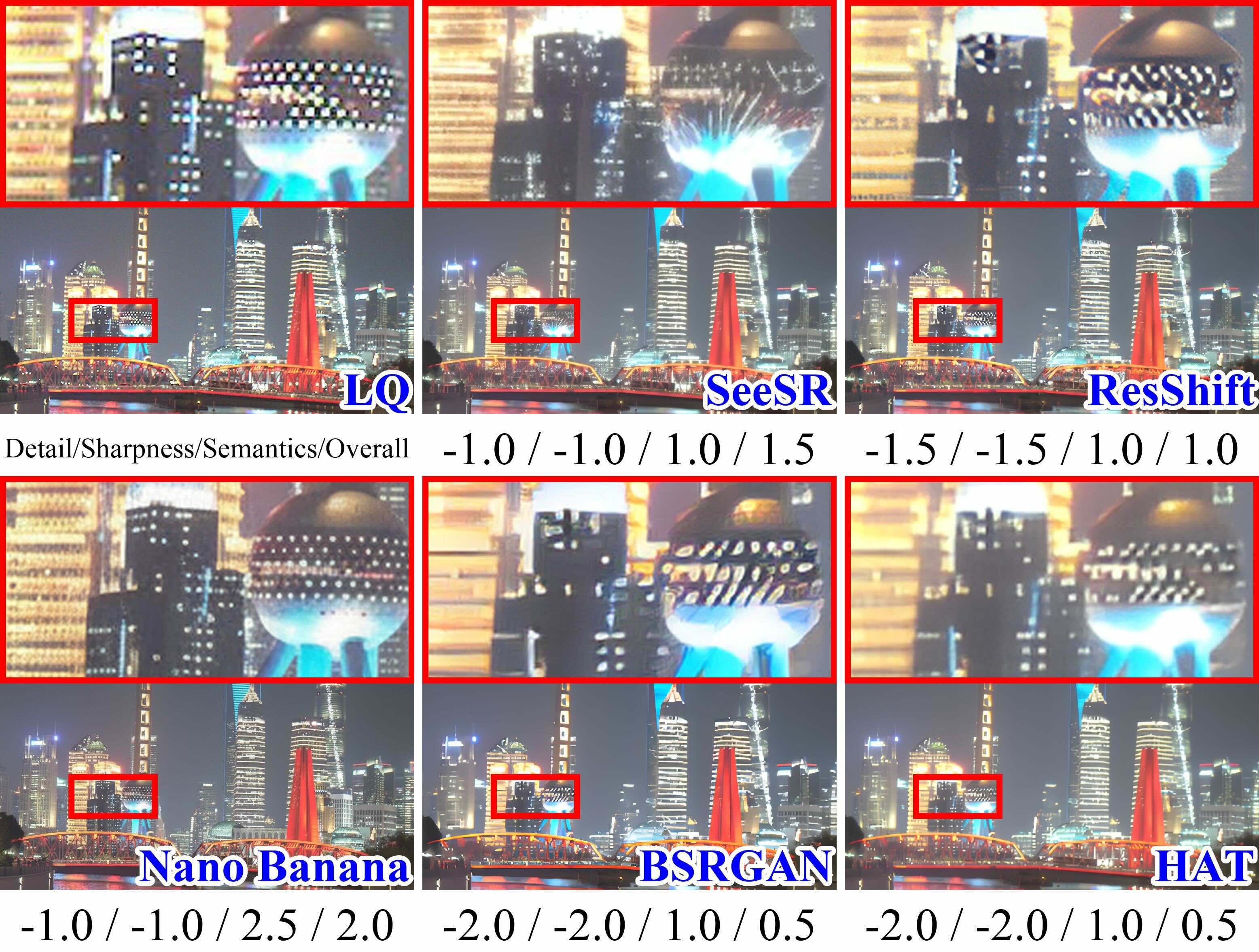}
   \hfill
   \includegraphics[width=0.48\linewidth]{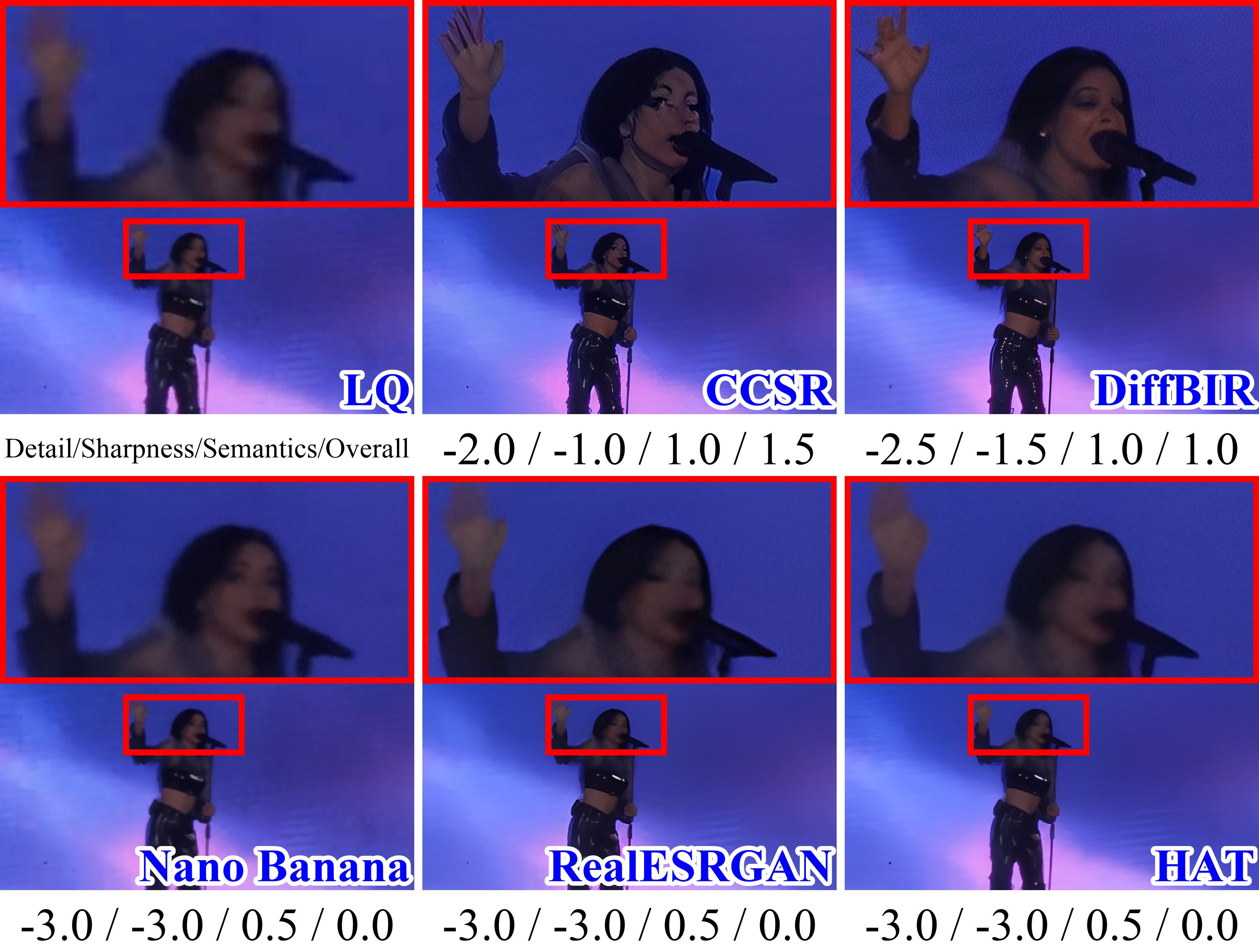}
   \hfill
   \includegraphics[width=0.48\linewidth]{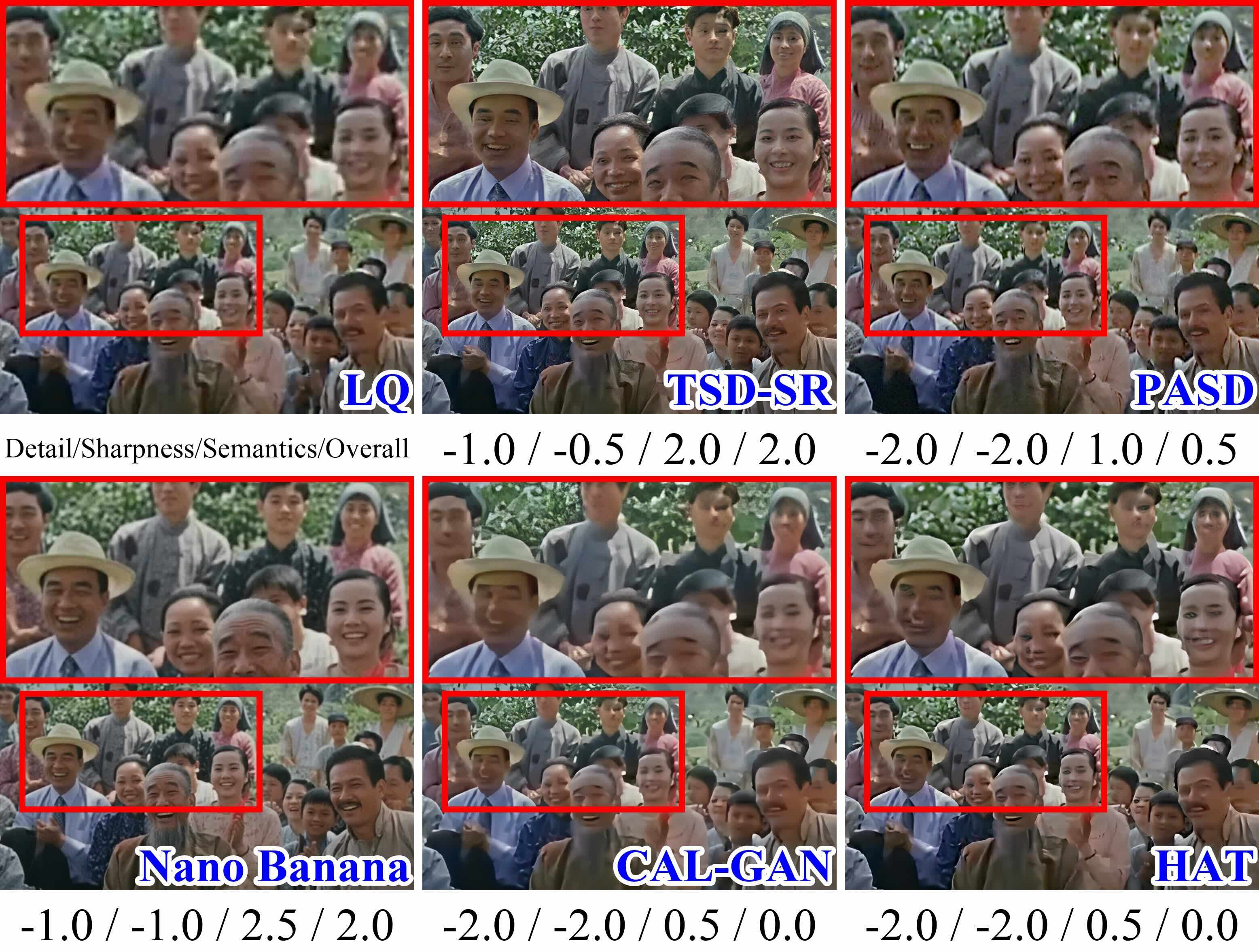}
   \hfill
   \includegraphics[width=0.48\linewidth]{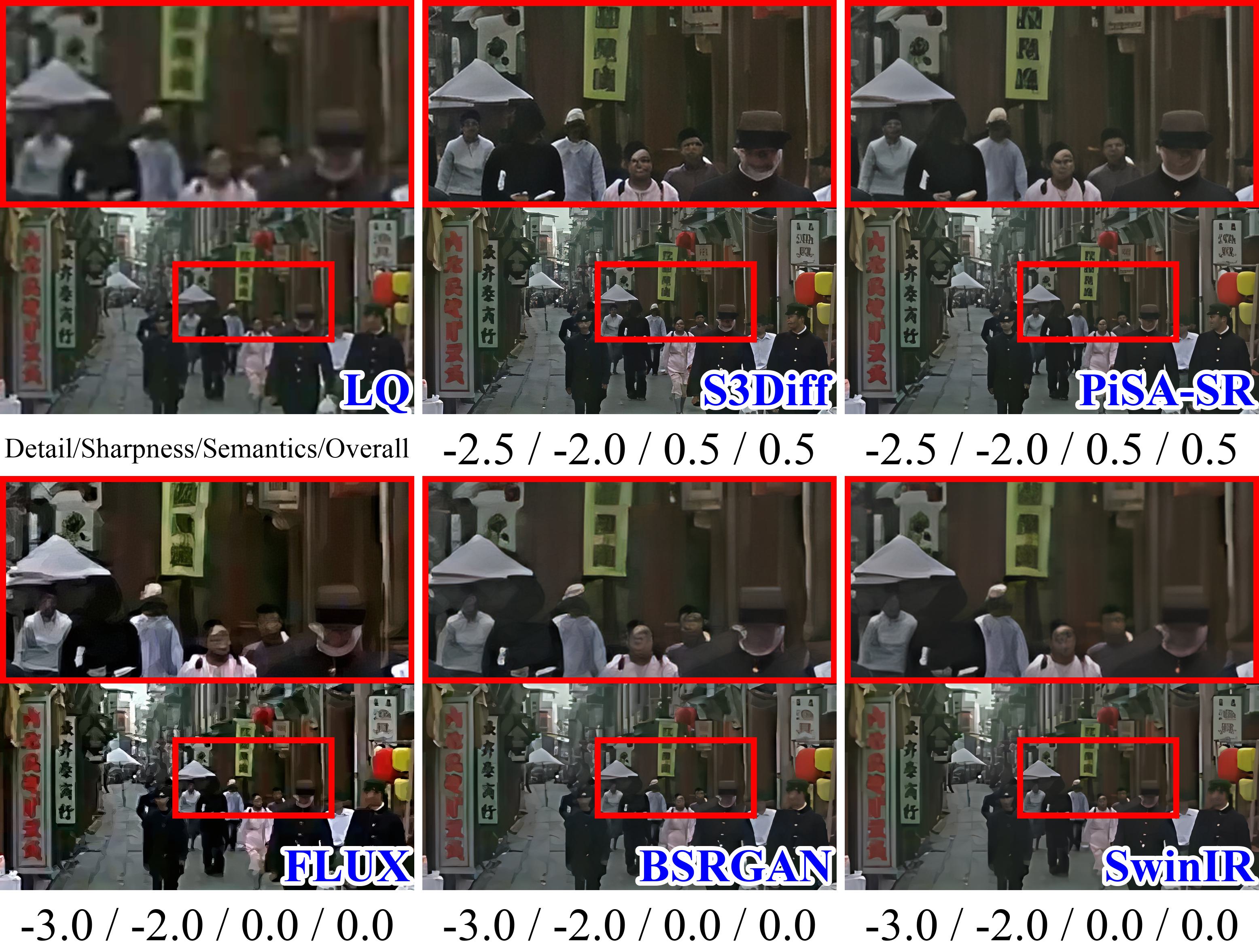}
    \caption{
    Failure cases of hard degradations. Zoom in for a better view.
    }
    \label{fig:failure_cases_supp4}
\end{figure*}

To further support the observations discussed in the main paper, we present additional failure cases of current GIR models across a wide range of semantic scenes and degradation types.
These examples cover various representative failure modes, including semantic errors in~\cref{fig:failure_cases_supp2,fig:failure_cases_supp3}, over-generation of details in~\cref{fig:failure_cases_supp1},  and insufficient degradation removal in~\cref{fig:failure_cases_supp4}, providing a more complete visual reference of the behaviors discussed in the main paper.

\section{Detail Results of each model}
\label{sec:model_details}
In the main paper, we primarily analyzed restoration behaviors at the model-family level.
Here, we provide model-specific results, reporting the detailed performance of all 20 restoration models across all evaluation dimensions.

We first present the overall performance ranking of each model in~\cref{tab:method_rank}, computed in two complementary ways:
(1) the average overall quality score across the entire dataset, and
(2) the average per-image ranking position, which reflects how frequently a model outperforms others on individual samples.

We then report per-degradation and per-semantic results for every evaluation dimension in~\cref{tab:per_method_Architecture_Food_Text,tab:per_method_Complex_Texture_Trees_&_leaves_Fabric_Texture,tab:per_method_Crowd_Hands/Feet_Animal_Fur,tab:per_method_Hand-drawn_Print_Media_Cartoon/Comic,tab:per_method_Large_Face_Medium_Face_Small_Face,tab:per_method_Leather_Surface_Reflective_Glass_Water_Flow,tab:per_method_Vehicles_Street_View_Aerial_View}.
Note that the Detail and Sharpness scores are calculated by taking the absolute value first, and then averaging the scores.
And since 0 is the point of balance, a lower score indicates a better performance for these two dimensions.
This includes the mean score of each model under every degradation type and semantic category, allowing comparison between different models.
\begin{table}[tp]
\centering
\footnotesize
\setlength\tabcolsep{9pt}
\begin{tabular}{l|cc}
\toprule
Model  & Avgerage Overall Score$\uparrow$ & Avgerage Rank$\downarrow$  \\
\midrule
HYPIR &2.775 & 3.127   \\
PiSA-SR &2.508 &5.547 \\
TSD-SR &2.467 &7.487    \\
DiffBIR &2.428 &5.181 \\
PASD & 2.38 & 6.382 \\
OSEDiff &2.377 &6.173 \\
Invsr &2.279 &7.635 \\
SUPIR  &2.235 &9.062   \\
SeeSR &2.201 &8.870 \\
CCSR &2.195 &10.365  \\
S3Diff &2.181 &8.484   \\
FLUX &1.935 &12.099   \\
StableSR &1.925 &11.133    \\
SwinIR &1.625 &15.569   \\
RealESRGAN &1.548 &15.734   \\
ResShift & 1.466 &13.858  \\
BSRGAN &1.435 &15.037  \\
HAT &1.426 &16.159   \\
Nano Banana &1.401 &15.552 \\
CAL-GAN &1.246 &16.544   \\
\bottomrule
\end{tabular}
\caption{
Overall performance ranking of all evaluated restoration models.
We report each model’s mean overall quality score across the full dataset and its average per-image ranking position.
}
\label{tab:method_rank}
\end{table}

\section{More Details of our IQA Model}
\label{sec:training}
\subsection{Training Details}
To evaluate the restored images across four distinct dimensions, we train a dedicated model for each specific direction. 
Notably, the original \textbf{Detail} and \textbf{Sharpness} assessments are bipolar metrics, with subjective scores ranging from $-3$ to $3$, where $0$ represents the optimal perceptual quality. 
To facilitate effective learning, we introduce a data preprocessing step that decouples these bipolar metrics into unipolar ones. 
For the \textbf{Detail} dimension, we utilize the data subset with \textbf{Detail} scores $\geq 0$ to train a model for predicting \textbf{Content Conciseness}, while the subset with scores $\leq 0$ is used to train the model for assessing \textbf{Detail Completeness}. 
For the \textbf{Sharpness} dimension, we utilize the data subset with \textbf{Sharpness} scores $\geq 0$ to train a model for predicting \textbf{Visual Clarity}, while the subset with scores $\leq 0$ is used to train the model for assessing \textbf{Visual Clarity}. The overall training protocol and hyperparameter configurations for our IQA models strictly follow those established in DeQA\cite{deqa_score}.
\subsection{Generalization}
We evaluated our model on three distinct external datasets; the results (SRCC/PLCC) are as shown in~\cref{tab:generalization}. 
While other models predict only overall quality, we provide a multi-dimensional IQA model that independently assesses Detail, Sharpness, and Semantics, enabling fine-grained evaluation.
\begin{table}[tp]
\centering
\footnotesize
\setlength\tabcolsep{0.8pt}
\begin{tabular}{c|ccccccc}
\toprule
Test set & NIQE & MANIQA & MUSIQ & CLIP-IQA & DeQA-Score & Ours \\
\midrule
LIVEW & 0.81/0.77 & 0.85/0.83 & 0.79/0.83 & \underline{0.83}/\underline{0.81} &  \textbf{0.89}/\textbf{0.88} & \underline{0.83}/\underline{0.81} \\
AGIQA & 0.72/0.65 & 0.72/0.64 & 0.72/0.63 & 0.74/0.69 &  \textbf{0.81}/\underline{0.73} & \textbf{0.81}/\textbf{0.76}  \\
CSIQ & 0.70/0.65 & 0.62/0.63 & 0.77/0.71 & 0.77/0.72 & \underline{0.79}/\underline{0.74} & \textbf{0.86}/\textbf{0.83} \\
\bottomrule
\end{tabular}
\caption{
Performance comparison of our proposed method against popular IQA models on three external datasets. Results are reported in terms of SRCC/PLCC. The best and second-best results are highlighted in \textbf{bold} and \underline{underlined} respectively.
}
\label{tab:generalization}
\end{table}



\begin{sidewaystable}[tp]
\centering
\footnotesize
\setlength\tabcolsep{4pt}
\begin{tabular}{l|cccc|cccc|cccc}
\toprule
\multirow{2}{*}{Model} & \multicolumn{4}{c|}{Large Face}& \multicolumn{4}{c|}{Medium Face}& \multicolumn{4}{c}{Small Face} \\
\cline{2-13}
& Overall$\uparrow$ & $\lvert$ Sharpness $\rvert\!\!\downarrow$  & $\lvert$ Detail $\rvert\!\!\downarrow$ & Semantics$\uparrow$ & Overall$\uparrow$ & $\lvert$ Sharpness $\rvert\!\!\downarrow$  & $\lvert$ Detail $\rvert\!\!\downarrow$ & Semantics$\uparrow$ & Overall$\uparrow$ & $\lvert$ Sharpness $\rvert\!\!\downarrow$  & $\lvert$ Detail $\rvert\!\!\downarrow$ & Semantics$\uparrow$ \\
\midrule
HYPIR & 3.07 & 0.57 & 0.57 & 3.14 & 3.00 & 0.94 & 0.78 & 3.17 & 2.78 & 1.17 & 1.00 & 2.89 \\
SUPIR & 2.57 & 1.14 & 0.93 & 2.79 & 2.67 & 1.39 & 1.06 & 2.72 & 2.33 & 1.56 & 1.06 & 2.56 \\
PiSA-SR & 2.64 & 1.07 & 1.36 & 2.86 & 2.78 & 1.22 & 1.06 & 2.94 & 2.50 & 1.44 & 1.44 & 2.67 \\
SeeSR & 2.57 & 1.43 & 1.36 & 2.71 & 2.39 & 1.44 & 1.44 & 2.72 & 2.28 & 1.39 & 1.39 & 2.22 \\
OSEDiff & 2.64 & 1.36 & 1.43 & 3.21 & 2.44 & 1.28 & 1.44 & 2.50 & 1.94 & 1.61 & 1.67 & 2.28 \\
CCSR & 2.57 & 0.29 & 1.00 & 2.93 & 2.56 & 0.67 & 0.94 & 2.72 & 2.06 & 1.28 & 1.50 & 2.17 \\
DiffBIR & 3.00 & 0.79 & 0.93 & 3.14 & 2.78 & 0.78 & 1.17 & 2.94 & 2.22 & 1.22 & 1.50 & 2.44 \\
StableSR & 2.21 & 1.64 & 1.50 & 2.79 & 2.06 & 1.83 & 1.61 & 2.44 & 1.94 & 2.06 & 1.83 & 2.22 \\
PASD & 2.71 & 0.79 & 0.86 & 2.93 & 2.22 & 1.11 & 1.11 & 2.44 & 2.06 & 0.72 & 0.67 & 2.11 \\
Invsr & 2.43 & 1.07 & 1.07 & 2.79 & 2.44 & 1.17 & 1.17 & 2.61 & 2.22 & 1.28 & 1.50 & 2.39 \\
S3Diff & 3.00 & 0.93 & 0.79 & 3.07 & 2.78 & 0.72 & 0.72 & 2.89 & 2.39 & 1.06 & 1.22 & 2.44 \\
TSD-SR & 2.50 & 1.07 & 1.07 & 2.71 & 2.56 & 1.11 & 1.06 & 2.72 & 2.39 & 1.33 & 1.28 & 2.50 \\
ResShift & 1.86 & 2.07 & 1.79 & 2.14 & 1.61 & 2.06 & 1.89 & 2.22 & 1.33 & 1.83 & 1.83 & 1.72 \\
\midrule
FLUX & 2.14 & 1.21 & 1.14 & 2.71 & 2.44 & 0.89 & 1.00 & 2.78 & 1.78 & 1.44 & 1.44 & 2.39 \\
Nano Banana & 1.71 & 1.71 & 1.57 & 2.79 & 1.89 & 2.00 & 1.39 & 2.50 & 0.78 & 2.44 & 2.28 & 2.11 \\
\midrule
BSRGAN & 2.00 & 2.00 & 1.71 & 2.71 & 1.83 & 2.06 & 2.11 & 2.39 & 1.22 & 2.17 & 2.11 & 1.39 \\
CAL-GAN & 1.79 & 2.00 & 1.71 & 2.50 & 1.50 & 2.22 & 1.94 & 2.06 & 0.83 & 2.17 & 2.28 & 1.67 \\
RealESRGAN & 1.71 & 2.14 & 2.00 & 2.43 & 1.83 & 1.89 & 1.83 & 2.39 & 1.39 & 1.94 & 2.17 & 1.56 \\
\midrule
HAT & 1.71 & 2.21 & 1.93 & 2.36 & 1.56 & 2.11 & 2.06 & 2.06 & 0.78 & 2.39 & 2.39 & 1.44 \\
SwinIR & 2.07 & 1.71 & 1.64 & 2.57 & 1.83 & 1.94 & 1.83 & 2.28 & 1.50 & 1.78 & 2.17 & 1.67 \\
\bottomrule
\end{tabular}
\vspace{-10pt}
\caption{The result of average overall, sharpness, detail, and semantics scores for all restoration models in Large Face, Large Face and Small Face. Note that the Detail and Sharpness scores are calculated by taking the absolute value first, and then averaging the scores.}
\label{tab:per_method_Large_Face_Medium_Face_Small_Face}
\end{sidewaystable}

\begin{sidewaystable}[tp]
\centering
\footnotesize
\setlength\tabcolsep{4pt}
\begin{tabular}{l|cccc|cccc|cccc}
\toprule
\multirow{2}{*}{Model} & \multicolumn{4}{c|}{Crowd}& \multicolumn{4}{c|}{Hands/Feet}& \multicolumn{4}{c}{Animal Fur} \\
\cline{2-13}
& Overall$\uparrow$ & $\lvert$ Sharpness $\rvert\!\!\downarrow$  & $\lvert$ Detail $\rvert\!\!\downarrow$ & Semantics$\uparrow$ & Overall$\uparrow$ & $\lvert$ Sharpness $\rvert\!\!\downarrow$  & $\lvert$ Detail $\rvert\!\!\downarrow$ & Semantics$\uparrow$ & Overall$\uparrow$ & $\lvert$ Sharpness $\rvert\!\!\downarrow$  & $\lvert$ Detail $\rvert\!\!\downarrow$ & Semantics$\uparrow$ \\
\midrule
HYPIR & 2.40 & 1.30 & 1.50 & 2.40 & 3.05 & 0.50 & 0.80 & 3.15 & 2.94 & 0.50 & 0.67 & 3.11 \\
SUPIR & 1.80 & 1.70 & 1.80 & 1.70 & 2.55 & 1.25 & 1.10 & 2.75 & 2.89 & 1.00 & 0.78 & 3.11 \\
PiSA-SR & 1.90 & 1.60 & 1.60 & 1.70 & 2.40 & 1.15 & 1.15 & 2.60 & 3.06 & 0.72 & 0.56 & 3.22 \\
SeeSR & 1.70 & 1.60 & 1.80 & 1.40 & 2.30 & 1.40 & 1.35 & 2.45 & 2.50 & 1.17 & 1.17 & 2.83 \\
OSEDiff & 1.20 & 1.80 & 1.90 & 1.50 & 2.40 & 1.15 & 1.25 & 2.50 & 2.72 & 1.06 & 0.83 & 3.06 \\
CCSR & 0.90 & 1.80 & 2.00 & 1.40 & 2.25 & 1.30 & 1.60 & 2.35 & 2.83 & 0.72 & 0.61 & 3.00 \\
DiffBIR & 1.60 & 1.70 & 1.70 & 1.60 & 2.40 & 0.80 & 1.20 & 2.60 & 3.06 & 0.50 & 0.72 & 3.11 \\
StableSR & 1.30 & 1.80 & 1.90 & 1.20 & 1.70 & 1.75 & 1.70 & 2.20 & 2.28 & 1.67 & 1.39 & 2.61 \\
PASD & 1.70 & 1.40 & 1.50 & 1.90 & 2.25 & 1.10 & 1.30 & 2.50 & 2.78 & 0.89 & 0.94 & 2.94 \\
Invsr & 1.80 & 1.80 & 1.70 & 1.60 & 2.40 & 1.10 & 1.20 & 2.60 & 3.00 & 0.56 & 0.72 & 3.00 \\
S3Diff & 2.00 & 1.40 & 1.60 & 1.80 & 2.80 & 0.95 & 0.85 & 2.90 & 3.17 & 0.67 & 0.67 & 3.22 \\
TSD-SR & 1.80 & 1.70 & 1.80 & 1.40 & 2.30 & 1.25 & 1.25 & 2.25 & 2.67 & 0.89 & 0.83 & 2.78 \\
ResShift & 1.10 & 1.80 & 2.00 & 1.30 & 1.55 & 1.75 & 1.75 & 2.00 & 2.11 & 1.44 & 1.28 & 2.61 \\
\midrule
FLUX & 1.40 & 1.60 & 1.60 & 1.50 & 1.95 & 1.45 & 1.25 & 2.30 & 1.67 & 1.94 & 1.67 & 2.22 \\
Nano Banana & 0.20 & 2.70 & 2.50 & 1.30 & 1.05 & 2.40 & 2.15 & 1.90 & 1.78 & 1.94 & 1.56 & 2.22 \\
\midrule
BSRGAN & 1.00 & 2.00 & 2.00 & 0.90 & 1.45 & 1.60 & 1.85 & 1.95 & 1.78 & 1.67 & 1.72 & 2.44 \\
CAL-GAN & 0.80 & 2.20 & 2.10 & 1.20 & 1.35 & 1.80 & 1.90 & 1.90 & 1.83 & 1.67 & 1.28 & 2.44 \\
RealESRGAN & 1.30 & 2.00 & 2.20 & 1.40 & 1.40 & 1.70 & 2.05 & 1.85 & 2.11 & 1.50 & 1.50 & 2.56 \\
\midrule
HAT & 0.80 & 2.10 & 2.20 & 0.90 & 1.60 & 1.85 & 1.80 & 1.95 & 2.28 & 1.72 & 1.50 & 2.72 \\
SwinIR & 0.90 & 1.40 & 1.80 & 1.00 & 1.50 & 2.00 & 2.00 & 2.05 & 2.28 & 1.39 & 1.39 & 2.67 \\
\bottomrule
\end{tabular}
\vspace{-10pt}
\caption{The result of average overall, sharpness, detail, and semantics scores for all restoration models in Crowd, Crowd and Animal Fur. Note that the Detail and Sharpness scores are calculated by taking the absolute value first, and then averaging the scores.}
\label{tab:per_method_Crowd_Hands/Feet_Animal_Fur}
\end{sidewaystable}

\begin{sidewaystable}[tp]
\centering
\footnotesize
\setlength\tabcolsep{4pt}
\begin{tabular}{l|cccc|cccc|cccc}
\toprule
\multirow{2}{*}{Model} & \multicolumn{4}{c|}{Complex Texture}& \multicolumn{4}{c|}{Trees \& leaves}& \multicolumn{4}{c}{Fabric Texture} \\
\cline{2-13}
& Overall$\uparrow$ & $\lvert$ Sharpness $\rvert\!\!\downarrow$  & $\lvert$ Detail $\rvert\!\!\downarrow$ & Semantics$\uparrow$ & Overall$\uparrow$ & $\lvert$ Sharpness $\rvert\!\!\downarrow$  & $\lvert$ Detail $\rvert\!\!\downarrow$ & Semantics$\uparrow$ & Overall$\uparrow$ & $\lvert$ Sharpness $\rvert\!\!\downarrow$  & $\lvert$ Detail $\rvert\!\!\downarrow$ & Semantics$\uparrow$ \\
\midrule
HYPIR & 3.00 & 0.92 & 0.75 & 3.17 & 2.79 & 1.07 & 1.14 & 2.86 & 2.70 & 0.30 & 0.70 & 2.70 \\
SUPIR & 2.58 & 0.67 & 0.67 & 2.67 & 2.86 & 1.07 & 0.93 & 2.86 & 3.10 & 0.50 & 0.10 & 3.20 \\
PiSA-SR & 2.58 & 1.00 & 1.08 & 2.83 & 2.57 & 1.00 & 1.07 & 2.71 & 2.50 & 0.60 & 1.10 & 2.40 \\
SeeSR & 2.25 & 1.17 & 1.25 & 2.58 & 2.43 & 1.57 & 1.43 & 2.86 & 2.80 & 0.90 & 0.70 & 3.00 \\
OSEDiff & 2.67 & 1.08 & 1.17 & 2.75 & 2.36 & 1.21 & 1.14 & 2.57 & 2.60 & 0.60 & 0.60 & 2.60 \\
CCSR & 2.25 & 1.33 & 1.33 & 2.33 & 2.36 & 1.43 & 1.14 & 2.79 & 2.40 & 0.90 & 1.40 & 2.40 \\
DiffBIR & 2.75 & 0.67 & 0.92 & 2.92 & 2.50 & 1.43 & 1.21 & 2.64 & 2.60 & 0.80 & 0.80 & 3.00 \\
StableSR & 2.17 & 1.58 & 1.33 & 2.58 & 1.86 & 2.07 & 1.86 & 2.29 & 2.00 & 1.80 & 1.10 & 2.50 \\
PASD & 2.33 & 0.58 & 0.58 & 2.50 & 2.57 & 1.29 & 1.07 & 2.93 & 2.70 & 1.00 & 0.80 & 2.90 \\
Invsr & 2.00 & 0.75 & 1.17 & 2.08 & 2.64 & 0.93 & 1.07 & 2.71 & 2.10 & 0.50 & 1.30 & 2.20 \\
S3Diff & 2.83 & 0.83 & 0.83 & 2.92 & 3.00 & 0.86 & 1.07 & 3.07 & 2.70 & 0.50 & 1.00 & 2.70 \\
TSD-SR & 2.58 & 0.67 & 0.75 & 2.50 & 2.71 & 1.07 & 0.93 & 3.07 & 2.50 & 0.70 & 0.70 & 2.50 \\
ResShift & 1.67 & 1.33 & 1.42 & 2.08 & 1.50 & 1.50 & 1.50 & 2.29 & 1.60 & 1.40 & 1.50 & 2.20 \\
\midrule
FLUX & 1.33 & 1.75 & 1.42 & 1.67 & 0.79 & 2.07 & 2.00 & 1.71 & 0.60 & 2.40 & 1.80 & 1.90 \\
Nano Banana & 0.83 & 2.25 & 2.00 & 2.17 & 1.21 & 2.14 & 2.00 & 1.71 & 1.50 & 1.70 & 1.50 & 2.60 \\
\midrule
BSRGAN & 1.92 & 1.58 & 1.33 & 2.17 & 1.36 & 2.14 & 1.86 & 2.36 & 1.70 & 1.50 & 1.80 & 2.20 \\
CAL-GAN & 1.58 & 1.42 & 1.17 & 2.00 & 1.14 & 1.71 & 1.79 & 2.07 & 1.60 & 1.30 & 1.50 & 1.90 \\
RealESRGAN & 1.92 & 1.08 & 1.33 & 2.17 & 1.79 & 1.50 & 1.29 & 2.50 & 2.10 & 1.40 & 2.00 & 1.90 \\
\midrule
HAT & 1.75 & 1.67 & 1.42 & 2.17 & 2.07 & 1.71 & 1.57 & 2.50 & 1.80 & 2.00 & 1.80 & 2.20 \\
SwinIR & 2.00 & 1.25 & 1.33 & 2.00 & 2.29 & 1.50 & 1.14 & 2.43 & 2.10 & 0.90 & 1.50 & 2.30 \\
\bottomrule
\end{tabular}
\vspace{-10pt}
\caption{The result of average overall, sharpness, detail, and semantics scores for all restoration models in Complex Texture, Complex Texture and Fabric Texture. Note that the Detail and Sharpness scores are calculated by taking the absolute value first, and then averaging the scores.}
\label{tab:per_method_Complex_Texture_Trees_&_leaves_Fabric_Texture}
\end{sidewaystable}

\begin{sidewaystable}[tp]
\centering
\footnotesize
\setlength\tabcolsep{4pt}
\begin{tabular}{l|cccc|cccc|cccc}
\toprule
\multirow{2}{*}{Model} & \multicolumn{4}{c|}{Leather Surface}& \multicolumn{4}{c|}{Reflective Glass}& \multicolumn{4}{c}{Water Flow} \\
\cline{2-13}
& Overall$\uparrow$ & $\lvert$ Sharpness $\rvert\!\!\downarrow$  & $\lvert$ Detail $\rvert\!\!\downarrow$ & Semantics$\uparrow$ & Overall$\uparrow$ & $\lvert$ Sharpness $\rvert\!\!\downarrow$  & $\lvert$ Detail $\rvert\!\!\downarrow$ & Semantics$\uparrow$ & Overall$\uparrow$ & $\lvert$ Sharpness $\rvert\!\!\downarrow$  & $\lvert$ Detail $\rvert\!\!\downarrow$ & Semantics$\uparrow$ \\
\midrule
HYPIR & 3.50 & 0.33 & 0.17 & 3.58 & 3.21 & 0.64 & 0.64 & 3.36 & 2.40 & 1.30 & 1.30 & 2.80 \\
SUPIR & 3.00 & 0.92 & 0.83 & 3.17 & 2.50 & 0.93 & 0.86 & 2.86 & 2.30 & 1.30 & 1.00 & 2.60 \\
PiSA-SR & 3.25 & 0.50 & 0.58 & 3.25 & 2.43 & 1.14 & 1.07 & 2.64 & 2.50 & 1.20 & 1.30 & 2.70 \\
SeeSR & 2.42 & 0.92 & 1.25 & 2.67 & 2.43 & 1.14 & 1.21 & 2.71 & 2.40 & 1.30 & 1.30 & 2.70 \\
OSEDiff & 3.42 & 0.50 & 0.58 & 3.50 & 2.50 & 1.14 & 1.14 & 2.86 & 2.40 & 1.50 & 1.50 & 2.70 \\
CCSR & 2.75 & 0.42 & 0.75 & 2.83 & 1.93 & 1.43 & 1.50 & 2.36 & 2.00 & 1.40 & 1.60 & 2.50 \\
DiffBIR & 3.33 & 0.25 & 0.42 & 3.58 & 2.57 & 0.79 & 0.86 & 2.64 & 2.60 & 1.10 & 1.20 & 2.90 \\
StableSR & 2.25 & 1.67 & 1.33 & 2.75 & 1.86 & 1.64 & 1.57 & 2.29 & 1.80 & 2.00 & 1.60 & 2.10 \\
PASD & 3.42 & 0.33 & 0.42 & 3.50 & 3.07 & 0.50 & 0.36 & 3.00 & 3.00 & 0.90 & 0.80 & 3.20 \\
Invsr & 2.92 & 0.75 & 0.58 & 2.83 & 2.57 & 0.93 & 1.07 & 2.64 & 2.50 & 1.00 & 1.20 & 2.50 \\
S3Diff & 3.00 & 0.42 & 0.50 & 3.17 & 2.64 & 0.79 & 0.79 & 2.86 & 2.50 & 1.00 & 1.00 & 2.50 \\
TSD-SR & 3.08 & 0.58 & 0.42 & 3.17 & 2.64 & 0.71 & 0.86 & 2.57 & 2.40 & 0.90 & 1.10 & 2.40 \\
ResShift & 2.25 & 1.50 & 1.42 & 2.75 & 1.79 & 1.57 & 1.50 & 2.07 & 2.00 & 1.50 & 1.10 & 2.60 \\
\midrule
FLUX & 1.75 & 1.75 & 1.33 & 2.75 & 1.86 & 1.79 & 1.43 & 2.50 & 0.90 & 1.90 & 1.90 & 1.90 \\
Nano Banana & 1.75 & 1.08 & 0.58 & 2.25 & 1.21 & 2.43 & 1.93 & 2.00 & 1.80 & 1.80 & 1.60 & 2.40 \\
\midrule
BSRGAN & 2.00 & 1.58 & 1.50 & 2.58 & 2.29 & 1.21 & 1.36 & 2.71 & 1.30 & 1.90 & 2.00 & 1.90 \\
CAL-GAN & 2.25 & 1.67 & 1.42 & 2.67 & 1.64 & 1.50 & 1.64 & 2.00 & 1.20 & 1.90 & 1.70 & 2.20 \\
RealESRGAN & 2.33 & 1.33 & 1.25 & 2.75 & 2.36 & 1.36 & 1.29 & 2.79 & 1.40 & 1.80 & 1.60 & 2.30 \\
\midrule
HAT & 2.25 & 1.33 & 1.58 & 2.83 & 1.93 & 1.71 & 1.71 & 2.64 & 1.60 & 2.00 & 1.80 & 2.00 \\
SwinIR & 2.33 & 1.42 & 1.25 & 2.67 & 2.14 & 1.36 & 1.21 & 2.57 & 1.80 & 1.90 & 2.00 & 2.20 \\
\bottomrule
\end{tabular}
\vspace{-10pt}
\caption{The result of average overall, sharpness, detail, and semantics scores for all restoration models in Leather Surface, Leather Surface and Water Flow. Note that the Detail and Sharpness scores are calculated by taking the absolute value first, and then averaging the scores.}
\label{tab:per_method_Leather_Surface_Reflective_Glass_Water_Flow}
\end{sidewaystable}

\begin{sidewaystable}[tp]
\centering
\footnotesize
\setlength\tabcolsep{4pt}
\begin{tabular}{l|cccc|cccc|cccc}
\toprule
\multirow{2}{*}{Model} & \multicolumn{4}{c|}{Vehicles}& \multicolumn{4}{c|}{Street View}& \multicolumn{4}{c}{Aerial View} \\
\cline{2-13}
& Overall$\uparrow$ & $\lvert$ Sharpness $\rvert\!\!\downarrow$  & $\lvert$ Detail $\rvert\!\!\downarrow$ & Semantics$\uparrow$ & Overall$\uparrow$ & $\lvert$ Sharpness $\rvert\!\!\downarrow$  & $\lvert$ Detail $\rvert\!\!\downarrow$ & Semantics$\uparrow$ & Overall$\uparrow$ & $\lvert$ Sharpness $\rvert\!\!\downarrow$  & $\lvert$ Detail $\rvert\!\!\downarrow$ & Semantics$\uparrow$ \\
\midrule
HYPIR & 2.80 & 1.00 & 1.00 & 2.80 & 2.64 & 1.07 & 1.36 & 2.93 & 2.50 & 1.30 & 1.40 & 2.80 \\
SUPIR & 2.70 & 1.00 & 1.00 & 2.60 & 2.43 & 1.07 & 1.07 & 2.43 & 2.40 & 1.60 & 1.00 & 2.40 \\
PiSA-SR & 2.70 & 1.00 & 1.60 & 3.00 & 2.36 & 1.36 & 1.29 & 2.43 & 1.40 & 1.60 & 1.70 & 2.10 \\
SeeSR & 2.20 & 1.00 & 1.40 & 2.70 & 1.93 & 1.64 & 1.71 & 2.07 & 1.70 & 1.70 & 1.40 & 2.40 \\
OSEDiff & 2.50 & 1.50 & 1.60 & 2.90 & 1.79 & 1.64 & 1.71 & 2.00 & 1.80 & 1.90 & 1.70 & 2.20 \\
CCSR & 2.50 & 1.30 & 1.20 & 2.60 & 2.21 & 1.43 & 1.36 & 2.71 & 1.70 & 2.20 & 1.80 & 1.90 \\
DiffBIR & 2.40 & 1.40 & 1.40 & 2.60 & 2.14 & 1.36 & 1.36 & 2.43 & 2.10 & 1.10 & 1.60 & 2.20 \\
StableSR & 2.00 & 1.70 & 1.60 & 2.50 & 1.64 & 2.00 & 1.71 & 2.14 & 1.80 & 1.90 & 1.50 & 2.50 \\
PASD & 3.10 & 0.60 & 0.70 & 2.90 & 2.21 & 1.00 & 0.93 & 2.29 & 2.90 & 0.70 & 0.60 & 3.00 \\
Invsr & 2.70 & 1.20 & 1.30 & 2.80 & 2.57 & 1.14 & 1.14 & 2.64 & 2.30 & 1.30 & 1.30 & 2.50 \\
S3Diff & 2.60 & 0.70 & 0.70 & 2.50 & 2.21 & 1.07 & 1.36 & 2.57 & 2.30 & 1.20 & 1.20 & 2.50 \\
TSD-SR & 2.90 & 1.10 & 1.40 & 2.80 & 2.43 & 0.64 & 1.07 & 2.50 & 2.30 & 1.00 & 1.00 & 2.30 \\
ResShift & 1.00 & 1.90 & 1.90 & 1.60 & 1.29 & 2.14 & 2.00 & 1.93 & 1.10 & 1.80 & 2.00 & 1.60 \\
\midrule
FLUX & 2.40 & 1.00 & 1.00 & 2.60 & 1.50 & 1.86 & 1.57 & 2.14 & 1.30 & 2.10 & 1.60 & 2.10 \\
Nano Banana & 1.50 & 2.10 & 1.90 & 2.20 & 0.79 & 2.50 & 2.36 & 2.50 & 0.70 & 2.70 & 2.20 & 2.20 \\
\midrule
BSRGAN & 1.10 & 1.80 & 1.90 & 1.40 & 1.57 & 2.21 & 2.07 & 2.07 & 1.40 & 1.70 & 1.60 & 2.10 \\
CAL-GAN & 1.50 & 1.50 & 1.70 & 1.50 & 1.50 & 1.79 & 1.64 & 2.14 & 1.10 & 1.70 & 1.70 & 1.70 \\
RealESRGAN & 1.70 & 1.60 & 1.90 & 2.10 & 2.07 & 1.64 & 1.64 & 2.36 & 0.80 & 1.70 & 1.90 & 1.50 \\
\midrule
HAT & 1.20 & 2.00 & 2.10 & 1.80 & 1.29 & 1.86 & 1.86 & 1.50 & 1.00 & 1.90 & 2.00 & 1.80 \\
SwinIR & 1.60 & 1.60 & 1.90 & 2.00 & 1.57 & 1.43 & 1.71 & 2.21 & 1.70 & 1.50 & 1.80 & 2.20 \\
\bottomrule
\end{tabular}
\vspace{-10pt}
\caption{The result of average overall, sharpness, detail, and semantics scores for all restoration models in Vehicles, Vehicles and Aerial View. Note that the Detail and Sharpness scores are calculated by taking the absolute value first, and then averaging the scores.}
\label{tab:per_method_Vehicles_Street_View_Aerial_View}
\end{sidewaystable}

\begin{sidewaystable}[tp]
\centering
\footnotesize
\setlength\tabcolsep{4pt}
\begin{tabular}{l|cccc|cccc|cccc}
\toprule
\multirow{2}{*}{Model} & \multicolumn{4}{c|}{Architecture}& \multicolumn{4}{c|}{Food}& \multicolumn{4}{c}{Text} \\
\cline{2-13}
& Overall$\uparrow$ & $\lvert$ Sharpness $\rvert\!\!\downarrow$  & $\lvert$ Detail $\rvert\!\!\downarrow$ & Semantics$\uparrow$ & Overall$\uparrow$ & $\lvert$ Sharpness $\rvert\!\!\downarrow$  & $\lvert$ Detail $\rvert\!\!\downarrow$ & Semantics$\uparrow$ & Overall$\uparrow$ & $\lvert$ Sharpness $\rvert\!\!\downarrow$  & $\lvert$ Detail $\rvert\!\!\downarrow$ & Semantics$\uparrow$ \\
\midrule
HYPIR & 2.88 & 0.75 & 0.88 & 3.06 & 2.93 & 0.57 & 0.57 & 3.00 & 2.56 & 0.78 & 0.94 & 2.56 \\
SUPIR & 2.62 & 0.44 & 0.81 & 2.69 & 3.00 & 0.71 & 0.86 & 3.00 & 2.50 & 0.94 & 0.61 & 2.39 \\
PiSA-SR & 2.44 & 1.06 & 1.19 & 2.69 & 3.07 & 1.07 & 0.79 & 3.14 & 2.33 & 0.94 & 0.94 & 2.61 \\
SeeSR & 2.38 & 1.12 & 1.25 & 2.69 & 2.64 & 0.79 & 0.86 & 2.79 & 1.78 & 1.00 & 0.89 & 1.89 \\
OSEDiff & 2.81 & 1.06 & 0.94 & 2.81 & 2.57 & 1.14 & 0.93 & 2.79 & 1.94 & 1.28 & 1.28 & 2.06 \\
CCSR & 2.38 & 0.94 & 1.06 & 2.50 & 2.93 & 0.93 & 0.86 & 3.00 & 2.22 & 0.94 & 1.22 & 2.33 \\
DiffBIR & 2.62 & 0.81 & 1.00 & 2.75 & 2.93 & 0.71 & 0.79 & 3.00 & 2.22 & 0.72 & 0.72 & 2.22 \\
StableSR & 1.62 & 1.88 & 1.69 & 2.38 & 2.43 & 1.29 & 1.14 & 2.86 & 1.61 & 1.39 & 1.22 & 2.11 \\
PASD & 2.75 & 0.81 & 0.62 & 3.00 & 3.14 & 0.21 & 0.43 & 2.93 & 2.11 & 0.78 & 0.83 & 2.33 \\
Invsr & 2.88 & 1.00 & 1.06 & 3.00 & 2.93 & 0.64 & 0.64 & 2.79 & 2.06 & 0.89 & 0.67 & 2.06 \\
S3Diff & 3.00 & 0.62 & 0.50 & 3.06 & 2.86 & 0.50 & 0.71 & 3.00 & 2.33 & 0.78 & 0.67 & 2.28 \\
TSD-SR & 2.88 & 1.06 & 1.06 & 2.94 & 3.07 & 0.36 & 0.43 & 3.07 & 2.33 & 0.94 & 0.83 & 2.33 \\
ResShift & 2.00 & 1.44 & 1.44 & 2.38 & 1.93 & 1.79 & 1.21 & 2.57 & 1.28 & 1.56 & 1.33 & 1.61 \\
\midrule
FLUX & 1.75 & 1.56 & 1.12 & 2.19 & 2.14 & 1.43 & 1.36 & 2.43 & 1.50 & 1.22 & 1.06 & 1.89 \\
Nano Banana & 1.44 & 2.06 & 1.88 & 2.38 & 2.00 & 1.64 & 1.57 & 2.50 & 2.00 & 1.39 & 0.94 & 2.33 \\
\midrule
BSRGAN & 1.62 & 1.75 & 1.56 & 2.12 & 2.07 & 1.71 & 1.43 & 3.07 & 1.33 & 2.00 & 1.39 & 1.83 \\
CAL-GAN & 1.69 & 1.25 & 1.62 & 2.12 & 1.36 & 1.86 & 1.86 & 1.86 & 1.22 & 1.78 & 1.33 & 1.50 \\
RealESRGAN & 2.12 & 1.25 & 1.50 & 2.44 & 1.79 & 1.64 & 1.43 & 2.36 & 1.72 & 1.61 & 1.28 & 1.83 \\
\midrule
HAT & 1.75 & 1.62 & 1.56 & 2.12 & 1.64 & 1.93 & 1.79 & 2.36 & 1.61 & 1.44 & 1.44 & 1.83 \\
SwinIR & 1.81 & 1.44 & 1.56 & 2.25 & 2.21 & 1.36 & 1.21 & 2.50 & 1.56 & 1.61 & 1.28 & 1.89 \\
\bottomrule
\end{tabular}
\vspace{-10pt}
\caption{The result of average overall, sharpness, detail, and semantics scores for all restoration models in Architecture, Architecture and Text. Note that the Detail and Sharpness scores are calculated by taking the absolute value first, and then averaging the scores.}
\label{tab:per_method_Architecture_Food_Text}
\end{sidewaystable}

\begin{sidewaystable}[tp]
\centering
\footnotesize
\setlength\tabcolsep{4pt}
\begin{tabular}{l|cccc|cccc|cccc}
\toprule
\multirow{2}{*}{Model} & \multicolumn{4}{c|}{Hand-drawn}& \multicolumn{4}{c|}{Print Media}& \multicolumn{4}{c}{Cartoon/Comic} \\
\cline{2-13}
& Overall$\uparrow$ & $\lvert$ Sharpness $\rvert\!\!\downarrow$  & $\lvert$ Detail $\rvert\!\!\downarrow$ & Semantics$\uparrow$ & Overall$\uparrow$ & $\lvert$ Sharpness $\rvert\!\!\downarrow$  & $\lvert$ Detail $\rvert\!\!\downarrow$ & Semantics$\uparrow$ & Overall$\uparrow$ & $\lvert$ Sharpness $\rvert\!\!\downarrow$  & $\lvert$ Detail $\rvert\!\!\downarrow$ & Semantics$\uparrow$ \\
\midrule
HYPIR & 3.00 & 0.79 & 0.64 & 2.93 & 2.64 & 0.93 & 0.86 & 2.50 & 3.17 & 0.50 & 0.67 & 3.17 \\
SUPIR & 2.00 & 1.57 & 1.14 & 2.64 & 2.29 & 1.21 & 1.21 & 2.07 & 2.58 & 1.08 & 1.00 & 2.75 \\
PiSA-SR & 2.71 & 1.21 & 1.29 & 2.79 & 2.29 & 1.21 & 0.93 & 2.29 & 2.83 & 0.67 & 1.08 & 2.92 \\
SeeSR & 2.50 & 1.36 & 1.21 & 2.57 & 2.21 & 1.43 & 1.14 & 2.50 & 2.25 & 0.92 & 1.33 & 2.42 \\
OSEDiff & 2.21 & 1.64 & 1.57 & 2.21 & 1.93 & 1.71 & 1.71 & 1.86 & 3.00 & 0.83 & 1.00 & 2.92 \\
CCSR & 2.07 & 0.93 & 1.21 & 2.07 & 2.07 & 1.00 & 1.07 & 2.29 & 2.83 & 0.83 & 0.92 & 2.83 \\
DiffBIR & 2.29 & 1.29 & 1.43 & 2.36 & 2.14 & 0.93 & 0.86 & 2.14 & 2.00 & 0.75 & 1.33 & 2.08 \\
StableSR & 1.71 & 2.14 & 1.43 & 2.14 & 1.71 & 1.79 & 1.36 & 2.00 & 2.08 & 1.67 & 1.17 & 2.58 \\
PASD & 2.50 & 0.86 & 0.86 & 2.71 & 2.86 & 0.79 & 0.86 & 2.71 & 2.75 & 0.75 & 0.75 & 2.83 \\
Invsr & 2.36 & 1.07 & 1.21 & 2.43 & 2.00 & 1.71 & 1.43 & 2.14 & 2.67 & 0.67 & 0.83 & 2.67 \\
S3Diff & 2.71 & 0.93 & 0.93 & 2.64 & 2.07 & 1.43 & 1.29 & 2.21 & 2.92 & 0.50 & 0.75 & 3.25 \\
TSD-SR & 2.29 & 1.14 & 1.07 & 2.36 & 2.57 & 0.79 & 0.93 & 2.71 & 2.92 & 0.50 & 0.67 & 2.92 \\
ResShift & 1.43 & 1.93 & 1.79 & 1.71 & 1.43 & 1.79 & 1.57 & 1.64 & 1.58 & 1.42 & 1.42 & 2.00 \\
\midrule
FLUX & 1.79 & 1.36 & 1.29 & 2.43 & 2.43 & 0.93 & 1.14 & 2.50 & 2.50 & 1.08 & 1.08 & 2.67 \\
Nano Banana & 1.43 & 1.86 & 1.50 & 2.00 & 1.43 & 2.14 & 1.71 & 1.93 & 1.58 & 2.00 & 1.58 & 2.33 \\
\midrule
BSRGAN & 1.86 & 1.86 & 1.71 & 2.36 & 1.21 & 2.14 & 2.00 & 1.64 & 2.00 & 1.33 & 1.50 & 2.17 \\
CAL-GAN & 1.64 & 1.93 & 1.86 & 2.29 & 1.29 & 1.71 & 1.50 & 1.14 & 1.58 & 1.50 & 1.67 & 2.17 \\
RealESRGAN & 1.93 & 1.57 & 1.50 & 2.00 & 1.29 & 1.79 & 1.57 & 1.50 & 2.08 & 1.25 & 1.67 & 2.42 \\
\midrule
HAT & 1.36 & 1.93 & 2.07 & 1.64 & 1.57 & 1.57 & 1.29 & 1.86 & 1.58 & 1.67 & 1.58 & 1.92 \\
SwinIR & 1.93 & 1.43 & 1.43 & 2.07 & 1.43 & 1.79 & 1.50 & 1.64 & 2.17 & 1.08 & 1.42 & 2.50 \\
\bottomrule
\end{tabular}
\vspace{-10pt}
\caption{The result of average overall, sharpness, detail, and semantics scores for all restoration models in Hand-drawn, Hand-drawn and Cartoon/Comic. Note that the Detail and Sharpness scores are calculated by taking the absolute value first, and then averaging the scores.}
\label{tab:per_method_Hand-drawn_Print_Media_Cartoon/Comic}
\end{sidewaystable}
\begin{sidewaystable}[tp]
\centering
\footnotesize
\setlength\tabcolsep{4pt}
\begin{tabular}{l|cccc|cccc|cccc}
\toprule
\multirow{2}{*}{Model} & \multicolumn{4}{c|}{Compression}& \multicolumn{4}{c|}{Defocus Blur}& \multicolumn{4}{c}{Digital Zoom} \\
\cline{2-13}
& Overall$\uparrow$ & $\lvert$ Sharpness $\rvert\!\!\downarrow$  & $\lvert$ Detail $\rvert\!\!\downarrow$ & Semantics$\uparrow$ & Overall$\uparrow$ & $\lvert$ Sharpness $\rvert\!\!\downarrow$  & $\lvert$ Detail $\rvert\!\!\downarrow$ & Semantics$\uparrow$ & Overall$\uparrow$ & $\lvert$ Sharpness $\rvert\!\!\downarrow$  & $\lvert$ Detail $\rvert\!\!\downarrow$ & Semantics$\uparrow$ \\
\midrule
HYPIR & 2.76 & 0.63 & 0.78 & 3.00 & 2.95 & 0.95 & 0.77 & 3.27 & 2.59 & 0.98 & 1.00 & 2.98 \\
SUPIR & 2.26 & 1.13 & 1.20 & 2.54 & 2.09 & 1.55 & 1.41 & 2.45 & 2.39 & 1.28 & 1.13 & 2.96 \\
PiSA-SR & 2.52 & 1.02 & 1.02 & 2.85 & 2.77 & 0.86 & 0.91 & 3.18 & 2.35 & 1.26 & 1.09 & 2.78 \\
SeeSR & 2.17 & 1.28 & 1.43 & 2.70 & 2.18 & 1.23 & 1.27 & 2.55 & 2.09 & 1.57 & 1.65 & 2.65 \\
OSEDiff & 2.41 & 1.02 & 1.04 & 2.74 & 2.73 & 0.82 & 0.95 & 3.05 & 2.04 & 1.37 & 1.30 & 2.67 \\
CCSR & 2.24 & 1.04 & 1.22 & 2.50 & 2.36 & 0.91 & 1.05 & 2.64 & 2.15 & 1.17 & 1.28 & 2.61 \\
DiffBIR & 2.22 & 1.22 & 1.33 & 2.50 & 2.59 & 0.82 & 1.00 & 2.73 & 2.17 & 1.28 & 1.13 & 2.59 \\
StableSR & 2.11 & 1.26 & 1.17 & 2.63 & 1.77 & 1.68 & 1.41 & 2.68 & 2.33 & 1.33 & 1.11 & 2.80 \\
PASD & 2.37 & 1.17 & 1.09 & 2.74 & 1.95 & 1.45 & 1.32 & 2.64 & 2.28 & 1.33 & 1.20 & 2.74 \\
Invsr & 2.30 & 1.04 & 1.09 & 2.63 & 2.32 & 1.09 & 1.05 & 2.73 & 2.09 & 1.41 & 1.33 & 2.76 \\
S3Diff & 1.54 & 1.80 & 1.54 & 2.37 & 2.23 & 1.45 & 1.41 & 2.77 & 1.28 & 2.17 & 2.11 & 2.39 \\
TSD-SR & 2.33 & 1.04 & 1.11 & 2.76 & 2.18 & 1.23 & 1.36 & 2.73 & 2.63 & 0.96 & 0.89 & 2.83 \\
ResShift & 1.39 & 1.78 & 1.65 & 2.20 & 1.36 & 2.00 & 1.91 & 2.23 & 1.26 & 1.89 & 1.85 & 2.11 \\
\midrule
FLUX & 1.91 & 1.63 & 1.50 & 2.57 & 2.27 & 1.32 & 1.27 & 2.73 & 2.26 & 1.28 & 1.20 & 2.83 \\
Nano Banana & 2.00 & 1.52 & 1.41 & 2.43 & 1.14 & 2.41 & 2.32 & 1.91 & 1.52 & 1.85 & 1.52 & 2.59 \\
\midrule
BSRGAN & 1.15 & 1.87 & 1.74 & 2.00 & 1.59 & 1.64 & 1.41 & 2.32 & 1.20 & 1.96 & 1.80 & 2.26 \\
CAL-GAN & 1.15 & 1.87 & 1.74 & 1.91 & 1.27 & 2.05 & 1.91 & 2.09 & 0.93 & 2.28 & 2.11 & 2.04 \\
RealESRGAN & 1.48 & 1.70 & 1.74 & 2.24 & 1.73 & 1.82 & 1.64 & 2.41 & 1.39 & 1.98 & 1.85 & 2.24 \\
\midrule
HAT & 1.52 & 1.80 & 1.74 & 1.98 & 1.55 & 2.09 & 1.95 & 2.41 & 1.33 & 1.98 & 1.65 & 2.17 \\
SwinIR & 1.57 & 1.57 & 1.48 & 2.28 & 1.59 & 1.73 & 1.64 & 2.50 & 1.57 & 1.76 & 1.67 & 2.52 \\
\bottomrule
\end{tabular}
\vspace{-10pt}
\caption{The result of average overall, sharpness, detail, and semantics scores for all restoration models in Compression, Compression and Digital Zoom. Note that the Detail and Sharpness scores are calculated by taking the absolute value first, and then averaging the scores.}
\label{tab:per_method_Compression_Defocus_Blur_Digital_Zoom}
\end{sidewaystable}

\begin{sidewaystable}[tp]
\centering
\footnotesize
\setlength\tabcolsep{4pt}
\begin{tabular}{l|cccc|cccc|cccc}
\toprule
\multirow{2}{*}{Model} & \multicolumn{4}{c|}{ISP Noise}& \multicolumn{4}{c|}{Low light}& \multicolumn{4}{c}{Motion Blur} \\
\cline{2-13}
& Overall$\uparrow$ & $\lvert$ Sharpness $\rvert\!\!\downarrow$  & $\lvert$ Detail $\rvert\!\!\downarrow$ & Semantics$\uparrow$ & Overall$\uparrow$ & $\lvert$ Sharpness $\rvert\!\!\downarrow$  & $\lvert$ Detail $\rvert\!\!\downarrow$ & Semantics$\uparrow$ & Overall$\uparrow$ & $\lvert$ Sharpness $\rvert\!\!\downarrow$  & $\lvert$ Detail $\rvert\!\!\downarrow$ & Semantics$\uparrow$ \\
\midrule
HYPIR & 3.12 & 0.54 & 0.71 & 3.25 & 2.39 & 1.13 & 1.05 & 2.61 & 2.43 & 1.50 & 1.64 & 2.79 \\
SUPIR & 2.08 & 1.71 & 1.42 & 2.71 & 1.66 & 1.97 & 1.66 & 2.39 & 0.86 & 2.50 & 2.43 & 2.50 \\
PiSA-SR & 2.75 & 0.96 & 0.88 & 2.88 & 2.45 & 1.21 & 1.08 & 2.66 & 1.43 & 2.14 & 1.57 & 2.21 \\
SeeSR & 1.92 & 1.33 & 1.38 & 2.50 & 1.61 & 1.71 & 1.63 & 2.05 & 1.57 & 2.07 & 2.07 & 2.21 \\
OSEDiff & 2.67 & 0.71 & 0.92 & 2.83 & 1.97 & 1.34 & 1.21 & 2.26 & 2.00 & 1.86 & 1.64 & 2.71 \\
CCSR & 2.29 & 1.29 & 1.08 & 2.58 & 1.74 & 1.42 & 1.24 & 2.21 & 1.00 & 2.21 & 2.14 & 2.43 \\
DiffBIR & 2.67 & 1.00 & 0.92 & 2.96 & 2.05 & 1.42 & 1.37 & 2.45 & 2.14 & 1.57 & 1.50 & 2.79 \\
StableSR & 2.54 & 0.96 & 0.83 & 3.04 & 1.71 & 1.76 & 1.84 & 2.37 & 0.71 & 2.43 & 2.57 & 2.36 \\
PASD & 2.33 & 1.17 & 1.08 & 2.79 & 1.97 & 1.47 & 1.42 & 2.26 & 1.57 & 2.21 & 1.71 & 2.71 \\
Invsr & 2.33 & 1.12 & 1.21 & 2.50 & 1.71 & 1.68 & 1.63 & 2.34 & 1.50 & 2.07 & 2.00 & 2.64 \\
S3Diff & 1.46 & 1.79 & 1.50 & 2.33 & 1.24 & 2.16 & 1.76 & 2.05 & 1.14 & 2.29 & 1.86 & 2.43 \\
TSD-SR & 2.71 & 0.79 & 0.79 & 2.96 & 2.18 & 1.39 & 1.45 & 2.50 & 1.50 & 2.00 & 1.79 & 2.50 \\
ResShift & 1.46 & 2.00 & 1.88 & 2.33 & 1.13 & 2.03 & 1.92 & 2.05 & 0.64 & 2.64 & 2.50 & 1.86 \\
\midrule
FLUX & 2.75 & 0.79 & 0.88 & 3.17 & 1.79 & 1.55 & 1.53 & 2.68 & 2.14 & 1.50 & 1.43 & 2.86 \\
Nano Banana & 1.62 & 1.50 & 1.38 & 2.92 & 1.66 & 1.84 & 1.55 & 2.24 & 0.57 & 2.64 & 2.29 & 2.21 \\
\midrule
BSRGAN & 1.38 & 1.83 & 1.92 & 1.88 & 1.03 & 2.16 & 1.97 & 1.45 & 0.64 & 2.71 & 2.64 & 1.64 \\
CAL-GAN & 1.17 & 1.96 & 1.83 & 1.88 & 0.82 & 2.26 & 2.16 & 1.50 & 0.43 & 2.79 & 2.50 & 1.79 \\
RealESRGAN & 1.54 & 1.71 & 1.79 & 2.17 & 1.08 & 2.13 & 2.03 & 1.87 & 0.57 & 2.50 & 2.50 & 1.50 \\
\midrule
HAT & 1.38 & 1.92 & 1.83 & 2.08 & 0.89 & 2.13 & 1.97 & 1.68 & 0.00 & 3.00 & 2.86 & 1.79 \\
SwinIR & 1.29 & 1.75 & 1.83 & 2.25 & 1.39 & 2.16 & 1.82 & 1.82 & 0.21 & 2.79 & 2.64 & 2.36 \\
\bottomrule
\end{tabular}
\vspace{-10pt}
\caption{The result of average overall, sharpness, detail, and semantics scores for all restoration models in ISP Noise, ISP Noise and Motion Blur. Note that the Detail and Sharpness scores are calculated by taking the absolute value first, and then averaging the scores.}
\label{tab:per_method_ISP_Noise_Low_light_Motion_Blur}
\end{sidewaystable}

\begin{sidewaystable}[tp]
\centering
\footnotesize
\setlength\tabcolsep{4pt}
\begin{tabular}{l|cccc|cccc|cccc}
\toprule
\multirow{2}{*}{Model} & \multicolumn{4}{c|}{Old Film}& \multicolumn{4}{c|}{Old Photo (B/W)}& \multicolumn{4}{c}{Old Photo (Color)} \\
\cline{2-13}
& Overall$\uparrow$ & $\lvert$ Sharpness $\rvert\!\!\downarrow$  & $\lvert$ Detail $\rvert\!\!\downarrow$ & Semantics$\uparrow$ & Overall$\uparrow$ & $\lvert$ Sharpness $\rvert\!\!\downarrow$  & $\lvert$ Detail $\rvert\!\!\downarrow$ & Semantics$\uparrow$ & Overall$\uparrow$ & $\lvert$ Sharpness $\rvert\!\!\downarrow$  & $\lvert$ Detail $\rvert\!\!\downarrow$ & Semantics$\uparrow$ \\
\midrule
HYPIR & 2.68 & 0.96 & 0.99 & 2.85 & 2.79 & 0.88 & 0.95 & 3.03 & 2.90 & 0.77 & 0.77 & 3.00 \\
SUPIR & 1.69 & 1.90 & 1.65 & 2.59 & 2.12 & 1.38 & 1.19 & 2.60 & 2.16 & 1.47 & 1.37 & 2.66 \\
PiSA-SR & 2.46 & 1.25 & 1.32 & 2.79 & 2.67 & 0.86 & 1.02 & 2.88 & 2.61 & 0.84 & 0.89 & 2.91 \\
SeeSR & 2.09 & 1.60 & 1.59 & 2.51 & 2.57 & 1.07 & 1.19 & 2.81 & 2.33 & 1.26 & 1.17 & 2.74 \\
OSEDiff & 2.38 & 1.22 & 1.22 & 2.68 & 2.64 & 0.90 & 0.95 & 2.83 & 2.59 & 1.03 & 1.19 & 2.83 \\
CCSR & 2.07 & 1.38 & 1.46 & 2.35 & 2.41 & 1.07 & 1.10 & 2.83 & 2.33 & 1.14 & 1.21 & 2.66 \\
DiffBIR & 2.35 & 1.19 & 1.25 & 2.68 & 2.78 & 0.97 & 0.98 & 2.86 & 2.51 & 1.03 & 1.13 & 2.83 \\
StableSR & 1.63 & 1.90 & 1.74 & 2.47 & 2.50 & 1.22 & 1.21 & 2.76 & 1.89 & 1.53 & 1.50 & 2.63 \\
PASD & 1.81 & 1.59 & 1.47 & 2.54 & 2.64 & 0.98 & 0.93 & 2.76 & 2.49 & 1.03 & 1.01 & 2.71 \\
Invsr & 2.10 & 1.22 & 1.22 & 2.50 & 2.52 & 1.16 & 1.16 & 2.83 & 2.31 & 0.99 & 1.06 & 2.69 \\
S3Diff & 2.32 & 1.32 & 1.25 & 2.82 & 2.03 & 1.57 & 1.31 & 2.47 & 2.06 & 1.47 & 1.37 & 2.69 \\
TSD-SR & 2.43 & 1.19 & 1.24 & 2.60 & 2.76 & 0.86 & 0.98 & 2.91 & 2.49 & 1.10 & 1.10 & 2.79 \\
ResShift & 1.46 & 2.03 & 1.87 & 2.37 & 1.60 & 1.64 & 1.53 & 2.14 & 1.46 & 1.80 & 1.86 & 2.20 \\
\midrule
FLUX & 1.97 & 1.54 & 1.41 & 2.44 & 2.21 & 0.81 & 0.78 & 2.83 & 2.19 & 1.23 & 1.13 & 2.69 \\
Nano Banana & 1.00 & 2.21 & 2.09 & 2.24 & 1.81 & 1.69 & 1.45 & 2.52 & 1.29 & 2.13 & 1.97 & 2.36 \\
\midrule
BSRGAN & 1.46 & 2.00 & 1.94 & 2.22 & 1.47 & 1.71 & 1.97 & 1.86 & 1.29 & 1.83 & 1.89 & 2.06 \\
CAL-GAN & 1.31 & 2.03 & 1.99 & 2.15 & 1.16 & 1.74 & 1.81 & 1.76 & 1.20 & 2.03 & 1.89 & 2.01 \\
RealESRGAN & 1.38 & 2.10 & 1.94 & 2.35 & 1.52 & 1.62 & 1.69 & 2.03 & 1.39 & 1.87 & 1.90 & 2.11 \\
\midrule
HAT & 1.53 & 2.12 & 1.85 & 2.29 & 1.57 & 1.83 & 1.81 & 2.10 & 1.14 & 1.96 & 1.80 & 2.07 \\
SwinIR & 1.38 & 2.01 & 1.93 & 2.21 & 1.88 & 1.55 & 1.55 & 2.26 & 1.51 & 1.80 & 1.83 & 2.27 \\
\bottomrule
\end{tabular}
\vspace{-10pt}
\caption{The result of average overall, sharpness, detail, and semantics scores for all restoration models in Old Film, Old Film and Old Photo (Color). Note that the Detail and Sharpness scores are calculated by taking the absolute value first, and then averaging the scores.}
\label{tab:per_method_Old_Film_Old_Photo_(B/W)_Old_Photo_(Color)}
\end{sidewaystable}

\begin{sidewaystable}[tp]
\centering
\footnotesize
\setlength\tabcolsep{0pt}
\begin{tabular}{
l|
>{\centering\arraybackslash}p{1.9cm}
>{\centering\arraybackslash}p{1.9cm}
>{\centering\arraybackslash}p{1.9cm}
>{\centering\arraybackslash}p{1.9cm}}
\toprule
\multirow{2}{*}{Model} & \multicolumn{4}{c}{Surveillance} \\
\cline{2-5}
& Overall$\uparrow$ & $\lvert$ Sharpness $\rvert\!\!\downarrow$  & $\lvert$ Detail $\rvert\!\!\downarrow$ & Semantics$\uparrow$ \\
\midrule
HYPIR & 2.25 & 1.43 & 1.36 & 2.50 \\
SUPIR & 2.00 & 1.57 & 1.39 & 2.29 \\
PiSA-SR & 2.00 & 1.64 & 1.54 & 2.46 \\
SeeSR & 2.04 & 1.64 & 1.68 & 2.32 \\
OSEDiff & 2.00 & 1.57 & 1.57 & 2.29 \\
CCSR & 1.61 & 1.82 & 1.64 & 1.96 \\
DiffBIR & 1.96 & 1.61 & 1.68 & 2.25 \\
StableSR & 1.25 & 2.18 & 2.11 & 1.89 \\
PASD & 2.25 & 1.18 & 1.29 & 2.68 \\
Invsr & 1.57 & 1.64 & 1.64 & 1.79 \\
S3Diff & 2.21 & 1.43 & 1.61 & 2.39 \\
TSD-SR & 1.79 & 1.61 & 1.61 & 2.11 \\
ResShift & 1.21 & 2.43 & 2.21 & 1.71 \\
\midrule
FLUX & 1.25 & 2.21 & 1.93 & 1.89 \\
Nano Banana & 0.89 & 2.25 & 2.14 & 1.46 \\
\midrule
BSRGAN & 1.43 & 2.21 & 2.14 & 1.96 \\
CAL-GAN & 1.00 & 2.18 & 2.11 & 1.61 \\
RealESRGAN & 1.39 & 2.18 & 2.07 & 2.21 \\
\midrule
HAT & 1.25 & 2.36 & 2.11 & 1.89 \\
SwinIR & 1.21 & 2.18 & 2.11 & 2.00 \\
\bottomrule
\end{tabular}
\vspace{-10pt}
\caption{The result of average overall, sharpness, detail, and semantics scores for all restoration models in Surveillance. Note that the Detail and Sharpness scores are calculated by taking the absolute value first, and then averaging the scores.}
\label{tab:per_method_Surveillance}
\end{sidewaystable}
\end{document}